\pdfoutput=1

\documentclass[a4paper,conference]{IEEEtran}

\ifCLASSINFOpdf
  \usepackage[pdftex]{graphicx}
  \DeclareGraphicsExtensions{.pdf,.jpg,.png}
\else
\fi

\usepackage{multirow}
\usepackage{threeparttable}
\usepackage{makecell}
\usepackage{amsmath}
\usepackage{amssymb}
\usepackage{xcolor}

\begin{document}
%
\title{Deep Image Destruction:\\Vulnerability of Deep Image-to-Image Models against Adversarial Attacks}

\author{\IEEEauthorblockN{Jun-Ho Choi\IEEEauthorrefmark{1},
Huan Zhang\IEEEauthorrefmark{2},
Jun-Hyuk Kim\IEEEauthorrefmark{1},
Cho-Jui Hsieh\IEEEauthorrefmark{2}, and
Jong-Seok Lee\IEEEauthorrefmark{1}}
\IEEEauthorblockA{\IEEEauthorrefmark{1}School of Integrated Technology, Yonsei University, Korea\\
Email: \{idearibosome, junhyuk.kim, jong-seok.lee\}@yonsei.ac.kr}
\IEEEauthorblockA{\IEEEauthorrefmark{2}Department of Computer Science, University of California, Los Angeles, California\\
Email: huanzhang@ucla.edu, chohsieh@cs.ucla.edu}}


\maketitle
\thispagestyle{plain}
\pagestyle{plain}

\begin{abstract}
  Recently, the vulnerability of deep image classification models to adversarial attacks has been investigated.
  However, such an issue has not been thoroughly studied for image-to-image tasks that take an input image and generate an output image (e.g., colorization, denoising, deblurring, etc.)
  This paper presents comprehensive investigations into the vulnerability of deep image-to-image models to adversarial attacks.
  For five popular image-to-image tasks, 16 deep models are analyzed from various standpoints such as output quality degradation due to attacks, transferability of adversarial examples across different tasks, and characteristics of perturbations.
  We show that unlike image classification tasks, the performance degradation on image-to-image tasks largely differs depending on various factors, e.g., attack methods and task objectives.
  In addition, we analyze the effectiveness of conventional defense methods used for classification models in improving the robustness of the image-to-image models.
\end{abstract}


%

\newcommand{\STAB}[1]{\begin{tabular}{@{}c@{}}#1\end{tabular}}

\section{Introduction}

The deep learning technology brings us tremendous advantages in various computer vision fields.
On the other side, recent studies have shown that deep learning-based algorithms are highly vulnerable to adversarial attacks, which add imperceptible noise to input images to fool the target deep model.
Such vulnerability has been investigated chiefly on image classification models \cite{goodfellow2014explaining,su2018robustness,dong2020benchmarking}.

Meanwhile, deep image-to-image models have also been developed, which receive an input image and generate an output image.
Unlike image classification models, image-to-image models cover much broader task objectives, including colorization \cite{zhang2016colorful}, super-resolution \cite{lim2017enhanced}, denoising \cite{zhang2017beyond}, deblurring \cite{nah2017deep}, and translation \cite{zhu2017unpaired}.
Image-to-image models are usually employed as pre-processing parts of computational systems or as standalone systems to generate images having better quality.
A malicious attack can cause the whole system containing an image-to-image model to malfunction \cite{yin2018deep}; e.g., in a system performing image enhancement and then classification, attacking the enhancement model can affect its output so that the classification model produces a wrong result.
Alternatively, an attack can change the visual content of the output image \cite{choi2019evaluating}, e.g., changing the text or the color of a traffic light in the image, which can induce erroneous judgment of both human viewers and computers.
Nevertheless, the vulnerability issue of deep models for image-to-image tasks has not been studied much compared to that of the image classification task.
Because the way that image-to-image models deal with image data largely differs from that for image classification models, the characteristics of such a vulnerability issue may also differ from those for image classification.

Let $\mathbf{X}$ and $\widetilde{\mathbf{X}}$ denote the original and attacked input images, respectively, i.e., $\widetilde{\mathbf{X}} = \mathbf{X} + \Delta$ for a small perturbation $\Delta$.
For a given classification model ${f}_{c}(\cdot)$, which outputs a class label $y$ for input $\mathbf{X}$, the goal of the adversarial attack is to achieve ${f}_{c}(\widetilde{\mathbf{X}}) \neq y$.
By contrast, because the image-to-image models output images instead of class probability distributions, the goal of an adversarial attack can be set to maximize the difference between the original output ${f}_{m}(\mathbf{X})$ and the attacked output ${f}_{m}(\widetilde{\mathbf{X}})$, where ${f}_{m}(\cdot)$ is the target image-to-image model, i.e.,
$\max_{\Delta} {d \big( {f}_{m}(\mathbf{X}), {f}_{m}(\widetilde{\mathbf{X}}) \big)}$,
where $d(\cdot, \cdot)$ is a distance measure.
Due to this fundamental difference of adversarial attacks between classification models and image-to-image models, the following essential research questions need to be answered, which is the objective of this paper.

\begin{itemize}
	\item
	\textbf{How can we measure the success of adversarial attacks for image-to-image models?}
	The metrics used in image classification models (e.g., success rate, fooling rate) cannot be used directly.
	
	\item
	\textbf{What are the characteristics of adversarial attacks for image-to-image models?}
	Because of the different properties of the models, the vulnerability of image-to-image models may appear differently in various factors.
	
	\item
	\textbf{What are the right ways to improve the robustness of image-to-image models?}
	The applicability of conventional defense methods used in image classification to image-to-image tasks needs to be examined.
\end{itemize}
We conduct comprehensive investigations on the vulnerability of deep image-to-image models against adversarial attacks.
We expose five popular image-to-image tasks to malicious attacks and examine various factors of the vulnerability.
In addition, we examine the feasibility of defense methods that are conventionally applied to image classification models.

\section{Related Work}
\label{sec:related_work}

Various adversarial attack methods for image classification models have been developed \cite{su2018robustness,dong2020benchmarking}.
Szegedy \textit{et al.} \cite{szegedy2013intriguing} developed an optimization-based approach to make a given model produce wrong classification results with minimum amounts of input perturbation.
Goodfellow \textit{et al.} \cite{goodfellow2014explaining} proposed the fast gradient sign method (FGSM), which calculates the perturbation from the sign of the gradients obtained for a given model, and Kurakin \textit{et al.} \cite{kurakin2016adversarial} extended it to an iterative approach, called I-FGSM.
Ganeshan and Babu \cite{ganeshan2019fda} proposed feature disruptive attack (FDA), which finds perturbation from the intermediate features of a given model.
Other vulnerability issues also have been investigated such as finding universal perturbation \cite{moosavi2017universal,zhang2021survey} and measuring transferability among different models \cite{liu2016delving,wang2021enhancing}.
These mostly focus on the image classification task, and an in-depth study on vulnerability of image-to-image models has not been conducted.

Defense methods on classification models have also been proposed.
One approach is to transform the input images before feeding them to a given model to reduce the amount of perturbation, including JPEG compression \cite{dziugaite2016study}, bit depth reduction \cite{xu2018feature}, and random resizing \cite{xie2018mitigating}.
Another effective way is adversarial training, which uses images containing adversarial perturbation as training data \cite{goodfellow2014explaining,kurakin2016adversarial,pang2021bag}.

\section{Methods}
\label{sec:method}

We consider 16 deep models for five popular image-to-image tasks: colorization (CIC \cite{zhang2016colorful}), deblurring (DeepDeblur \cite{nah2017deep}), denoising (three DnCNN models \cite{zhang2017beyond} for grayscale or color images with a single or multiple Gaussian noise levels ($\sigma$)), super-resolution (EDSR \cite{lim2017enhanced}, RCAN \cite{zhang2018image}, CARN \cite{ahn2018fast}, SRResNet \cite{ledig2017photo}, and SRGAN \cite{ledig2017photo}), and translation (three CycleGAN models \cite{zhu2017unpaired} for apple$\leftrightarrow$orange, horse$\leftrightarrow$zebra, and Van Gogh's paintings$\leftrightarrow$photos).
We consider two types of attack methods that are used for image classification models and are applicable to image-to-image models: feature-based attack (FDA \cite{ganeshan2019fda}) and gradient-based attack (I-FGSM \cite{choi2019evaluating,kurakin2016adversarial}).
The feature-based attack tries to reduce the variance of the intermediate activations in the target model.
It does not rely on the model output to find perturbation, so it can be used to attack image-to-image models as well as classification models.
The gradient-based attack iteratively finds the attacked input $\mathbf{\widetilde{X}}$ from the gradient sign of the L2 difference between the original output and attacked output images.
We set the pixel value scale of the images as $[0, 255]$, and limit the ${L}_{\infty}$ norm of perturbation $\epsilon$ to $\epsilon=8$ in all cases except the universal attacks using $\epsilon=16$.

\subsection{Quantitative Vulnerability Evaluation}
One big challenge to evaluate image-to-image models is that it is hard to use evaluation methods used for image classification models directly (e.g., success rate or fooling rate).
To this end, we design an evaluation metric based on the peak signal-to-noise ratio (PSNR), which is one of the most widely used metrics for evaluating the quality of image-to-image tasks.
Given that the aim of attacks is to minimize the amount of perturbation in the input image but maximize the amount of deterioration in the output image, we first measure PSNR for both input and output images, denoted by $Q_i$ and $Q_o$, respectively.
$Q_i$ is measured between the original input image $\mathbf{X}$ and the attacked one $\mathbf{\widetilde{X}}$, indicating the amount of injected perturbation.
$Q_o$ is measured between the output image for the original input image ${f}_{m}(\mathbf{X})$ and the output for the attacked input image ${f}_{m}(\mathbf{\widetilde{X}})$, quantifying the amount of deterioration due to the attack.

If an image-to-image model is more vulnerable to adversarial attacks than another, a smaller amount of perturbation injected into the input image can deteriorate the quality of its output image more significantly.
In that case, ${Q}_{i}$ would be larger, while ${Q}_{o}$ would be smaller.
Considering this, we define a \emph{vulnerability index (VI)} that measures the degree of vulnerability, which is calculated as $\mathrm{VI} = {Q}_{i} / {Q}_{o}$.
A large VI indicates high vulnerability.
We consider the degradations of both input and output images because we found that the amount of degradation in input images can significantly differ depending on the target image-to-image tasks and models, even under the same $\epsilon$ value.
Note that other metrics such as SSIM can be used instead of PSNR for VI.
Using SSIM, we observed similar trends to those using PSNR, which are omitted due to the space limit.

\subsection{Transferability of Universal Perturbations}
It is possible to find a \textit{universal} perturbation that can affect any input image for a given model \cite{moosavi2017universal}.
This can be obtained by applying a given attack method (i.e., FDA or I-FGSM) to the averaged deterioration, instead of the deterioration of each output image.
In addition, because the universal perturbation does not rely on a specific input image, it enables us to investigate its \textit{transferability} \cite{liu2016delving,wu2020towards}, which refers to the applicability of the universal perturbation found for a model to another model.
By measuring transferability of different perturbations, it is possible to determine whether some models have similar characteristics in terms of vulnerability.

We investigate the transferability of universal perturbations in two-fold.
We examine the transferability of universal perturbations from one image-to-image model to another.
In addition, we further investigate the transferability between image-to-image and classification models (VGG16 \cite{simonyan2014very}, ResNet-101 \cite{he2016deep}, MobileNetV2 \cite{sandler2018mobilenetv2}).
These can clarify whether the perturbations are transferable across different tasks, not only between image-to-image tasks but also from/to the image classification task that is a completely different type of task.

\subsection{Characteristics of Adversarial Examples}
Image-to-image models exhibit various patterns of qualitative degradation depending on the characteristics of the perturbations.
In classification, different characteristics of attack approaches are revealed mainly through changes in quantitative performance (e.g., success rate and computational complexity).
In image-to-image tasks, however, the differences according to the input perturbations can be directly observed through changes in the output images.
In this regard, we investigate the characteristics of adversarial perturbations by manipulating frequency components during the I-FGSM process, which was recently applied to the classification task \cite{guo2019low,sharma2019effectiveness}.
Let ${\Delta}^{(i)} \in \mathbb{R}^{{W}\times{H}\times{C}}$ denote the perturbation for a given input image $\mathbf{X}$ found at the $i$-th iteration, where $W$, $H$, and $C$ are the width, height, and number of channels of the input image, respectively.
We obtain low-frequency components of the perturbation by
$\widehat{\Delta}^{(i)} = \mathrm{IDCT} \big( \mathbf{M} \circ \mathrm{DCT} ( {\Delta}^{(i)} ) \big)$,
where $\mathrm{DCT}(\cdot)$ and $\mathrm{IDCT}(\cdot)$ are the 2-D discrete cosine transform (DCT) and inverse DCT, respectively, $\circ$ denotes the element-wise multiplication, and $\mathbf{M} \in \mathbb{R}^{{W}\times{H}\times{C}}$ is a binary mask that extracts certain frequency components of the perturbation.
For attacking the low-frequency subspace, the value of $\mathbf{M}$ at $(w, h, c)$ is set to one when $w$$\leq$$rW$ and $h$$\leq$$rH$, and zero otherwise.
The parameter $r \in [0, 1]$ controls the range of the low-frequency subspace.
Finally, the I-FGSM update rule is applied from the gradient sign of the L2 difference between the original output image and the output image obtained from $\mathbf{X} + \widehat{\Delta}^{(i)}$.
We also investigate the effectiveness of attacking the high-frequency subspace by setting the value of $\mathbf{M}$ at $(w, h, c)$ to one when $w$$\geq$$(1-r)W$ and $h$$\geq$$(1-r)H$, and zero otherwise.

\section{Experimental Results}
\label{sec:results}

\subsection{Vulnerability Evaluation}
\label{subsec:results_evaluation}

\begin{figure}[t]
	\begin{center}
		\centering
		\renewcommand{\arraystretch}{1.5}
		\renewcommand{\tabcolsep}{2.0pt}
		\footnotesize
		\begin{tabular}{cccc}
			\raisebox{-0.5\height}{\includegraphics[width=0.265\linewidth]{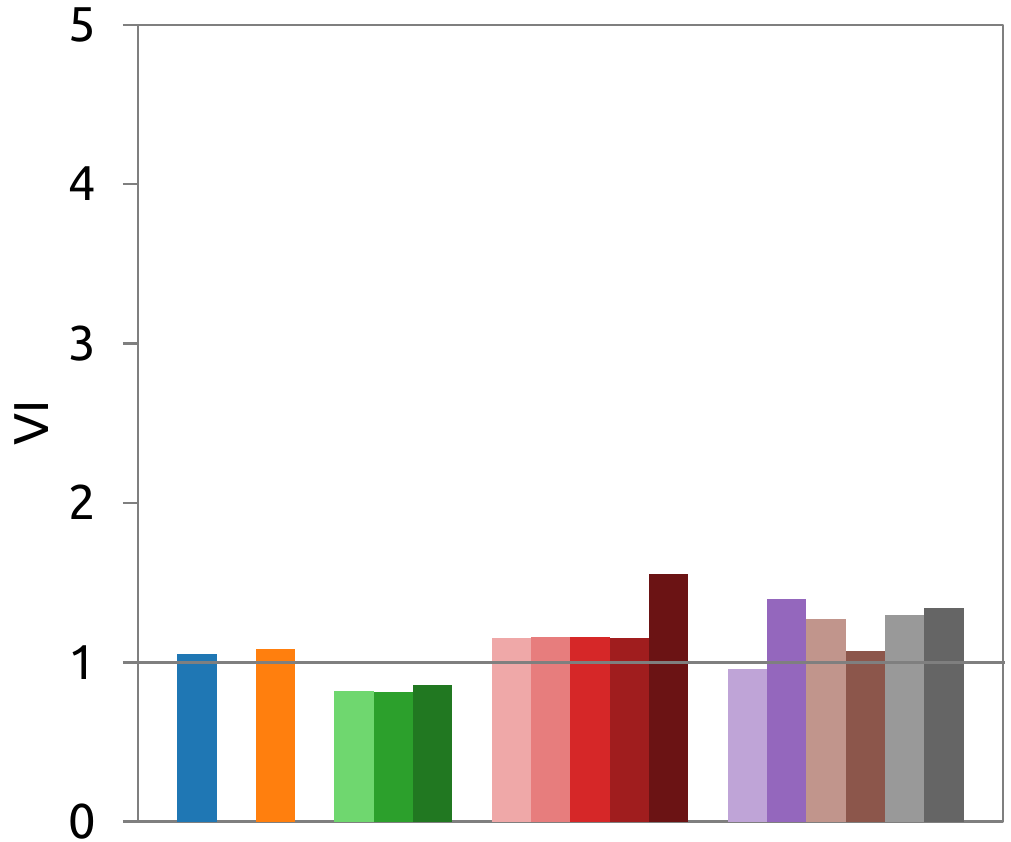}} &
			\raisebox{-0.5\height}{\includegraphics[width=0.265\linewidth]{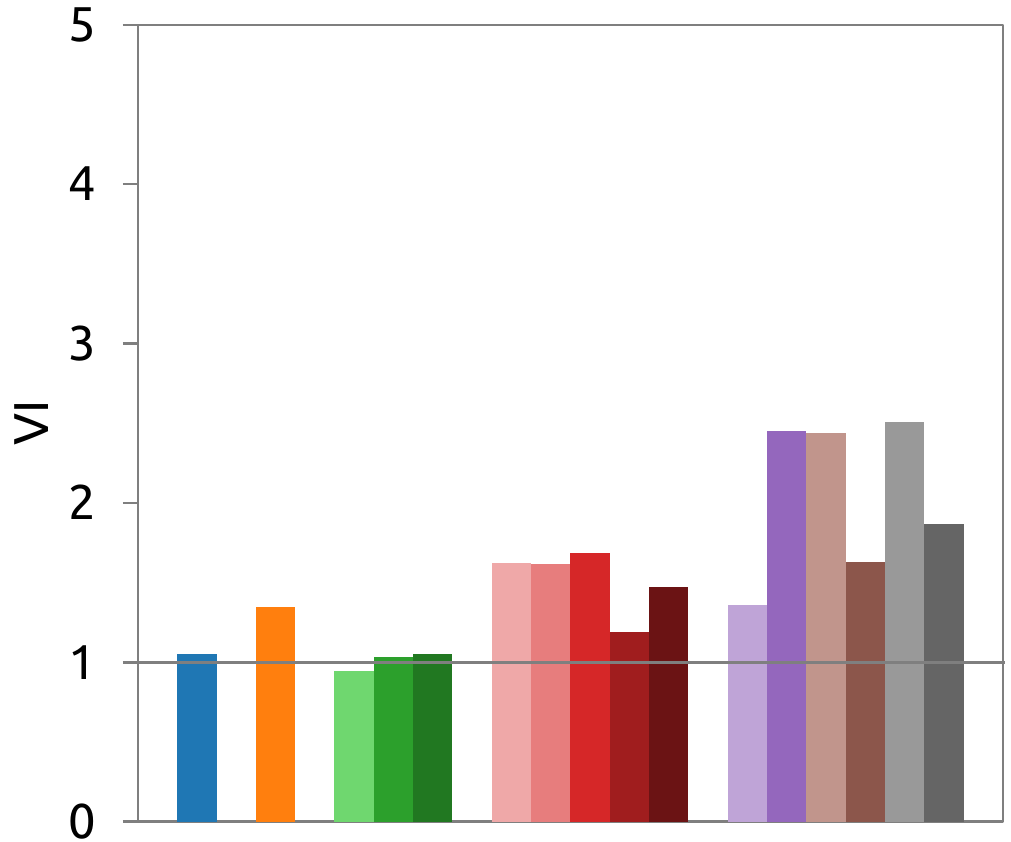}} & \raisebox{-0.5\height}{\includegraphics[width=0.265\linewidth]{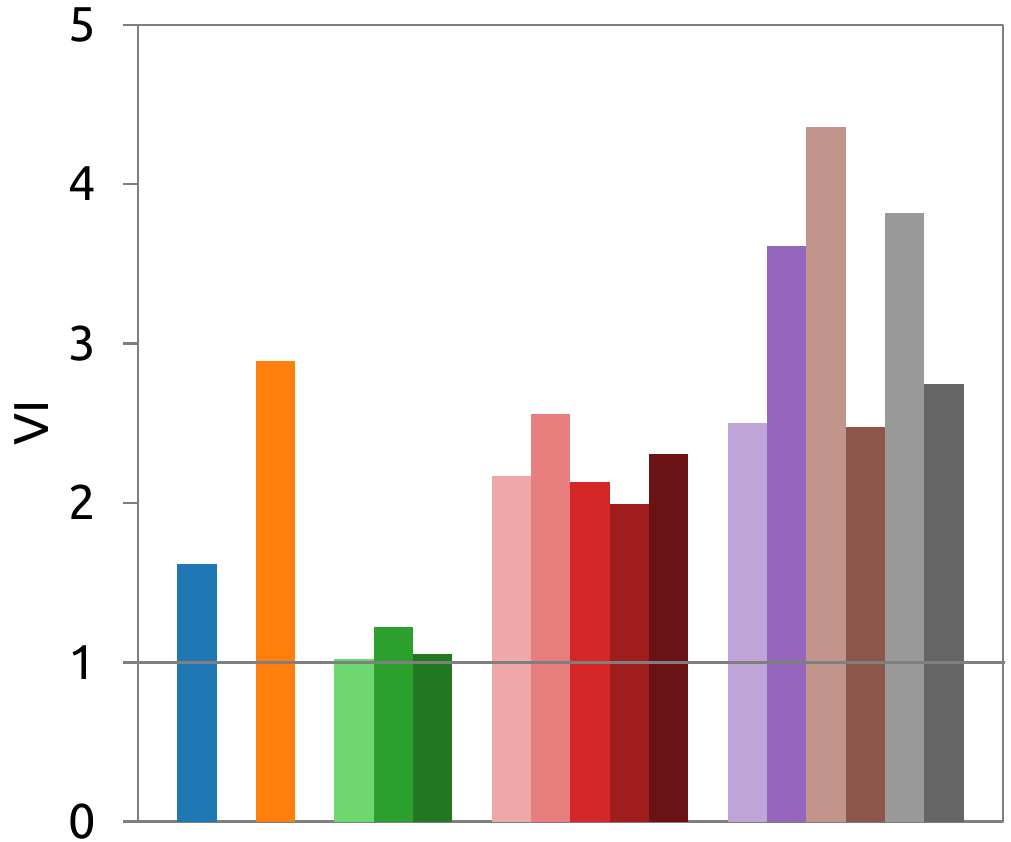}} &
			\multirow{2}{*}[17.5pt]{\raisebox{-0.5\height}{\includegraphics[width=0.146\linewidth]{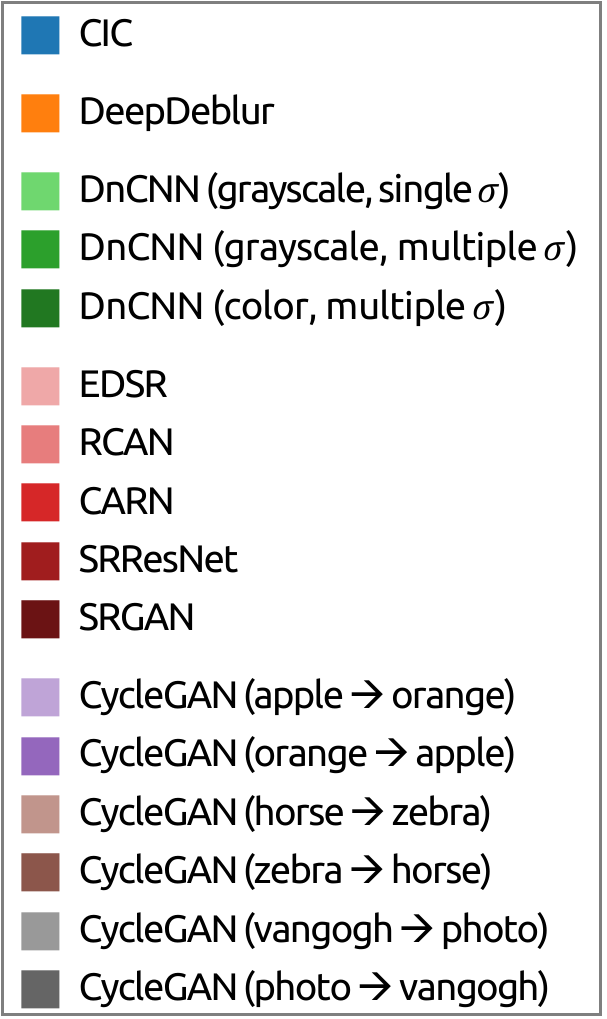}}}\\
			(a) Random noise & (b) FDA & (c) I-FGSM
		\end{tabular}
	\end{center}
	\caption{Performance in terms of VI for different attack methods.}
	\label{fig:basic_attack_psnrratio}
\end{figure}

\begin{figure}[t]
	\begin{center}
		\centering
		\renewcommand{\arraystretch}{0.2}
		\renewcommand{\tabcolsep}{0.5pt}
		\scriptsize
		\begin{tabular}{ccccccc}
			& & Colorization & Deblurring & Denoising & Super-resolution & Translation \\ \\
			\multirow{3}{*}[-4ex]{\rotatebox[origin=c]{90}{\footnotesize{Original}}} &
			\rotatebox[origin=c]{90}{Input} & 
			\raisebox{-0.5\height}{\includegraphics[width=0.182\linewidth]{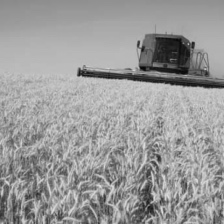}} &
			\raisebox{-0.5\height}{\includegraphics[width=0.182\linewidth]{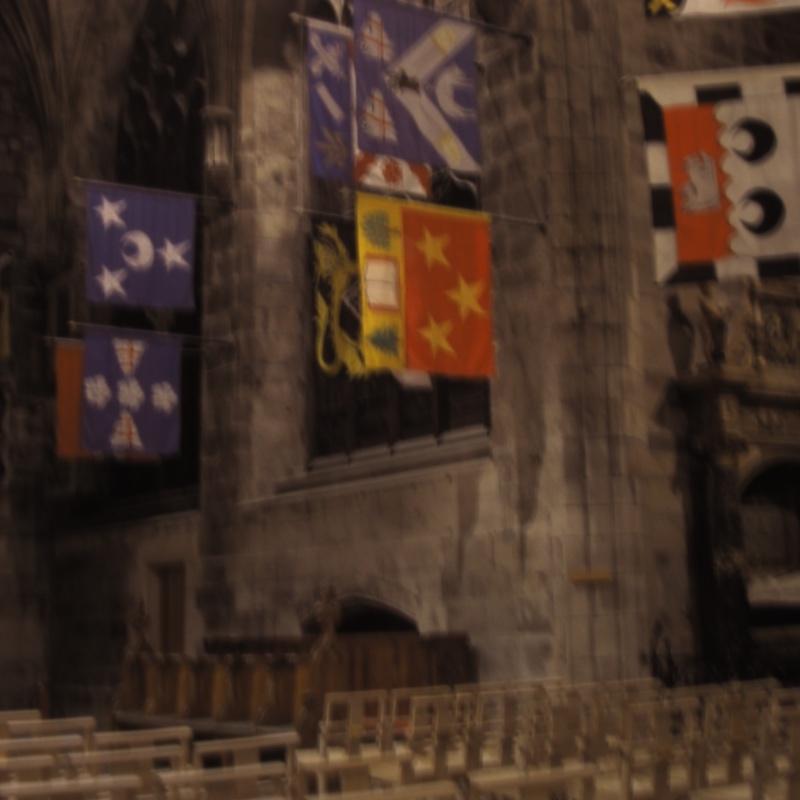}} &
			\raisebox{-0.5\height}{\includegraphics[trim={0 80px 0 80px},clip,width=0.182\linewidth]{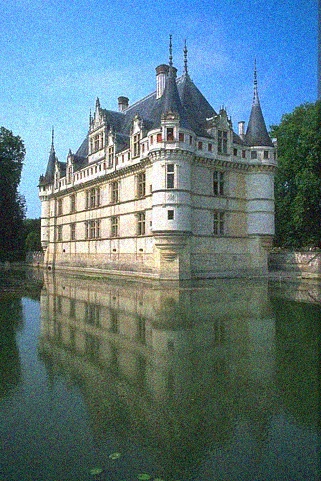}} &
			\raisebox{-0.5\height}{\includegraphics[trim={0 20px 0 20px},clip,width=0.182\linewidth]{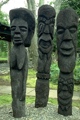}} &
			\raisebox{-0.5\height}{\includegraphics[width=0.182\linewidth]{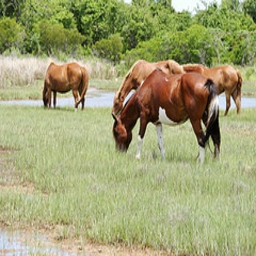}} \\ \\
			&
			\rotatebox[origin=c]{90}{Output} & 
			\raisebox{-0.5\height}{\includegraphics[width=0.182\linewidth]{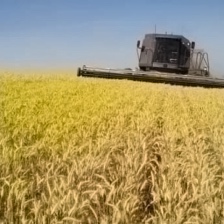}} &
			\raisebox{-0.5\height}{\includegraphics[width=0.182\linewidth]{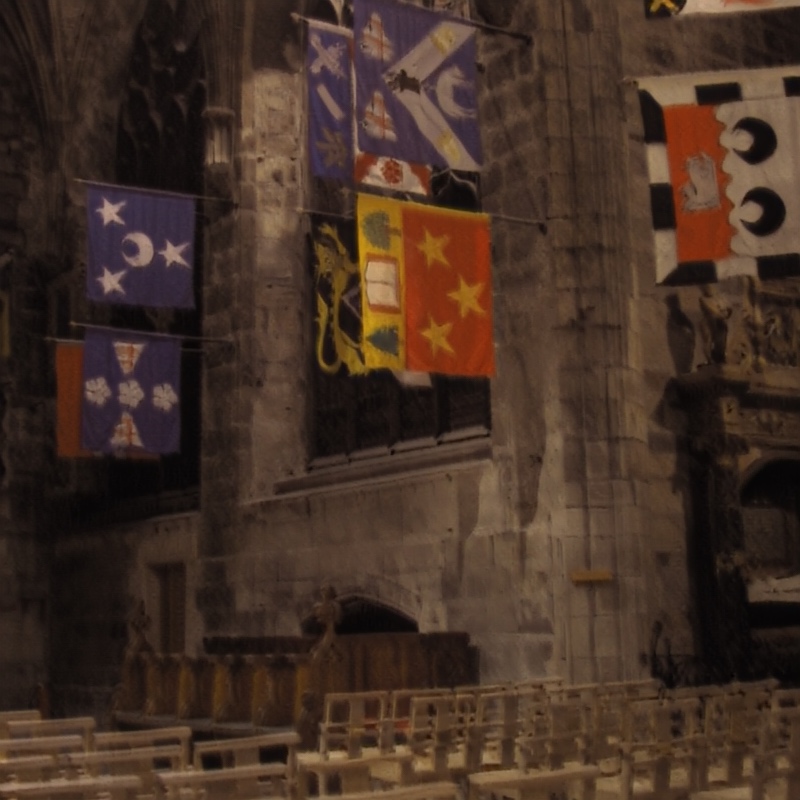}} &
			\raisebox{-0.5\height}{\includegraphics[trim={0 80px 0 80px},clip,width=0.182\linewidth]{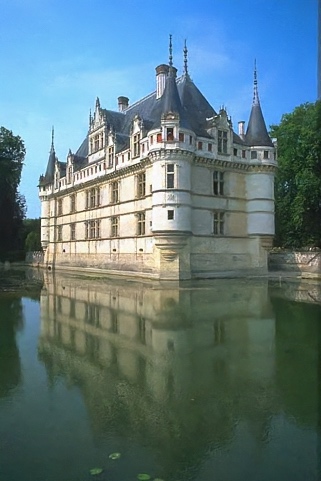}} &
			\raisebox{-0.5\height}{\includegraphics[trim={0 80px 0 80px},clip,width=0.182\linewidth]{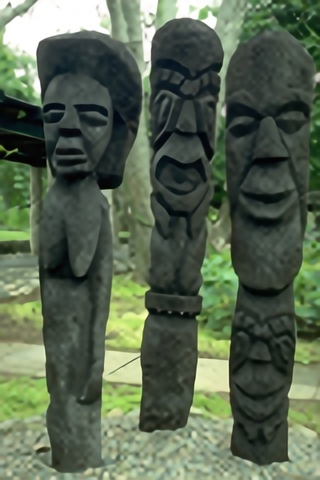}} &
			\raisebox{-0.5\height}{\includegraphics[width=0.182\linewidth]{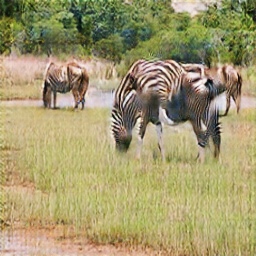}} \\ \\
			\\ \\
			\multirow{5}{*}[-4.5ex]{\rotatebox[origin=c]{90}{\footnotesize{Feature-based (FDA)}}} &
			\rotatebox[origin=c]{90}{Perturbation} & 
			\raisebox{-0.5\height}{\includegraphics[width=0.182\linewidth]{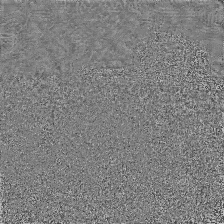}} &
			\raisebox{-0.5\height}{\includegraphics[width=0.182\linewidth]{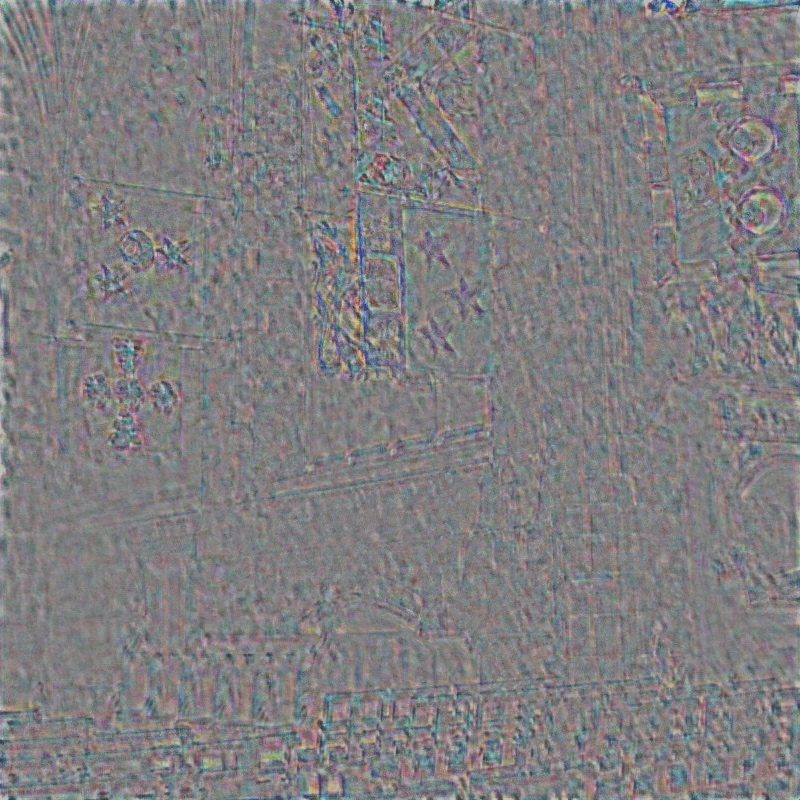}} &
			\raisebox{-0.5\height}{\includegraphics[trim={0 80px 0 80px},clip,width=0.182\linewidth]{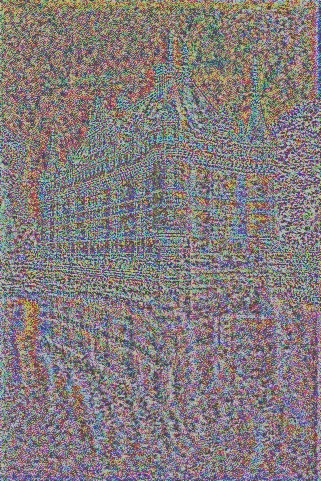}} &
			\raisebox{-0.5\height}{\includegraphics[trim={0 20px 0 20px},clip,width=0.182\linewidth]{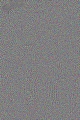}} &
			\raisebox{-0.5\height}{\includegraphics[width=0.182\linewidth]{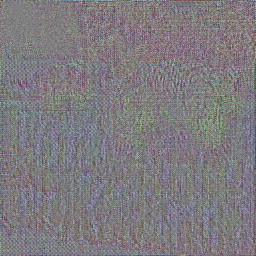}} \\ \\
			&
			\rotatebox[origin=c]{90}{Input} & 
			\raisebox{-0.5\height}{\includegraphics[width=0.182\linewidth]{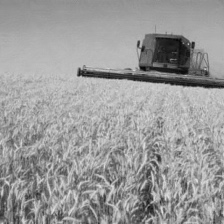}} &
			\raisebox{-0.5\height}{\includegraphics[width=0.182\linewidth]{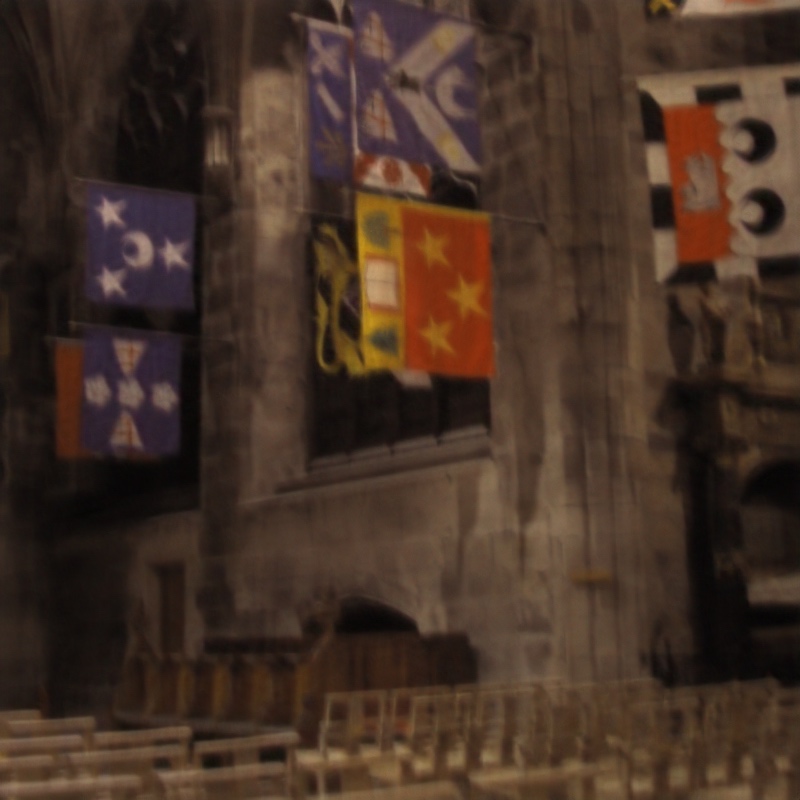}} &
			\raisebox{-0.5\height}{\includegraphics[trim={0 80px 0 80px},clip,width=0.182\linewidth]{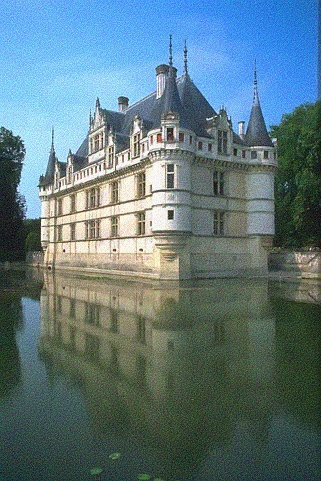}} &
			\raisebox{-0.5\height}{\includegraphics[trim={0 20px 0 20px},clip,width=0.182\linewidth]{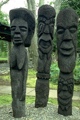}} &
			\raisebox{-0.5\height}{\includegraphics[width=0.182\linewidth]{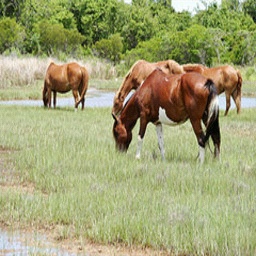}} \\ \\
			&
			\rotatebox[origin=c]{90}{Output} & 
			\raisebox{-0.5\height}{\includegraphics[width=0.182\linewidth]{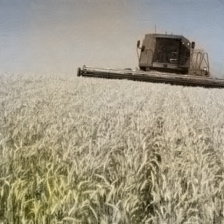}} &
			\raisebox{-0.5\height}{\includegraphics[width=0.182\linewidth]{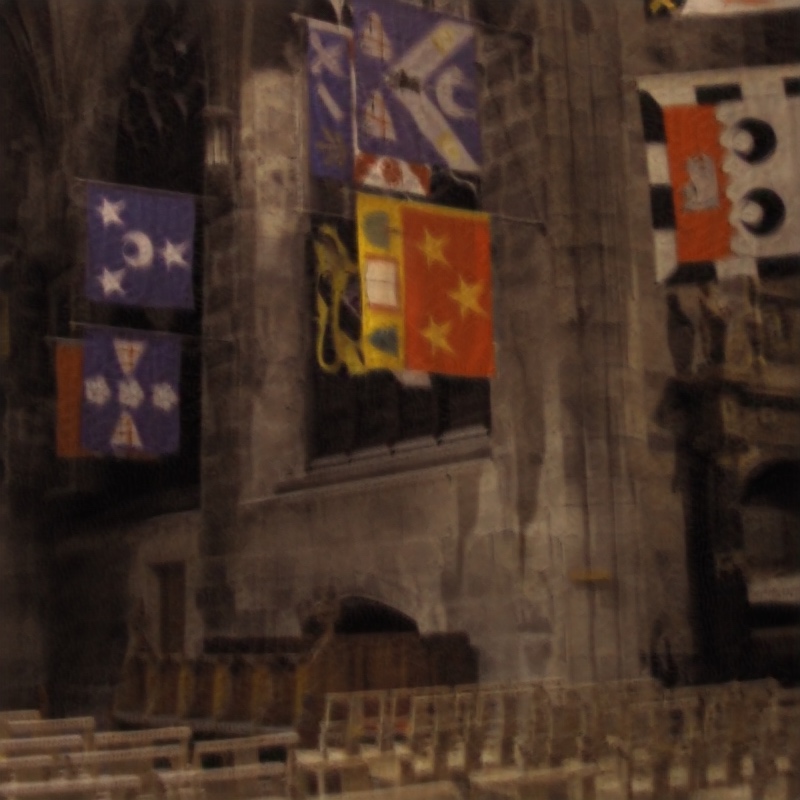}} &
			\raisebox{-0.5\height}{\includegraphics[trim={0 80px 0 80px},clip,width=0.182\linewidth]{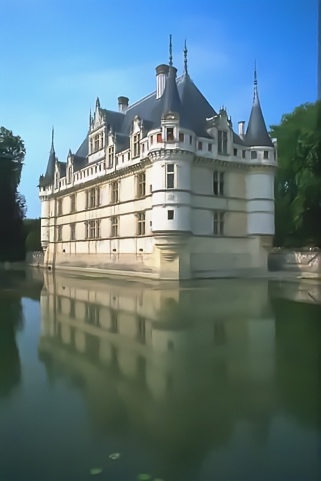}} &
			\raisebox{-0.5\height}{\includegraphics[trim={0 80px 0 80px},clip,width=0.182\linewidth]{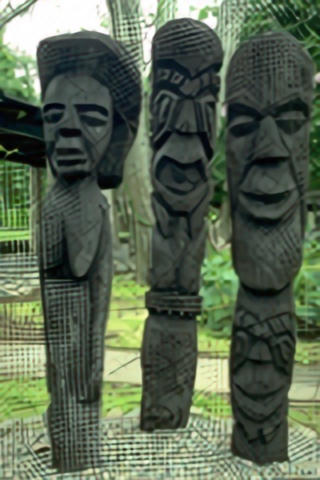}} &
			\raisebox{-0.5\height}{\includegraphics[width=0.182\linewidth]{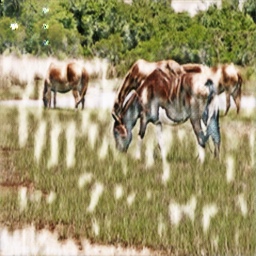}} \\ \\
			\\ \\
			\multirow{5}{*}[-2ex]{\rotatebox[origin=c]{90}{\footnotesize{Gradient-based (I-FGSM)}}} &
			\rotatebox[origin=c]{90}{Perturbation} & 
			\raisebox{-0.5\height}{\includegraphics[width=0.182\linewidth]{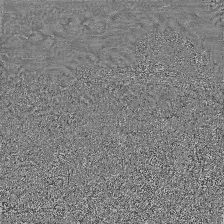}} &
			\raisebox{-0.5\height}{\includegraphics[width=0.182\linewidth]{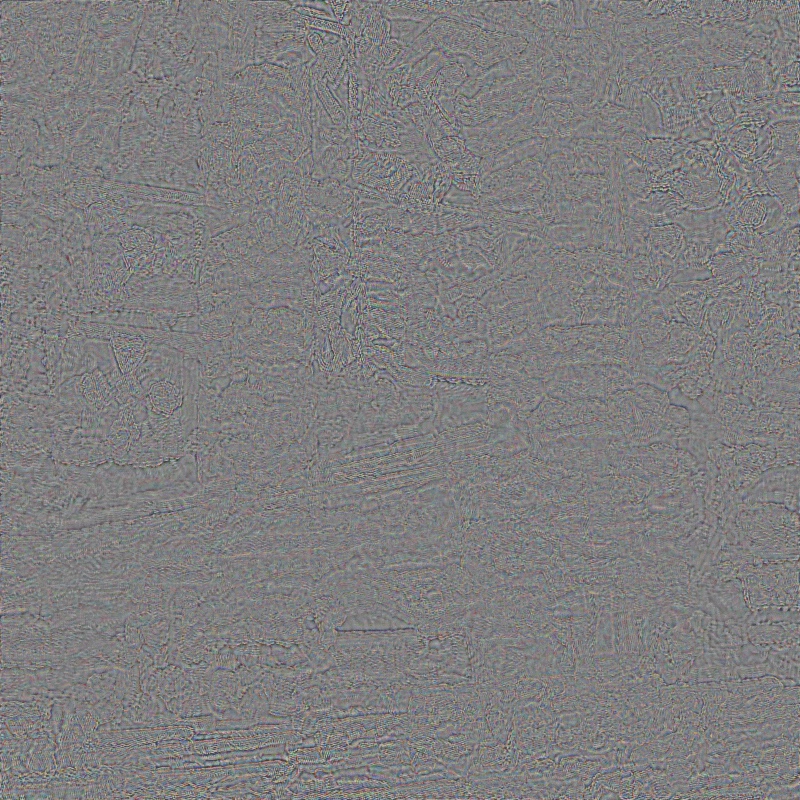}} &
			\raisebox{-0.5\height}{\includegraphics[trim={0 80px 0 80px},clip,width=0.182\linewidth]{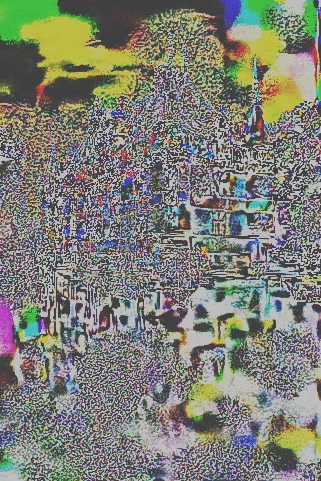}} &
			\raisebox{-0.5\height}{\includegraphics[trim={0 20px 0 20px},clip,width=0.182\linewidth]{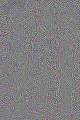}} &
			\raisebox{-0.5\height}{\includegraphics[width=0.182\linewidth]{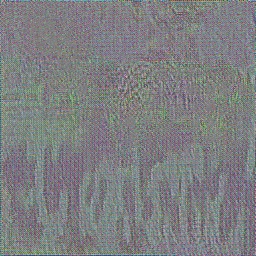}} \\ \\
			&
			\rotatebox[origin=c]{90}{Input} & 
			\raisebox{-0.5\height}{\includegraphics[width=0.182\linewidth]{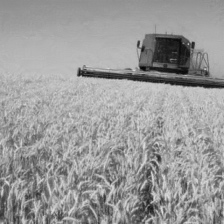}} &
			\raisebox{-0.5\height}{\includegraphics[width=0.182\linewidth]{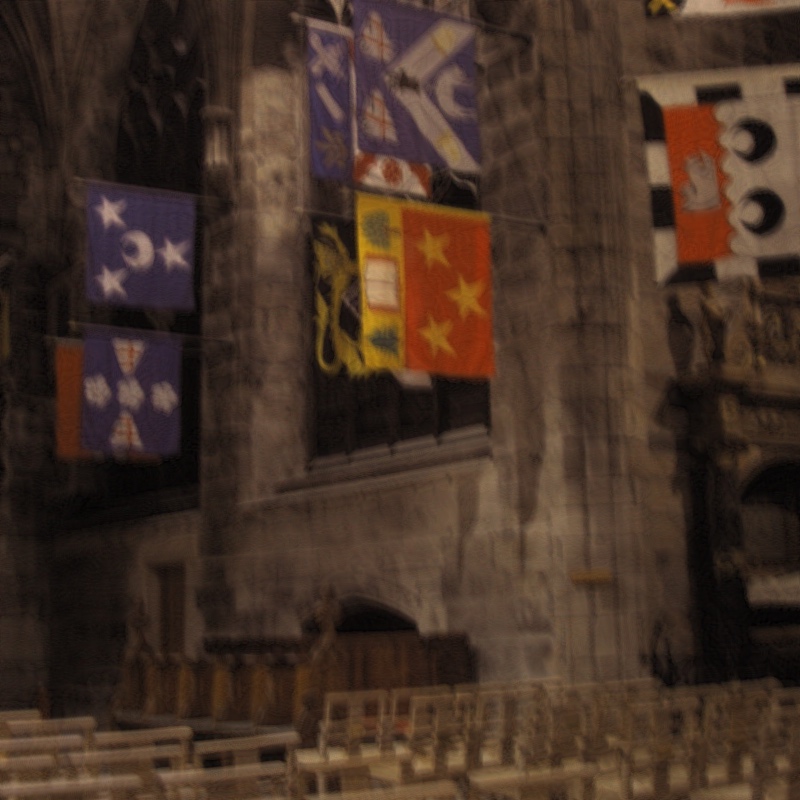}} &
			\raisebox{-0.5\height}{\includegraphics[trim={0 80px 0 80px},clip,width=0.182\linewidth]{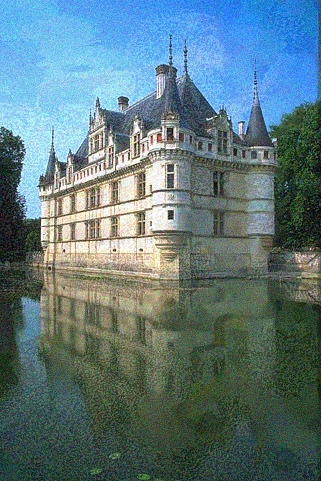}} &
			\raisebox{-0.5\height}{\includegraphics[trim={0 20px 0 20px},clip,width=0.182\linewidth]{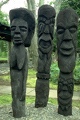}} &
			\raisebox{-0.5\height}{\includegraphics[width=0.182\linewidth]{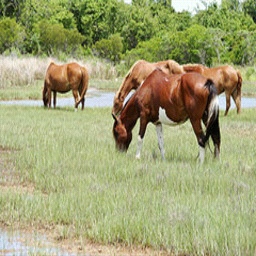}} \\ \\
			&
			\rotatebox[origin=c]{90}{Output} & 
			\raisebox{-0.5\height}{\includegraphics[width=0.182\linewidth]{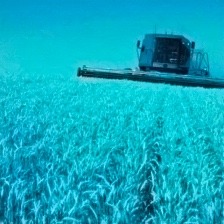}} &
			\raisebox{-0.5\height}{\includegraphics[width=0.182\linewidth]{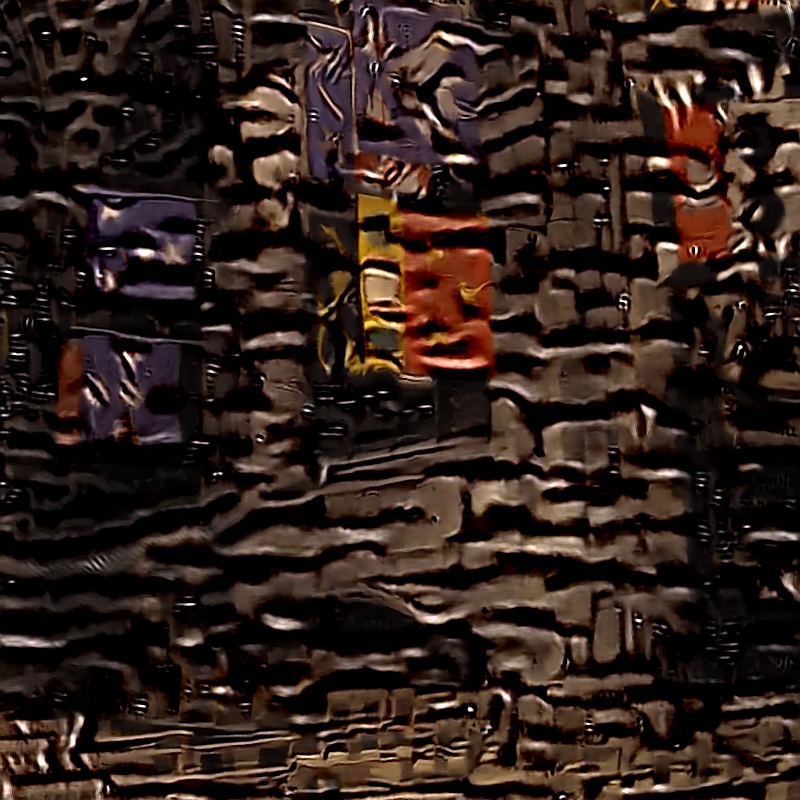}} &
			\raisebox{-0.5\height}{\includegraphics[trim={0 80px 0 80px},clip,width=0.182\linewidth]{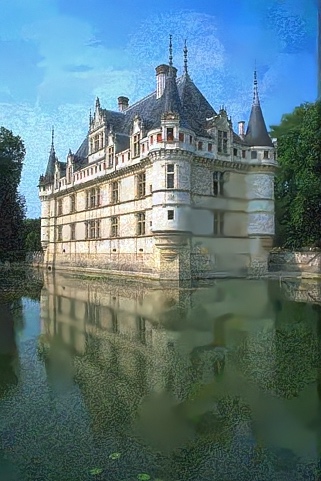}} &
			\raisebox{-0.5\height}{\includegraphics[trim={0 80px 0 80px},clip,width=0.182\linewidth]{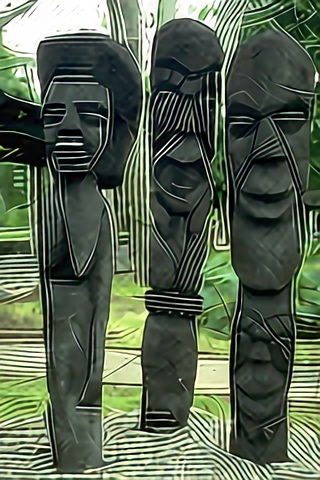}} &
			\raisebox{-0.5\height}{\includegraphics[width=0.182\linewidth]{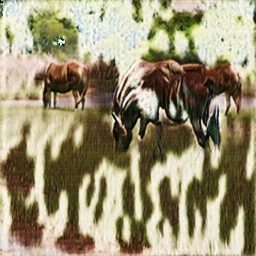}}
		\end{tabular}
	\end{center}
	\caption{Visual showcases of the input and output images obtained from CIC, DeepDeblur, DnCNN (color, multiple $\sigma$), RCAN, and CycleGAN (horse$\rightarrow$zebra). Each perturbation is magnified $\times$8 for better visualization.}
	\label{fig:basic_attack_examples}
\end{figure}

\figurename~\ref{fig:basic_attack_psnrratio} compares the performance of the models in terms of VI.
We also employ random uniform noise within $[-\epsilon, \epsilon]$ as a baseline attack method.
The VI values for the random noise are near 1, meaning that the input and output images deteriorate similarly, and it is barely effective as an attack.
By contrast, FDA and I-FGSM show significantly larger VI values than 1 except for DnCNN, proving that the adversarial attack approaches work well.
In addition, FDA shows the VI value close to 1 for DeepDeblur, while I-FGSM shows a much larger VI value.
These can also be observed in \figurename~\ref{fig:basic_attack_examples}: Both FDA and I-FGSM fail to conceal changes in the input images of DnCNN, and FDA fails to deteriorate the output images of DeepDeblur, while I-FGSM results in significant degradation.
It proves that VI is a reasonable measure to evaluate vulnerability of the image-to-image models.
In the following, we examine vulnerability patterns in more detail.

First, diverse degradation patterns are observed depending on the \emph{task}.
For instance, in \figurename~\ref{fig:basic_attack_examples}, the attacks add hardly perceivable perturbation to the input images for all tasks.
However, significant quality deterioration can be found in the output images with various deterioration textures except denoising (as shown in \figurename~\ref{fig:basic_attack_psnrratio}).
The failure to attack the denoising models is because they are trained to identify and remove the elements unrelated to the original visual content, which include the adversarial perturbations.
On the contrary, the models for the other tasks try to find latent information from a given image (e.g., color for colorization and high-frequency details for deblurring and super-resolution) and recover the corrupted content, during which the perturbations are amplified.
Besides, because the visual components that each model tries to recover differ, different patterns of destruction are observed in the output images of different models.

\begin{figure}[t]
	\begin{center}
		\centering
		\renewcommand{\arraystretch}{1.0}
		\renewcommand{\tabcolsep}{1.5pt}
		\footnotesize
		\begin{tabular}{ccccc}
			Input & Output & ~ & Input & Output \\
			\includegraphics[width=0.22\linewidth]{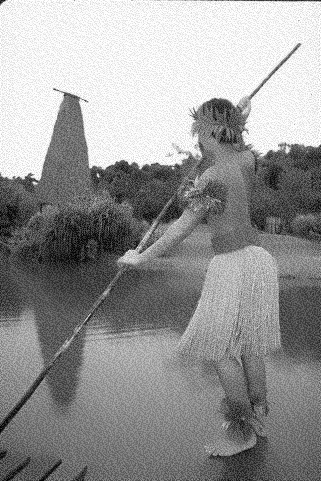} &
			\includegraphics[width=0.22\linewidth]{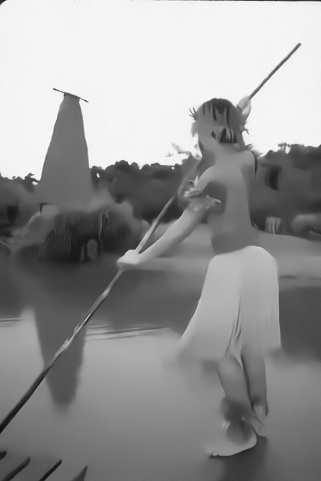} & &
			\includegraphics[width=0.22\linewidth]{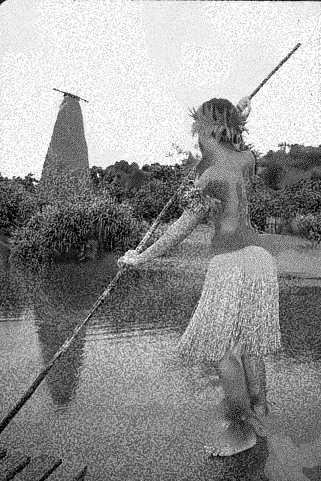} &
			\includegraphics[width=0.22\linewidth]{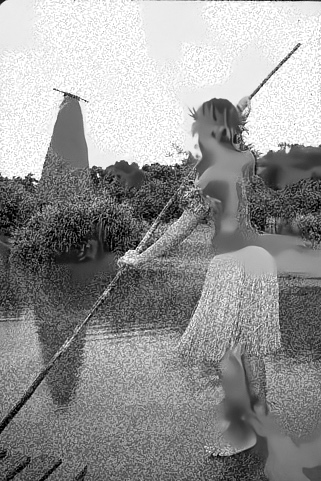}\\
			\multicolumn{2}{c}{\footnotesize{(a) FDA}} & & \multicolumn{2}{c}{\footnotesize{(b) I-FGSM}}\\
		\end{tabular}
	\end{center}
	\caption{Images obtained from DnCNN (multiple $\sigma$).}
	\label{fig:basic_attack_approaches}
\end{figure}

\begin{figure}[t]
	\begin{center}
		\centering
		\renewcommand{\arraystretch}{1.0}
		\renewcommand{\tabcolsep}{1.0pt}
		\footnotesize
		\begin{minipage}[b]{0.42\linewidth}
			\centering
			\centerline{\includegraphics[width=1.0\linewidth]{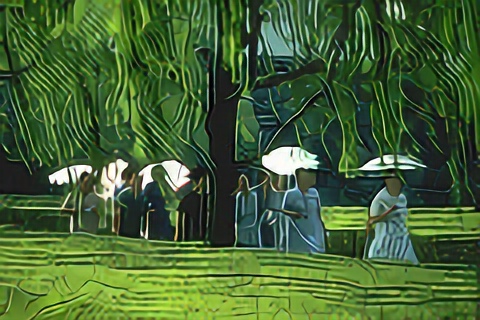}}
			\centerline{(a) SRResNet}
		\end{minipage}
		~~
		\begin{minipage}[b]{0.42\linewidth}
			\centering
			\centerline{\includegraphics[width=1.0\linewidth]{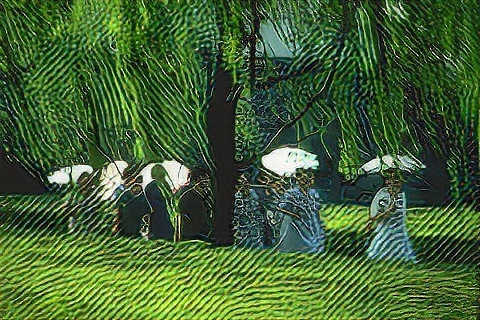}}
			\centerline{(b) SRGAN}
		\end{minipage}
		\caption{Images obtained from SRResNet and SRGAN under I-FGSM.}
		\label{fig:basic_attack_gans}
	\end{center}
\end{figure}

Second, degradation patterns in the output images reveal the characteristics of the \emph{attack methods}.
For example, in \figurename~\ref{fig:basic_attack_examples}, the output images under I-FGSM look more corrupted than those under FDA overall.
This is also shown as larger VI values for I-FGSM in \figurename~\ref{fig:basic_attack_psnrratio}.
\figurename~\ref{fig:basic_attack_approaches} shows another examples in denoising.
FDA largely removes the sharp textures in the output image, which is mainly due to the reduction of the activations.
By contrast, I-FGSM tries to deteriorate the quality of the output image as much as possible, resulting in noisy patterns.
These kinds of characteristics cannot be found in image classification, where the attack results are usually represented only as success rates regardless of attack methods.

\begin{figure}[t]
	\begin{center}
		\centering
		\renewcommand{\arraystretch}{1.0}
		\renewcommand{\tabcolsep}{0.25pt}
		\scriptsize
		\begin{tabular}{ccccccccc}
			\makecell[c]{Input \\ (original)} & \makecell[c]{Output \\ (original)} & \makecell[c]{Input \\ (FDA)} & \makecell[c]{Output \\ (FDA)} & & \makecell[c]{Input \\ (original)} & \makecell[c]{Output \\ (original)} & \makecell[c]{Input \\ (FDA)} & \makecell[c]{Output \\ (FDA)} \\
			\includegraphics[width=0.120\linewidth]{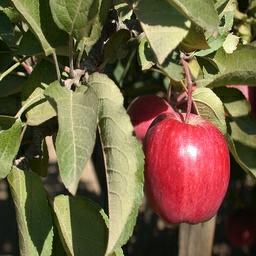} &
			\includegraphics[width=0.120\linewidth]{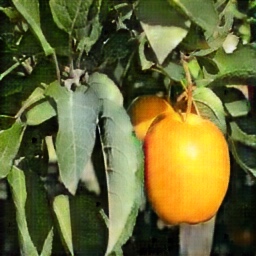} &
			\includegraphics[width=0.120\linewidth]{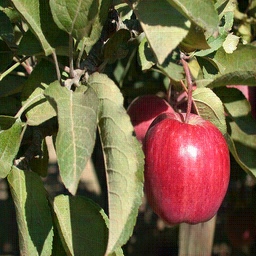} &
			\includegraphics[width=0.120\linewidth]{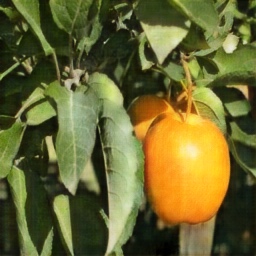} &
			&
			\includegraphics[width=0.120\linewidth]{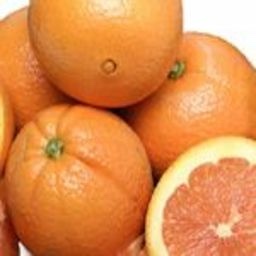} &
			\includegraphics[width=0.120\linewidth]{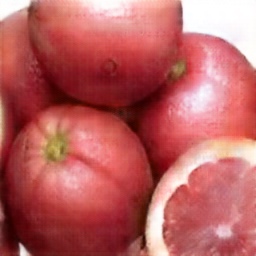} &
			\includegraphics[width=0.120\linewidth]{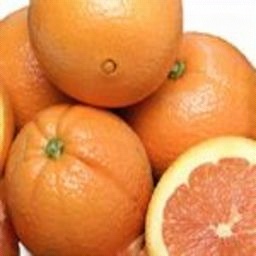} &
			\includegraphics[width=0.120\linewidth]{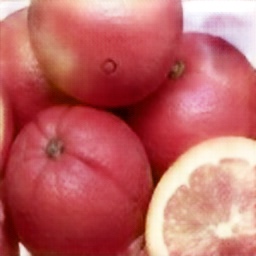}
			\\
			\multicolumn{4}{c}{\footnotesize{(a) Apple $\rightarrow$ Orange}}
			& &
			\multicolumn{4}{c}{\footnotesize{(b) Orange $\rightarrow$ Apple}}
		\end{tabular}
	\end{center}
	\caption{Images obtained from CycleGAN trained using different datasets.}
	\label{fig:basic_attack_datasets}
\end{figure}

Third, some image-to-image models employ an exquisite \emph{training mechanism} that is not so much used in image classification models, which is generative adversarial networks (GANs) \cite{goodfellow2014generative} to improve perceptual quality of the output images.
We observe that these models output more deteriorated images than the models without GANs when attacked.
In \figurename~\ref{fig:basic_attack_psnrratio}, DeepDeblur and CycleGAN, which employ GANs, show relatively larger VI values.
This is also confirmed when SRResNet and SRGAN are compared in \figurename~\ref{fig:basic_attack_gans}.
These two models share the same model structure, but SRGAN is trained with a GAN-based loss function for better perceptual quality.
The models trained with GANs usually produce perceptually appealing sharp textures \cite{blau2018perception}.
Thus, the perturbation in the input image also tends to be intensified in the output image.

Fourth, different characteristics in terms of vulnerability and distortion patterns can be observed depending on the \emph{training dataset}.
In \figurename~\ref{fig:basic_attack_psnrratio}, the model converting apples to oranges shows the least vulnerability, while the model converting horses to zebras is the most vulnerable among the translation models.
Note that these models share the same model structure and training mechanism.
\figurename~\ref{fig:basic_attack_datasets} depicts example images when FDA is employed.
For converting apples to oranges, the model simply finds red regions and converts their colors to yellowish ones, meaning that it is primarily sensitive to color information in the input image.
By contrast, due to grainy textures in the oranges, the model converting oranges to apples relies on complex analysis of the input image.
Thus, there is more room for the attack to find effective perturbation for these models.
This demonstrates that the vulnerability of the image-to-image models can differ depending on the employed dataset.

\subsection{Universal Perturbations and Transferability}
\label{subsec:results_transferability}

\begin{figure}[t]
	\begin{center}
		\centering
		\renewcommand{\arraystretch}{1.2}
		\renewcommand{\tabcolsep}{0.6pt}
		\scriptsize
		\begin{tabular}{ccccccc}
			\makecell[c]{Colorization} & \makecell[c]{Deblurring} & \makecell[c]{Denoising} & \makecell[c]{Super-\\resolution} & \makecell[c]{Translation} & ~~ & \makecell[c]{Classification}\\
			\includegraphics[width=0.155\linewidth]{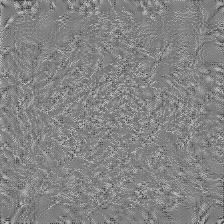} &
			\includegraphics[width=0.155\linewidth]{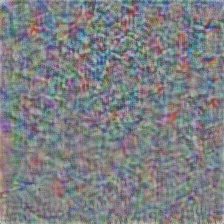} &
			\includegraphics[width=0.155\linewidth]{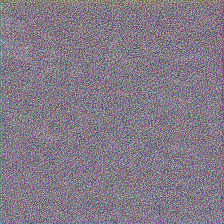} &
			\includegraphics[width=0.155\linewidth]{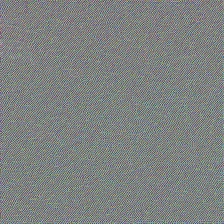} &
			\includegraphics[width=0.155\linewidth]{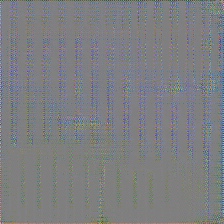} & &
			\includegraphics[width=0.155\linewidth]{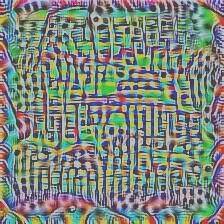}
			\\
			\multicolumn{7}{c}{\footnotesize{(a) FDA}}\smallskip\\
			\includegraphics[width=0.155\linewidth]{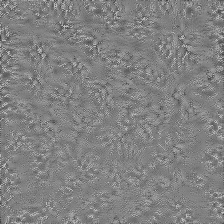} &
			\includegraphics[width=0.155\linewidth]{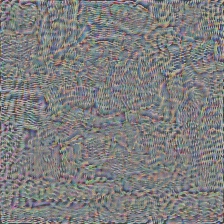} &
			\includegraphics[width=0.155\linewidth]{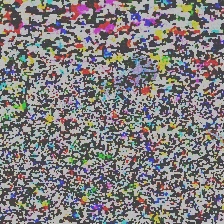} &
			\includegraphics[width=0.155\linewidth]{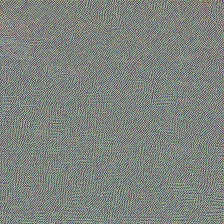} &
			\includegraphics[width=0.155\linewidth]{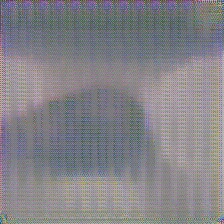} & &
			\includegraphics[width=0.155\linewidth]{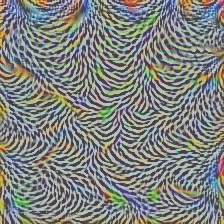}
			\\
			\multicolumn{7}{c}{\footnotesize{(b) I-FGSM}}
		\end{tabular}
	\end{center}
	\caption{Universal perturbations obtained from CIC, DeepDeblur, DnCNN (color, multiple $\sigma$), CARN, CycleGAN (Van Gogh$\rightarrow$photo), and VGG16. Each perturbation is magnified $\times$4 for better visualization.}
	\label{fig:universal_attack_images}
\end{figure}

\begin{figure}[t]
	\begin{center}
		\centering
		\renewcommand{\arraystretch}{1.5}
		\renewcommand{\tabcolsep}{3.0pt}
		\footnotesize
		\begin{tabular}{ccc}
			\raisebox{-0.5\height}{\includegraphics[width=0.265\linewidth]{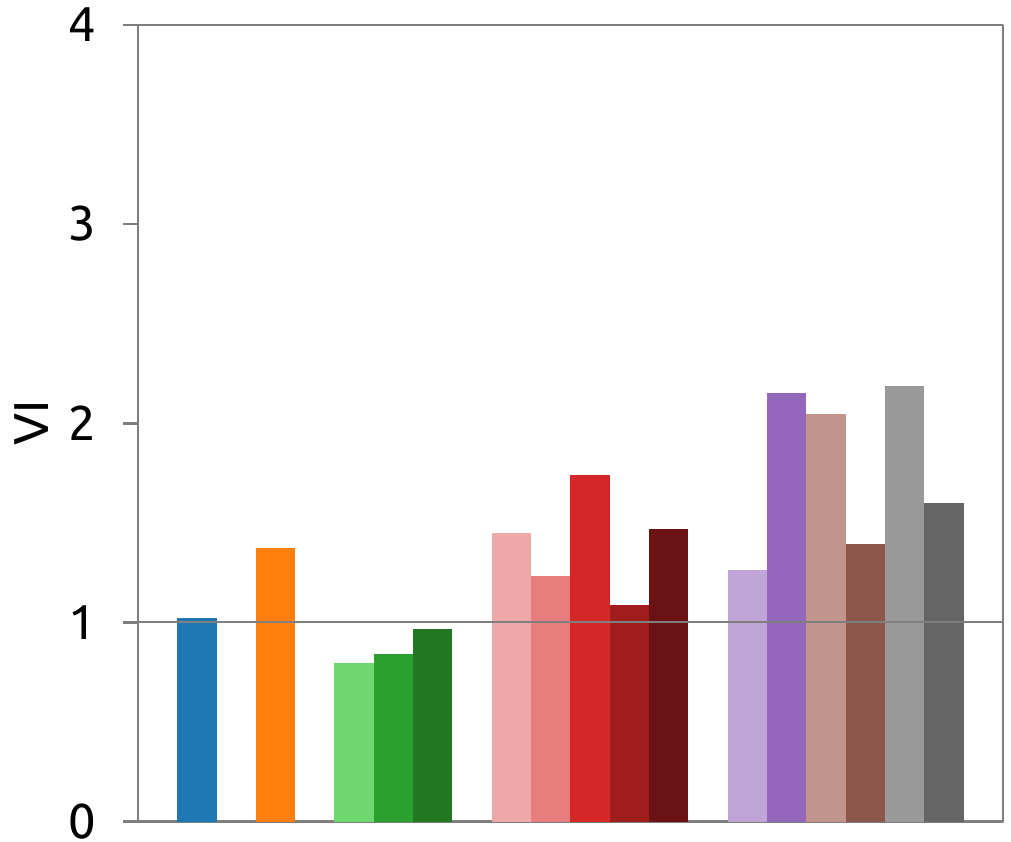}} &
			\raisebox{-0.5\height}{\includegraphics[width=0.265\linewidth]{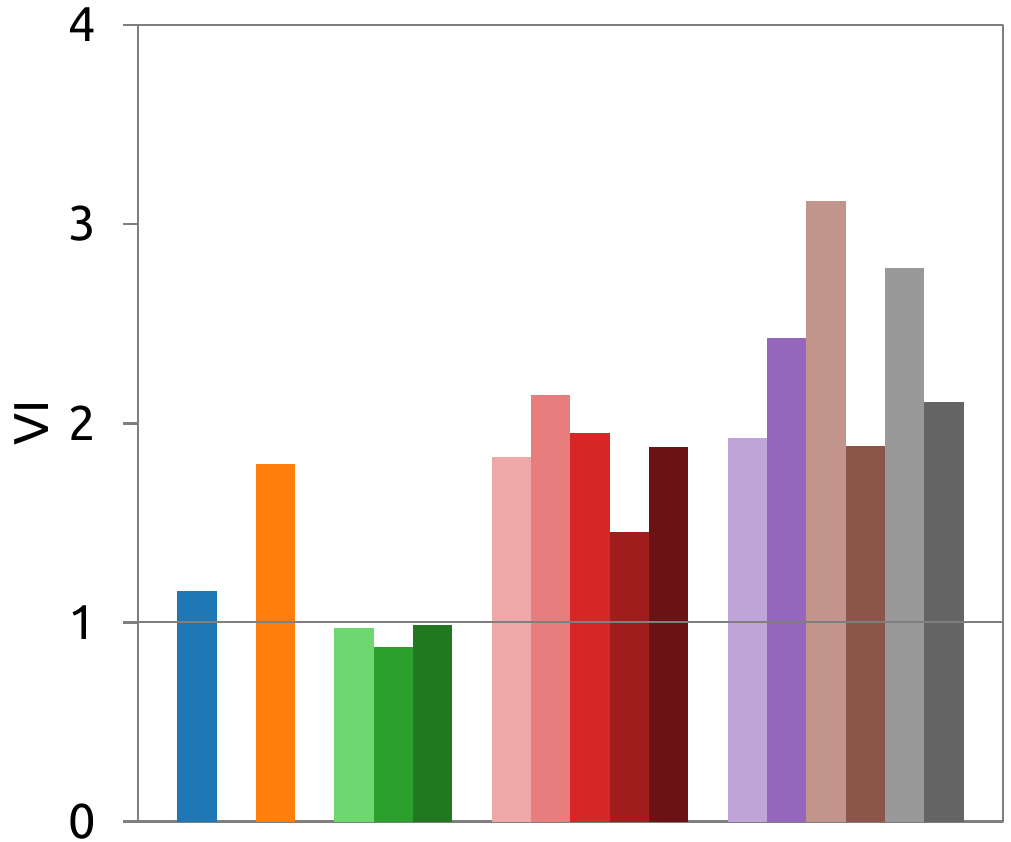}} &
			\multirow{2}{*}[17.0pt]{\raisebox{-0.5\height}{\includegraphics[width=0.146\linewidth]{figures/basic_vi_legend}}}\\
			(a) FDA & (b) I-FGSM &
		\end{tabular}
	\end{center}
	\caption{Performance in terms of VI for universal perturbation.}
	\label{fig:universal_attack}
\end{figure}

\figurename~\ref{fig:universal_attack_images} shows the universal perturbation for each attack method and each task.
First of all, the perturbation patterns differ depending on the attack method because of the differences in the attack objectives, i.e., while FDA tries to find perturbations that can reduce the intermediate activations, I-FGSM tries to maximize the amount of output deterioration directly.
In addition, the overall patterns of the perturbation differ depending on the task.
For example, grouped colorful pixels can be found in the perturbation of the denoising model for I-FGSM because Gaussian noise-like perturbations, obtained in the other tasks, tend to be removed in the output images due to the denoising process.

It is interesting to compare the perturbations obtained from the image-to-image models with those obtained from the image classification models.
The former contains more granular textures, while the latter has thicker and bolder textures.
One possible reason is due to the different architectures: The image classification models employ spatial aggregation operations (e.g., pooling) to reduce the number of intermediate features, while the image-to-image models usually do not employ such operations to minimize information loss that can directly damage quality of output images.

\figurename~\ref{fig:universal_attack} compares VI for the universal attack.
The results are similar to those obtained for image-specific attacks in \figurename~\ref{fig:basic_attack_psnrratio}, showing that the image-agnostic universal attack is successful although it is slightly less effective than the image-specific attack.
Moreover, I-FGSM generates universal perturbations stronger than those of FDA because of the different objectives as in the image-specific attack.

\begin{figure}[t]
	\begin{center}
		\centering
		\footnotesize
		\begin{minipage}[b]{0.95\linewidth}
			\centering
			\centerline{\includegraphics[width=1.0\linewidth]{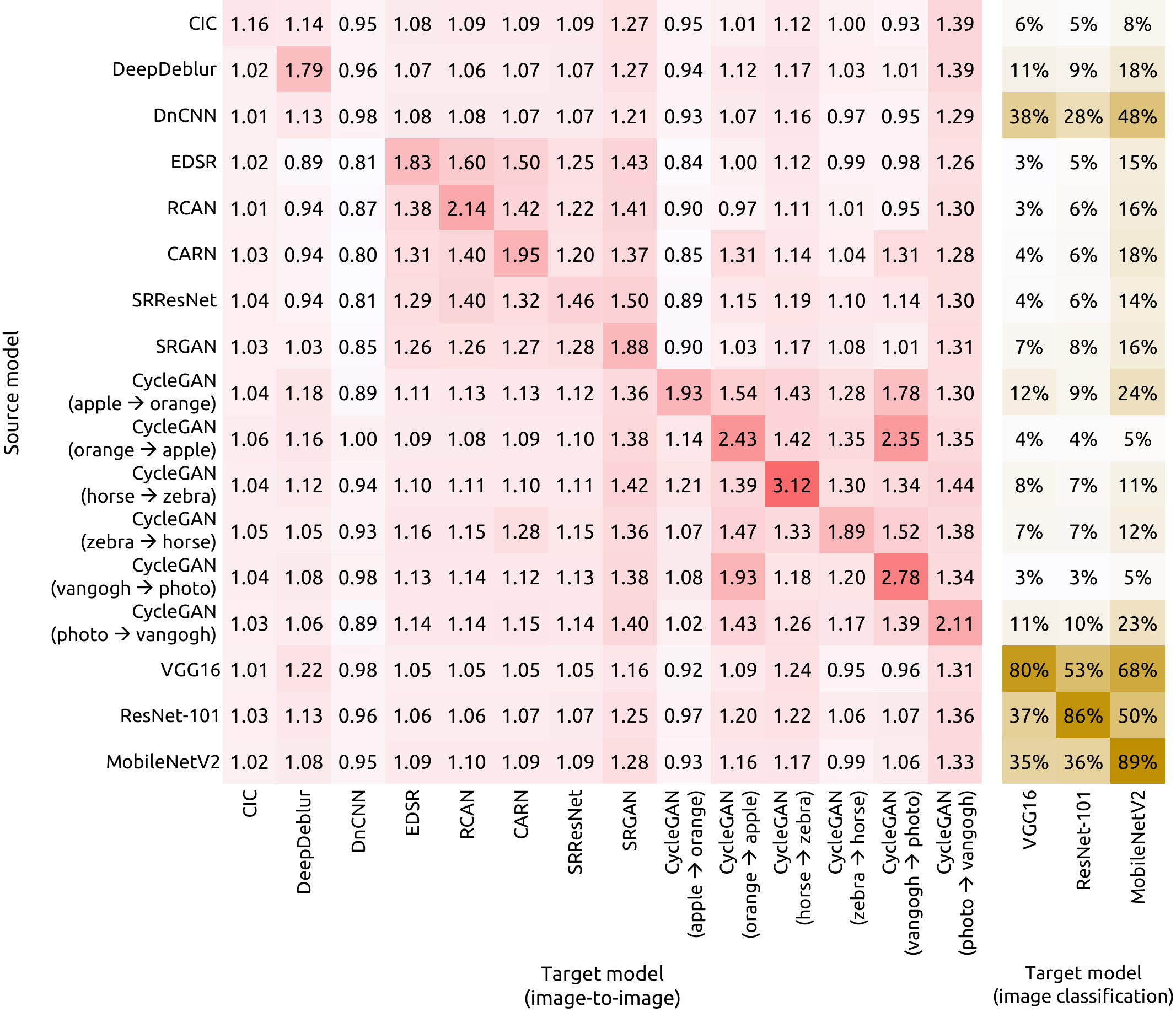}}
		\end{minipage}
	\end{center}
	\caption{Transferability for I-FGSM. The VI values are reported when an image-to-image model is used as a target model and the fooling rates are reported when an image classification model is used as a target model.}
	\label{fig:universal_attack_transfer_heatmap}
\end{figure}

We also investigate the transferability of universal perturbations.
Specifically, we measure VI or a fooling rate of a \textit{target} model for the universal perturbation found from another \textit{source} model.
\figurename~\ref{fig:universal_attack_transfer_heatmap} summarizes the results for I-FGSM.
The universal perturbations are transferable between different models to some extent even across different tasks, although they are relatively more transferable when the source and target models are for the same task.
For example, SRGAN for super-resolution is easily compromised by the universal perturbations found for the other tasks; especially, the VI values when the universal perturbations found for translation are used are as high as those when the universal perturbations found for the other super-resolution models are used.
A similar observation can be made when CycleGAN (photo$\rightarrow$Van Gogh) is set as a target model.
It is thought that these two models are particularly vulnerable to universal perturbations because they try to recover complex texture using GANs.

In most cases, the transferability of universal perturbations between an image-to-image model and an image classification model is lower than that between image-to-image models or image classification models, which can be easily understood.
Surprisingly, however, the universal perturbations are sometimes transferable between an image-to-image model and a classification model.
For instance, the perturbation found in DnCNN can fool MobileNetV2 with a fooling rate of almost 50\%.
In \figurename~\ref{fig:universal_attack_images}b, the perturbation of DnCNN looks the most similar to that of the classification model in terms of texture patterns and variety of colors.
However, the opposite is not applied; as the original attack on DnCNN does not work well, the perturbations of the classification models do not successfully attack DnCNN, either.

\subsection{Characteristics of Adversarial Examples}
\label{subsec:perturbation_characteristics}

\begin{figure}[t]
	\begin{center}
		\centering
		\renewcommand{\arraystretch}{1.5}
		\renewcommand{\tabcolsep}{3.0pt}
		\footnotesize
		\begin{tabular}{ccc}
			\raisebox{-0.5\height}{\includegraphics[width=0.265\linewidth]{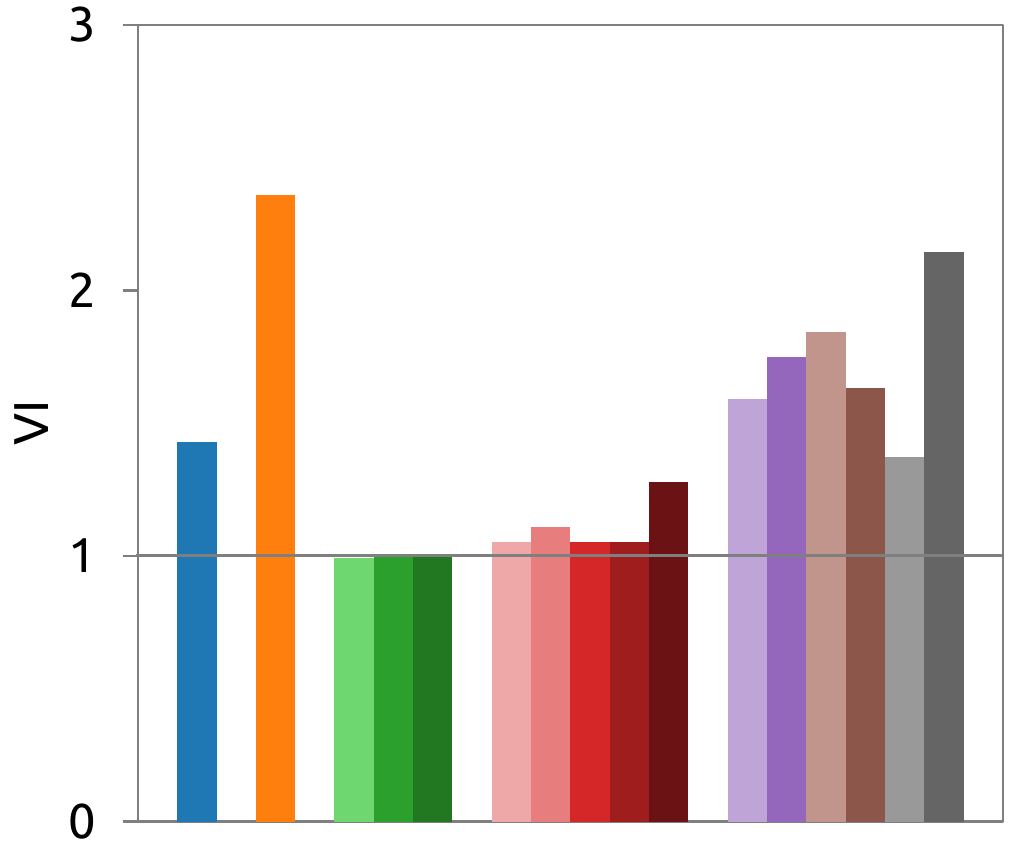}} &
			\raisebox{-0.5\height}{\includegraphics[width=0.265\linewidth]{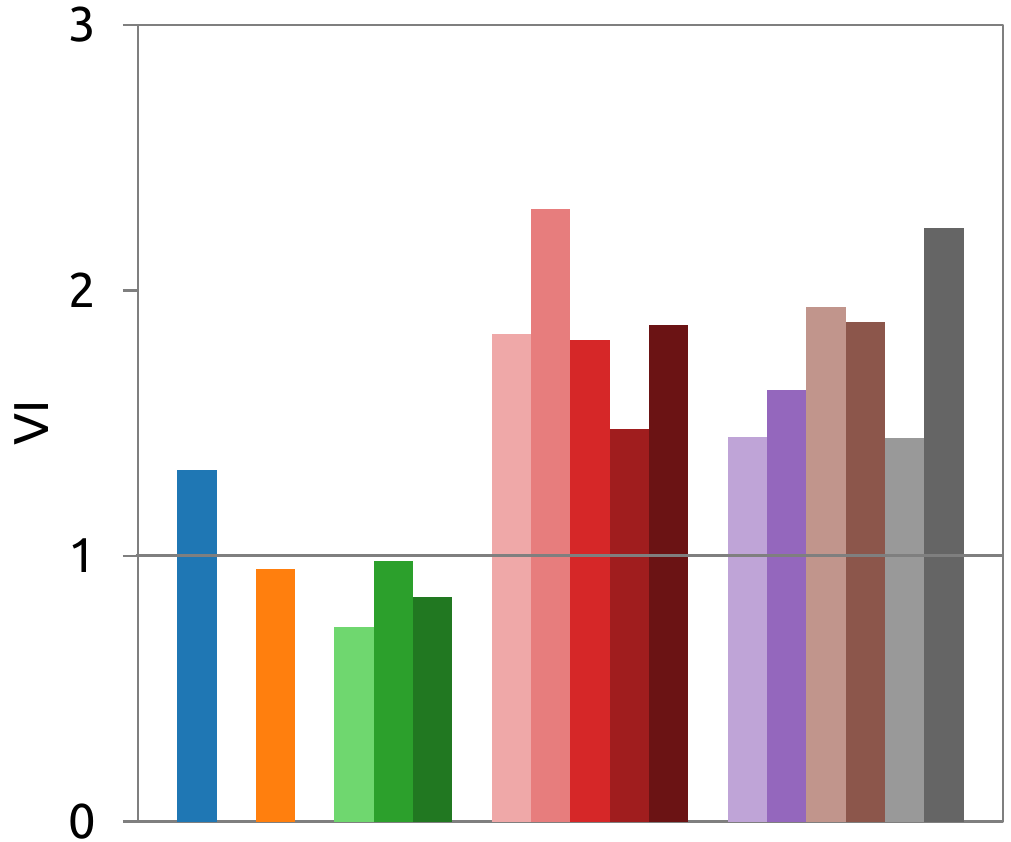}} &
			\multirow{2}{*}[17pt]{\raisebox{-0.5\height}{\includegraphics[width=0.146\linewidth]{figures/basic_vi_legend}}}\\
			(a) Low-frequency attack & (b) High-frequency attack &
		\end{tabular}
	\end{center}
	\caption{Performance of the frequency-aware attacks in terms of VI for I-FGSM.}
	\label{fig:frequency_attack}
\end{figure}

\begin{figure}[t]
	\begin{center}
		\centering
		\renewcommand{\arraystretch}{1.0}
		\renewcommand{\tabcolsep}{0pt}
		\footnotesize
		\begin{tabular}{ccccc}
			\makecell[c]{Input\\(low-frequency)} & \makecell[c]{Output\\(low-frequency)} & ~~~ & \makecell[c]{Input\\(high-frequency)} & \makecell[c]{Output\\(high-frequency)} \smallskip \\
			~\includegraphics[width=0.215\linewidth]{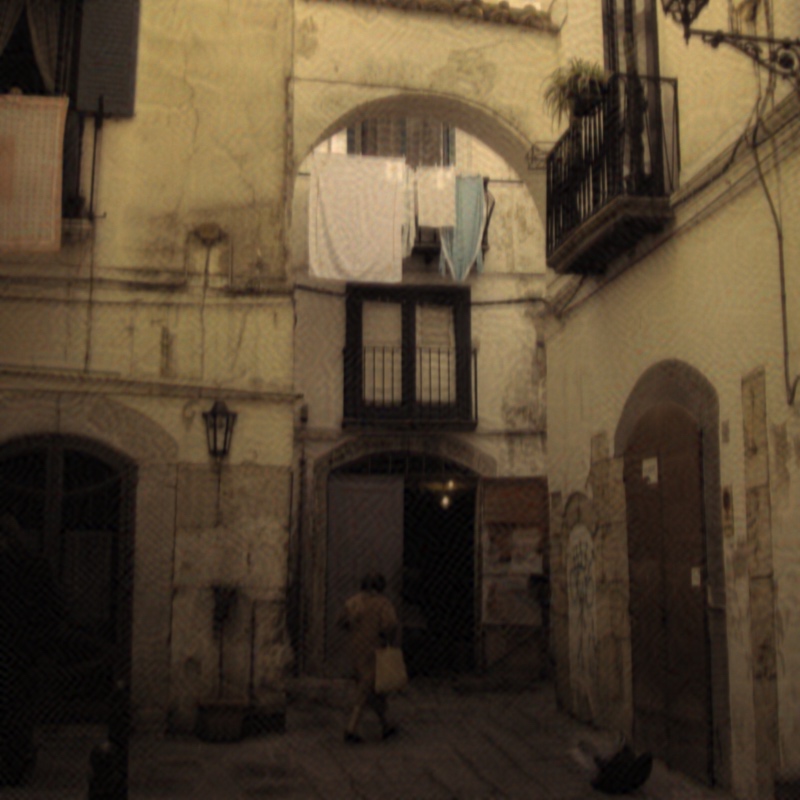}~ &
			~\includegraphics[width=0.215\linewidth]{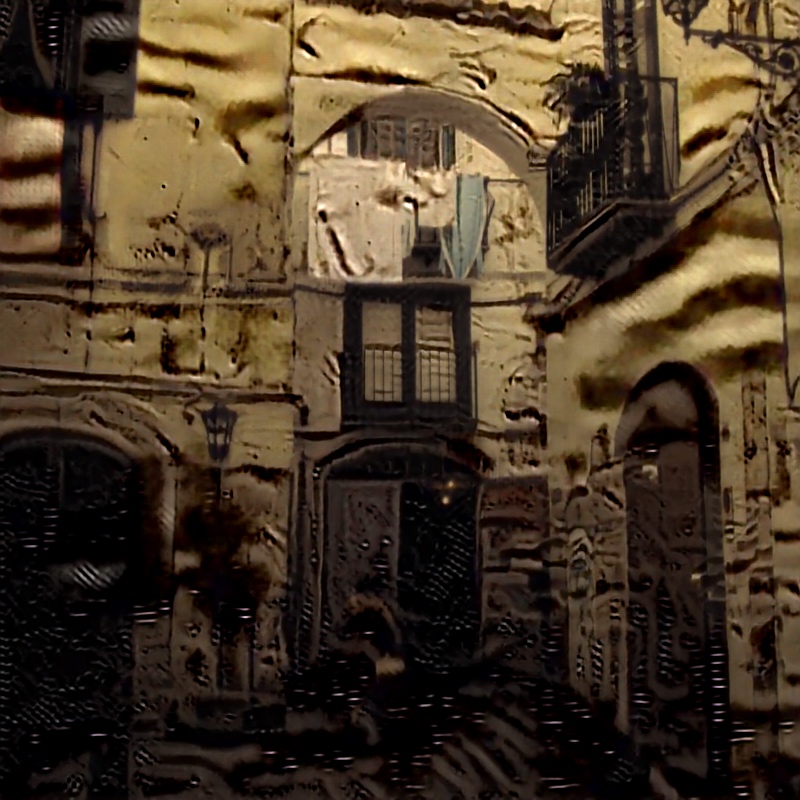}~ & &
			~\includegraphics[width=0.215\linewidth]{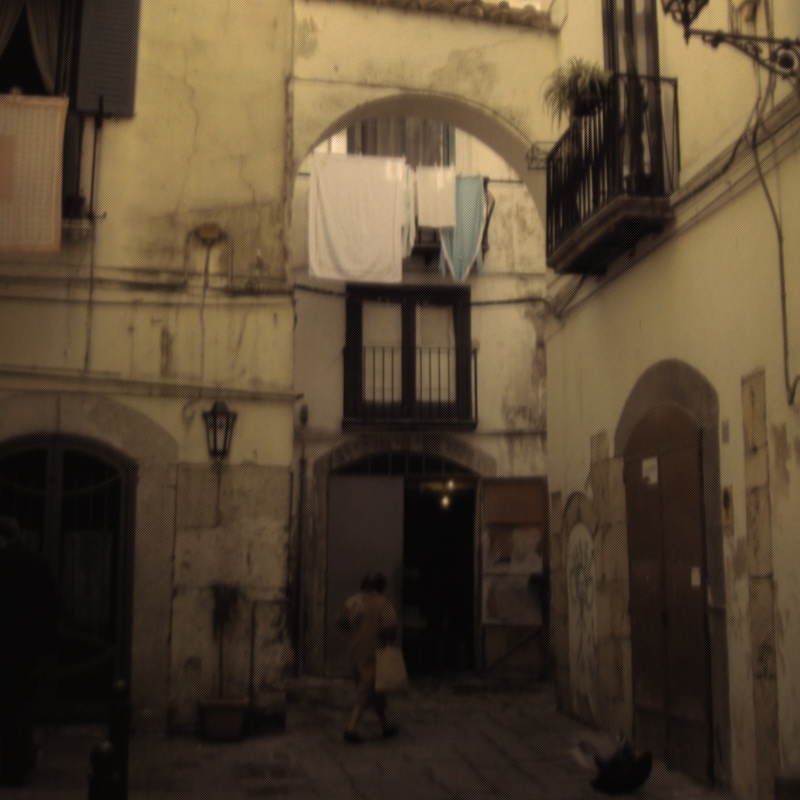}~ &
			~\includegraphics[width=0.215\linewidth]{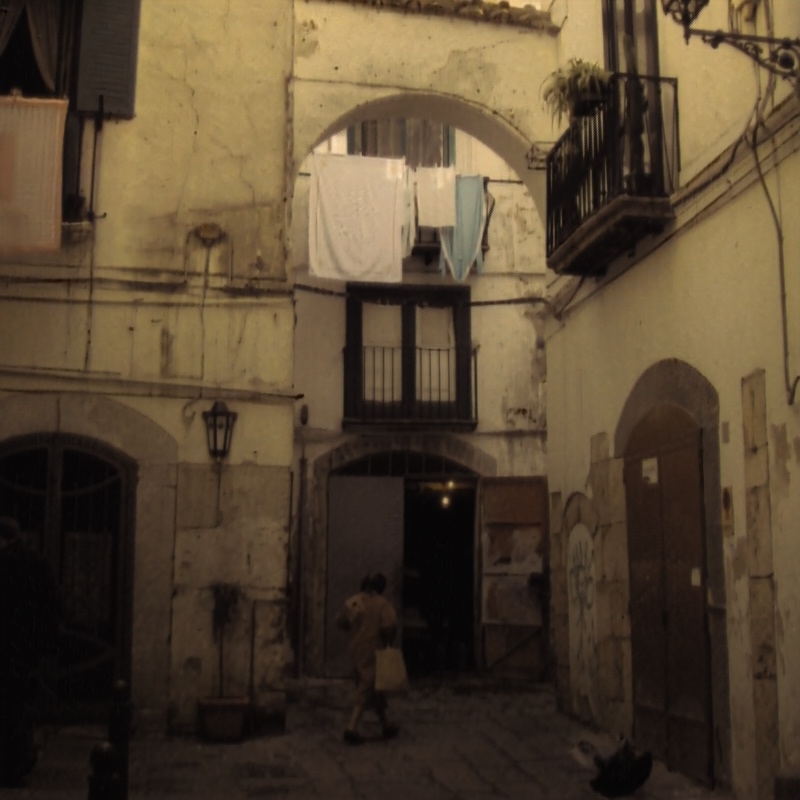}~\\
			\multicolumn{5}{c}{\footnotesize{(a) DeepDeblur}} \smallskip \\
			~\includegraphics[width=0.215\linewidth]{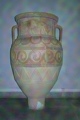}~ &
			~\includegraphics[width=0.215\linewidth]{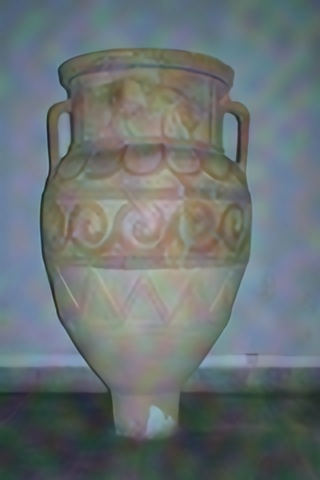}~ & &
			~\includegraphics[width=0.215\linewidth]{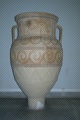}~ &
			~\includegraphics[width=0.215\linewidth]{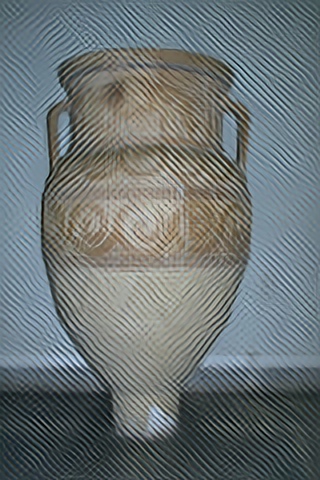}~\\
			\multicolumn{5}{c}{\footnotesize{(b) EDSR}}\\
		\end{tabular}
	\end{center}
	\caption{Images obtained from (a) DeepDeblur and (b) EDSR.}
	\label{fig:frequency_attack_images}
\end{figure}

We further investigate characteristics of perturbations by employing the frequency-aware attacks with $r=1/4$.
\figurename~\ref{fig:frequency_attack} shows the VI values for the low-frequency and high-frequency attacks.
The results explain that the effectiveness of the frequency-aware attacks varies depending on the task.
In particular, the deblurring model (DeepDeblur) is more vulnerable to the low-frequency attack than the high-frequency attack, while the super-resolution models are more vulnerable to the high-frequency attack.

\figurename~\ref{fig:frequency_attack_images} shows example images obtained from these models.
In DeepDeblur, only the low-frequency attack significantly deteriorates the quality of the output image because the model exploits information in the low-frequency components (i.e., blurred regions) to recover missing details.
By contrast, the output image of EDSR is affected by both the low-frequency and high-frequency attacks.
However, the low-frequency attack also significantly deteriorates the quality of the input image.
A super-resolution model finds latent textures through the high-frequency subspace of an input image; thus, the high-frequency attack works well.
However, the low-frequency features in the input image, which do not have notable textures, appear in the output image without much change.
Hence, the low-frequency attack tries to add a large amount of perturbation to degrade the output, which leads to clearly visible deterioration in both the input and output images.
These show that the frequency-dependent degree of vulnerability is determined by the model's mechanism optimized for the task.

\section{Defending Image-to-image Models}
\label{sec:defenses}

Finally, we evaluate the feasibility of defending image-to-image-models by employing defense methods that have been widely used for image classification models: image transformation and adversarial training.
For transformation-based approaches, we consider JPEG compression \cite{dziugaite2016study}, random resizing \cite{xie2018mitigating}, and bit reduction \cite{xu2018feature}.
In addition, we consider geometric self-ensemble that was introduced in the super-resolution \cite{lim2017enhanced} to improve quality of the output images without additional training.
For adversarial training, adversarial examples are obtained by I-FGSM.

\begin{figure}[t]
	\begin{center}
		\centering
		\renewcommand{\arraystretch}{1.5}
		\renewcommand{\tabcolsep}{0pt}
		\scriptsize
		\begin{tabular}{ccccc}
			\raisebox{-0.5\height}{\includegraphics[width=0.199\linewidth]{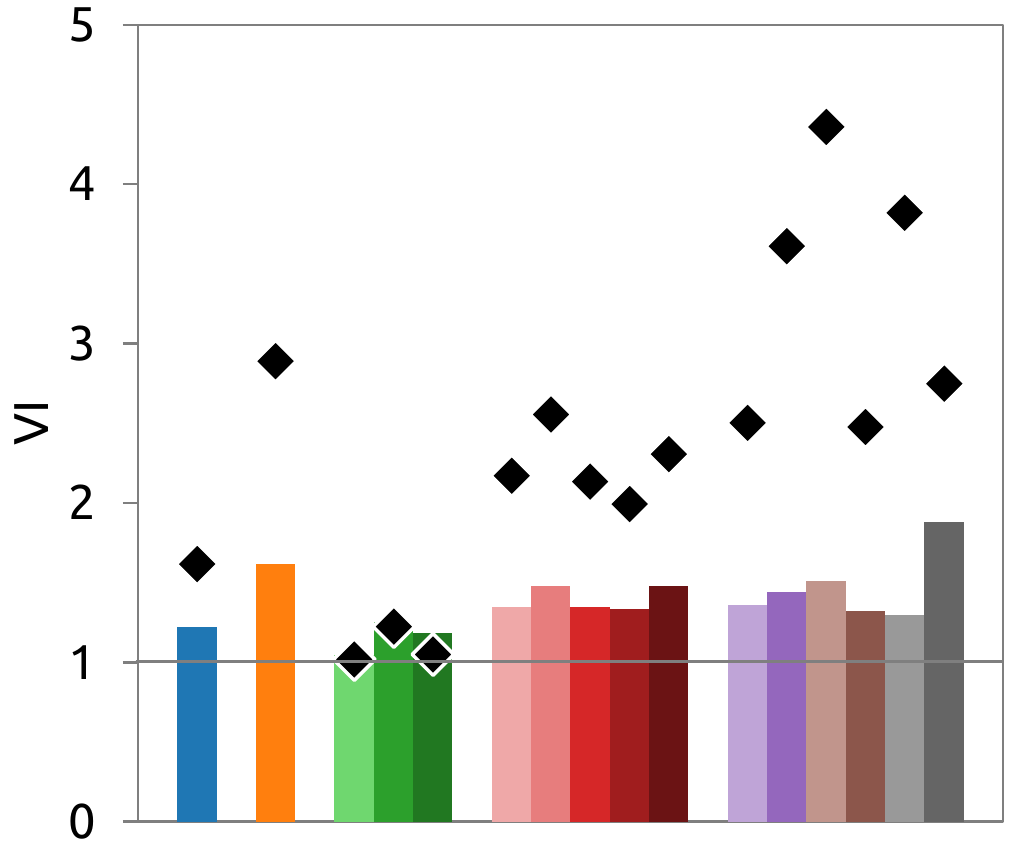}} &
			\raisebox{-0.5\height}{\includegraphics[width=0.199\linewidth]{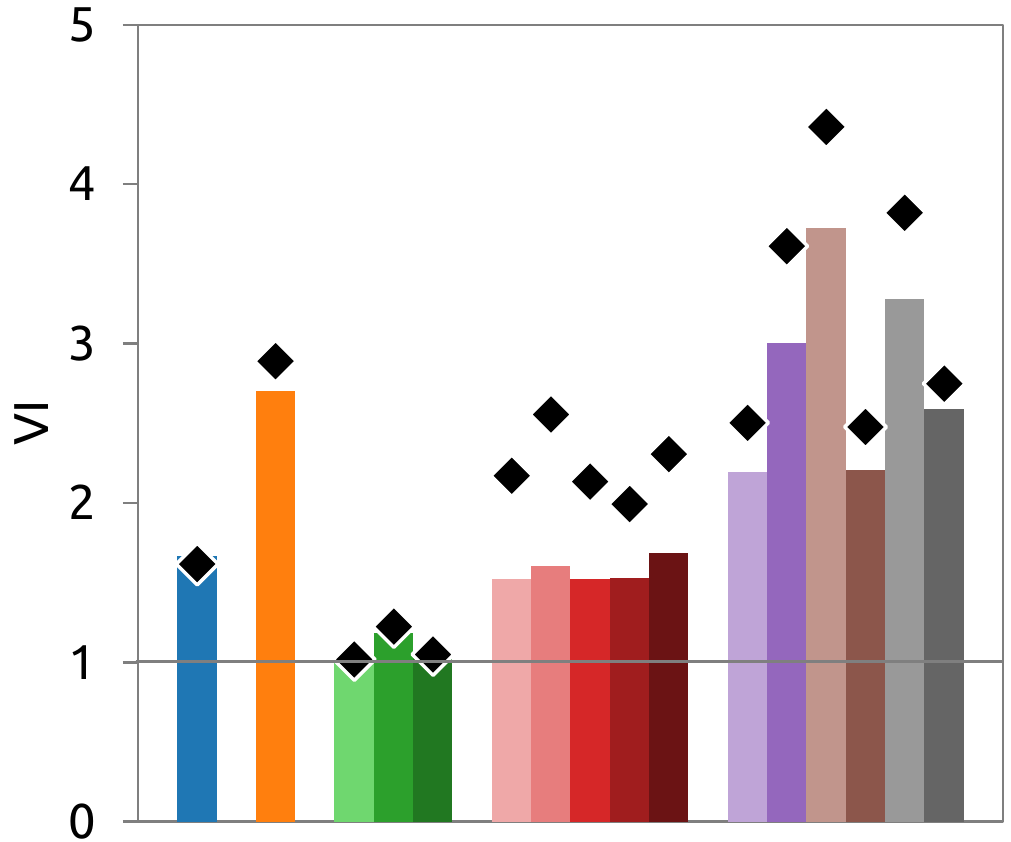}} &
			\raisebox{-0.5\height}{\includegraphics[width=0.199\linewidth]{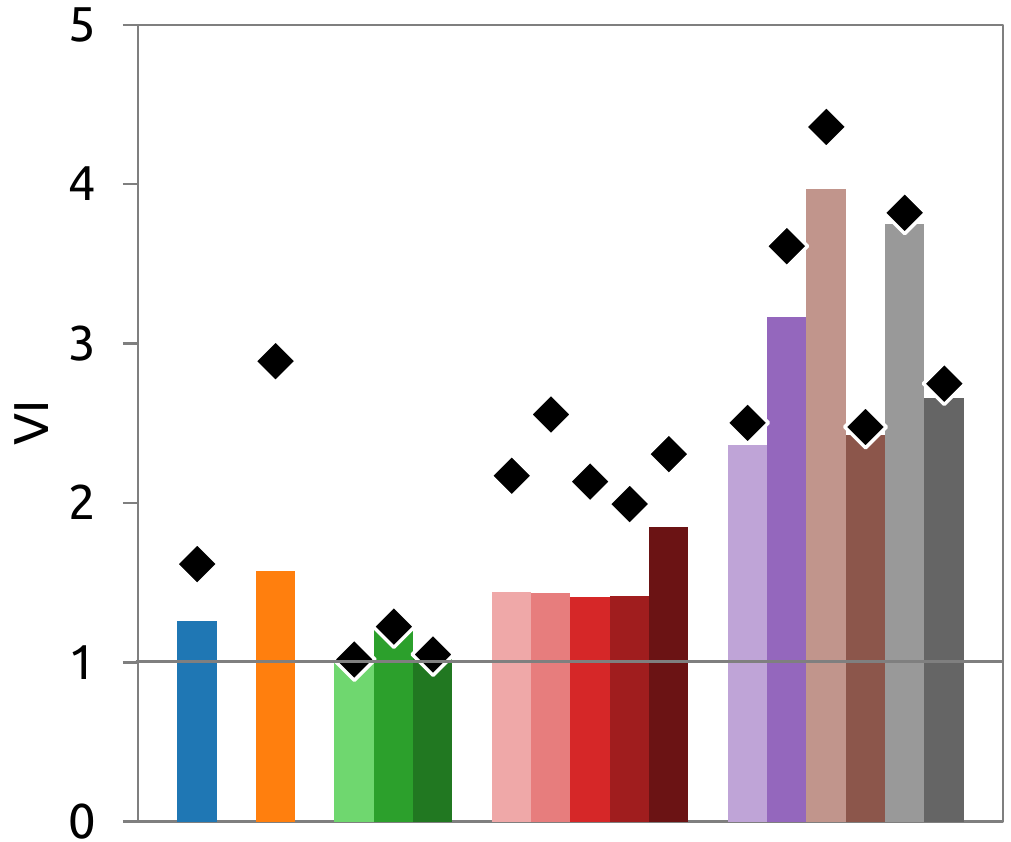}} &
			\raisebox{-0.5\height}{\includegraphics[width=0.199\linewidth]{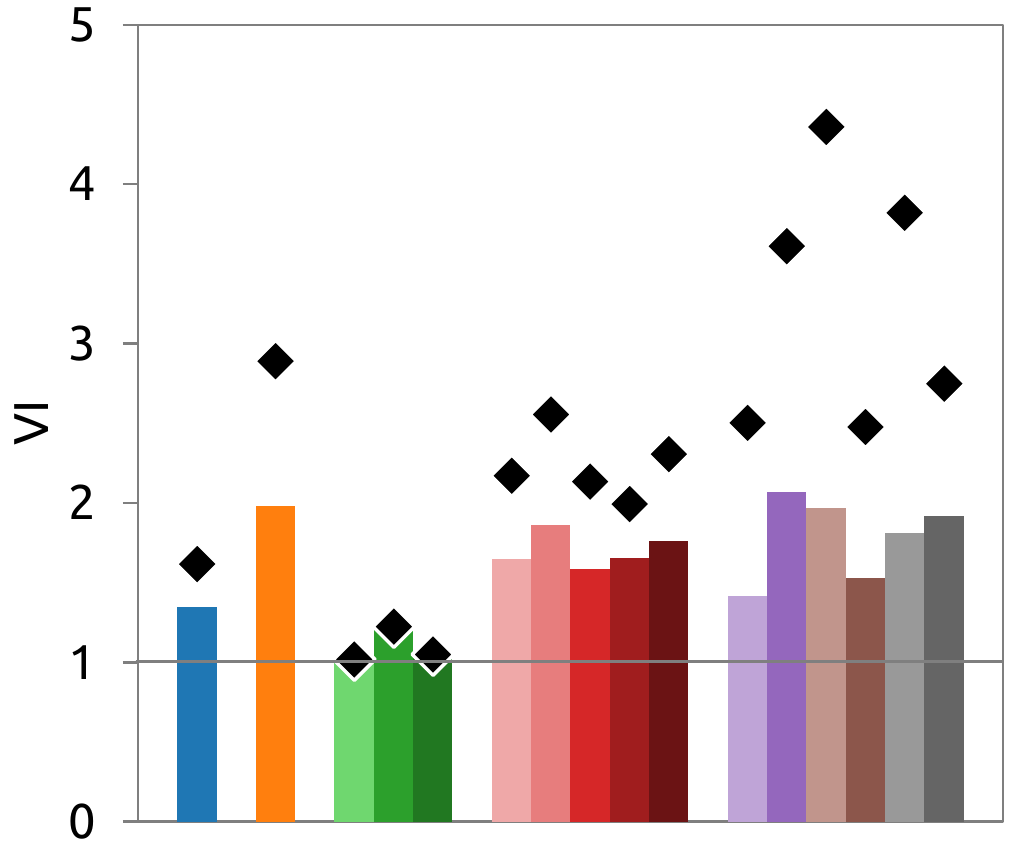}} &
			\raisebox{-0.5\height}{\includegraphics[width=0.199\linewidth]{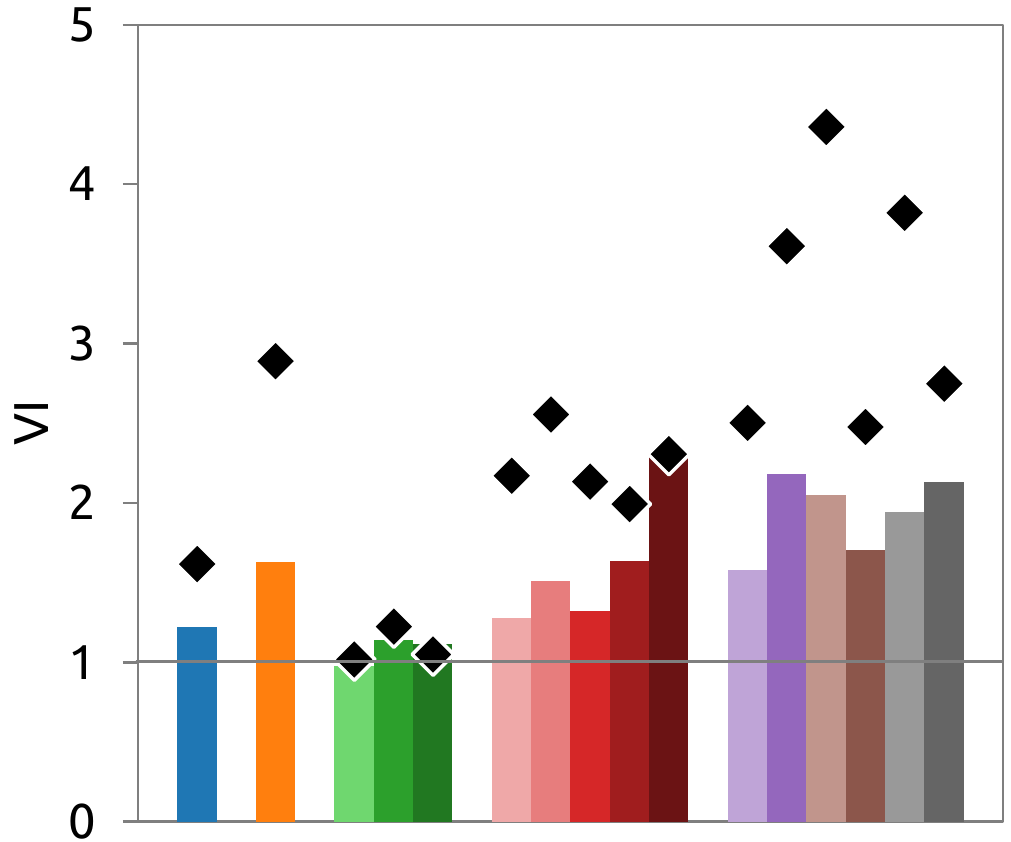}}
			\\
			\makecell[c]{JPEG} & \makecell[c]{Resizing} & \makecell[c]{Bit\\reduction} & \makecell[c]{Self-\\ensemble} & \makecell[c]{Adversarial\\training}
			\smallskip \\
			\multicolumn{5}{c}{\raisebox{-0.5\height}{~\includegraphics[width=0.850\linewidth]{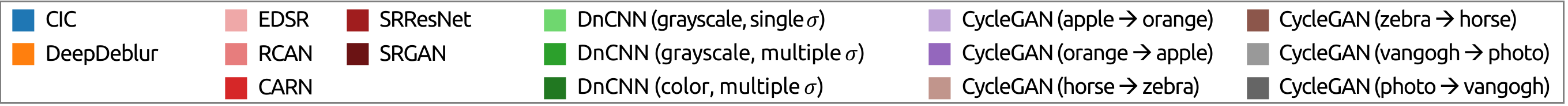}}}
		\end{tabular}
	\end{center}
	\caption{Performance of the defense methods in terms of VI for I-FGSM. Diamond markers refer to the VI values without defense.}
	\label{fig:defenses_vi}
\end{figure}

\figurename~\ref{fig:defenses_vi} compares the performance of the defense methods in terms of VI, when I-FGSM is employed.
Overall, the defense methods reduce influences of adversarial perturbations to some extent, i.e., the VI values are reduced.
The JPEG compression shows the highest effectiveness.
Nevertheless, the VI values are still larger than 1, which implies that deterioration in the output images is not completely removed.

\begin{figure}[t]
	\begin{center}
		\centering
		\renewcommand{\arraystretch}{1.5}
		\renewcommand{\tabcolsep}{0pt}
		\scriptsize
		\begin{tabular}{ccccc}
			\raisebox{-0.5\height}{\includegraphics[width=0.199\linewidth]{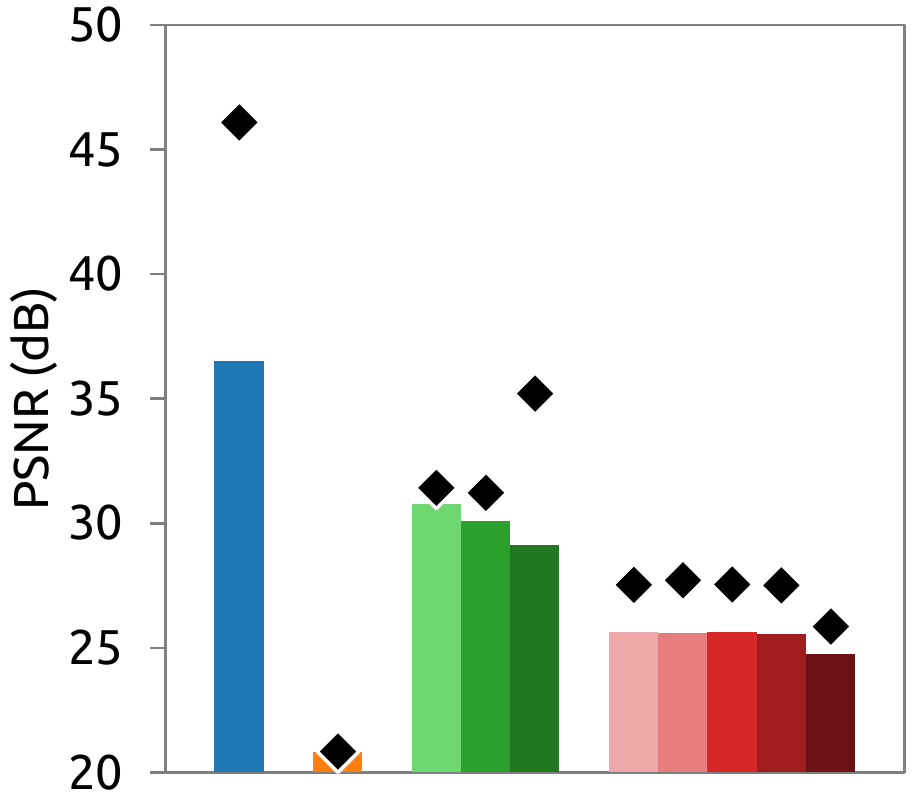}} &
			\raisebox{-0.5\height}{\includegraphics[width=0.199\linewidth]{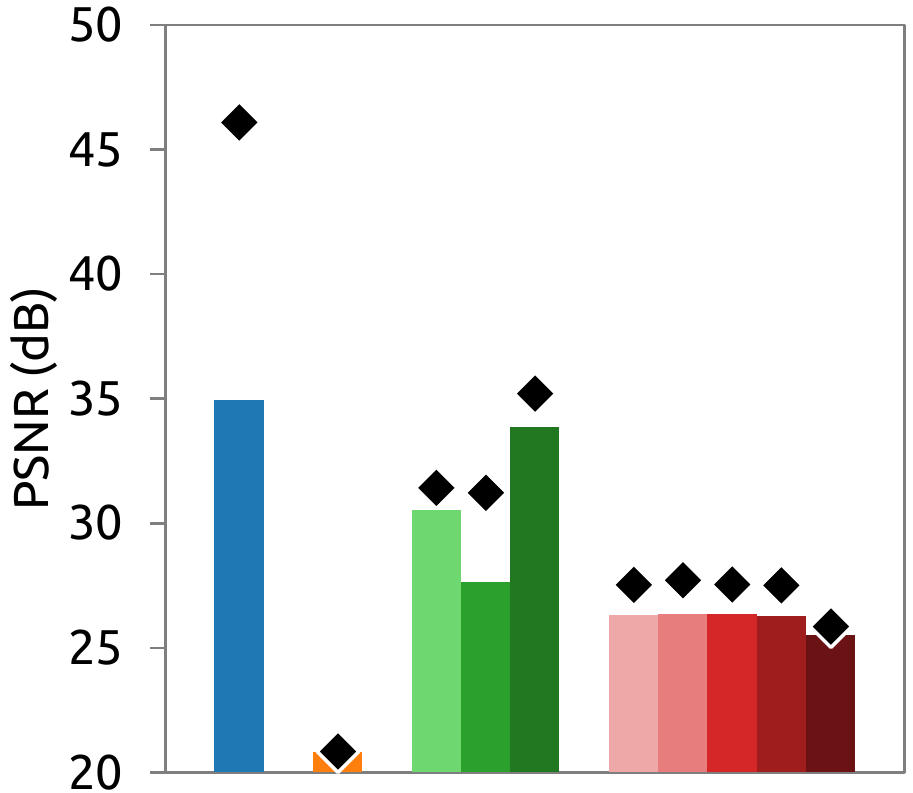}} &
			\raisebox{-0.5\height}{\includegraphics[width=0.199\linewidth]{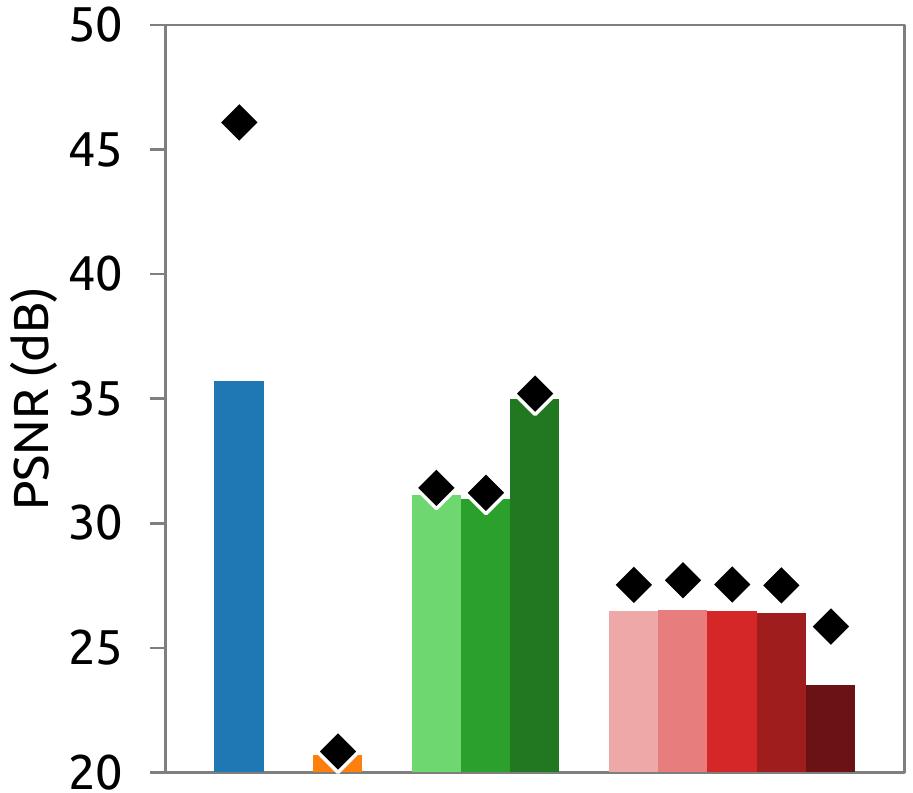}} &
			\raisebox{-0.5\height}{\includegraphics[width=0.199\linewidth]{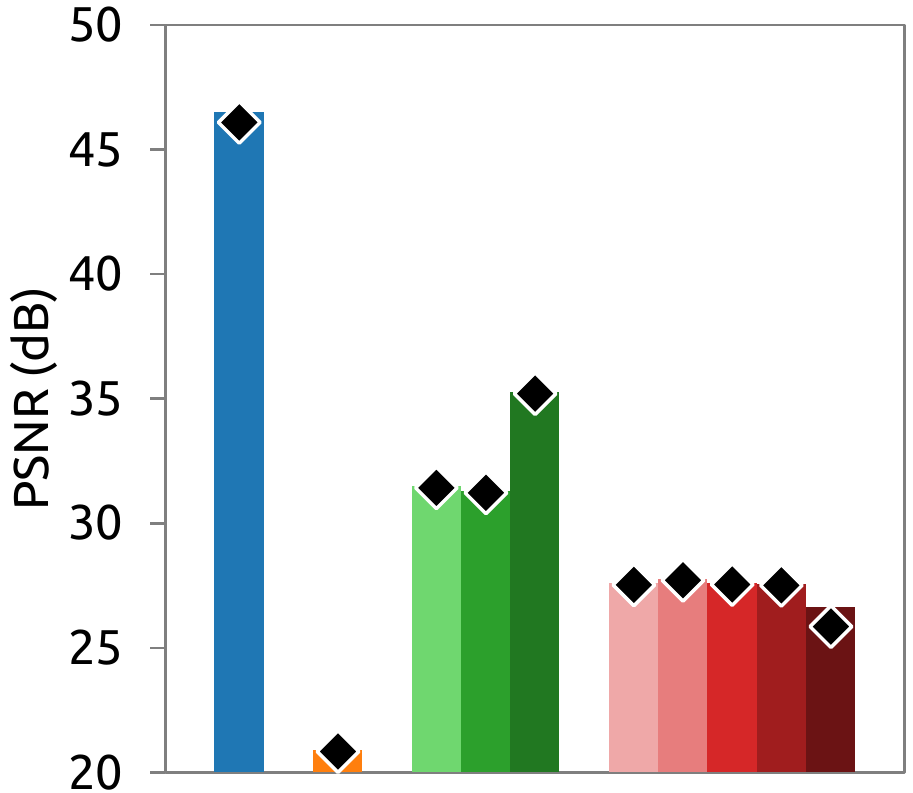}} &
			\raisebox{-0.5\height}{\includegraphics[width=0.199\linewidth]{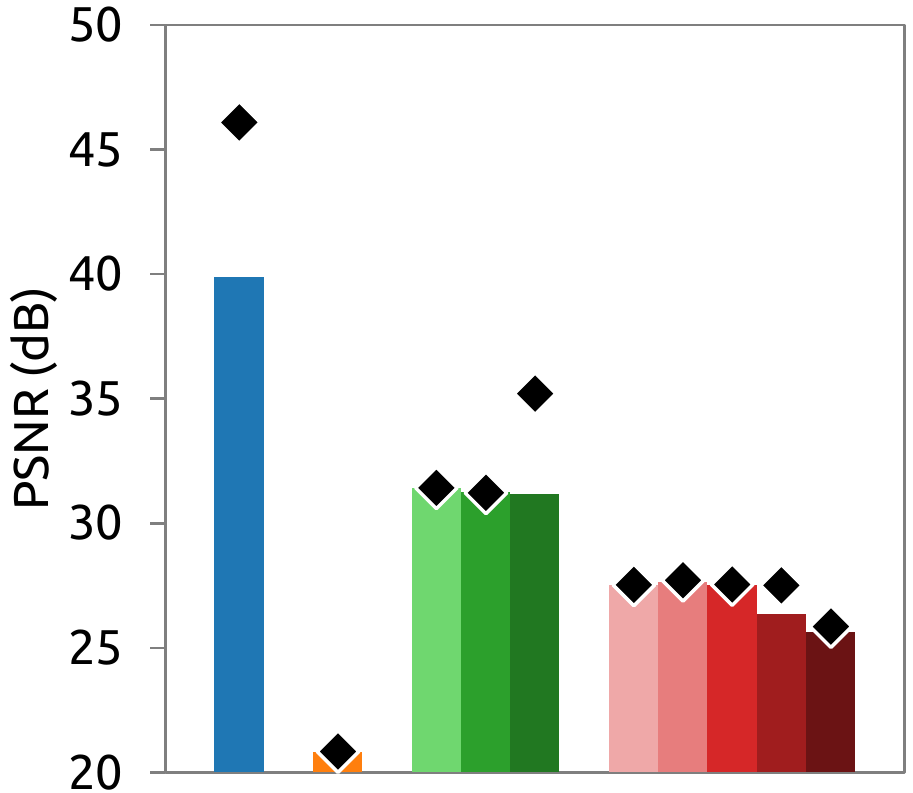}}
			\\
			\makecell[c]{JPEG} & \makecell[c]{Resizing} & \makecell[c]{Bit\\reduction} & \makecell[c]{Self-\\ensemble} & \makecell[c]{Adversarial\\training}
			\smallskip \\
			\multicolumn{5}{c}{\raisebox{-0.5\height}{~\includegraphics[width=0.480\linewidth]{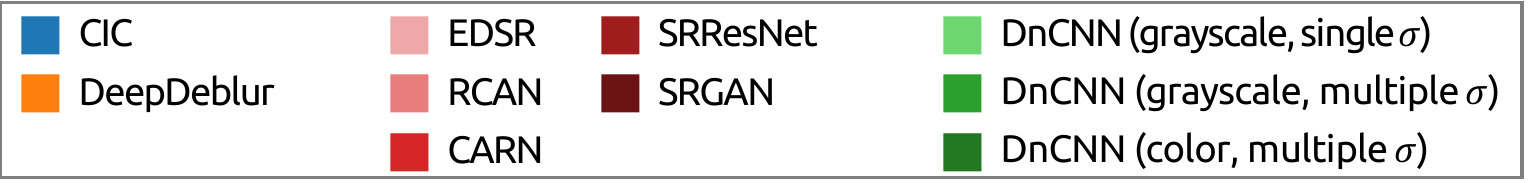}}}
		\end{tabular}
	\end{center}
	\caption{Comparison of the original performance under defense in terms of PSNR. Diamond markers refer to the original PSNR values without defense.}
	\label{fig:defenses_origpsnr}
\end{figure}

We note that defending image-to-image models brings another concern: It is not desirable if the original performance without attack is degraded due to the application of a defense method.
For classification models, slight changes in the unattacked input image due to, for instance, image transformation-based defense do not alter the image content much and thus the classification result is expected to remain the same.
However, for image-to-image models, small changes in the input image due to defense can significantly affect the quality of the output image.
In order to examine this, we compute original PSNR values from the ground-truth and output images when a defense is applied to unattacked input images.
\figurename~\ref{fig:defenses_origpsnr} shows that the defense methods tend to lower the original performance, especially when transformation-based approaches except the geometric self-ensemble are employed.
The geometric self-ensemble most successfully preserves the original performance, which is followed by adversarial training.
JPEG compression, which is the most effective defense method in \figurename~\ref{fig:defenses_vi}, harms the original performance the most.

\begin{figure}[t]
	\begin{center}
		\centering
		\renewcommand{\arraystretch}{1.0}
		\renewcommand{\tabcolsep}{0.5pt}
		\footnotesize
		\begin{tabular}{cccccc}
			\makecell[c]{No defense} & \makecell[c]{JPEG} & \makecell[c]{Resizing} & \makecell[c]{Bit\\reduction} & \makecell[c]{Self-\\ensemble} & \makecell[c]{Adversarial\\training} \smallskip \\
			\includegraphics[width=0.161\linewidth]{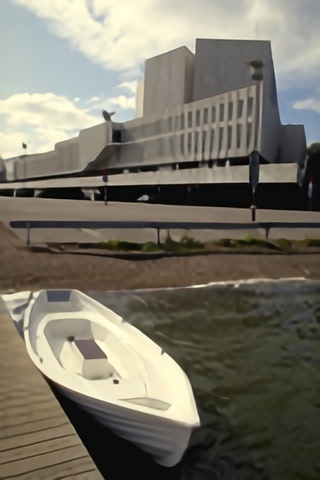} &
			\includegraphics[width=0.161\linewidth]{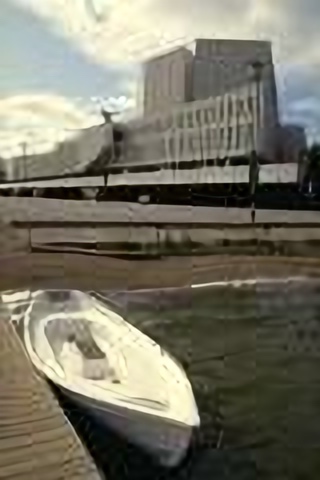} &
			\includegraphics[width=0.161\linewidth]{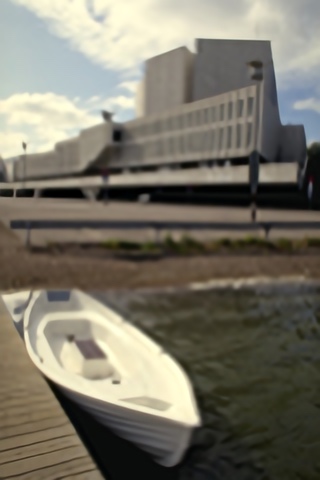} &
			\includegraphics[width=0.161\linewidth]{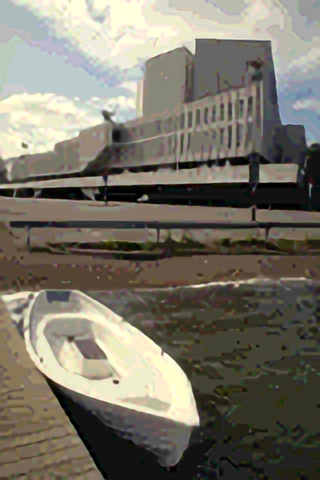} &
			\includegraphics[width=0.161\linewidth]{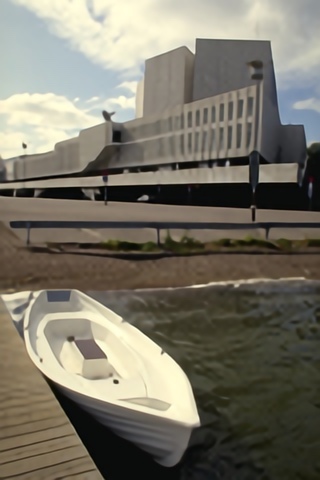} &
			\includegraphics[width=0.161\linewidth]{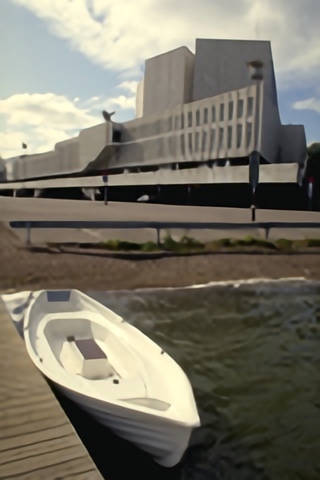}
			\\
			\multicolumn{6}{c}{\footnotesize{(a) Original outputs}} \smallskip \\
			\includegraphics[width=0.161\linewidth]{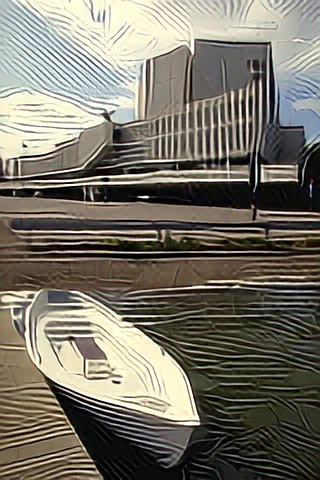} &
			\includegraphics[width=0.161\linewidth]{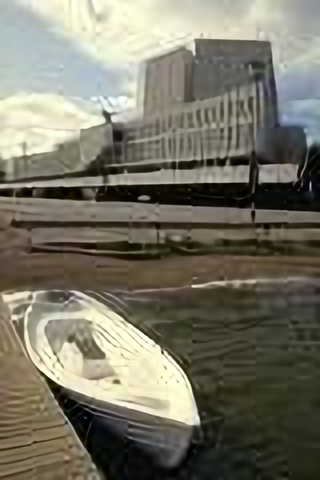} &
			\includegraphics[width=0.161\linewidth]{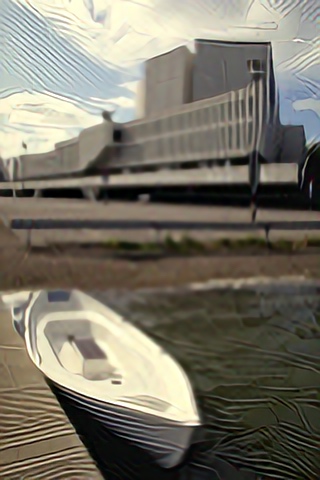} &
			\includegraphics[width=0.161\linewidth]{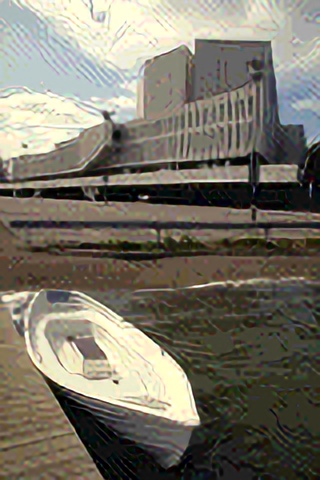} &
			\includegraphics[width=0.161\linewidth]{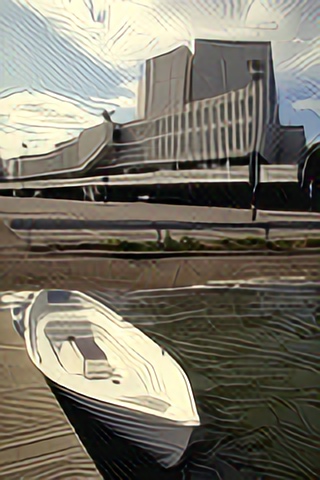} &
			\includegraphics[width=0.161\linewidth]{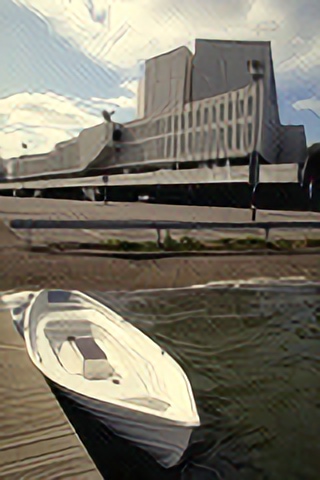}
			\\
			\multicolumn{6}{c}{\footnotesize{(b) I-FGSM}}
		\end{tabular}
	\end{center}
	\caption{Examples of the defenses for EDSR.}
	\label{fig:defense_images}
\end{figure}

\figurename~\ref{fig:defense_images} provides visual examples for super-resolution (EDSR).
It shows limitations of applying defense methods to image-to-image models, i.e., significant deterioration is introduced in the output images when the input image is not attacked, except the self-ensemble method that inherently aims to improve the original performance.
Furthermore, all defense methods are not so effective when an adversarial attack is applied; the deterioration is still clearly visible although the amount of deterioration is reduced.

We can conclude that the existing defense methods developed for image classification models are not well-suited for image-to-image models because of the distinguished characteristics of defense of image-to-image models compared to image classification models.
In other words, it is challenging for the defense methods to satisfy two conditions simultaneously: to completely remove the effect of the adversarial perturbation in the output image and to maintain the original performance when no attack exists.

\section{Conclusion}

We presented in-depth analysis on vulnerability of deep image-to-image models against adversarial attacks.
Our study aimed to 1) examine vulnerability patterns in various aspects with a proper measure of success of adversarial attacks, 2) find characteristics of image-to-image models in terms of vulnerability across different tasks, and 3) evaluate applicability of existing defense methods.
To this end, we defined a measure for image-to-image models, vulnerability index (VI), and our results showed that most state-of-the-art image-to-image models are highly vulnerable to adversarial attacks and universal perturbations are transferable even across tasks.
However, the attack results and the perturbation patterns have significantly different characteristics depending on the task type, attack method, training mechanism, and dataset.
Finally, our results showed that many of the existing defense methods used for classification models are not suitable to image-to-image models and suggested the necessity to develop effective defense methods for image-to-image models in the future.

\section*{Appendix}

\subsection{Attack Methods}

We consider two types of attack methods that are applicable to image-to-image models: a feature-based method and a gradient-based method.
As we discuss in Section~\ref{subsec:results_evaluation}, investigating vulnerability via two different approaches enables us to find various characteristics of attack methods that directly affect the patterns of the quality degradation in the output images.

As a feature-based attack, the FDA method \cite{ganeshan2019fda} is considered.
It tries to reduce the variance of the activation functions in the target model by maximizing the following function:
\begin{equation}
	\label{eq:fda}
	\begin{split}
		& \log \Big( { \big|\big| \{ \mathbf{\Phi}_{w, h, d} | \mathbf{\Phi}_{w, h, d} < C(w, h) \} \big|\big|_{2} } \Big) \\
		& - \log \Big( { \big|\big| \{ \mathbf{\Phi}_{w, h, d} | \mathbf{\Phi}_{w, h, d} > C(w, h) \} \big|\big|_{2} } \Big),
	\end{split}
\end{equation}
where $\mathbf{\Phi}$ is one of the intermediate features and $C(w, h)$ is the mean values across the channel dimension.
Because it does not rely on the model output to find a perturbation, it can be used to attack image-to-image models as well as classification models.
The perturbation for an input image is obtained iteratively while the ${L}_{\infty}$ norm of the perturbation is kept smaller than a constant $\epsilon$.
The number of iterations $T$ is set to 50 and the amount of the perturbation at each iteration is set to $\epsilon / T$.

As a gradient-based attack, the I-FGSM method \cite{choi2019evaluating} is considered.
Similarly to the original I-FGSM method for classification models \cite{kurakin2016adversarial}, it iteratively finds the attacked input $\mathbf{\widetilde{X}}$ as follows:
\begin{equation}
	\label{eq:ifgsm}
	\mathbf{\widetilde{X}}^{(i+1)} = \mathbf{\widetilde{X}}^{(i)} + \frac{\epsilon}{T} ~ { \mathrm{sgn} \Big( \nabla \big|\big| f_{m}( \mathbf{\widetilde{X}}^{(i)} ) - f_{m}( \mathbf{X} ) \big|\big|_{2} \Big) }
	,
\end{equation}
where $\mathbf{\widetilde{X}}^{(i)}$ is the attacked image at the $i$-th iteration.
Since $f_{m}(\cdot)$ can be any model that outputs an image, this method can be applied to the image-to-image models.
We set $\mathbf{\widetilde{X}}^{(0)}$$=$$\mathbf{X}$.
As in FDA, the number of iterations $T$ is set to 50.

For all the attack methods, we set $\epsilon \in \{1, 2, 4, 8, 16, 32\}$ in the pixel value scale of $[0, 255]$.

\subsection{Models and Datasets}

We consider 16 deep models for five popular image-to-image tasks: colorization, deblurring, denoising, super-resolution, and translation.
Table~\ref{table:target_models_properties} summarizes deep learning-based image-to-image models that are examined in this paper.
For colorization, CIC \cite{zhang2016colorful} is used.
For deblurring, DeepDeblur \cite{nah2017deep} is used.
For denoising, three DnCNN models \cite{zhang2017beyond} are used, which are trained on grayscale images having Gaussian noise with $\sigma$$=$$15$, grayscale images having various levels of Gaussian noise ($\sigma$$\in$$[0, 55]$), and RGB images having various levels of Gaussian noise ($\sigma$$\in$$[0, 55]$).
For super-resolution, EDSR \cite{lim2017enhanced}, RCAN \cite{zhang2018image}, CARN \cite{ahn2018fast}, SRResNet \cite{ledig2017photo}, and SRGAN \cite{ledig2017photo} are used.
For translation, three CycleGAN models \cite{zhu2017unpaired} trained on three pairs of datasets in different domains (apple $\leftrightarrow$ orange, horse $\leftrightarrow$ zebra, and Van Gogh's paintings $\leftrightarrow$ photos) are used.
The models are trained with the procedures reported in the original papers.
The training datasets are also the same to those used in the original papers.

Evaluation is performed using the datasets that are commonly used for evaluating each task in the literature.
For colorization, we use the 1,000 images of the validation split of the ImageNet dataset \cite{russakovsky2015imagenet} after cropping at the center regions and resizing to 224$\times$224 pixels (i.e., the same size as the training images).
For deblurring, we use the K{\"o}hler dataset \cite{kohler2012recording}.
For denoising, we employ 68 images in the BSD500 dataset \cite{arbelaez2010contour} that are not used for training the models, as in \cite{zhang2017beyond}.
For super-resolution, we employ the BSD100 dataset \cite{martin2001database}.
For translation, we employ the validation dataset provided by \cite{zhu2017unpaired} for each translation pair.

\subsection{Implementation Details}

We conduct our experiments on various CPUs (e.g., Intel Xeon CPU E5-1660v3, Intel Core i7-7700) and GPUs (e.g., NVIDIA GTX 1080, NVIDIA GTX 2080Ti).
We implement our code by using TensorFlow with Python.

\subsection{Additional Results}

\subsubsection{Performance Comparison in Terms of PSNR}

We first show the performance comparison in terms of VI for different attack methods employed with various $\epsilon$ values in \figurename~\ref{fig:basic_attack_psnrratio_appendix}.
In addition, we show the performance comparison in terms of the PSNR values for the input (${Q}_{i}$) and the output (${Q}_{o}$) in \figurename~\ref{fig:basic_attack_psnr}.
A curve closer to the lower right corner means that the corresponding model is more vulnerable, i.e., a small amount of perturbation in the input image results in a large amount of deterioration in the output image.

\subsubsection{Output Degradation on Other Translation Models}

To supplement the results in \figurename~\ref{fig:basic_attack_datasets}, we show additional example images obtained from the CycleGAN models converting horses to zebras and Van Gogh's paintings to photos, when FDA is employed, in \figurename~\ref{fig:basic_attack_datasets_appendix}.

\subsubsection{Performance Comparison in Terms of SSIM}

Along with the evaluation results in terms of PSNR, we also include the performance comparison in terms of structural similarity (SSIM) \cite{wang2004image}.
Similar to the performance comparison in terms of VI derived from PSNR, we calculate the vulnerability index calculated on SSIM, i.e., $\mathrm{VI}_{\mathrm{SSIM}}$$=$$\mathrm{SSIM}_{i}/\mathrm{SSIM}_{o}$, where $\mathrm{SSIM}_{i}$ and $\mathrm{SSIM}_{o}$ are the SSIM values for the input and output images, respectively.

The performance comparison is shown in \figurename~\ref{fig:basic_attack_ssimratio}.
The overall trend is similar to that shown in \figurename~\ref{fig:basic_attack_psnrratio}.
One noticeable difference is that $\mathrm{VI}_{\mathrm{SSIM}}$ penalizes the super-resolution models more than the translation models, while the translation models appear more vulnerable in terms of $\mathrm{VI}$ calculated on PSNR.
This is because although the pixel-wise changes in the output images are larger for the translation models than for the super-resolution models, which makes $\mathrm{VI}$ higher for the former, the changes for the translation models tend to alter the image content (e.g., textures) and be less perceptually annoying than noise-like distortion in the super-resolution models, which is considered by $\mathrm{VI}_{\mathrm{SSIM}}$.

In addition, the performance comparison in terms of $\mathrm{SSIM}_{i}$ and $\mathrm{SSIM}_{o}$ is shown in \figurename~\ref{fig:basic_attack_ssim}.

\subsubsection{Frequency-Aware Attack with Different Values of $r$}

In Section~\ref{sec:results}, we show the frequency-aware attack, which finds the perturbation only in a low-frequency or high-frequency subspace.
While the results obtained with $r$$=$$1/4$ are shown, we also report the results obtained with different values of $r$.
The quantitative performance comparison in terms of VI is shown in Figs.~\ref{fig:lowfreq_attack_psnrratio} and \ref{fig:highfreq_attack_psnrratio}.
In addition, example input and output images are also shown in \figurename~\ref{fig:frequency_attack_examples}.

\subsubsection{Defenses against Adversarial Attacks}

In Section~\ref{sec:defenses}, we discuss the effectiveness of conventional defense approaches in the image-to-image tasks.
We show additional example results obtained from the models of five image-to-image tasks.
\figurename~\ref{fig:defense_transformation_images} shows visual examples of the transformation-based defenses, including JPEG compression, random resizing, bit reduction, and geometric self-ensemble.
\figurename~\ref{fig:defense_adversarial_training_images} shows visual examples of the adversarial training-based defense.

\clearpage

\begin{table*}[t]
	\caption{Details of the examined image-to-image models and datasets.}
	\label{table:target_models_properties}
	\renewcommand{\arraystretch}{1.2}
	\centering
	\begin{center}
		\begin{tabular}{c|c|c|c|c}
			\hline
			\hline
			\textbf{Task} & \textbf{Model} & \textbf{With GAN} & \textbf{Training dataset} & \textbf{Evaluation dataset} \\
			\hline
      \multirow{1}{*}{Colorization} & CIC \cite{zhang2016colorful} & - & ImageNet \cite{russakovsky2015imagenet} & ImageNet \cite{russakovsky2015imagenet} subset \\
			\hline
      \multirow{1}{*}{Deblurring} & DeepDeblur \cite{nah2017deep} & Yes & GOPRO \cite{nah2017deep} & K{\"o}hler \cite{kohler2012recording} \\
			\hline
      \multirow{3}{*}{Denoising} & DnCNN \cite{zhang2017beyond} (grayscale, single $\sigma$) & - & BSD500 \cite{arbelaez2010contour} subset & BSD500 \cite{arbelaez2010contour} subset \\
      & DnCNN \cite{zhang2017beyond} (grayscale, multiple $\sigma$) & - & BSD500 \cite{arbelaez2010contour} subset & BSD500 \cite{arbelaez2010contour} subset \\
      & DnCNN \cite{zhang2017beyond} (color, multiple $\sigma$) & - & BSD500 \cite{arbelaez2010contour} subset & BSD500 \cite{arbelaez2010contour} subset \\
			\hline
      \multirow{5}{*}{\makecell[c]{Super-\\resolution}} & EDSR \cite{lim2017enhanced} & - & DIV2K \cite{agustsson2017ntire} & BSD100 \cite{martin2001database} \\
      & RCAN \cite{zhang2018image} & - & DIV2K \cite{agustsson2017ntire} & BSD100 \cite{martin2001database} \\
      & CARN \cite{ahn2018fast} & - & DIV2K \cite{agustsson2017ntire} & BSD100 \cite{martin2001database} \\
      & SRResNet \cite{ledig2017photo} & - & ImageNet \cite{russakovsky2015imagenet} subset & BSD100 \cite{martin2001database} \\
      & SRGAN \cite{ledig2017photo} & Yes & ImageNet \cite{russakovsky2015imagenet} subset & BSD100 \cite{martin2001database} \\
			\hline
      \multirow{3}{*}{Translation} & CycleGAN \cite{zhu2017unpaired} (apple $\leftrightarrow$ orange) & Yes & ImageNet \cite{russakovsky2015imagenet} subset & ImageNet \cite{russakovsky2015imagenet} subset \\
      & CycleGAN \cite{zhu2017unpaired} (horse $\leftrightarrow$ zebra) & Yes & ImageNet \cite{russakovsky2015imagenet} subset & ImageNet \cite{russakovsky2015imagenet} subset \\
      & CycleGAN \cite{zhu2017unpaired} (vangogh $\leftrightarrow$ photo) & Yes & Wikiart, Flickr \cite{zhu2017unpaired} & Wikiart, Flickr \cite{zhu2017unpaired} \\
			\hline
			\hline
		\end{tabular}
	\end{center}
\end{table*}

\begin{figure*}[t]
	\begin{center}
		\centering
		\renewcommand{\arraystretch}{1.5}
		\renewcommand{\tabcolsep}{0.8pt}
		\footnotesize
		\begin{tabular}{cccccc}
			$\epsilon=1$ & $\epsilon=2$ & $\epsilon=4$ & $\epsilon=8$ & $\epsilon=16$ & $\epsilon=32$ \\
			\raisebox{-0.5\height}{\includegraphics[width=0.160\linewidth]{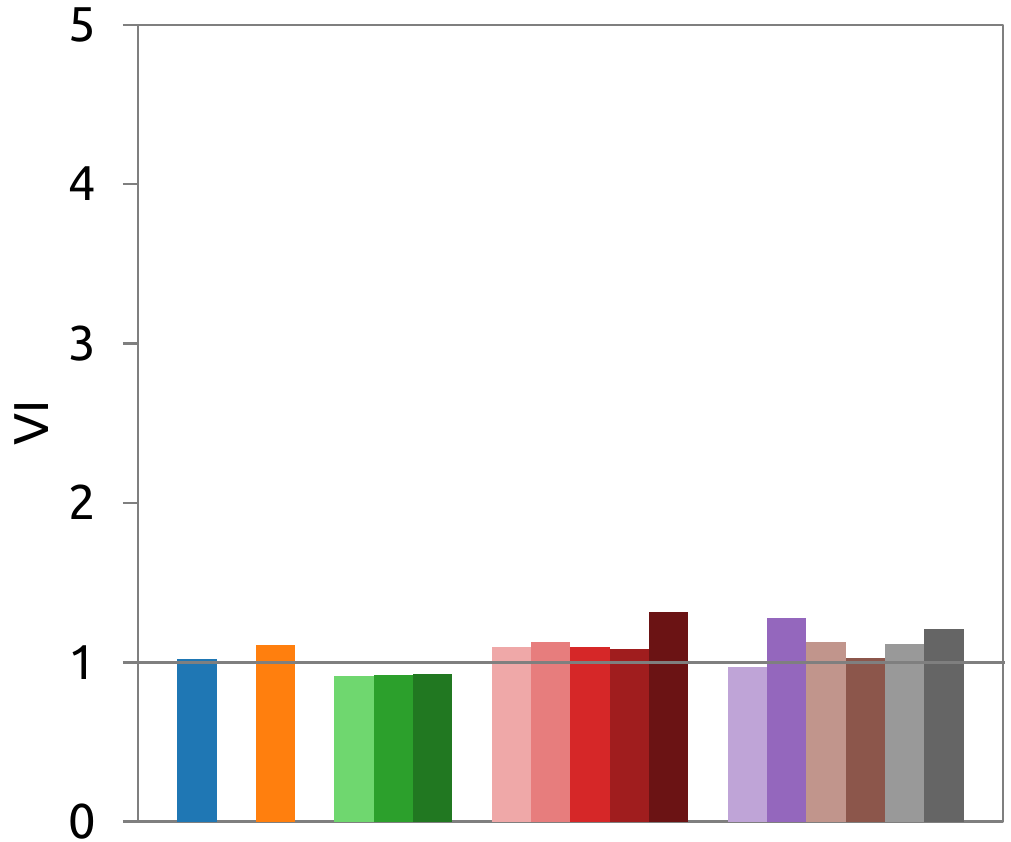}} &
			\raisebox{-0.5\height}{\includegraphics[width=0.160\linewidth]{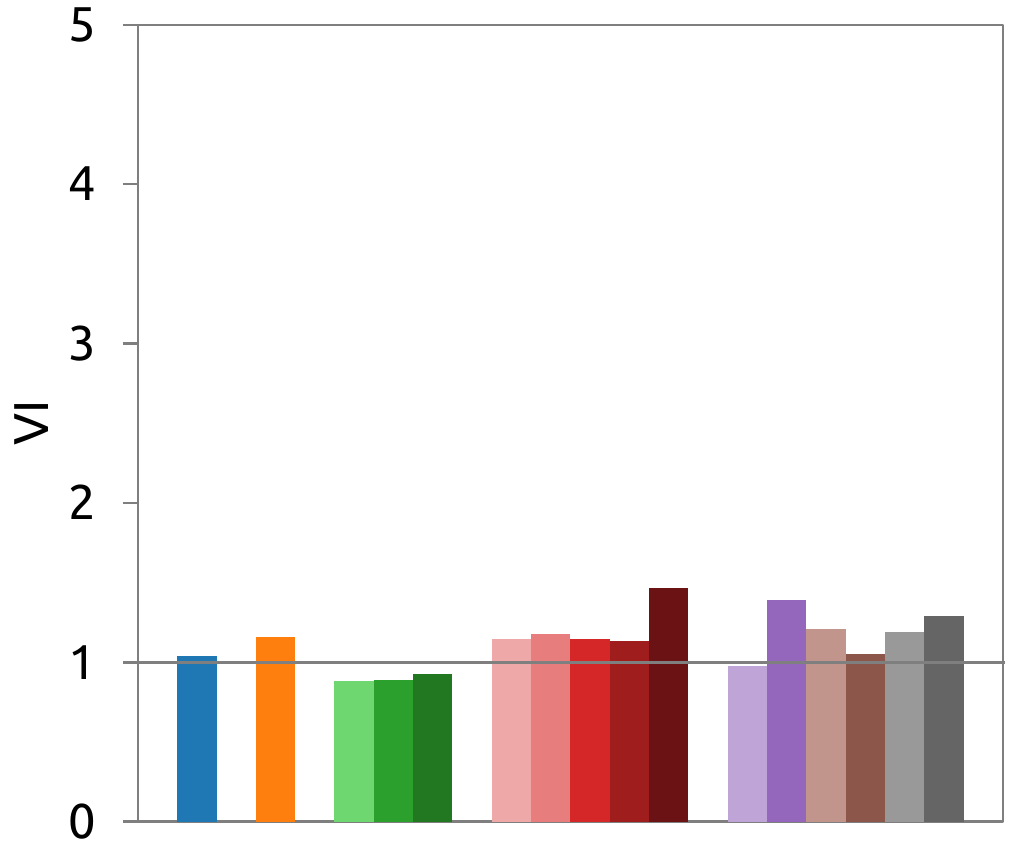}} &
			\raisebox{-0.5\height}{\includegraphics[width=0.160\linewidth]{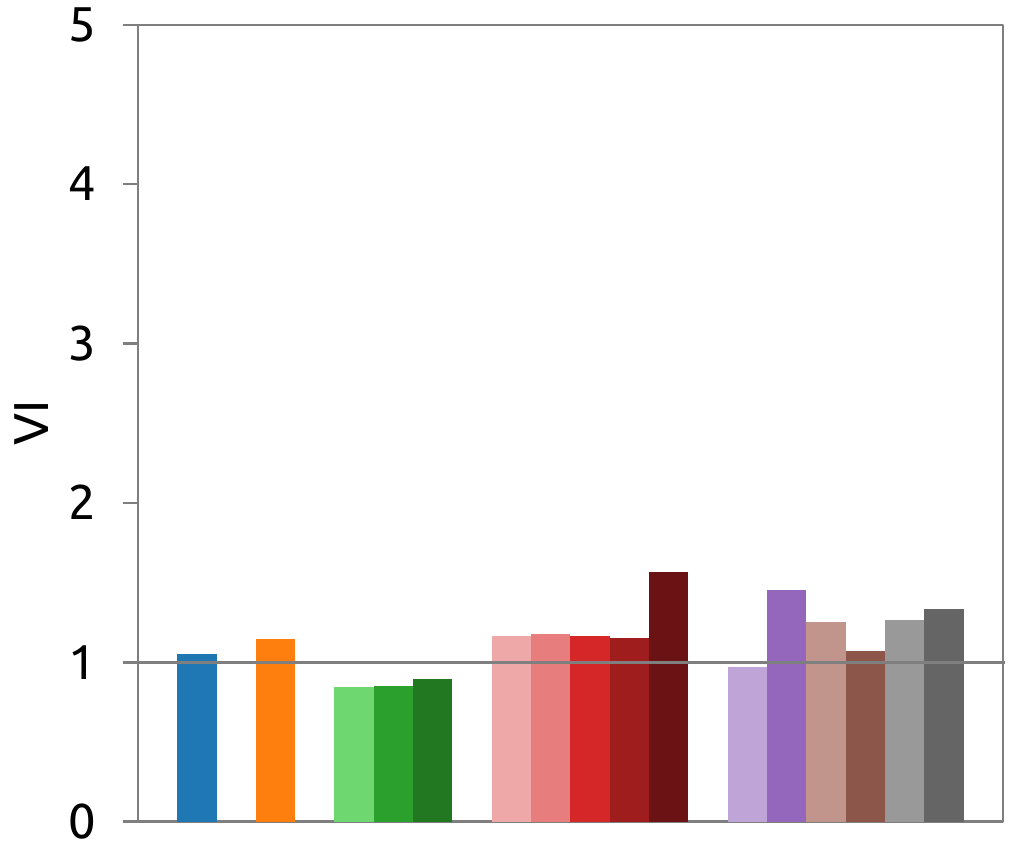}} &
			\raisebox{-0.5\height}{\includegraphics[width=0.160\linewidth]{figures/basic_vi_randuniform_e8}} &
			\raisebox{-0.5\height}{\includegraphics[width=0.160\linewidth]{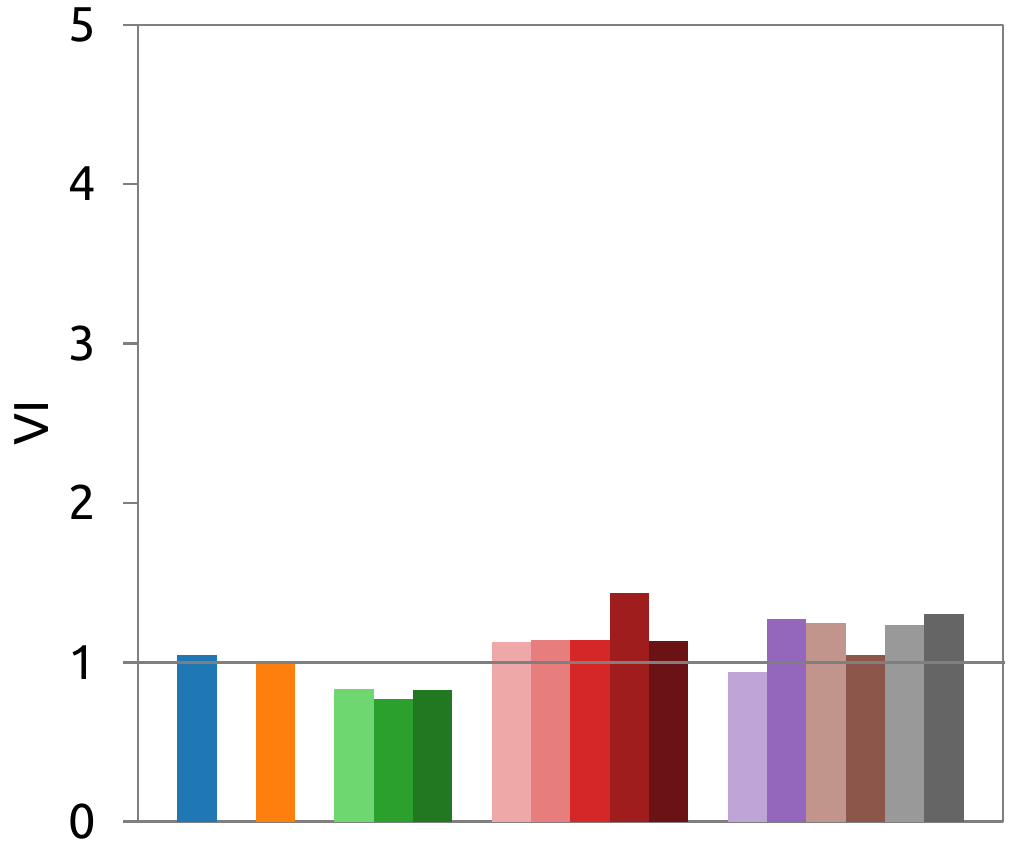}} &
			\raisebox{-0.5\height}{\includegraphics[width=0.160\linewidth]{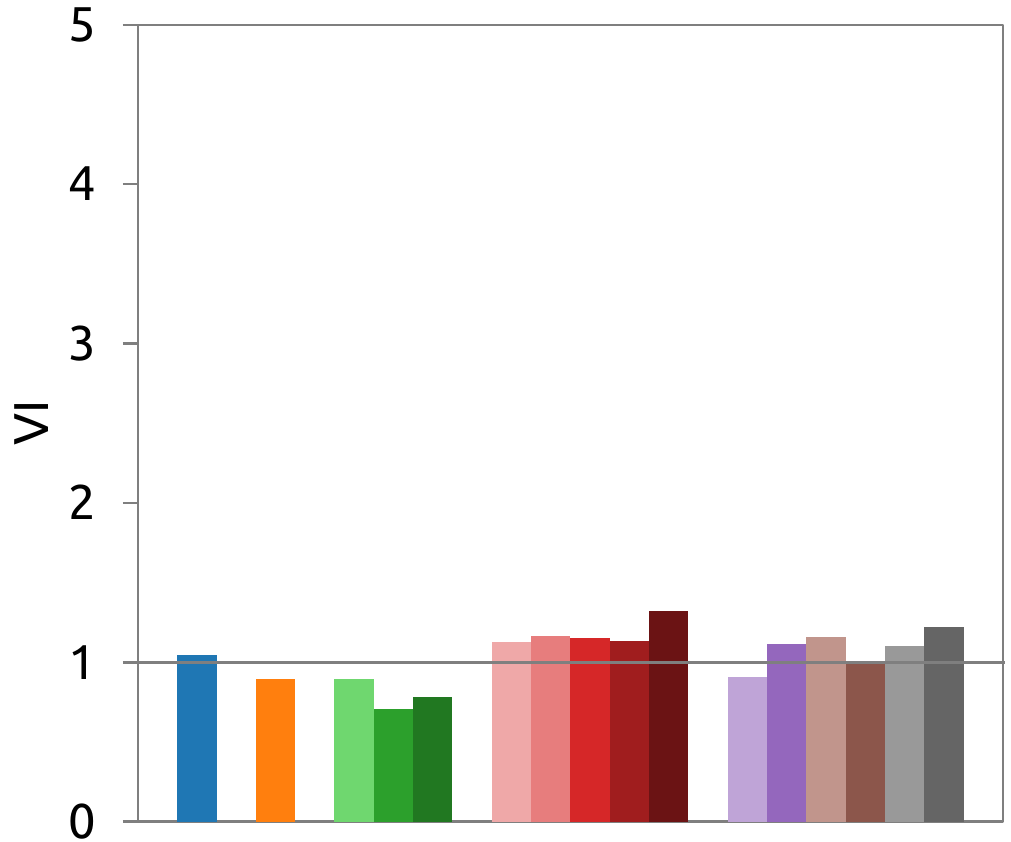}} \\
			\multicolumn{6}{c}{(a) Random uniform} \\
			\raisebox{-0.5\height}{\includegraphics[width=0.160\linewidth]{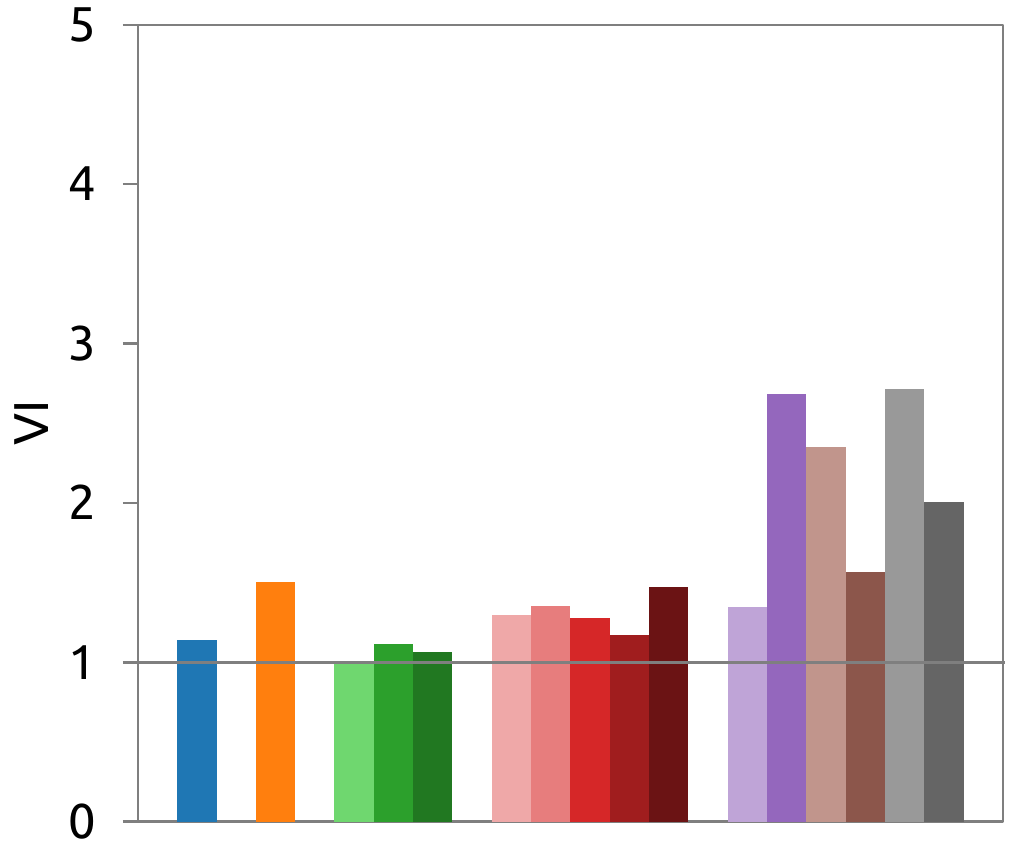}} &
			\raisebox{-0.5\height}{\includegraphics[width=0.160\linewidth]{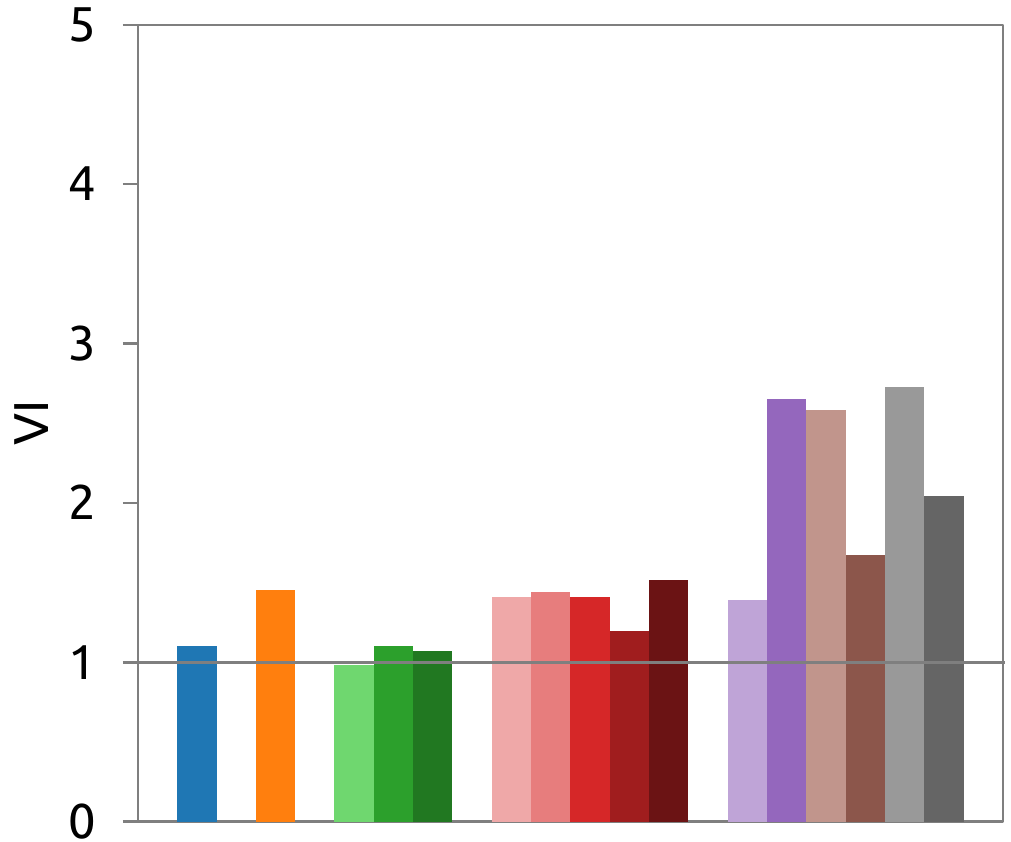}} &
			\raisebox{-0.5\height}{\includegraphics[width=0.160\linewidth]{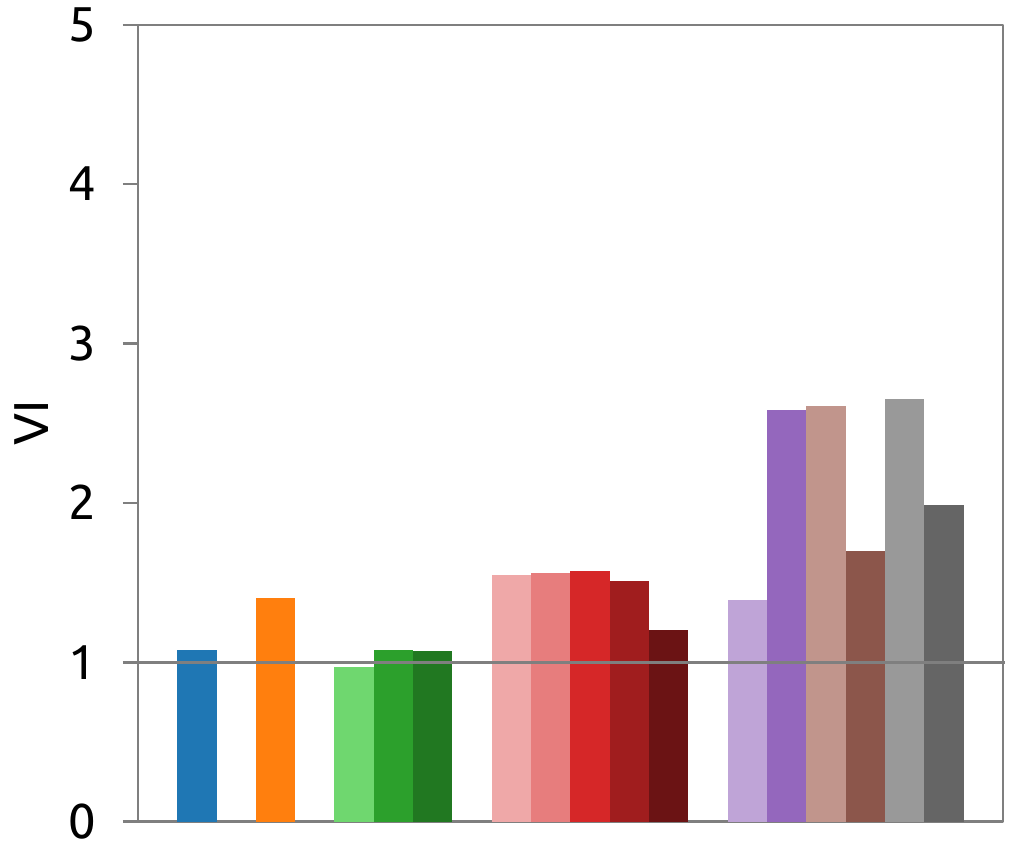}} &
			\raisebox{-0.5\height}{\includegraphics[width=0.160\linewidth]{figures/basic_vi_fda_e8}} &
			\raisebox{-0.5\height}{\includegraphics[width=0.160\linewidth]{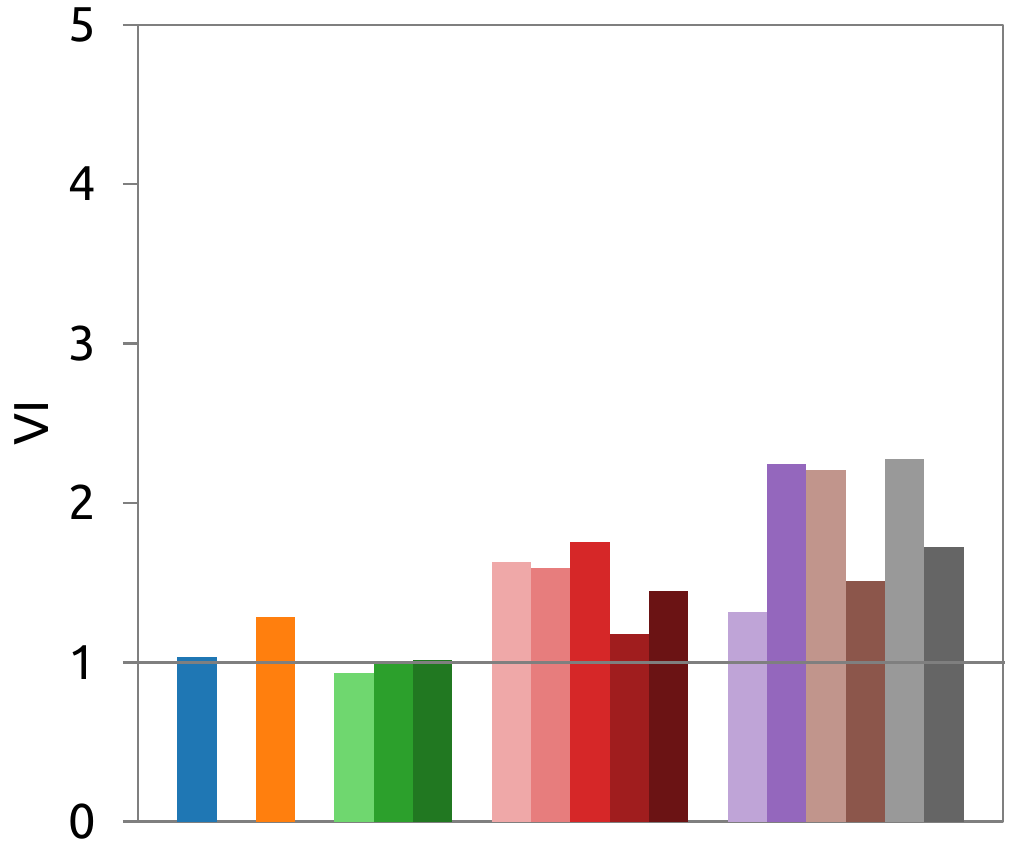}} &
			\raisebox{-0.5\height}{\includegraphics[width=0.160\linewidth]{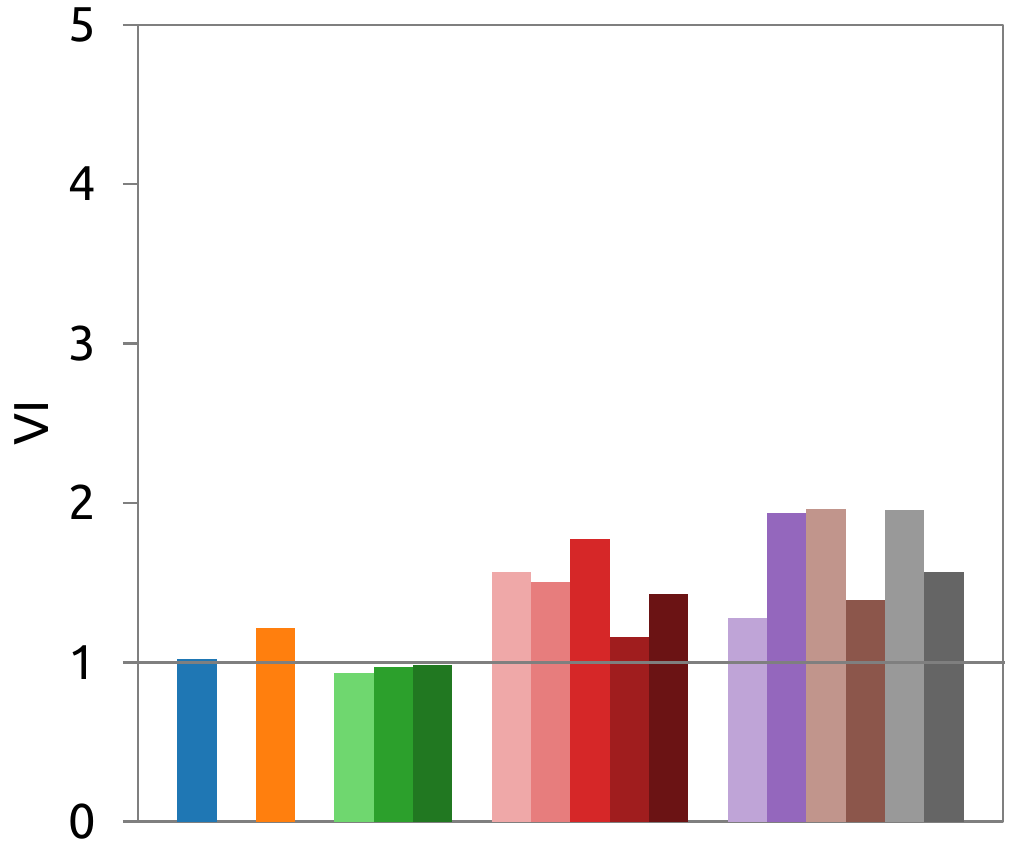}} \\
			\multicolumn{6}{c}{(b) FDA} \\
			\raisebox{-0.5\height}{\includegraphics[width=0.160\linewidth]{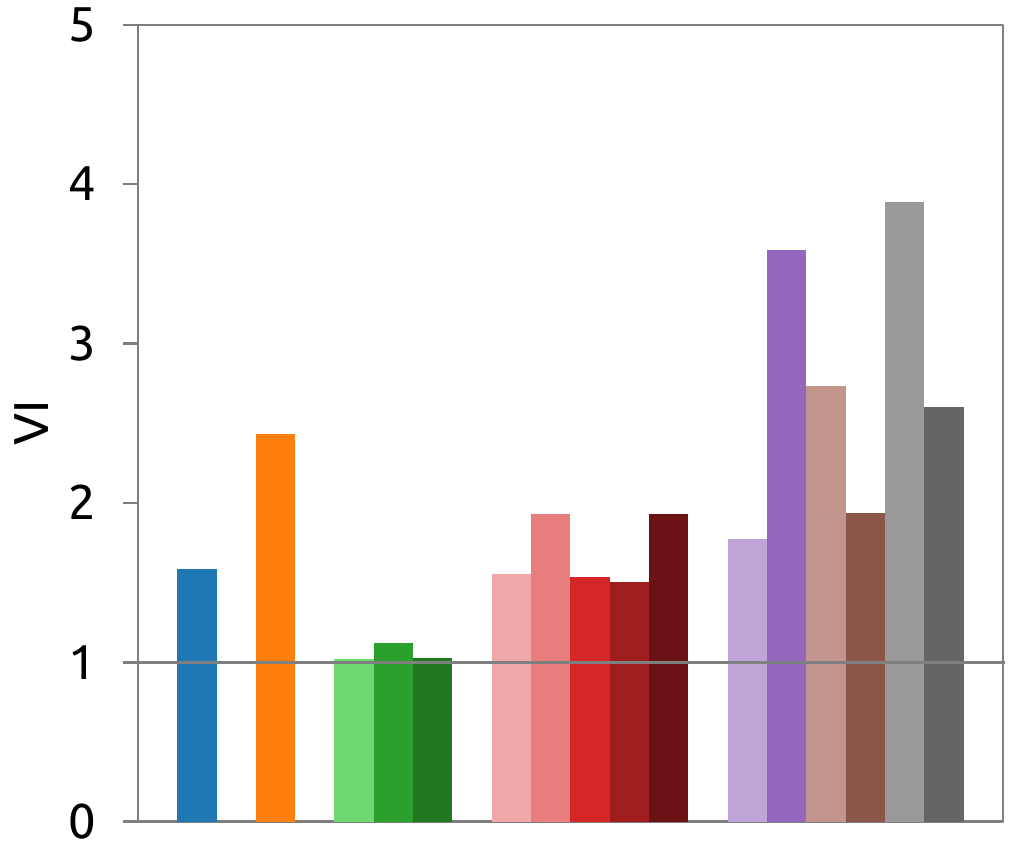}} &
			\raisebox{-0.5\height}{\includegraphics[width=0.160\linewidth]{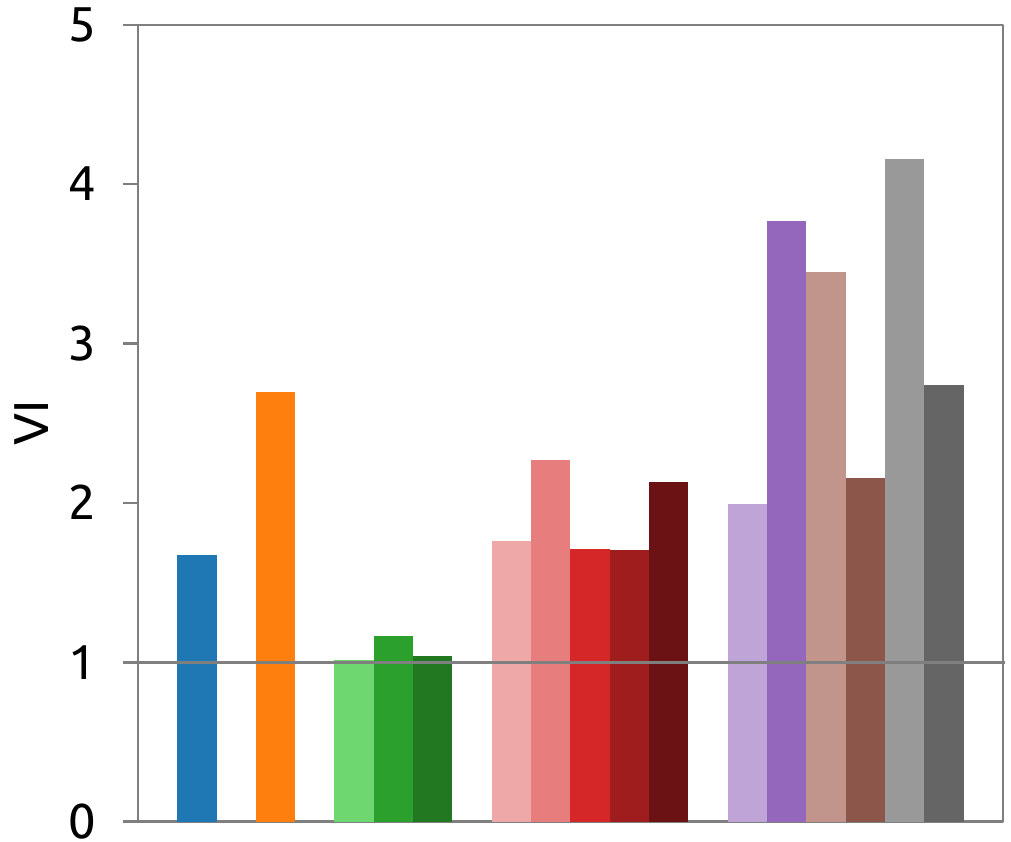}} &
			\raisebox{-0.5\height}{\includegraphics[width=0.160\linewidth]{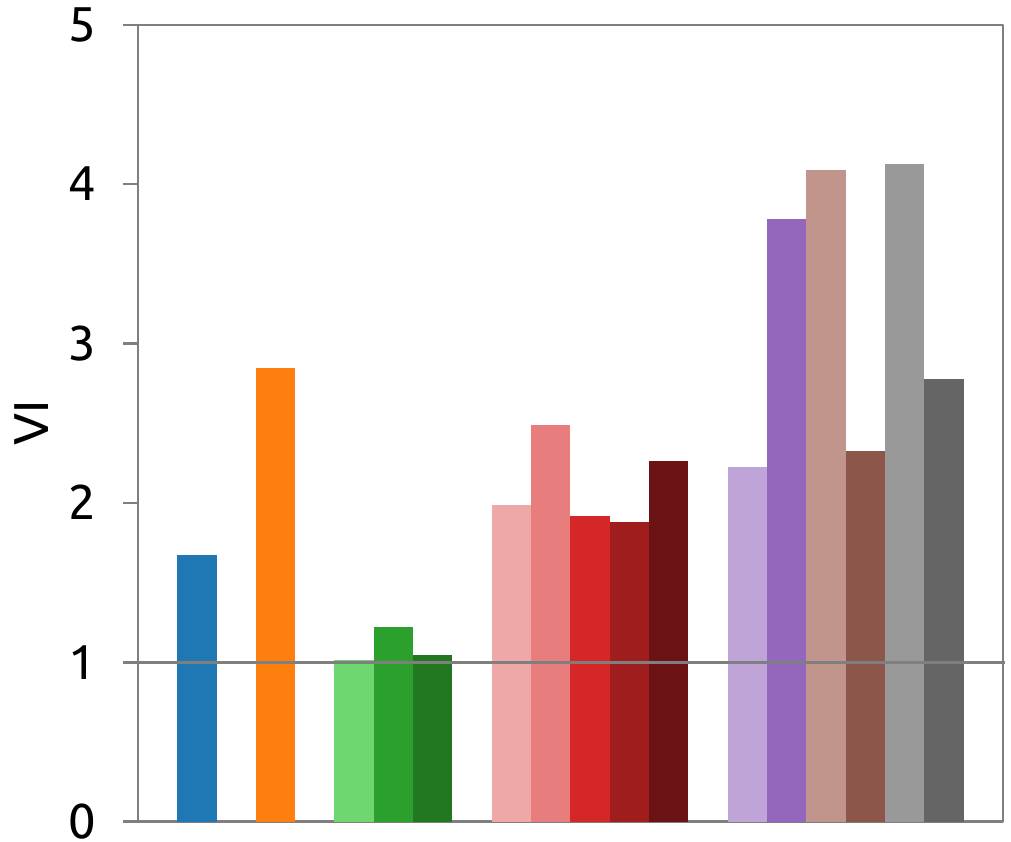}} &
			\raisebox{-0.5\height}{\includegraphics[width=0.160\linewidth]{figures/basic_vi_ifgsm_e8}} &
			\raisebox{-0.5\height}{\includegraphics[width=0.160\linewidth]{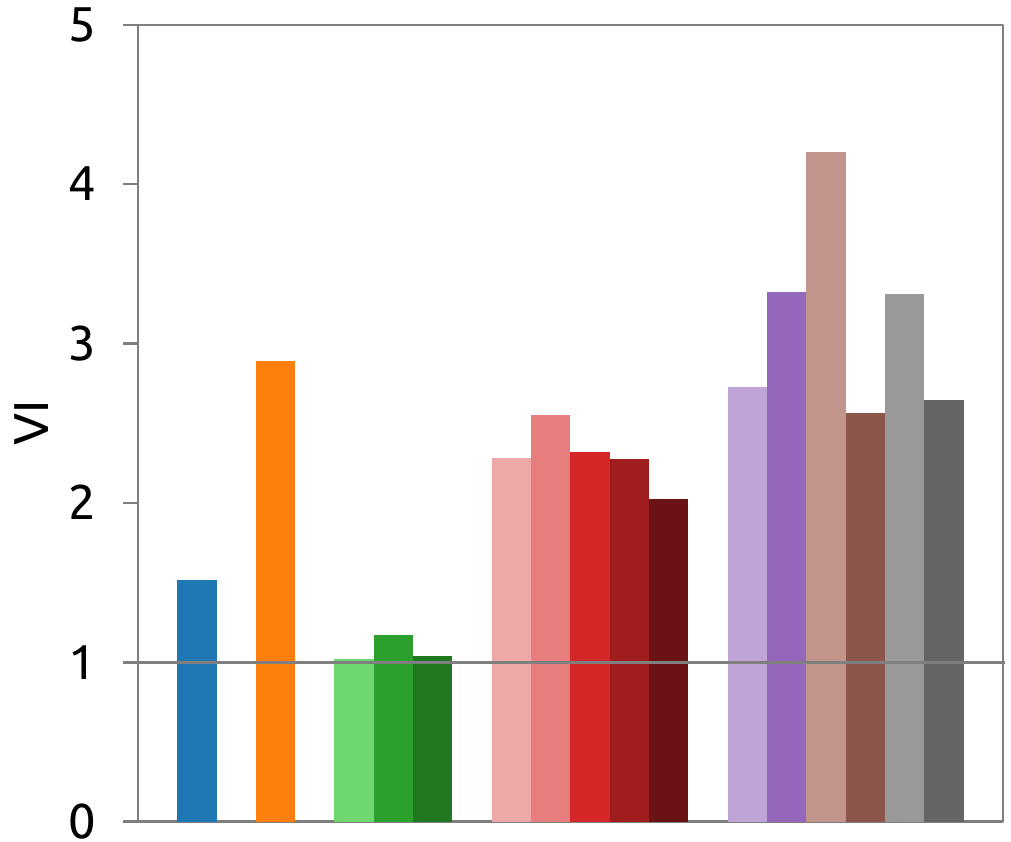}} &
			\raisebox{-0.5\height}{\includegraphics[width=0.160\linewidth]{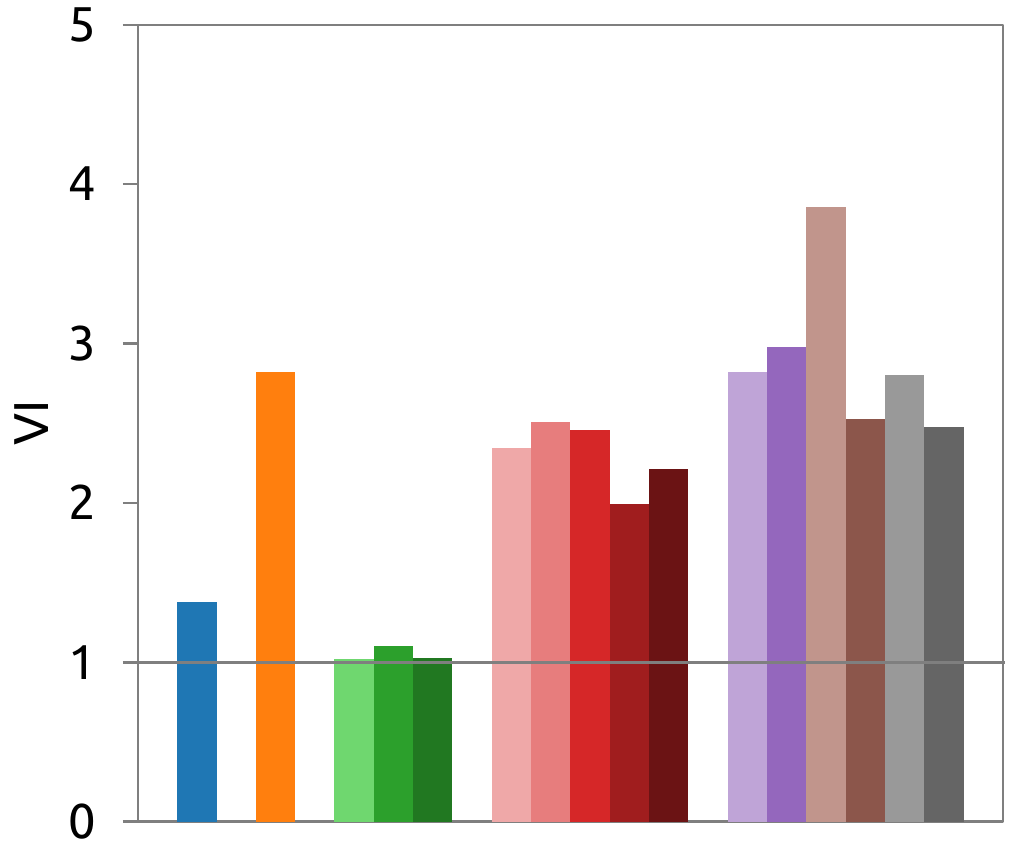}} \\
			\multicolumn{6}{c}{(c) I-FGSM} \\
			\multicolumn{6}{c}{\raisebox{-0.5\height}{\includegraphics[width=0.8\linewidth]{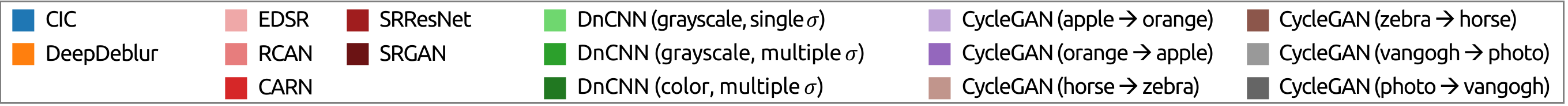}}}
		\end{tabular}
	\end{center}
	\caption{Performance comparison in terms of VI for different attack methods employed with $\epsilon \in \{1, 2, 4, 8, 16, 32\}$.}
	\label{fig:basic_attack_psnrratio_appendix}
\end{figure*}

\begin{figure*}[t]
	\begin{center}
		\centering
		\renewcommand{\arraystretch}{1.5}
		\renewcommand{\tabcolsep}{1.2pt}
		\footnotesize
		\begin{tabular}{ccccc}
			Colorization & Deblurring & Denoising & Super-resolution & Translation \\ \includegraphics[width=0.195\textwidth]{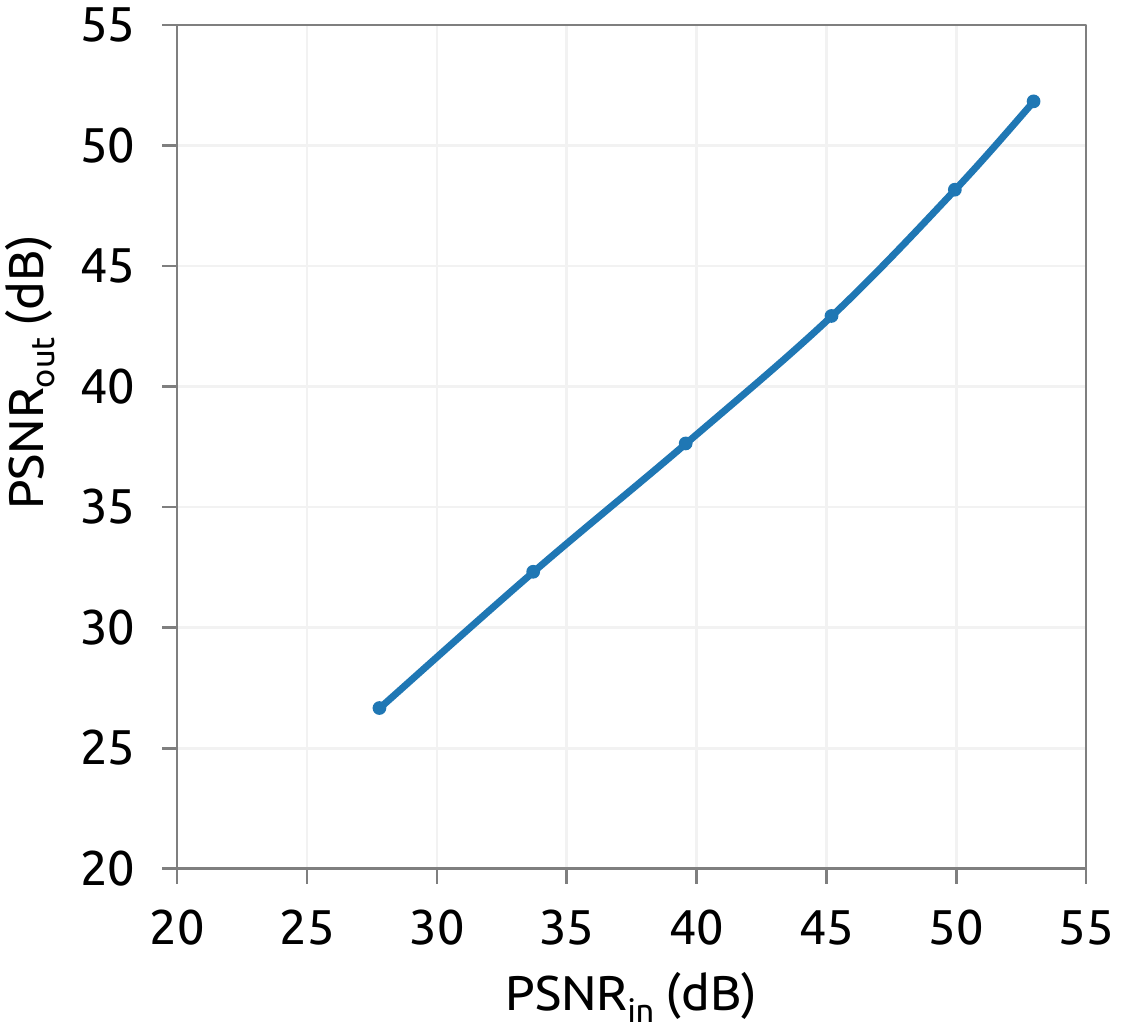} & \includegraphics[width=0.195\textwidth]{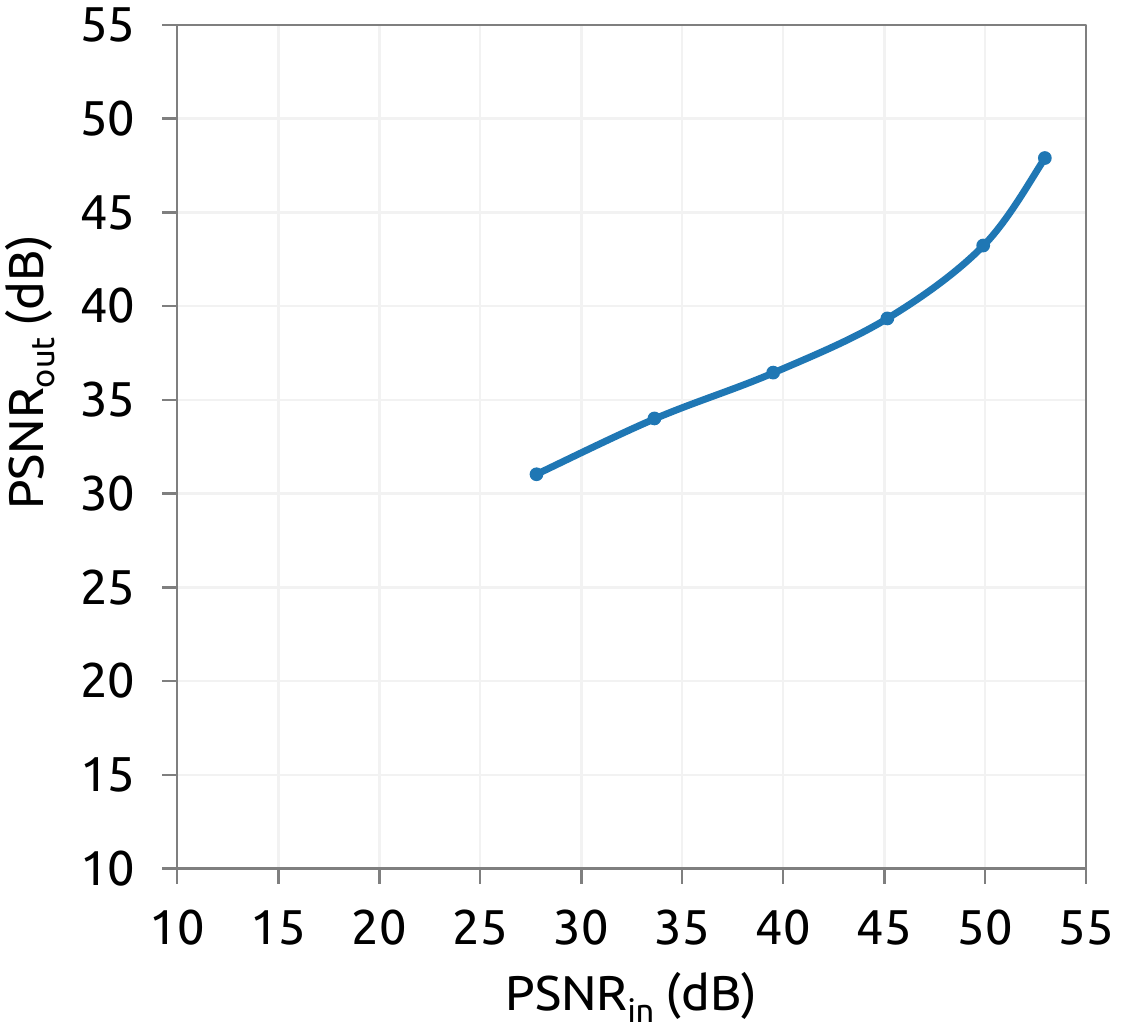} & \includegraphics[width=0.195\textwidth]{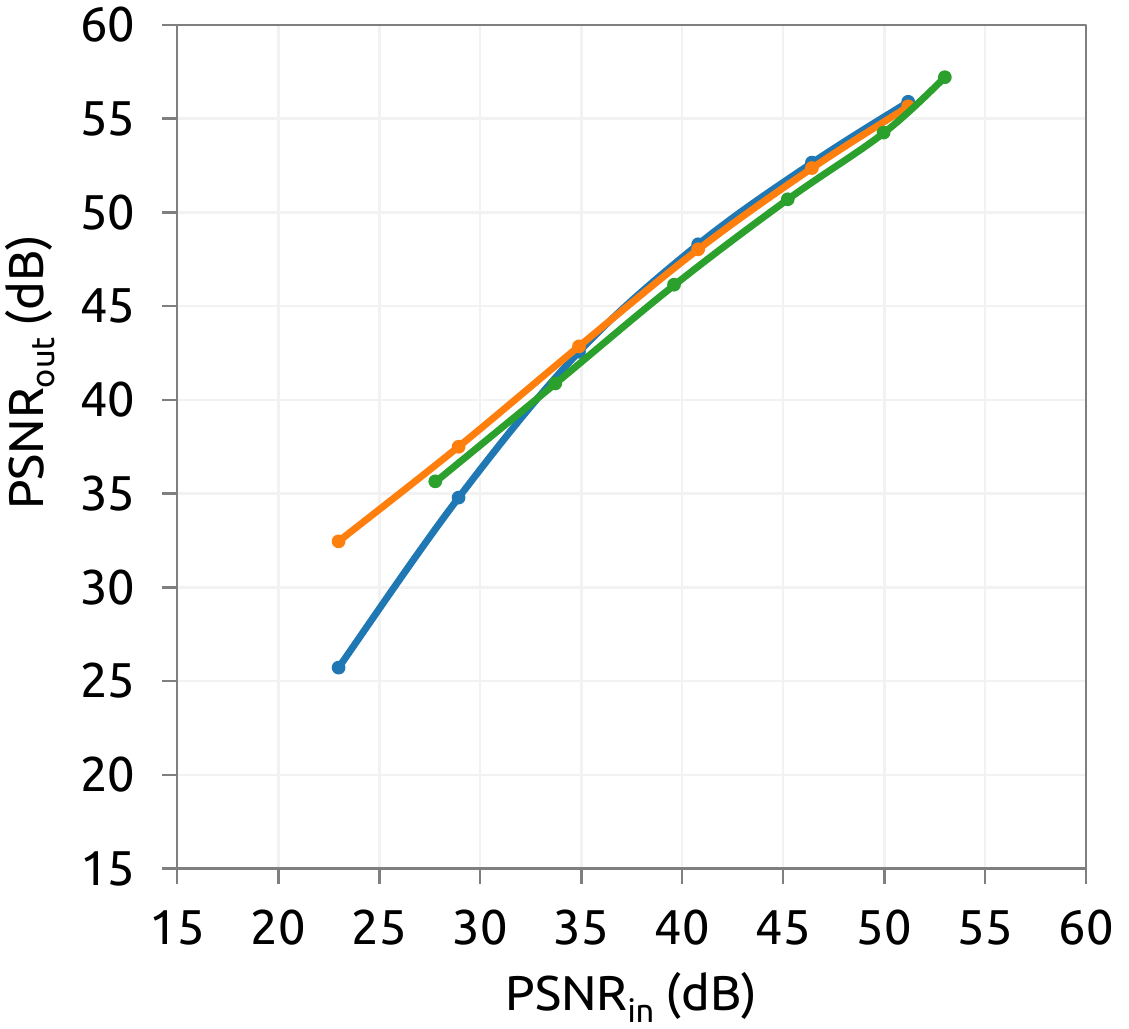} & \includegraphics[width=0.195\textwidth]{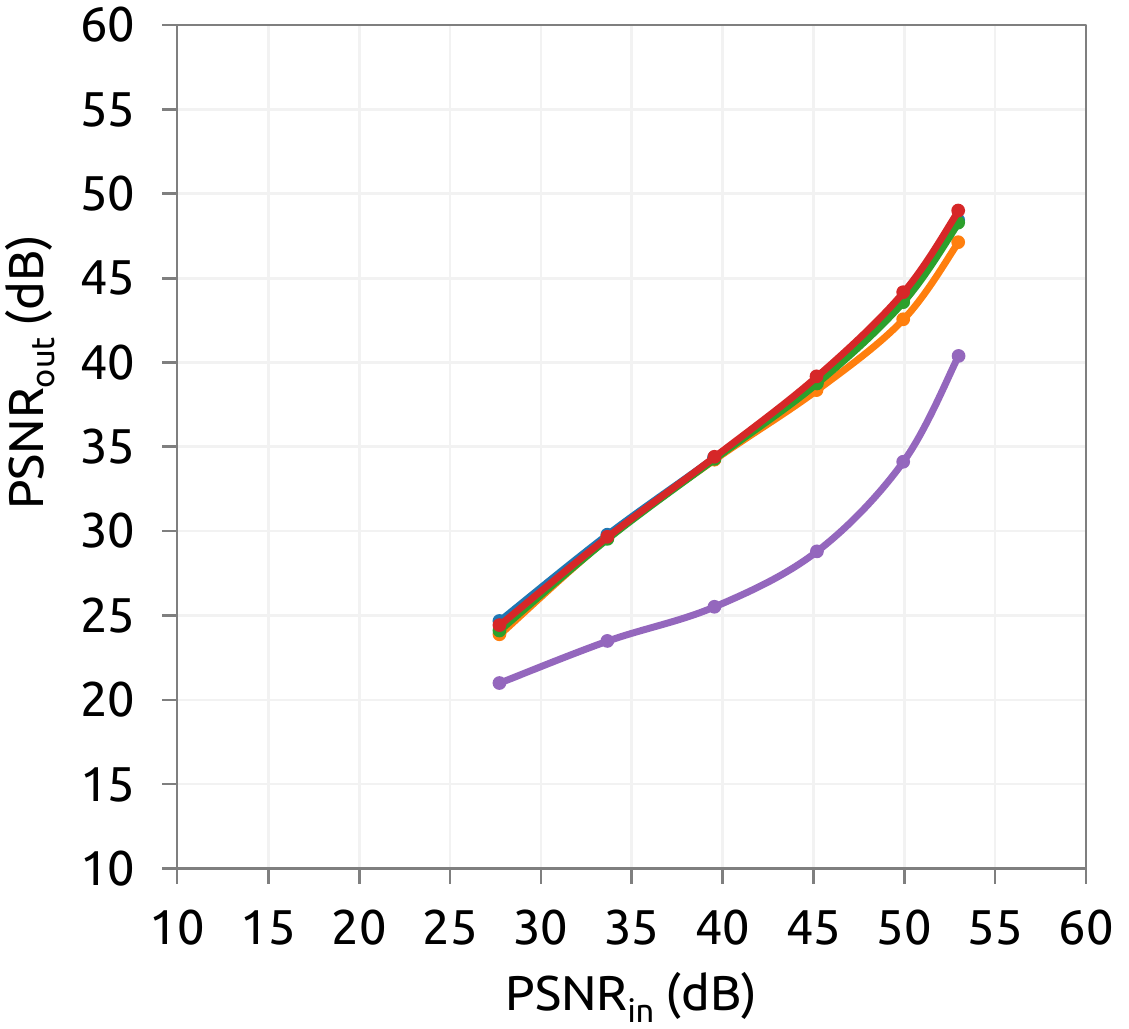} & \includegraphics[width=0.195\textwidth]{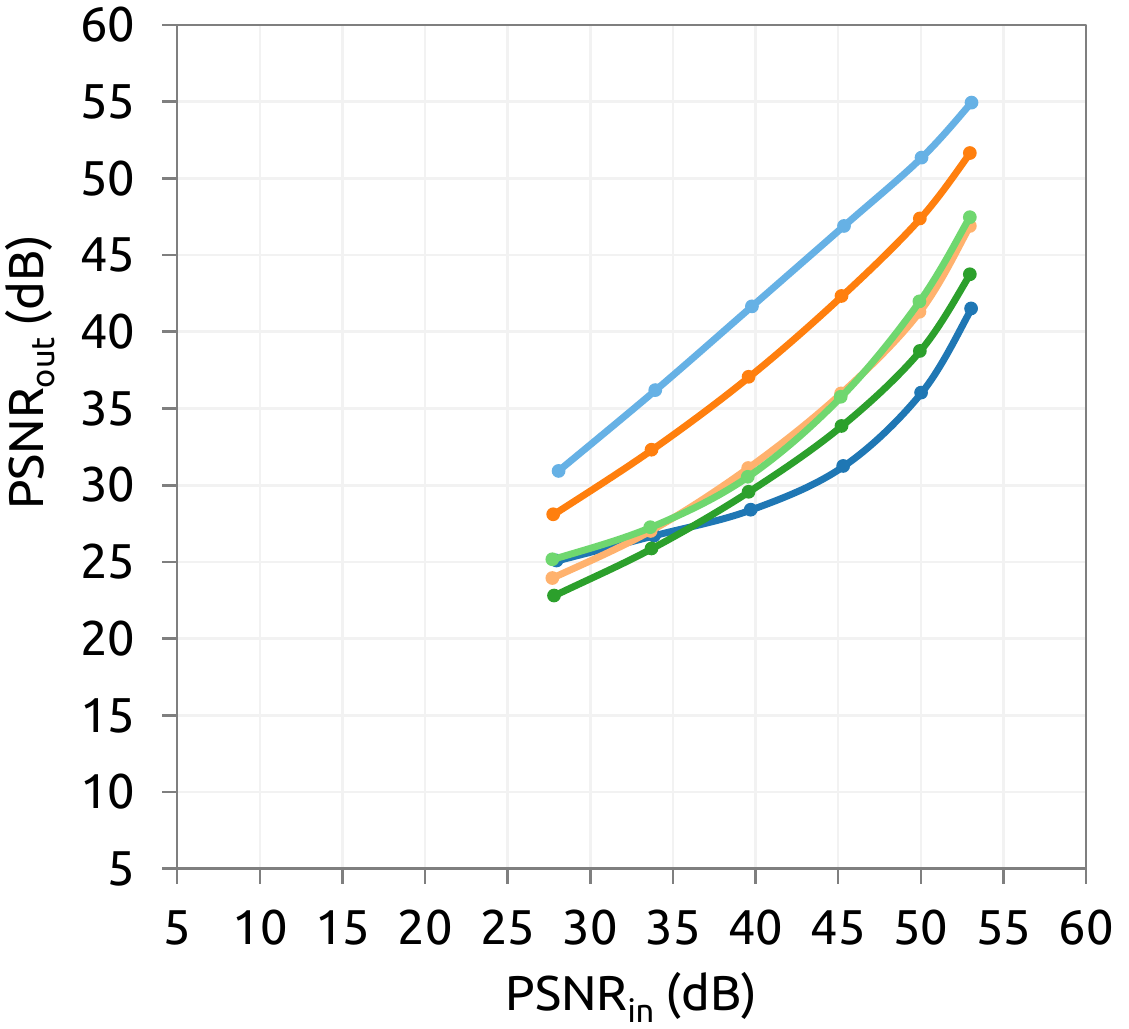} \\
			\multicolumn{5}{c}{\footnotesize{(a) Random uniform}} \\
			\includegraphics[width=0.195\textwidth]{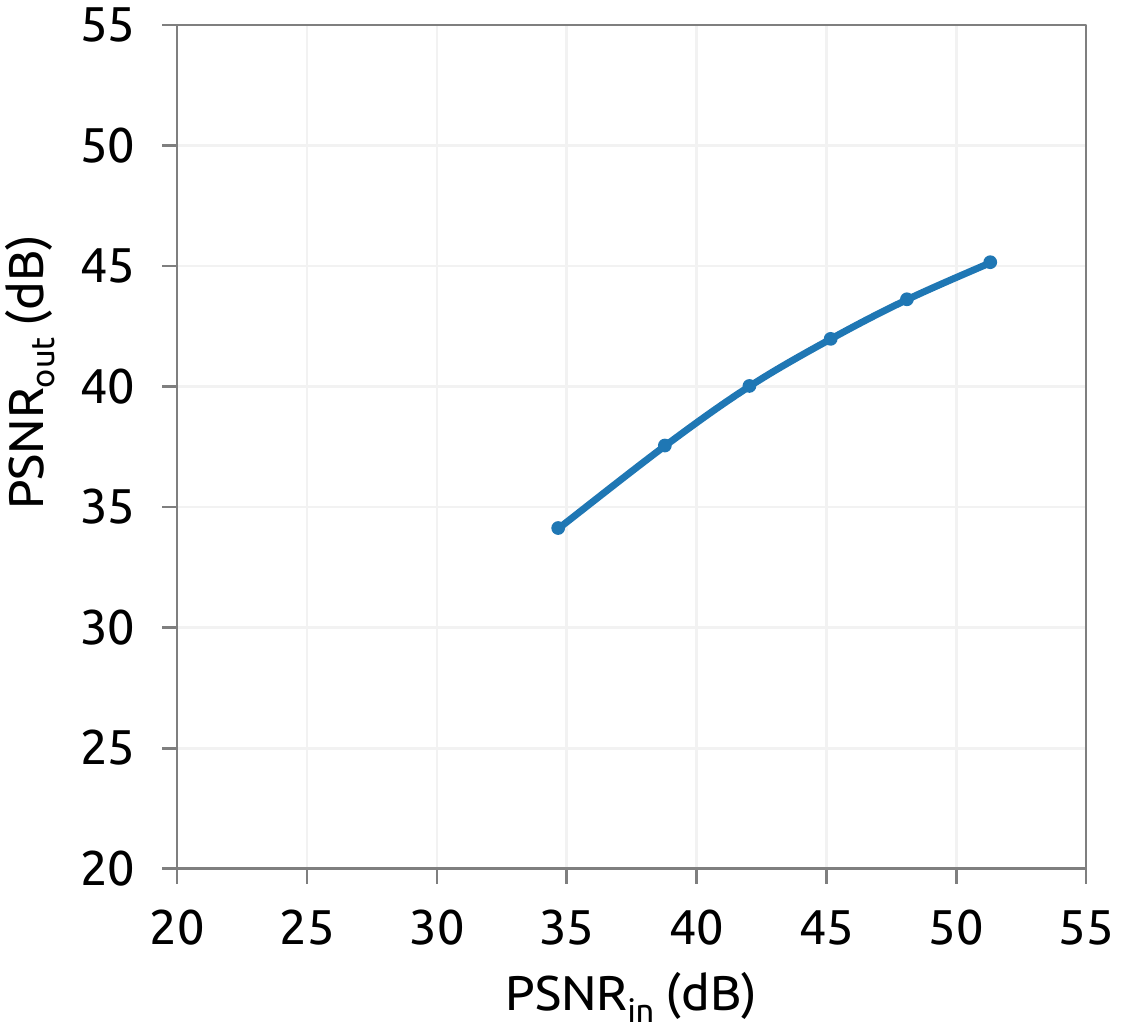} & \includegraphics[width=0.195\textwidth]{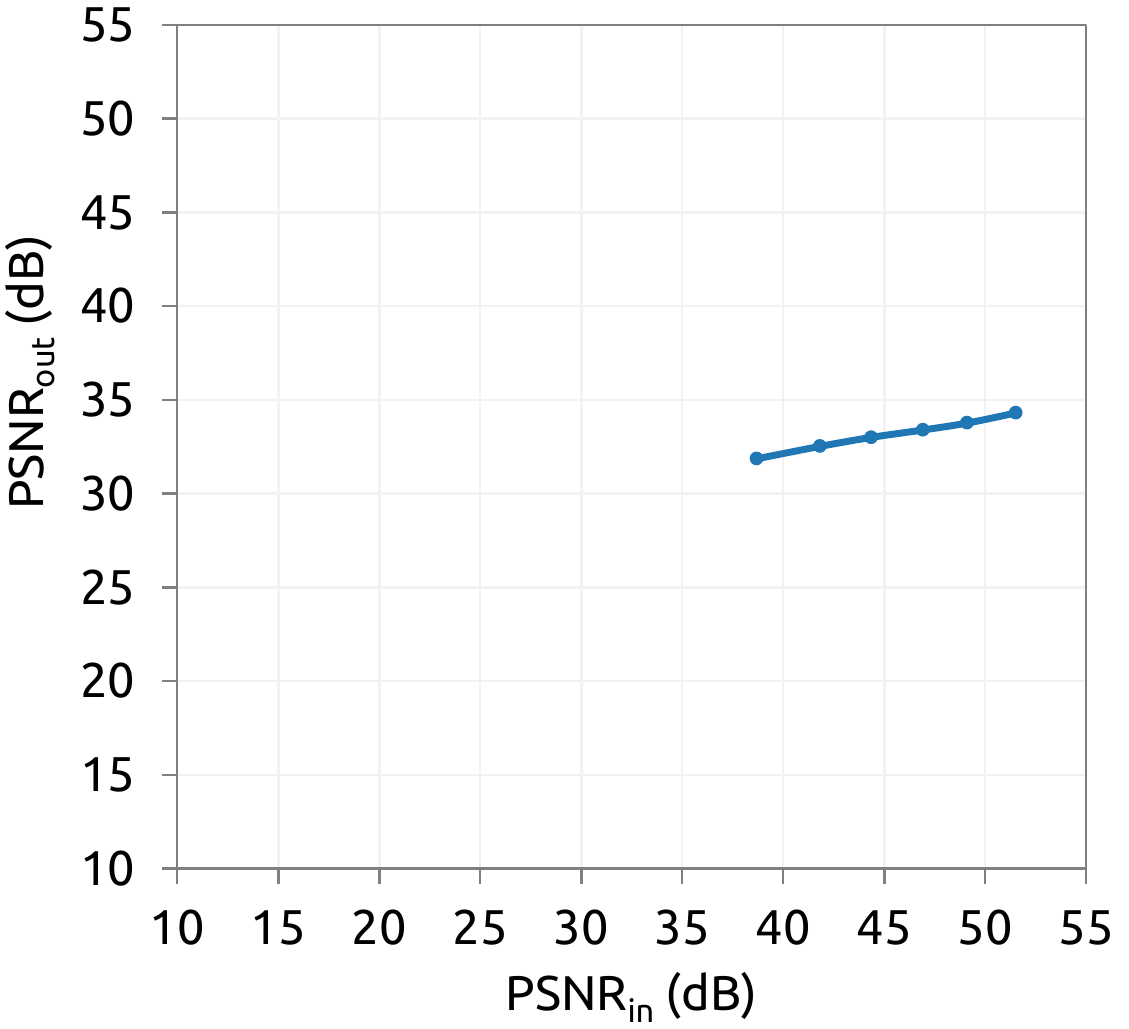} & \includegraphics[width=0.195\textwidth]{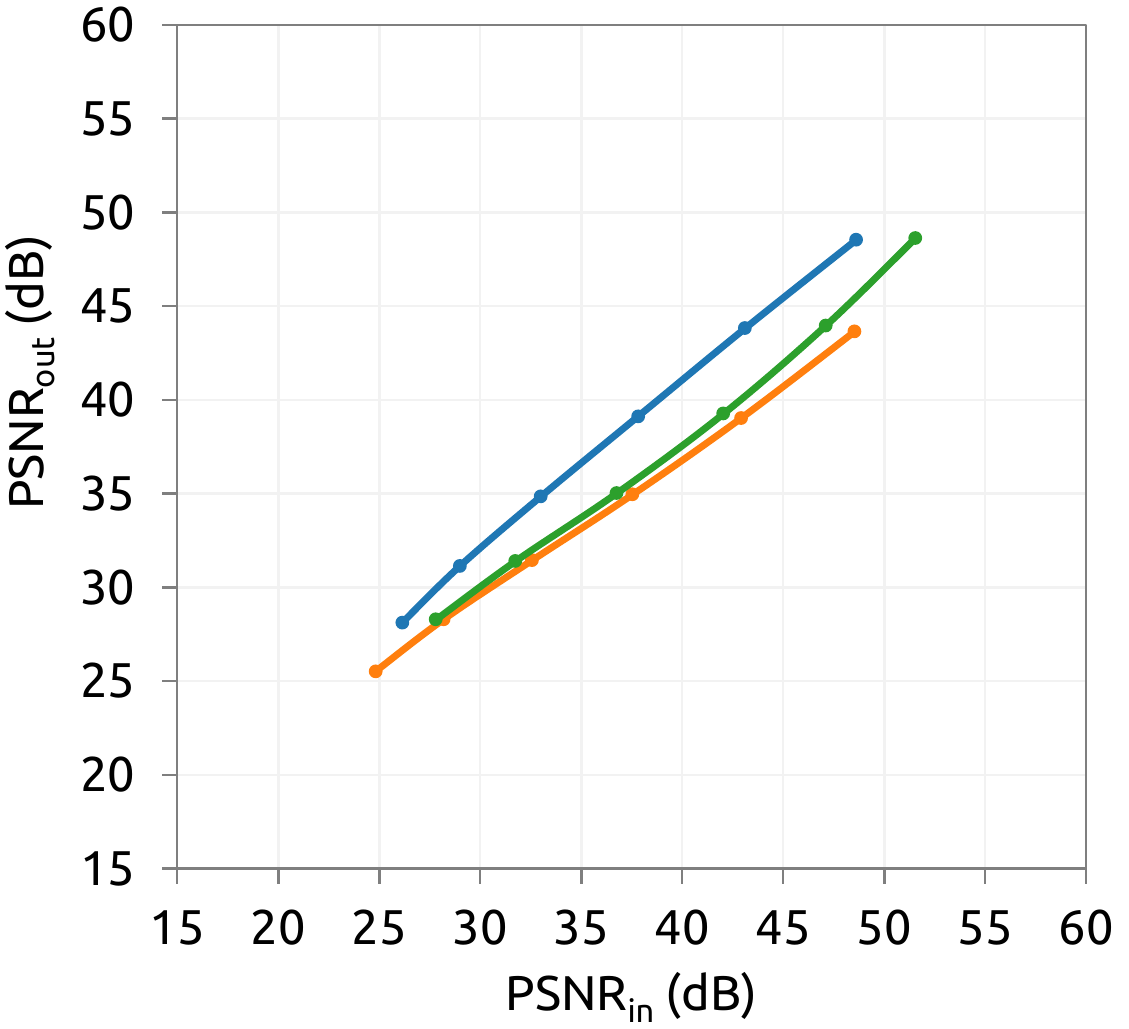} & \includegraphics[width=0.195\textwidth]{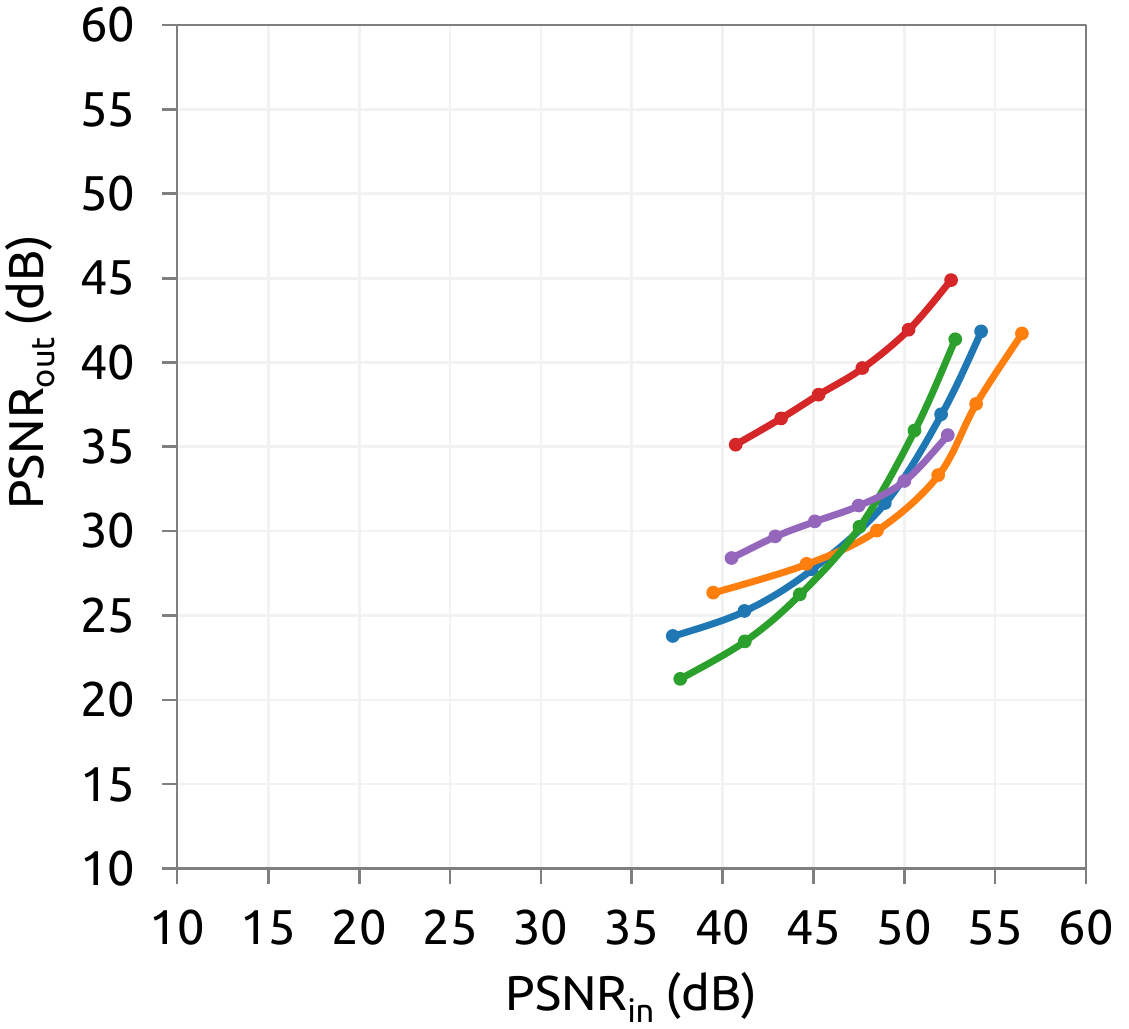} & \includegraphics[width=0.195\textwidth]{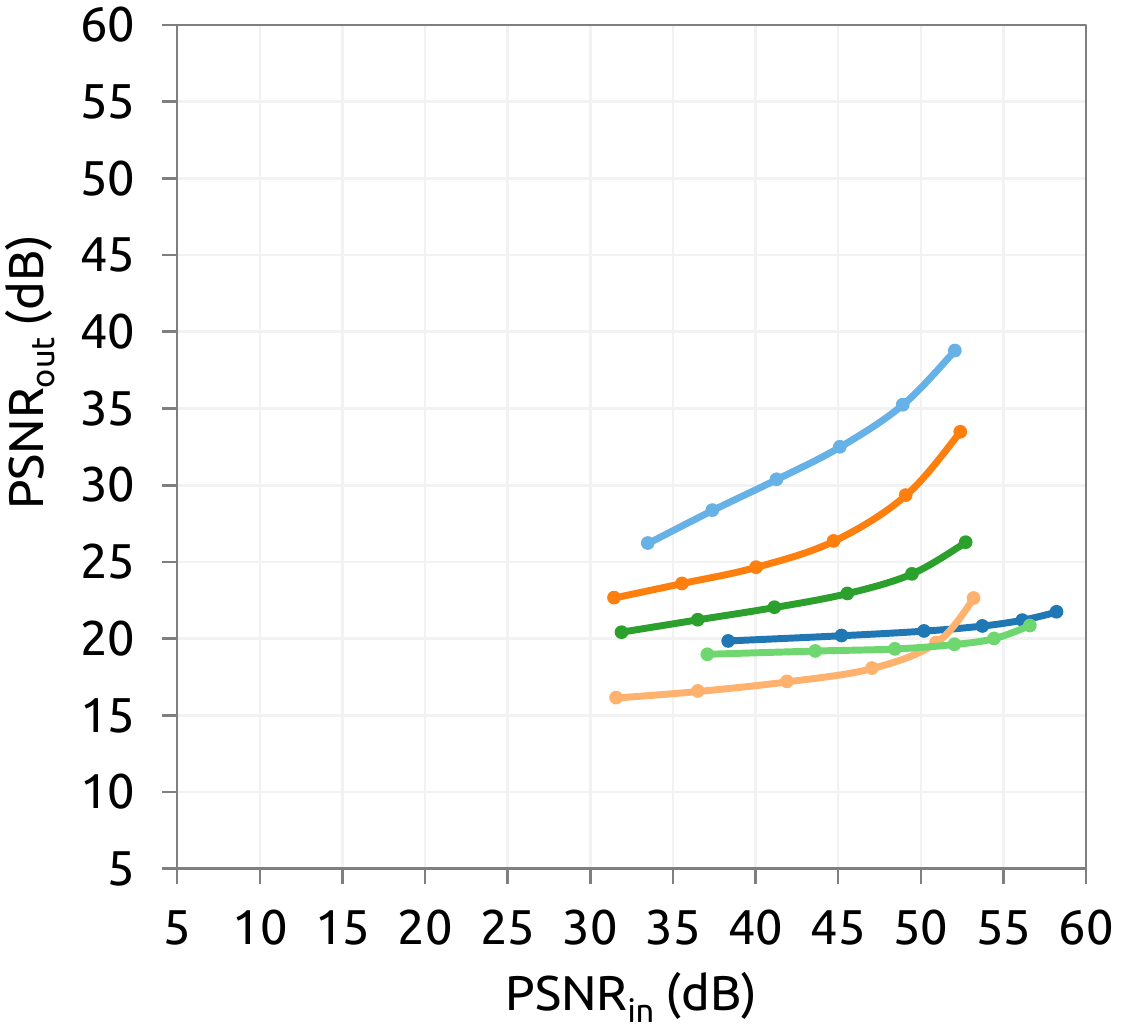} \\
			\multicolumn{5}{c}{\footnotesize{(b) FDA}} \\ \includegraphics[width=0.195\textwidth]{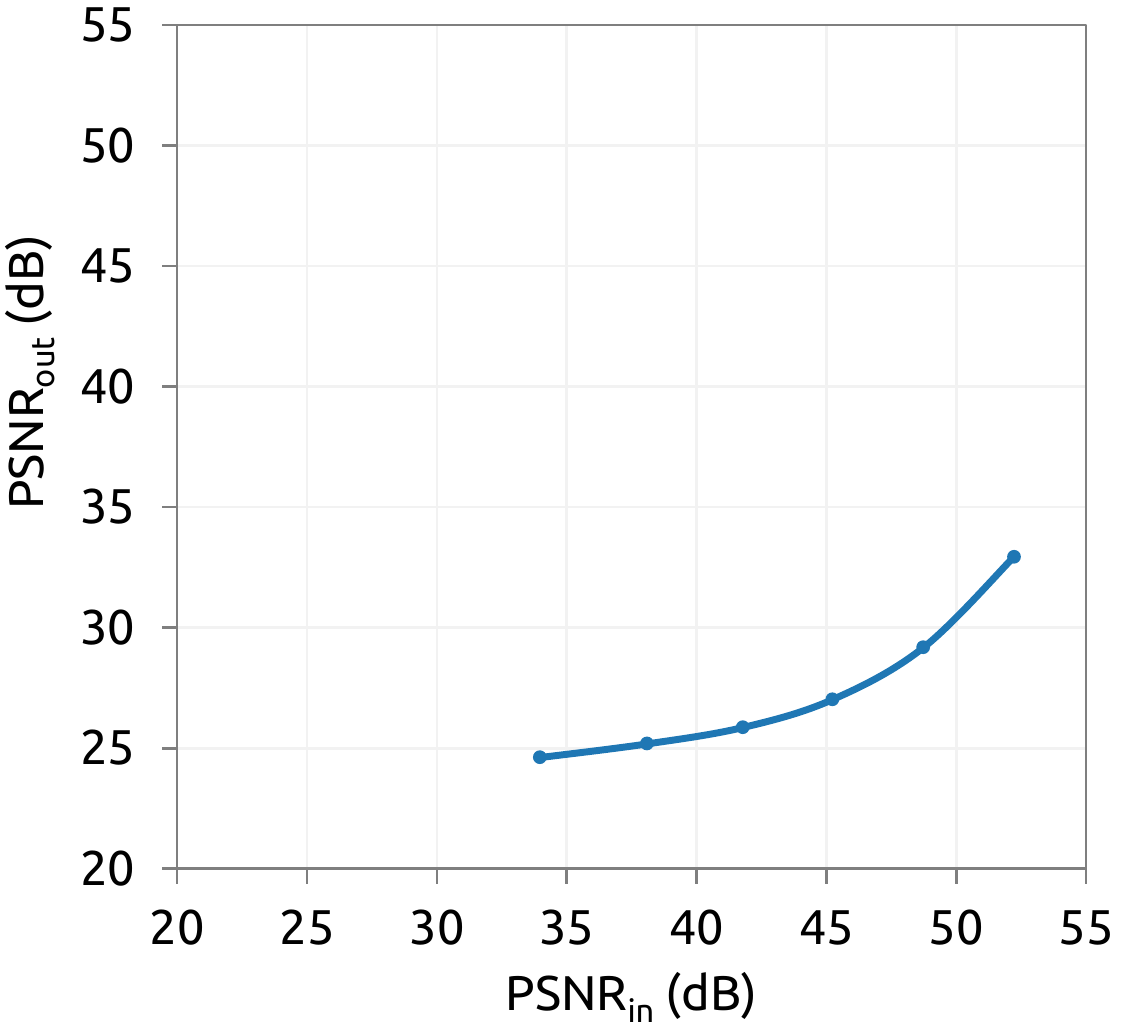} & \includegraphics[width=0.195\textwidth]{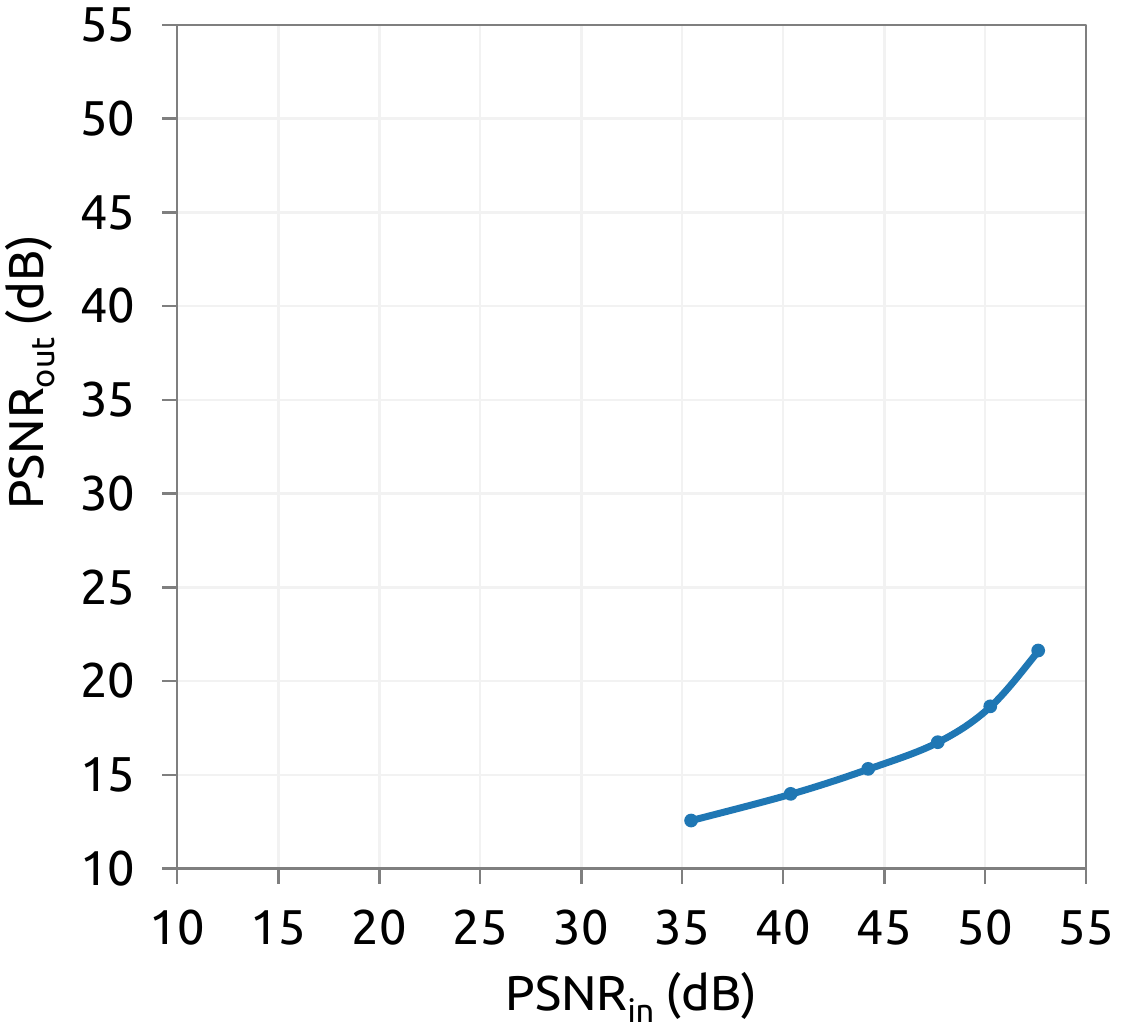} & \includegraphics[width=0.195\textwidth]{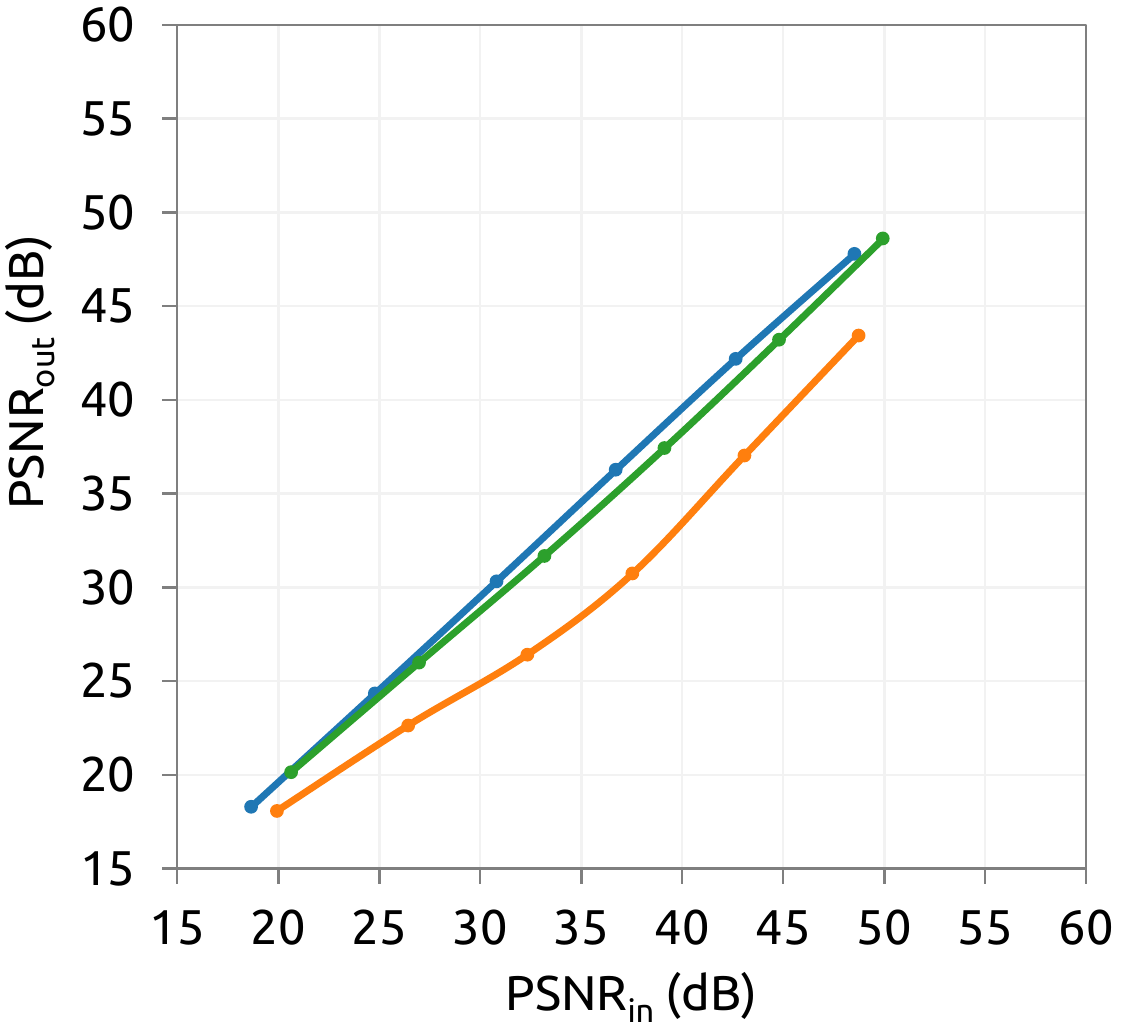} & \includegraphics[width=0.195\textwidth]{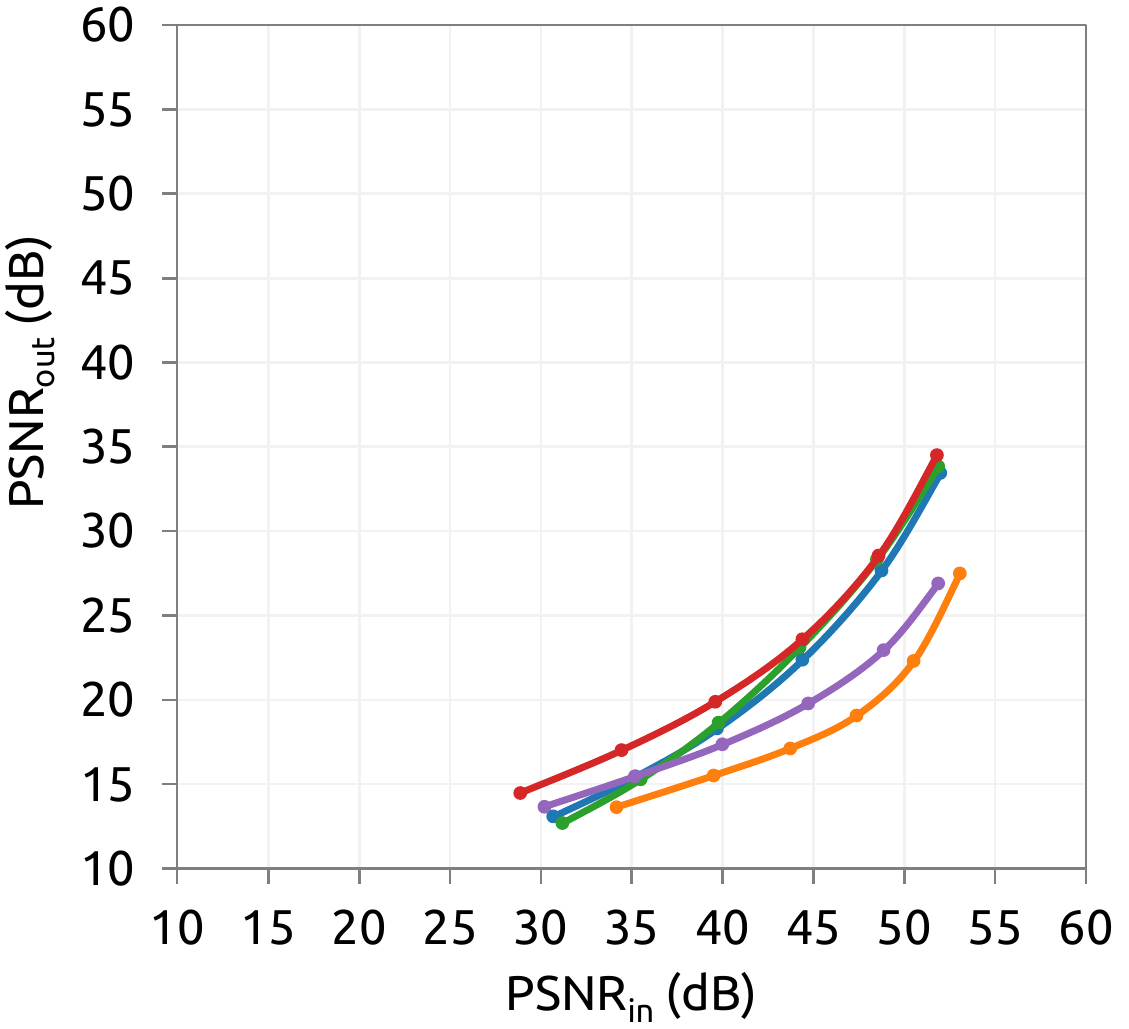} & \includegraphics[width=0.195\textwidth]{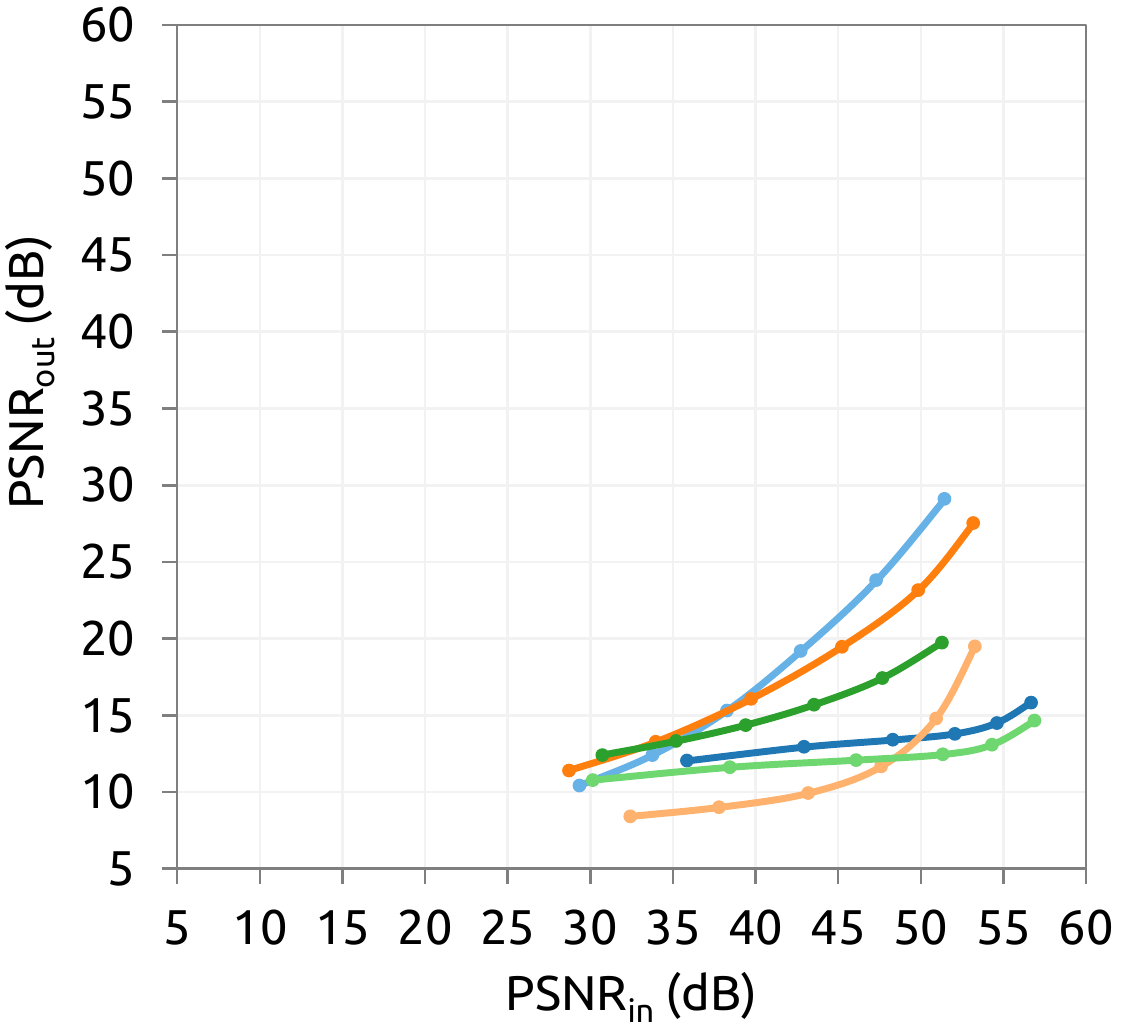} \\
			\multicolumn{5}{c}{\footnotesize{(c) I-FGSM}} \\
			\includegraphics[width=0.165\textwidth]{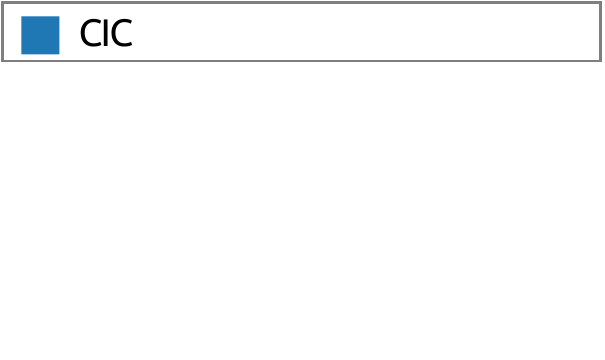} & \includegraphics[width=0.165\textwidth]{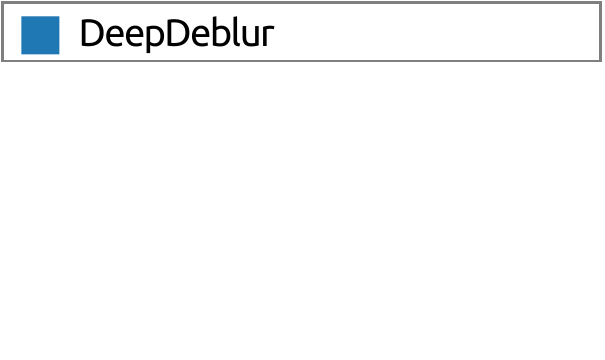} & \includegraphics[width=0.165\textwidth]{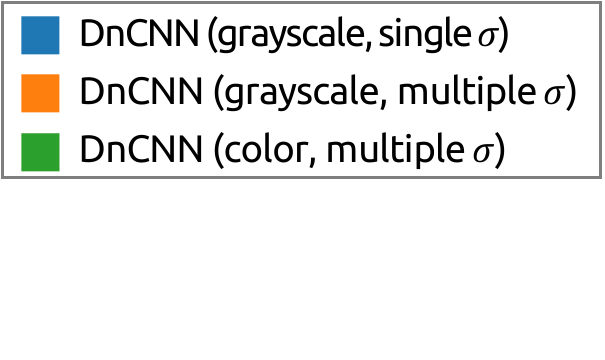} & \includegraphics[width=0.165\textwidth]{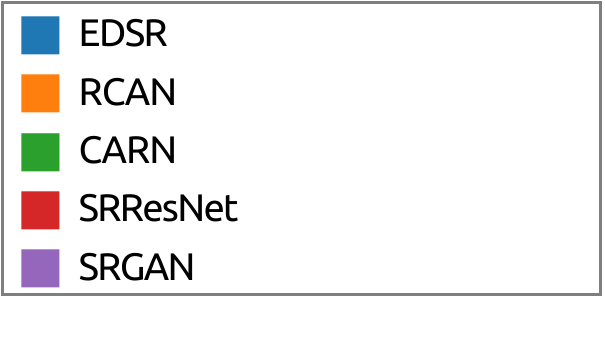} & \includegraphics[width=0.165\textwidth]{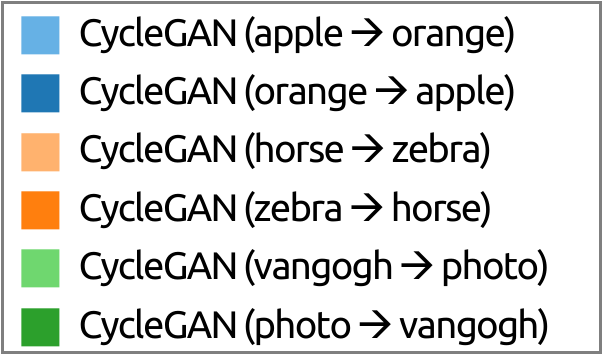}
		\end{tabular}
	\end{center}
	\caption{Performance comparison in terms of ${Q}_{i}$ and ${Q}_{o}$. Six points of each curve correspond to six different values of $\epsilon$.}
	\label{fig:basic_attack_psnr}
\end{figure*}

\begin{figure*}[t]
	\begin{center}
		\centering
		\renewcommand{\arraystretch}{1.0}
		\renewcommand{\tabcolsep}{1.5pt}
		\footnotesize
		\begin{tabular}{ccccccccc}
			\makecell[c]{Input (original)} & \makecell[c]{Output (original)} & \makecell[c]{Input (FDA)} & \makecell[c]{Output (FDA)} & ~ & \makecell[c]{Input (original)} & \makecell[c]{Output (original)} & \makecell[c]{Input (FDA)} & \makecell[c]{Output (FDA)}\\
			\includegraphics[width=0.113\linewidth]{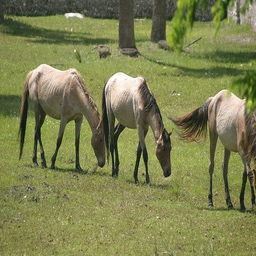} &
			\includegraphics[width=0.113\linewidth]{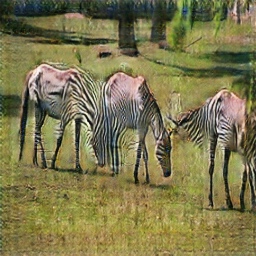} &
			\includegraphics[width=0.113\linewidth]{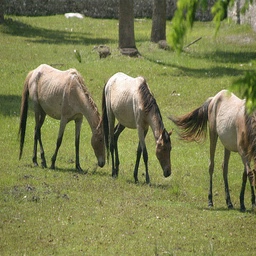} &
			\includegraphics[width=0.113\linewidth]{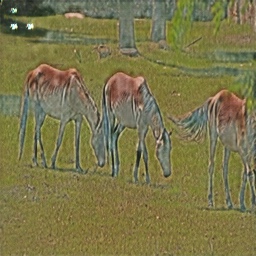} & &
			\includegraphics[width=0.113\linewidth]{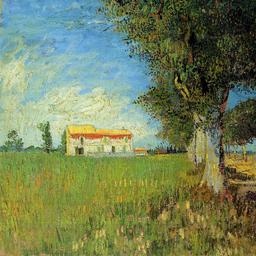} &
			\includegraphics[width=0.113\linewidth]{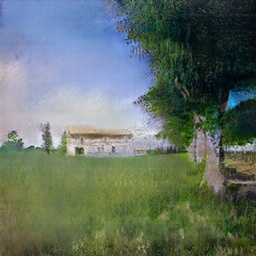} &
			\includegraphics[width=0.113\linewidth]{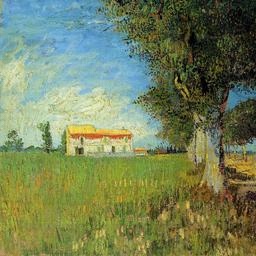} &
			\includegraphics[width=0.113\linewidth]{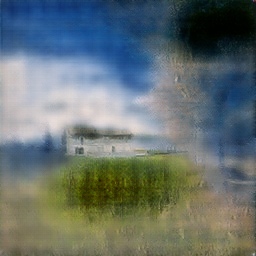}\\
			\multicolumn{4}{c}{\footnotesize{(a) Horse $\rightarrow$ Zebra}} & &
			\multicolumn{4}{c}{\footnotesize{(b) Van Gogh $\rightarrow$ Photo}}
		\end{tabular}
	\end{center}
	\caption{Images obtained from CycleGAN trained using different datasets. FDA is employed with $\epsilon=8$.}
	\label{fig:basic_attack_datasets_appendix}
\end{figure*}

\begin{figure*}[t]
	\begin{center}
		\centering
		\renewcommand{\arraystretch}{1.5}
		\renewcommand{\tabcolsep}{2.0pt}
		\footnotesize
		\begin{tabular}{cccc}
			\raisebox{-0.5\height}{\includegraphics[width=0.250\linewidth]{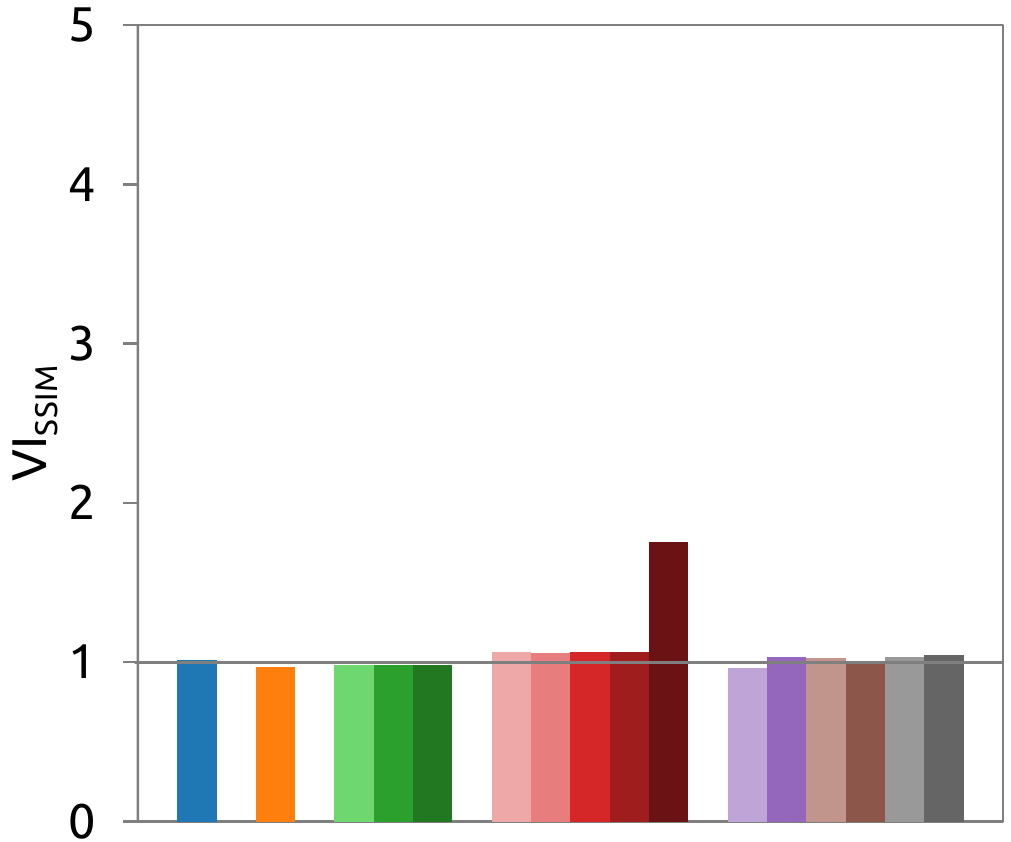}} &
			\raisebox{-0.5\height}{\includegraphics[width=0.250\linewidth]{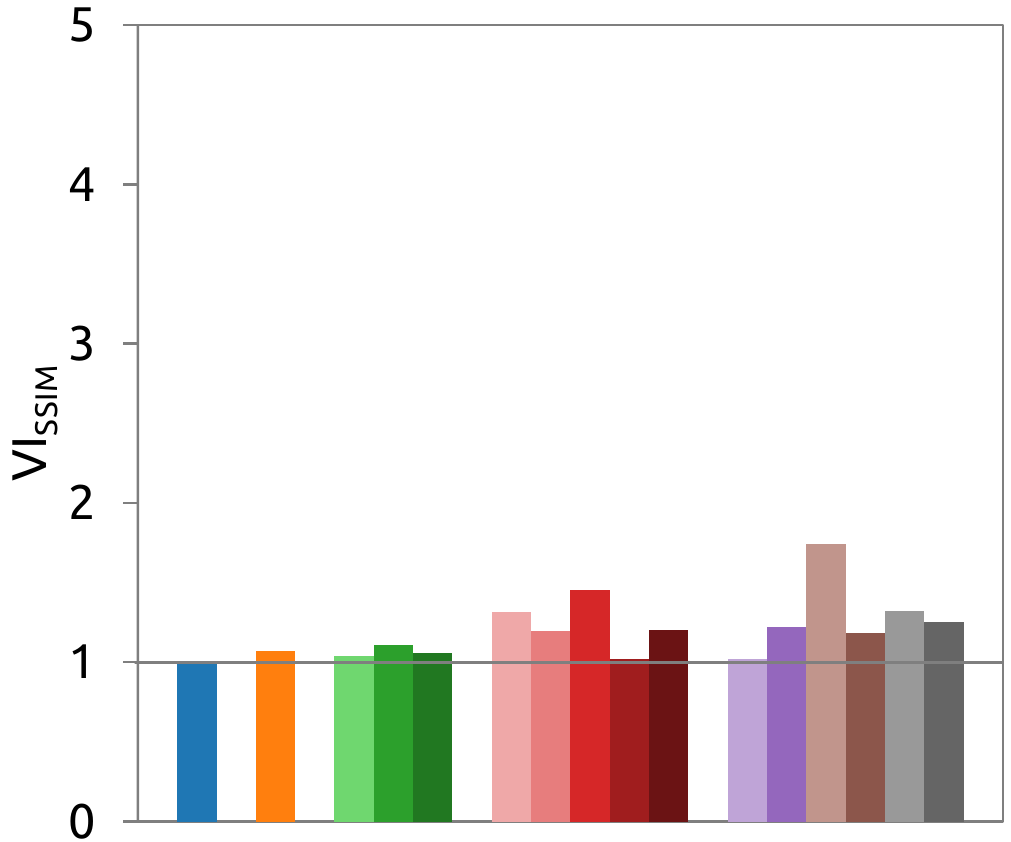}} &
			\raisebox{-0.5\height}{\includegraphics[width=0.250\linewidth]{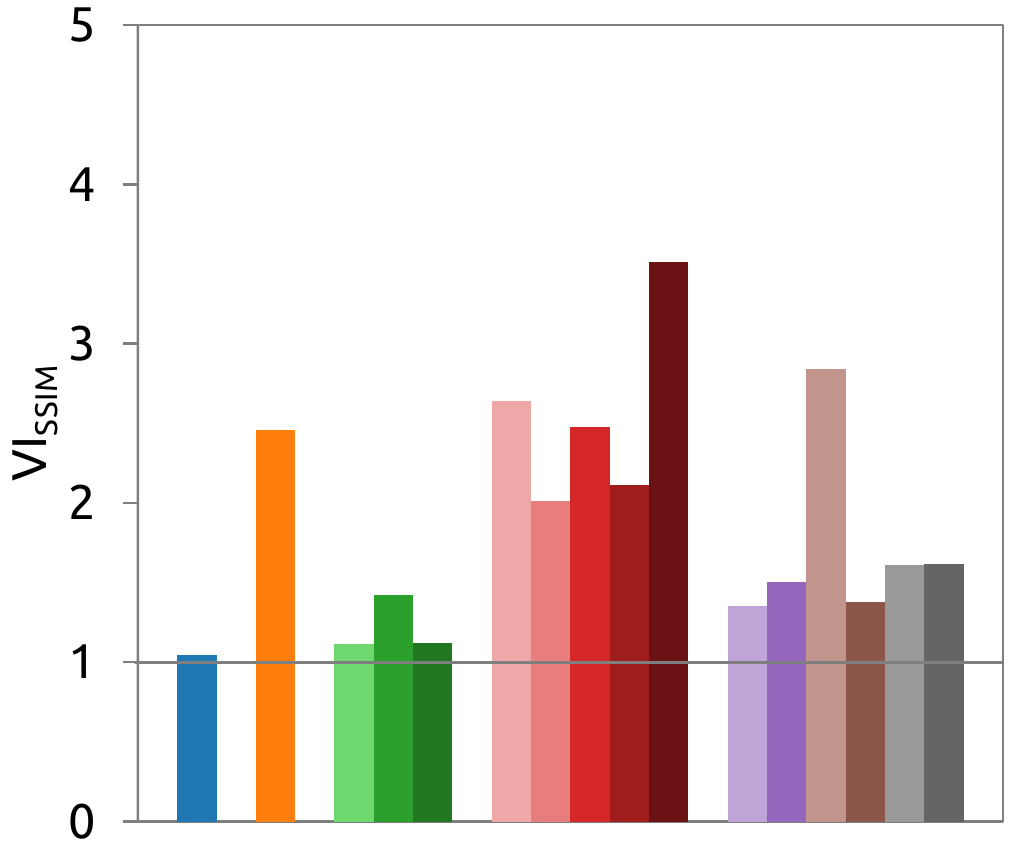}} &
			\multirow{2}{*}[41.5pt]{\raisebox{-0.5\height}{\includegraphics[width=0.135\linewidth]{figures/basic_vi_legend}}}\\
			(a) Random noise & (b) FDA & (c) I-FGSM &
		\end{tabular}
	\end{center}
	\caption{Performance comparison in terms of $\mathrm{VI}_{\mathrm{SSIM}}$ for different attack methods employed with $\epsilon=8$.}
	\label{fig:basic_attack_ssimratio}
\end{figure*}

\begin{figure*}[t]
	\begin{center}
		\centering
		\renewcommand{\arraystretch}{1.5}
		\renewcommand{\tabcolsep}{1.2pt}
		\footnotesize
		\begin{tabular}{ccccc}
			Colorization & Deblurring & Denoising & Super-resolution & Translation \\ \includegraphics[width=0.195\textwidth]{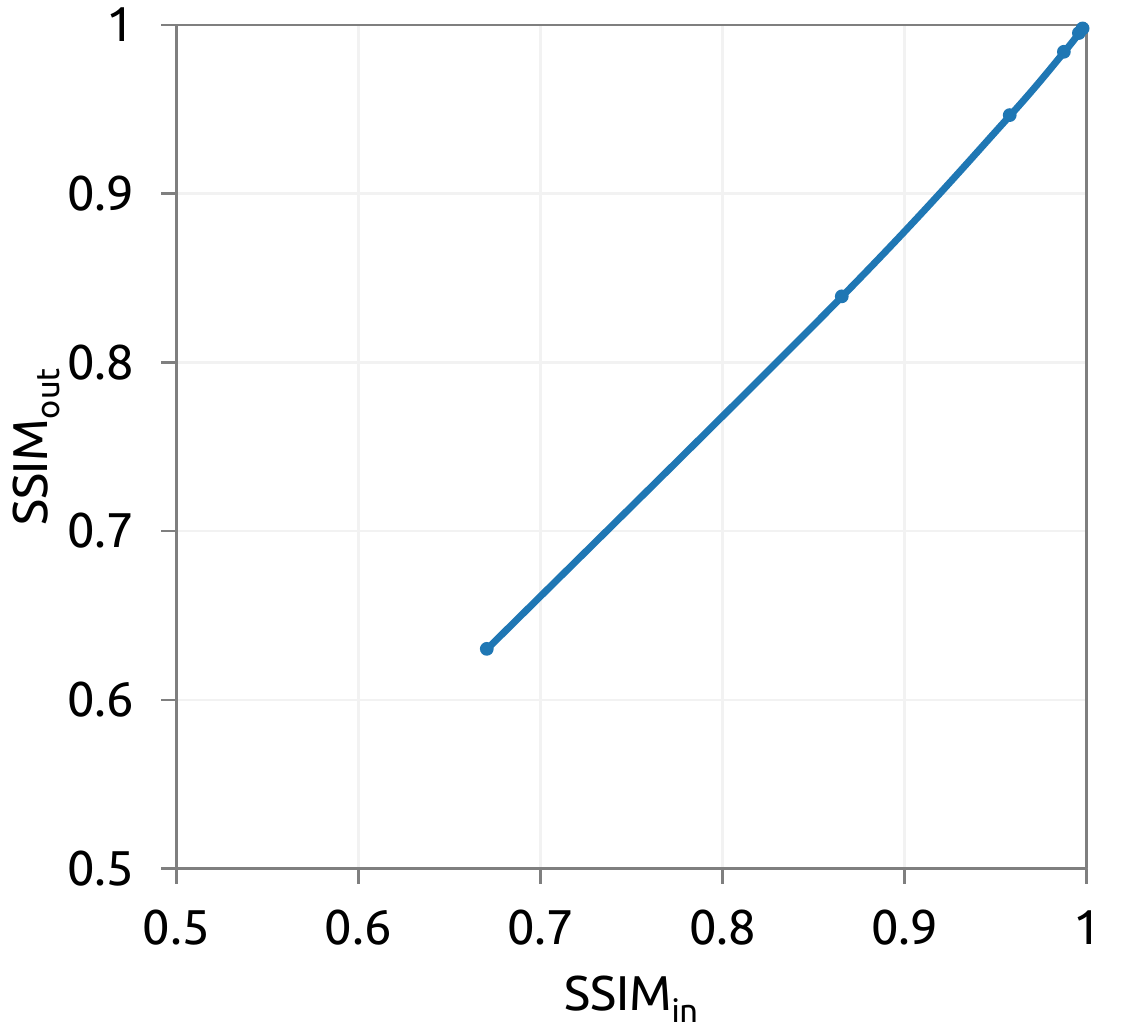} & \includegraphics[width=0.195\textwidth]{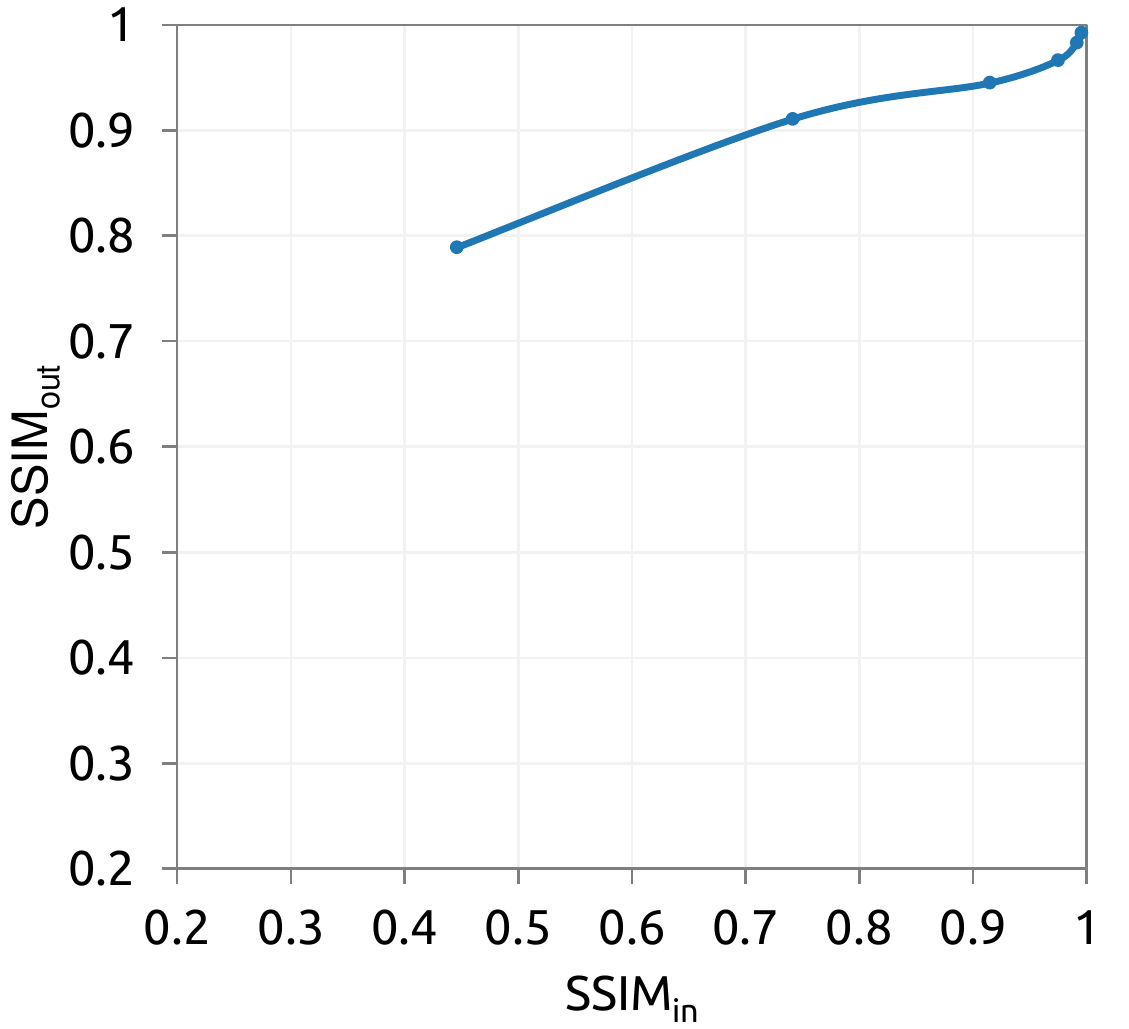} & \includegraphics[width=0.195\textwidth]{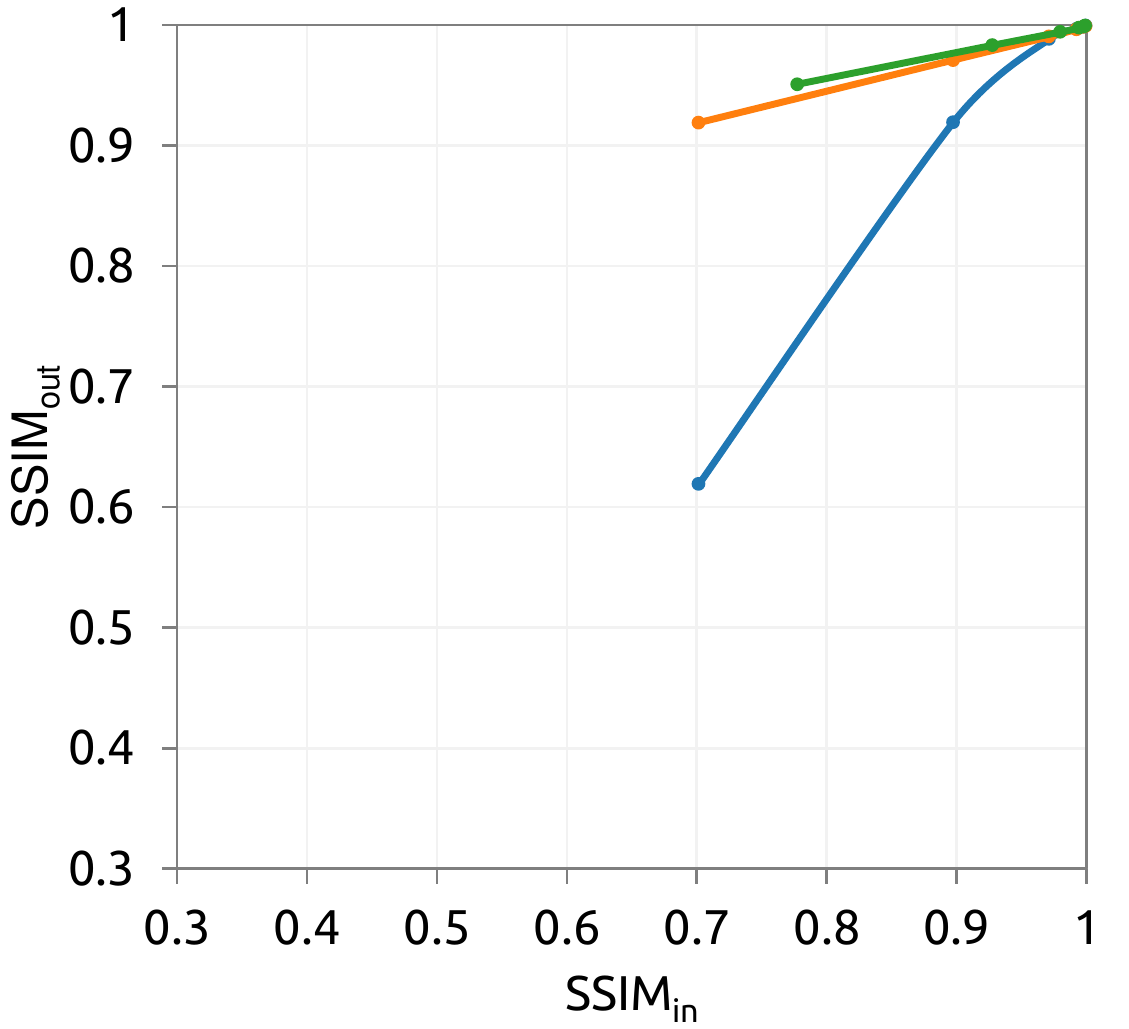} & \includegraphics[width=0.195\textwidth]{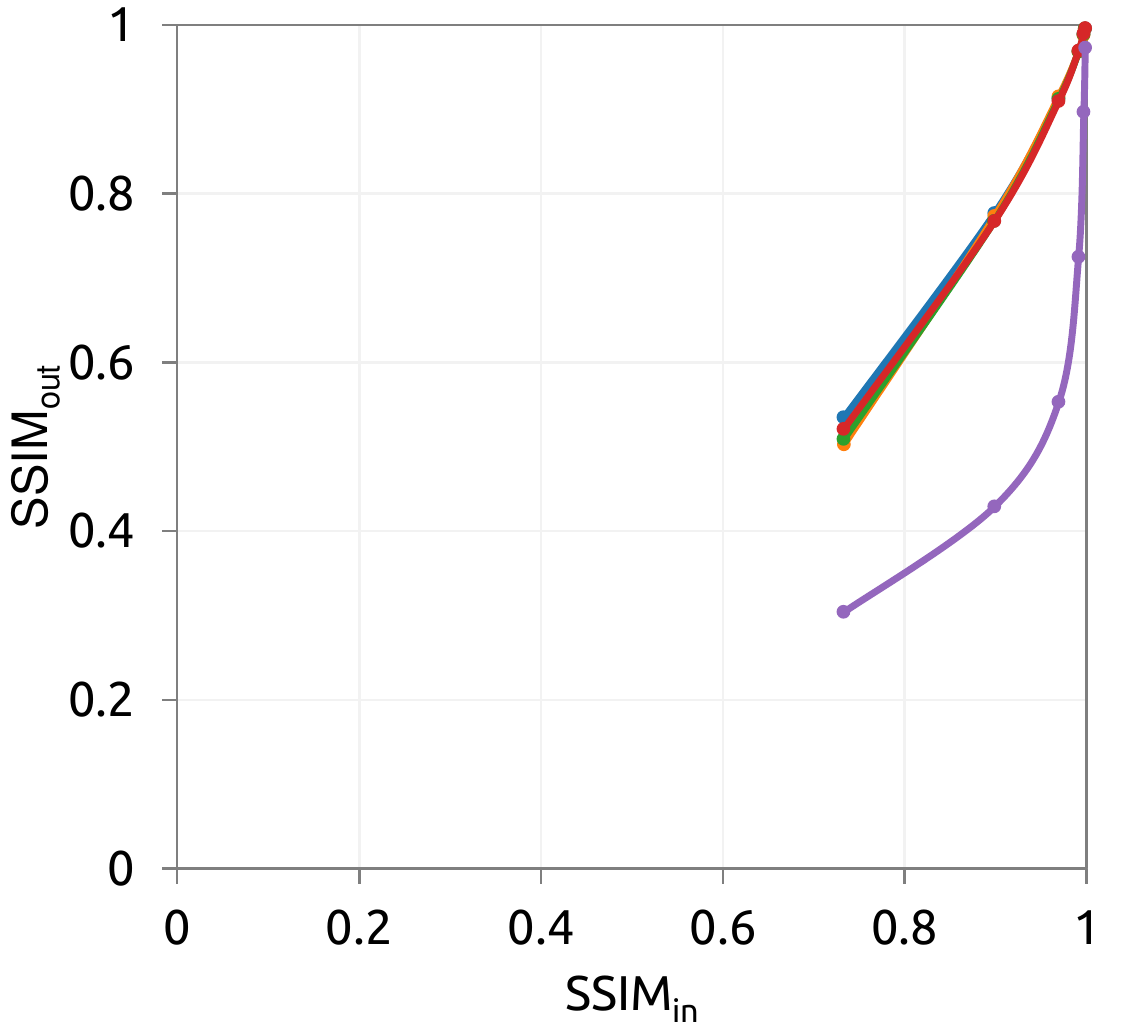} & \includegraphics[width=0.195\textwidth]{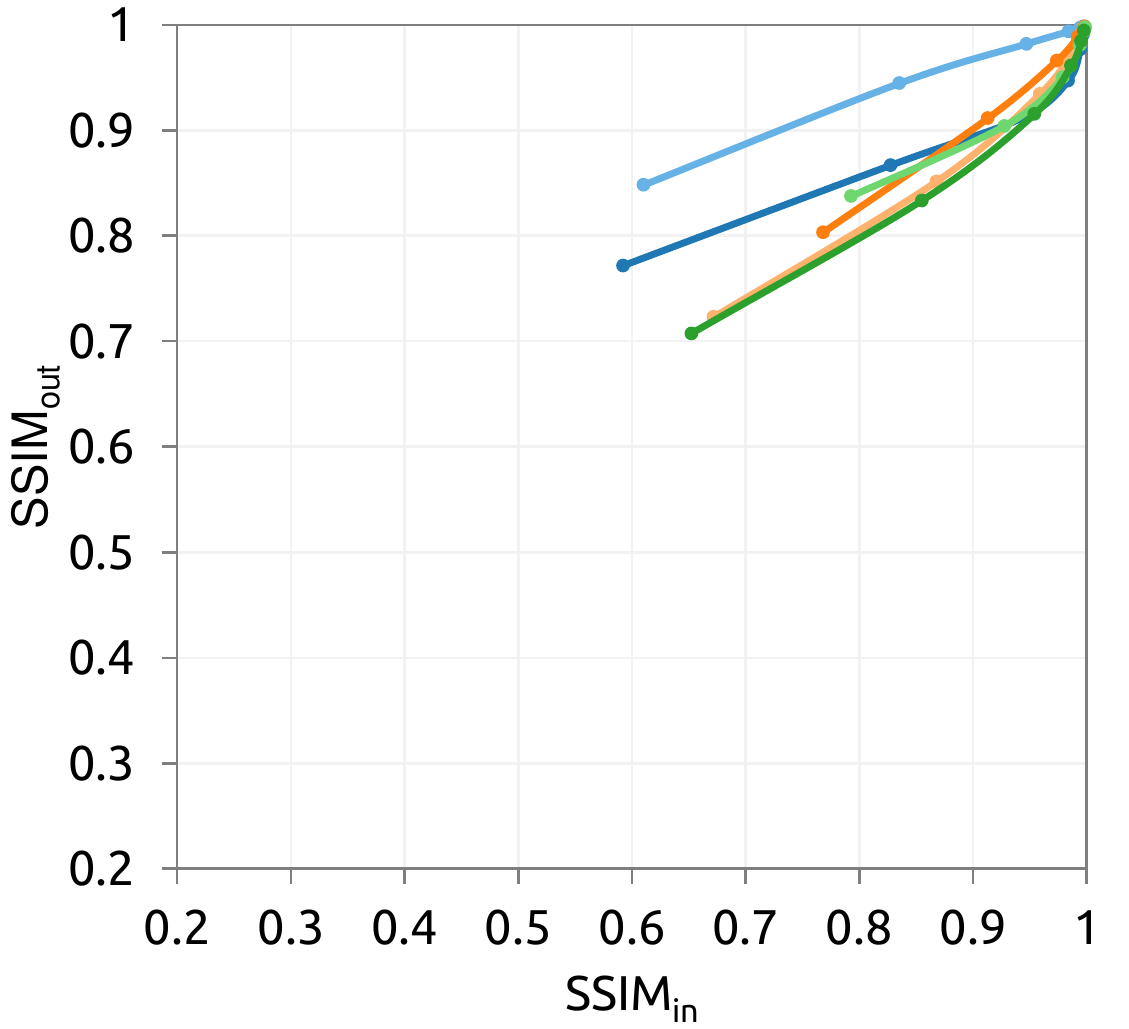} \\
			\multicolumn{5}{c}{\footnotesize{(a) Random uniform}} \\
			\includegraphics[width=0.195\textwidth]{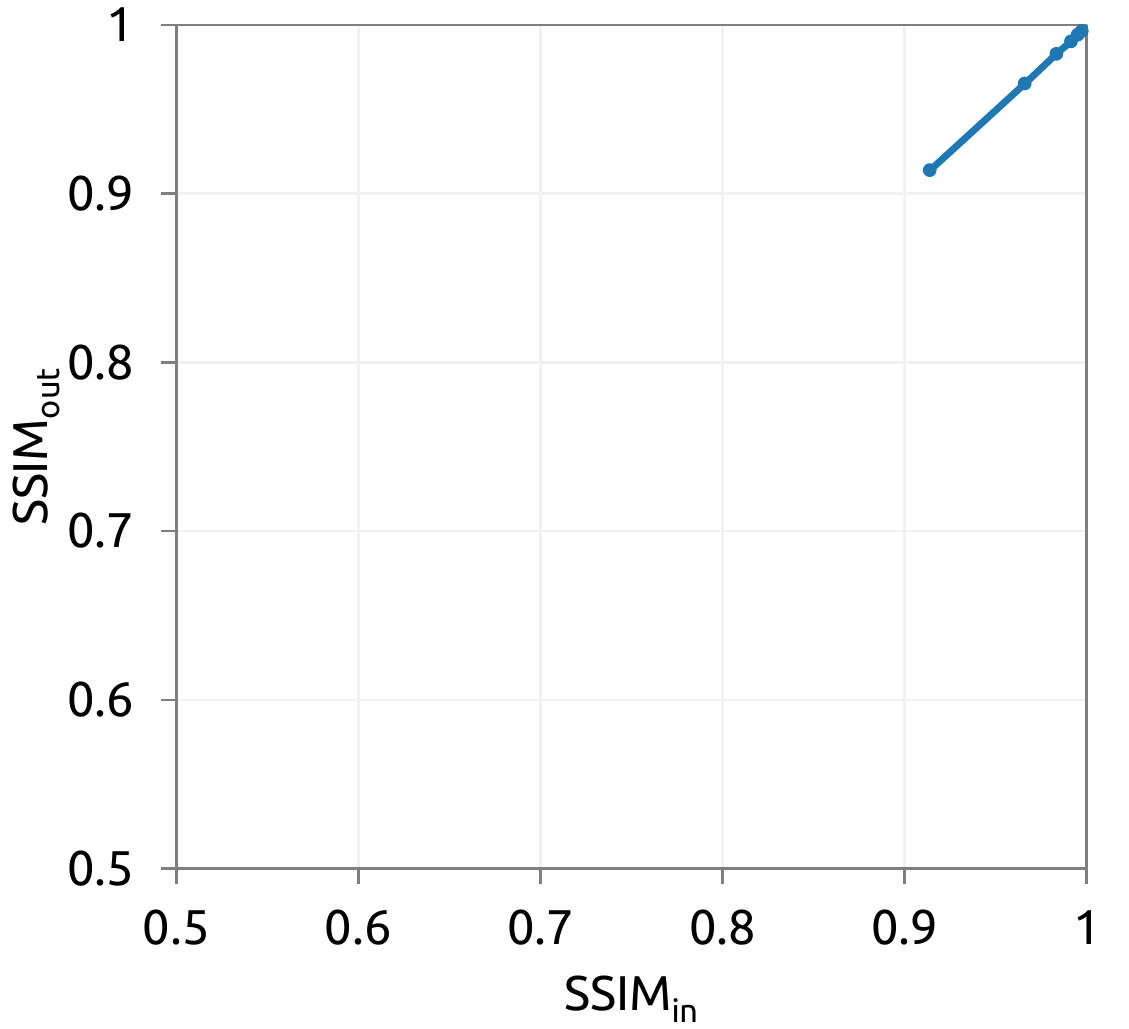} & \includegraphics[width=0.195\textwidth]{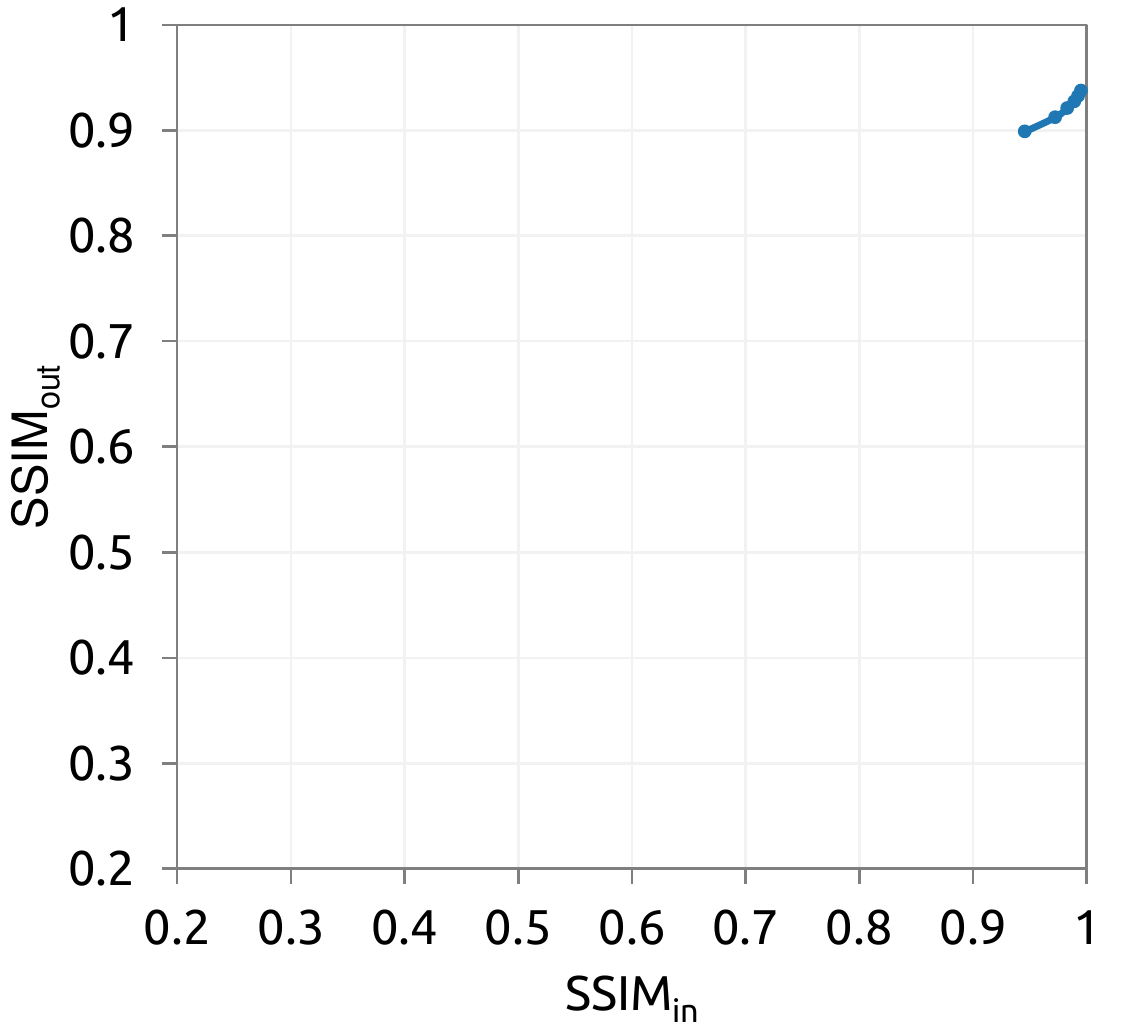} & \includegraphics[width=0.195\textwidth]{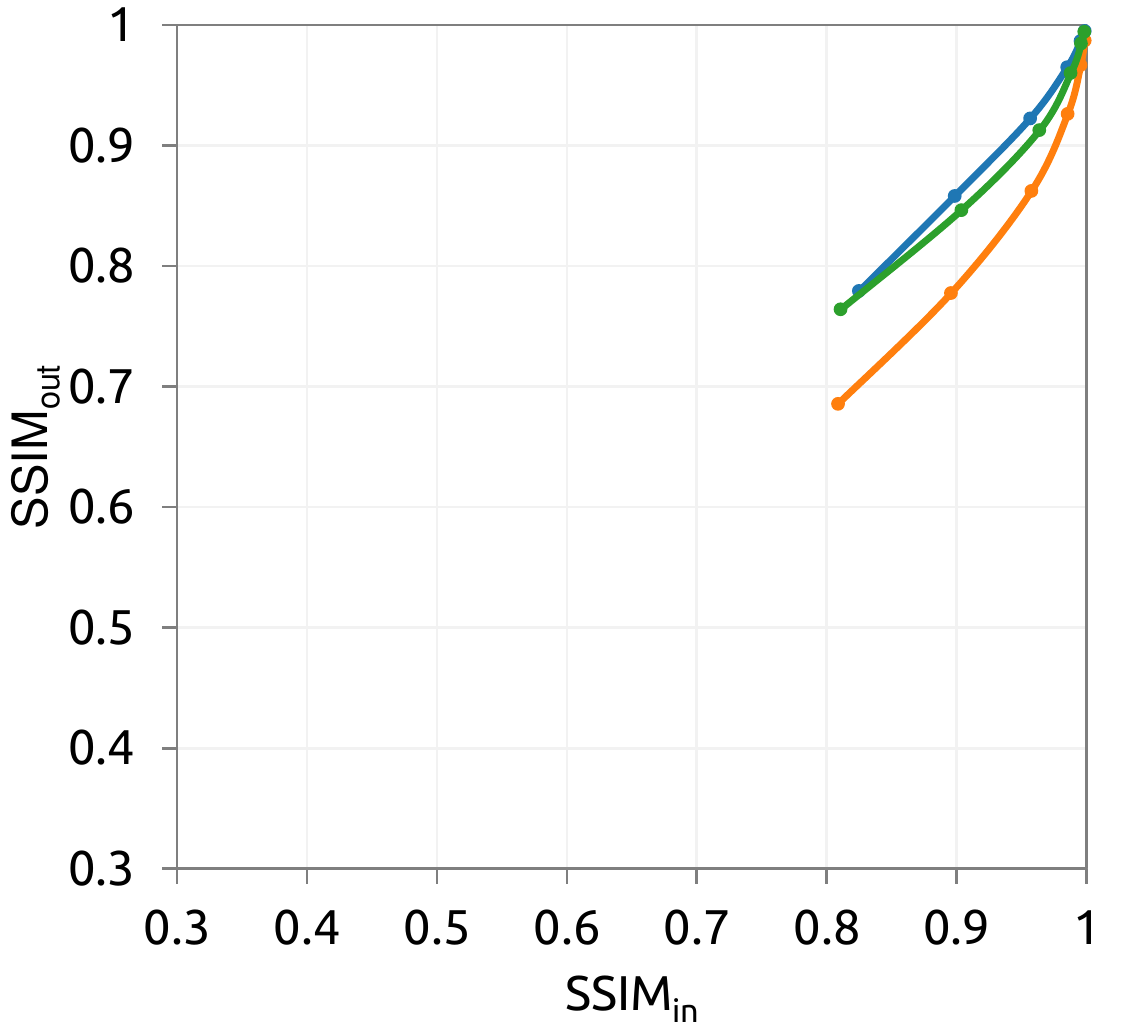} & \includegraphics[width=0.195\textwidth]{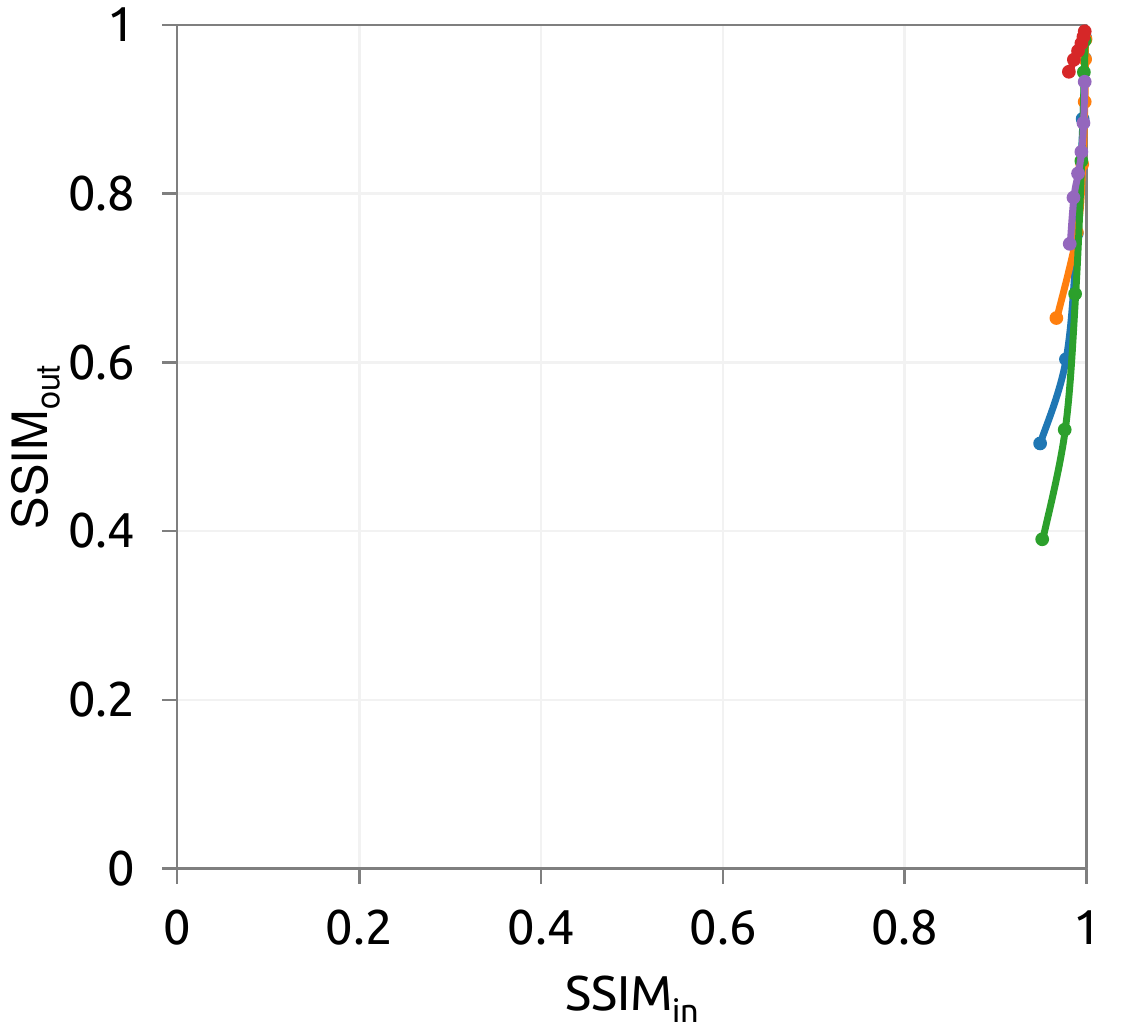} & \includegraphics[width=0.195\textwidth]{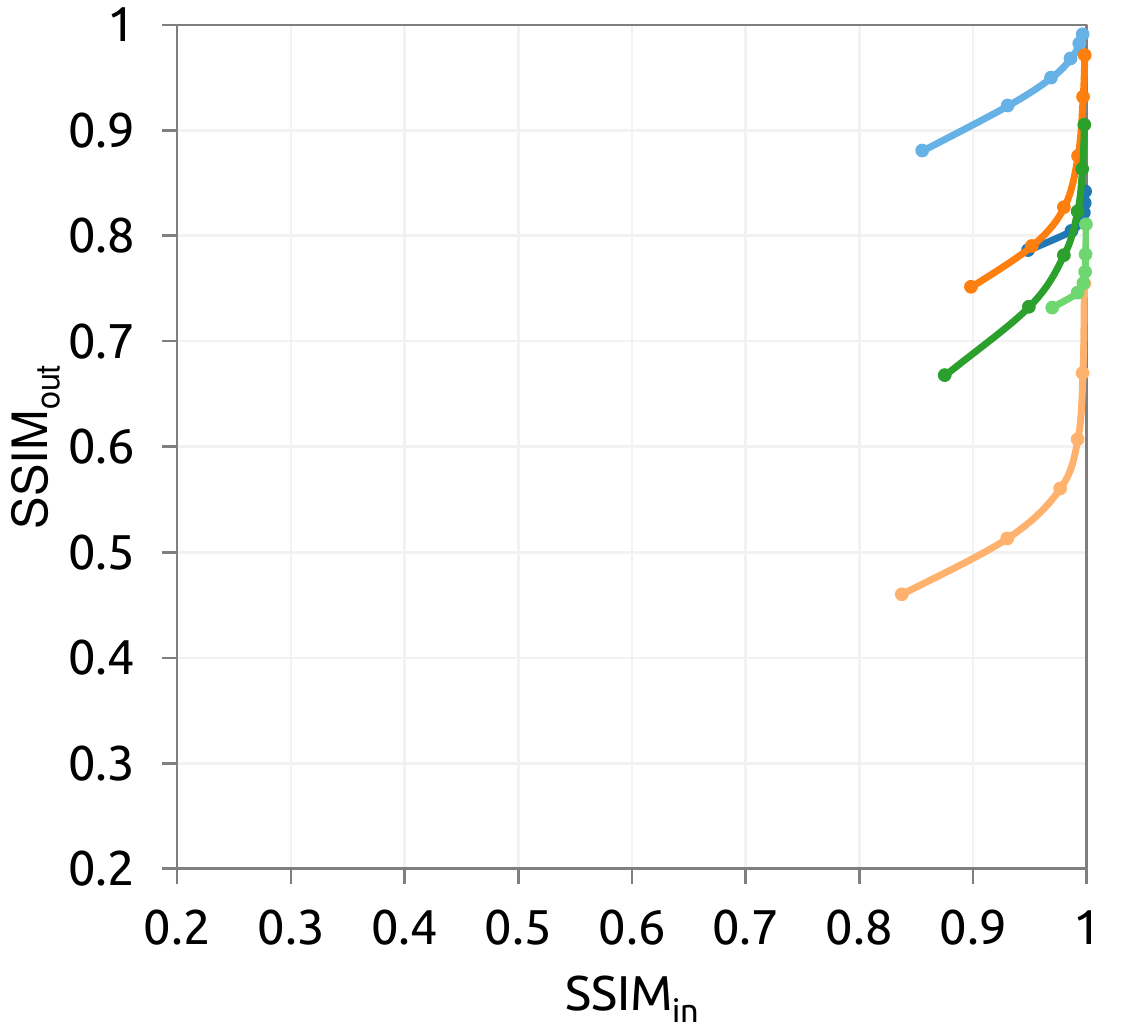} \\
			\multicolumn{5}{c}{\footnotesize{(b) FDA}} \\ \includegraphics[width=0.195\textwidth]{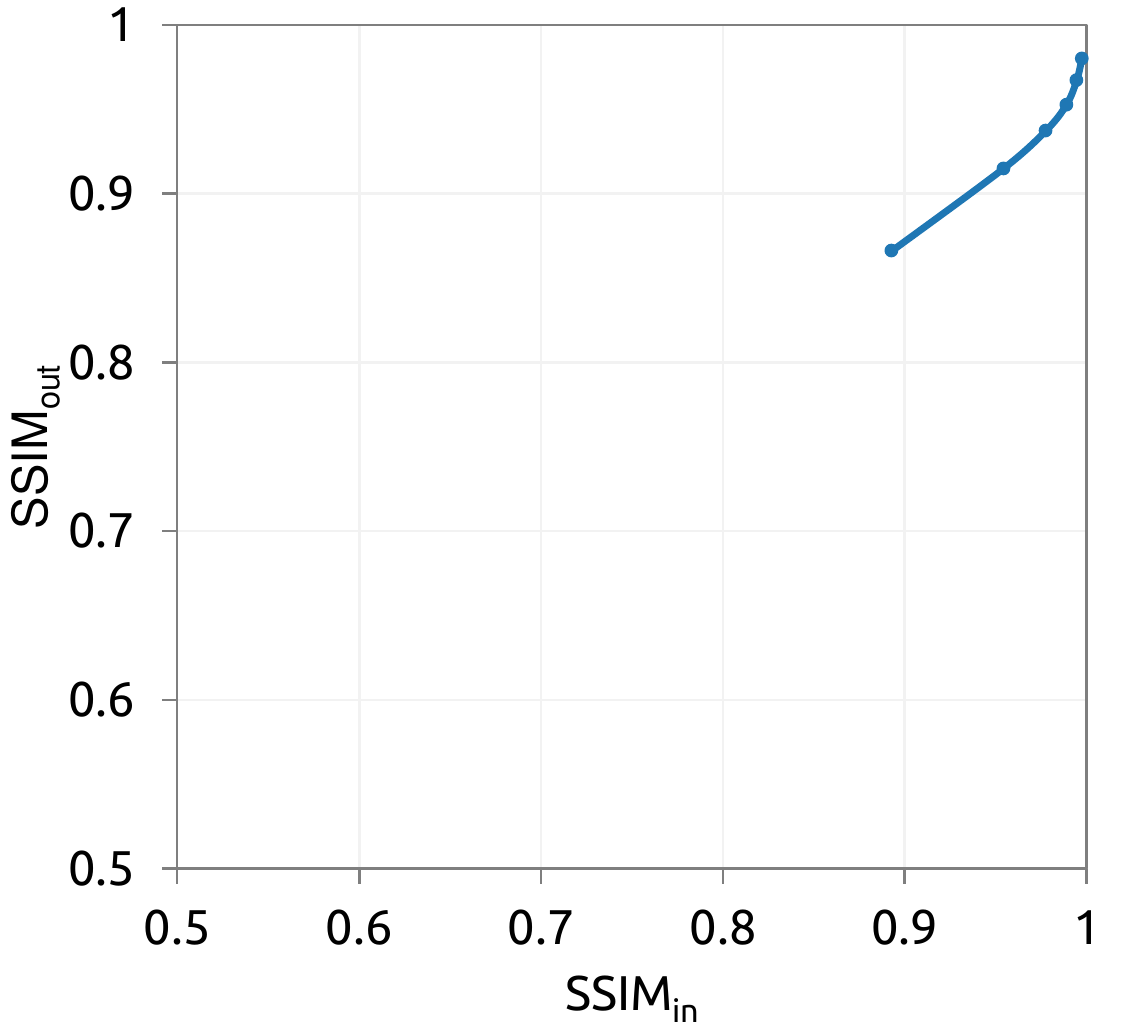} & \includegraphics[width=0.195\textwidth]{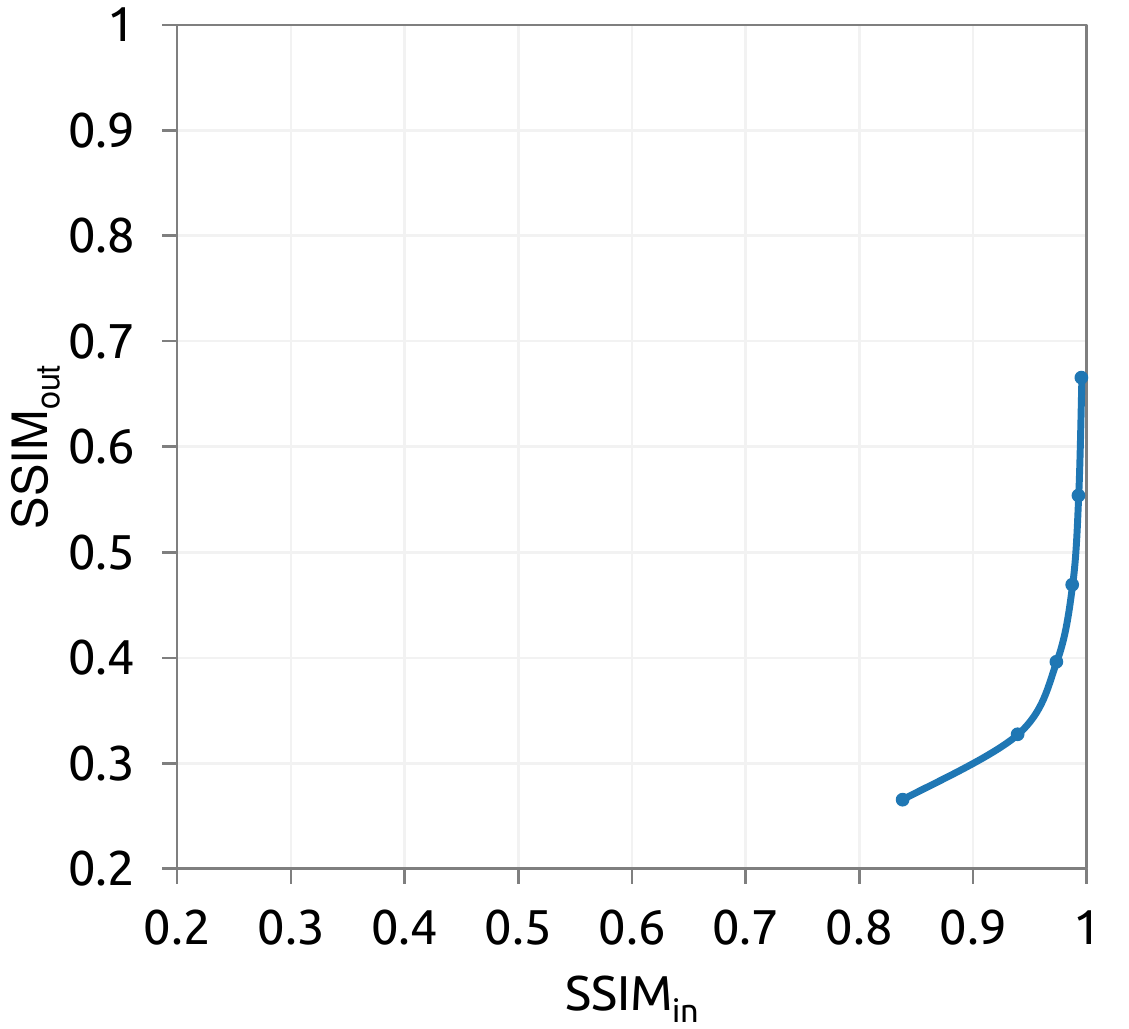} & \includegraphics[width=0.195\textwidth]{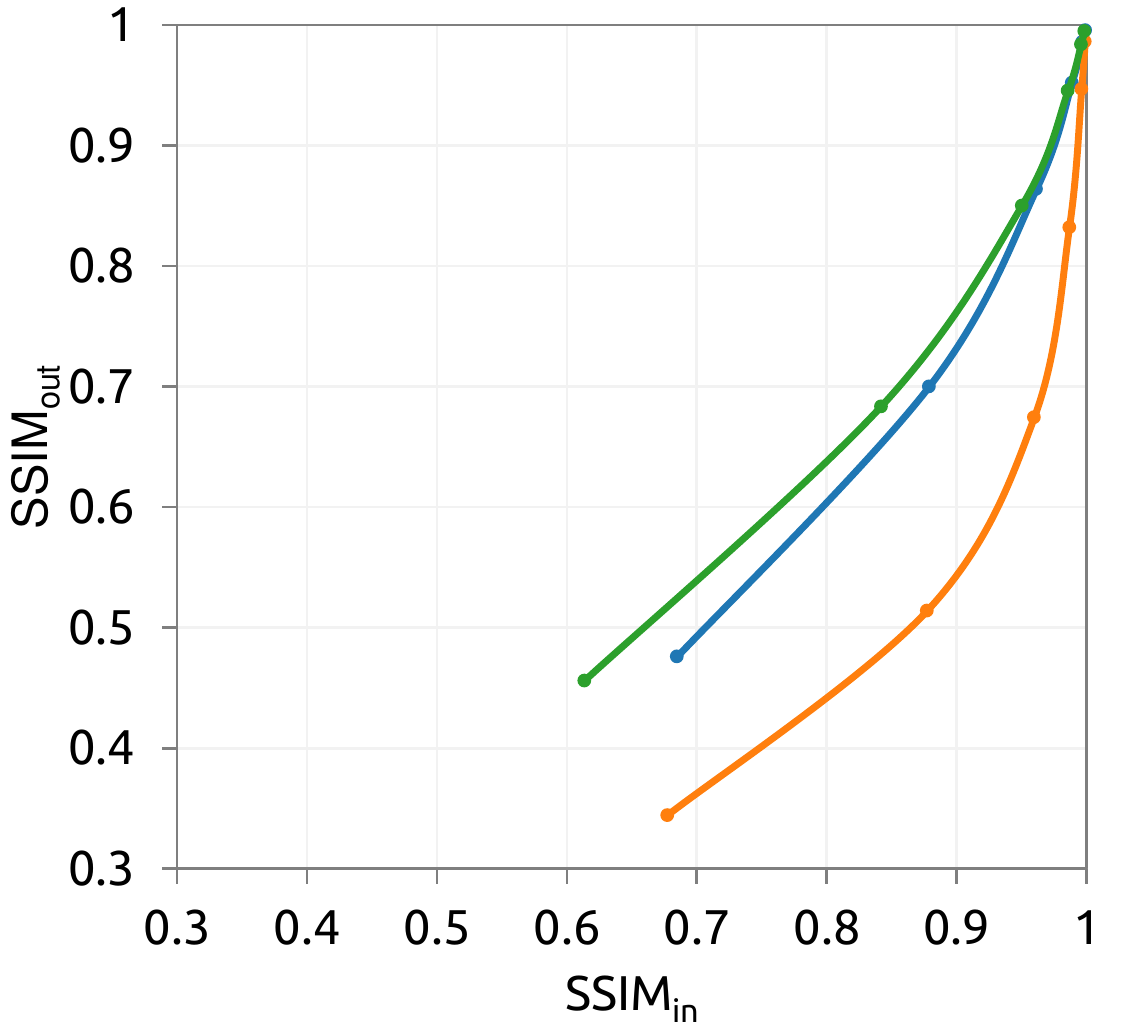} & \includegraphics[width=0.195\textwidth]{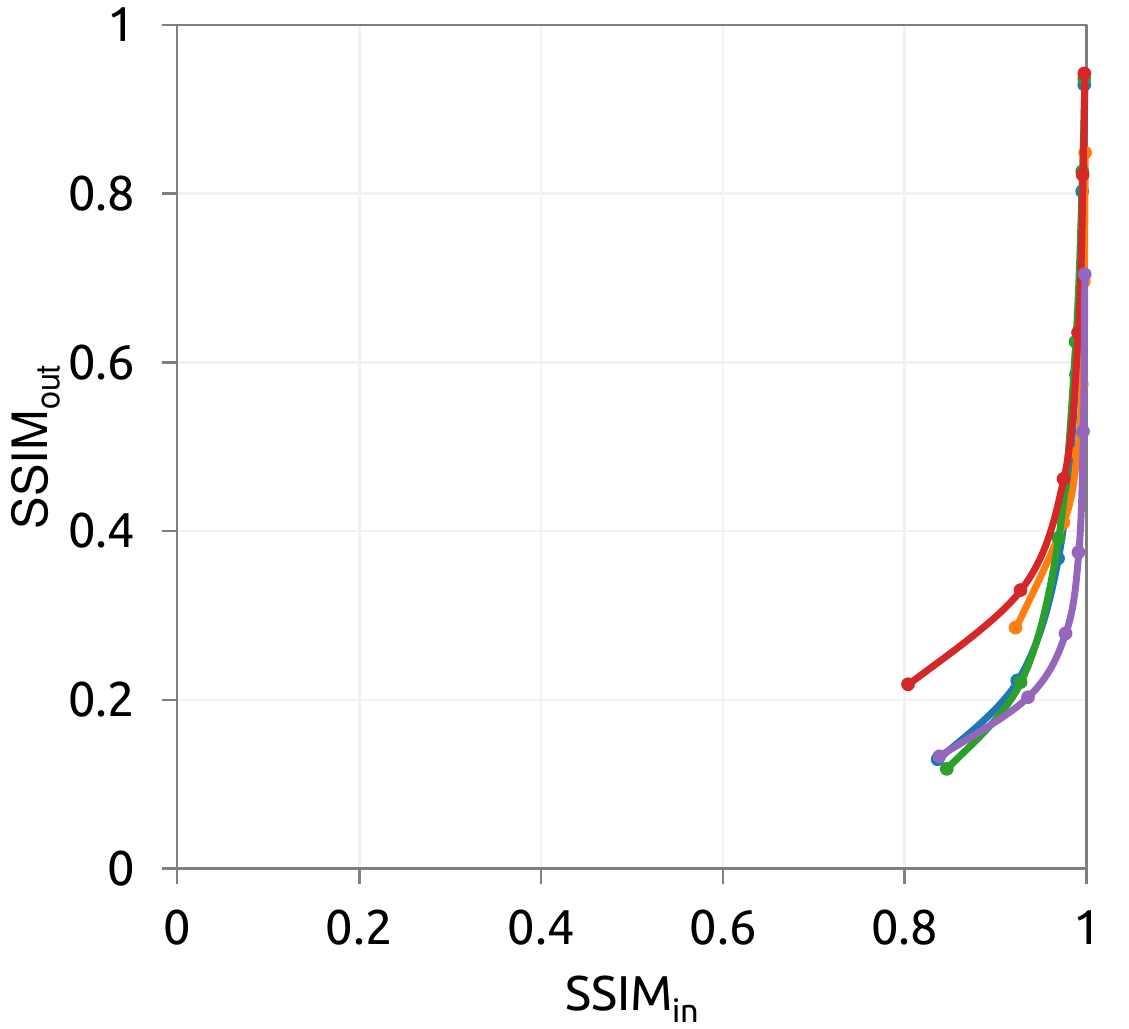} & \includegraphics[width=0.195\textwidth]{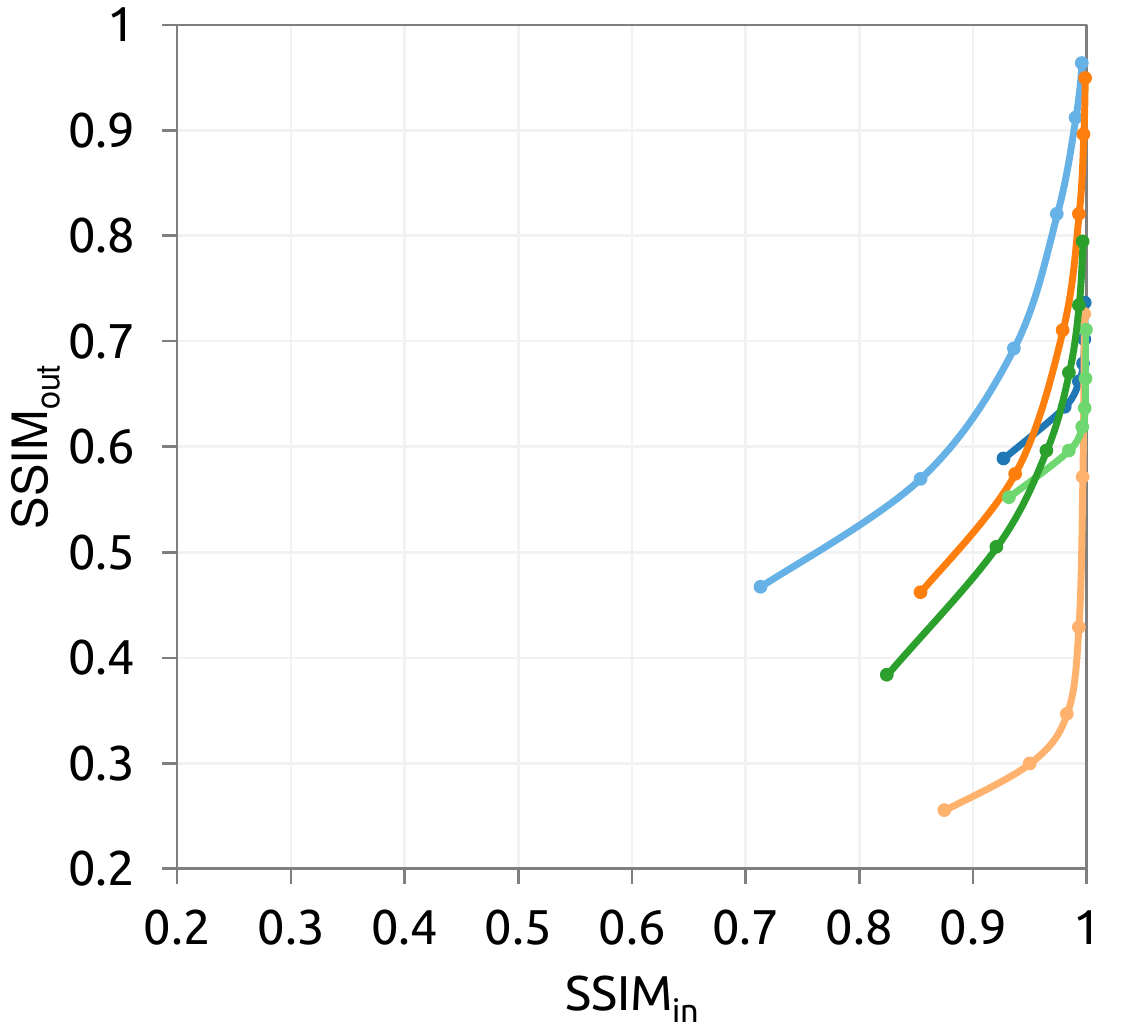} \\
			\multicolumn{5}{c}{\footnotesize{(c) I-FGSM}} \\
			\includegraphics[width=0.165\textwidth]{figures/basic_psnrssim_colorization_legend} & \includegraphics[width=0.165\textwidth]{figures/basic_psnrssim_deblurring_legend} & \includegraphics[width=0.165\textwidth]{figures/basic_psnrssim_denoising_legend} & \includegraphics[width=0.165\textwidth]{figures/basic_psnrssim_superresolution_legend} & \includegraphics[width=0.165\textwidth]{figures/basic_psnrssim_translation_legend}
		\end{tabular}
	\end{center}
	\caption{Performance comparison in terms of $\mathrm{SSIM}_{i}$ and $\mathrm{SSIM}_{o}$. Six points of each curve correspond to six different values of $\epsilon$.}
	\label{fig:basic_attack_ssim}
\end{figure*}

\clearpage

\begin{figure*}[t]
	\begin{center}
		\centering
		\renewcommand{\arraystretch}{1.5}
		\renewcommand{\tabcolsep}{2.0pt}
		\footnotesize
		\begin{tabular}{ccccc}
			\raisebox{-0.5\height}{\includegraphics[width=0.210\linewidth]{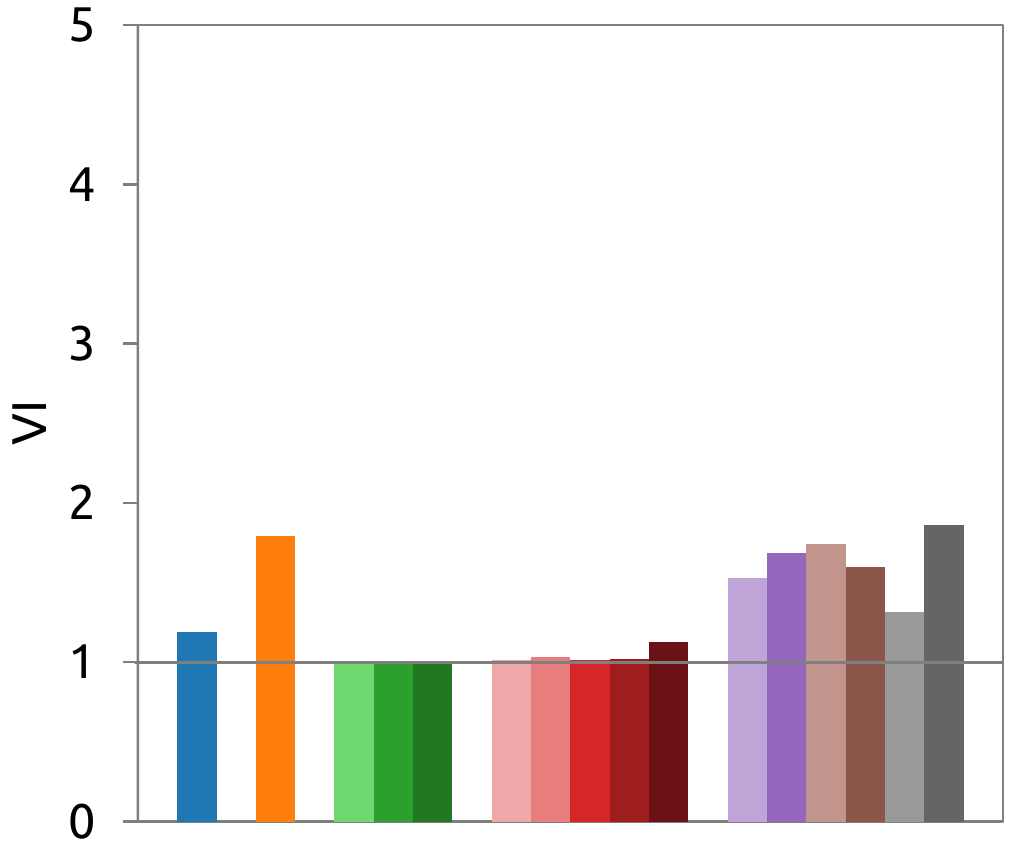}} &
			\raisebox{-0.5\height}{\includegraphics[width=0.210\linewidth]{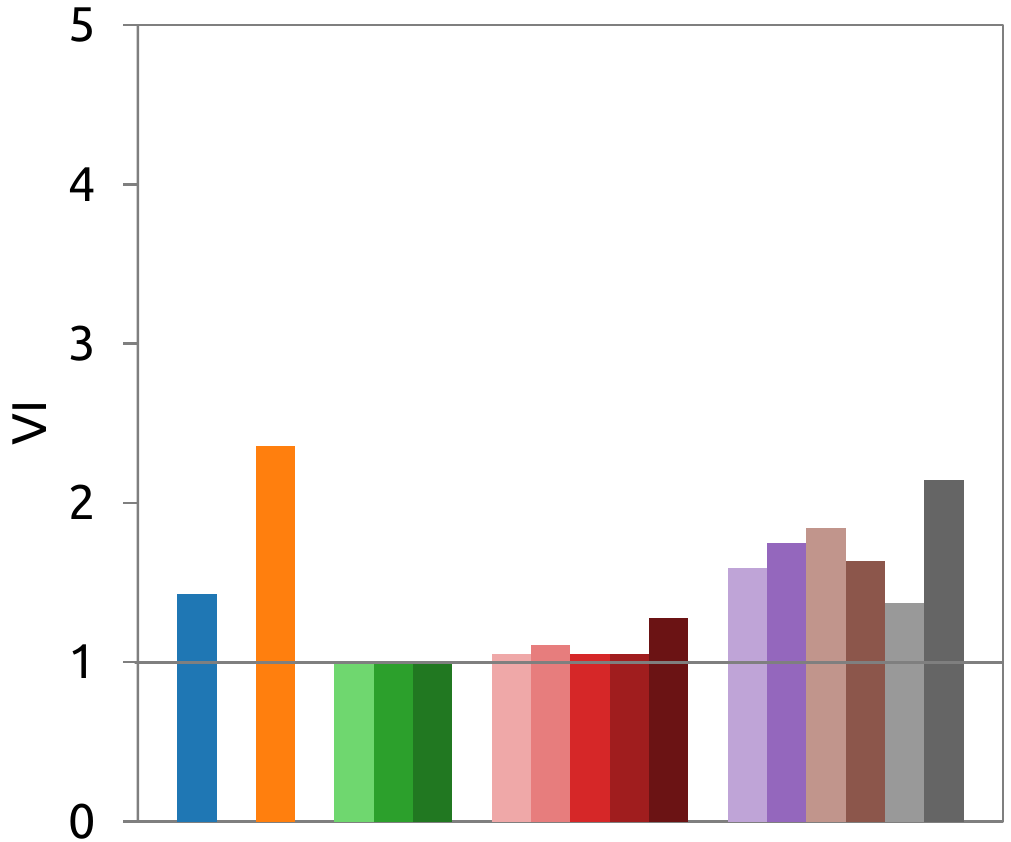}} &
			\raisebox{-0.5\height}{\includegraphics[width=0.210\linewidth]{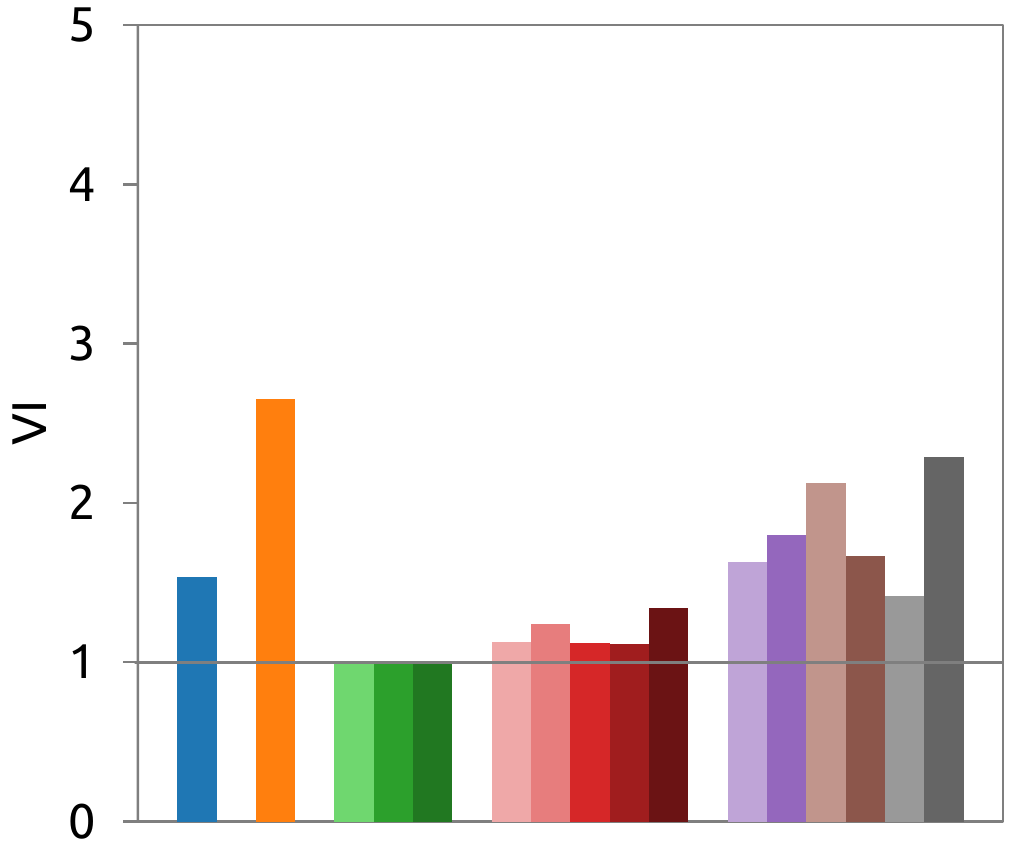}} &
			\raisebox{-0.5\height}{\includegraphics[width=0.210\linewidth]{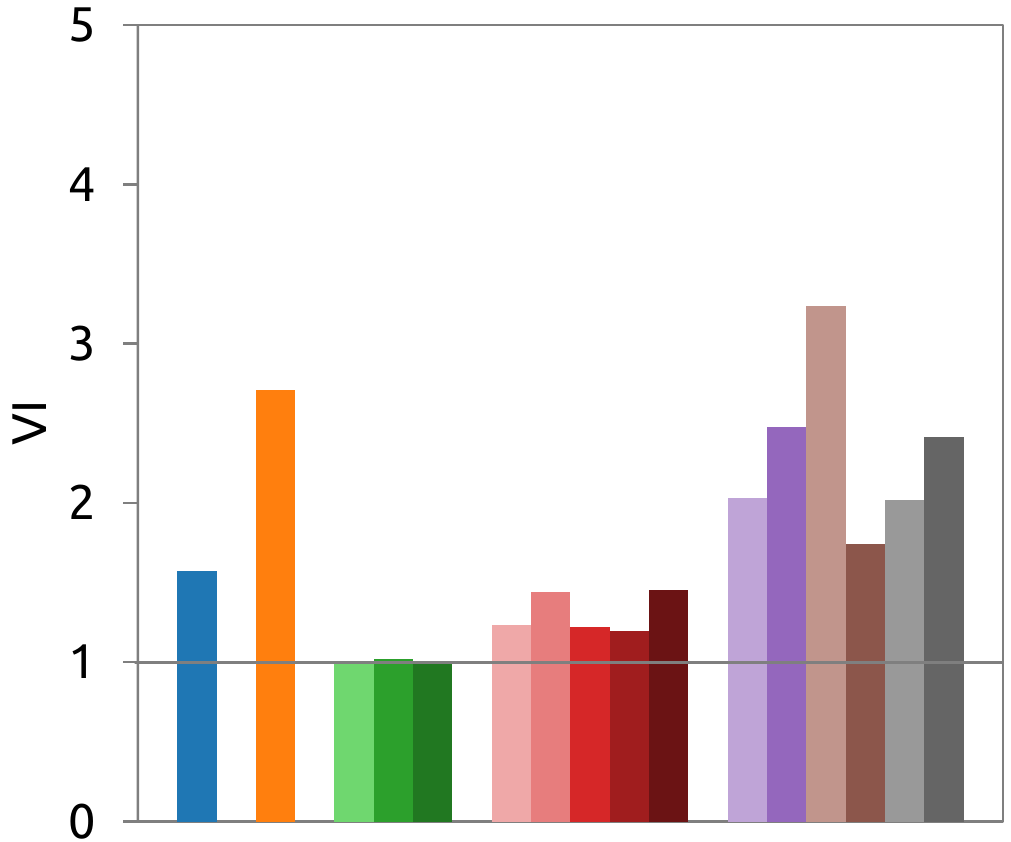}} &
			\multirow{2}{*}[33.5pt]{\raisebox{-0.5\height}{\includegraphics[width=0.115\linewidth]{figures/basic_vi_legend}}}\\
			(a) $r$$=$$1/8$ & (b) $r$$=$$2/8$ & (c) $r$$=$$3/8$ & (d) $r$$=$$4/8$ &
		\end{tabular}
	\end{center}
	\caption{Performance comparison in terms of VI for the low-frequency attack with $\epsilon=8$.}
	\label{fig:lowfreq_attack_psnrratio}
\end{figure*}

\begin{figure*}[t]
	\begin{center}
		\centering
		\renewcommand{\arraystretch}{1.5}
		\renewcommand{\tabcolsep}{2.0pt}
		\footnotesize
		\begin{tabular}{ccccc}
			\raisebox{-0.5\height}{\includegraphics[width=0.210\linewidth]{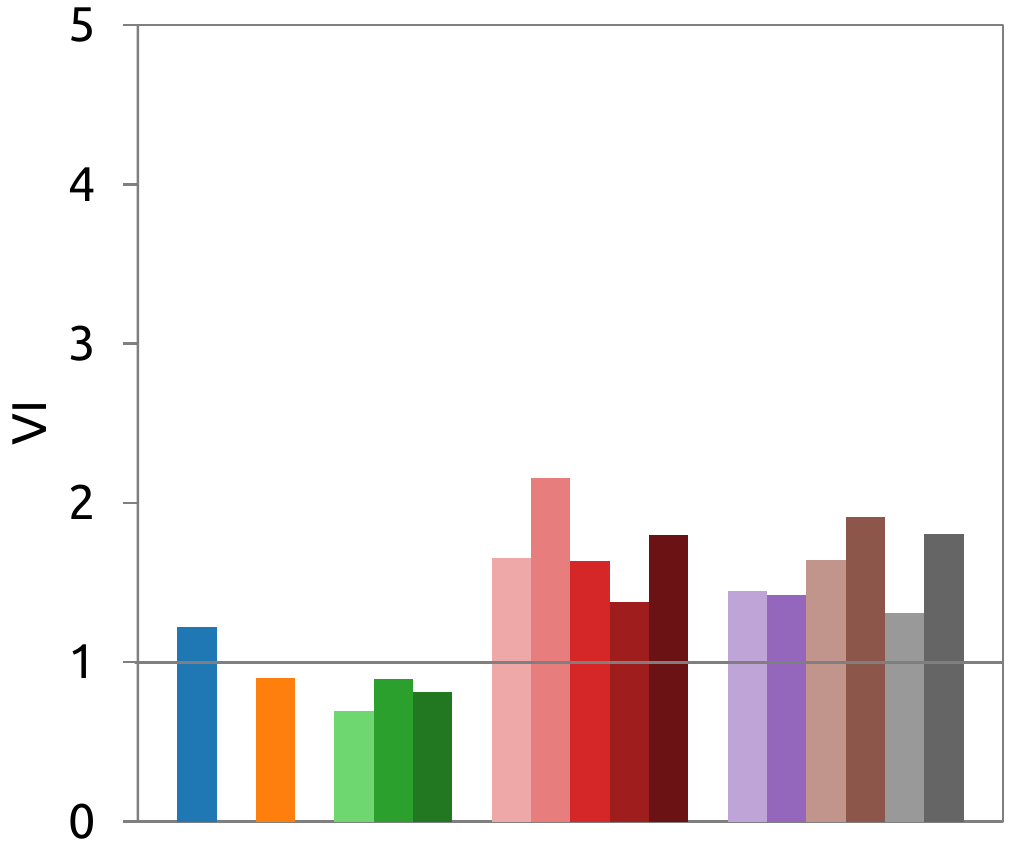}} &
			\raisebox{-0.5\height}{\includegraphics[width=0.210\linewidth]{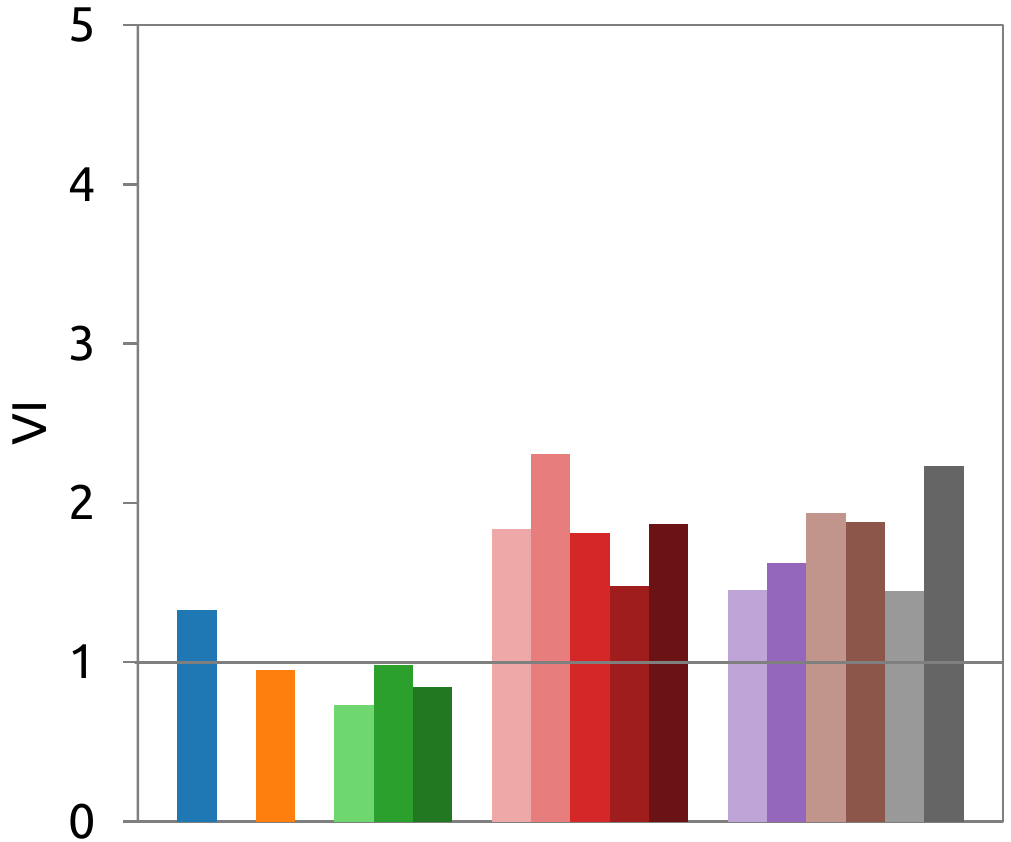}} &
			\raisebox{-0.5\height}{\includegraphics[width=0.210\linewidth]{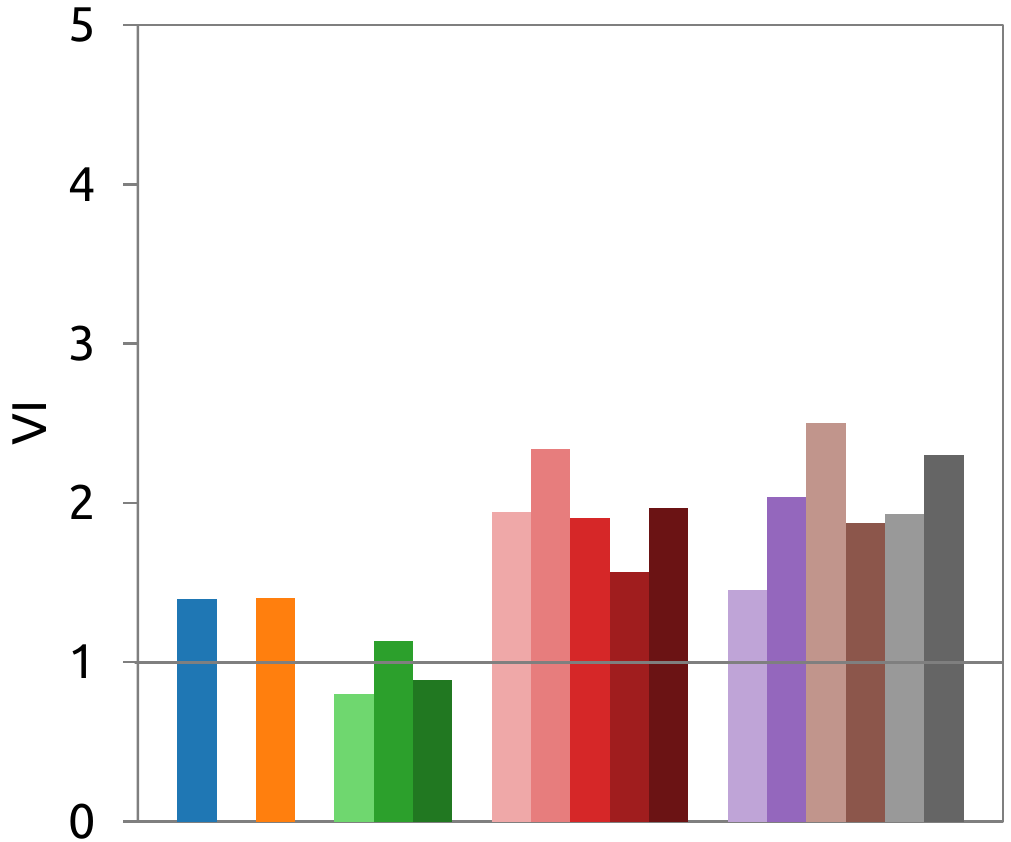}} &
			\raisebox{-0.5\height}{\includegraphics[width=0.210\linewidth]{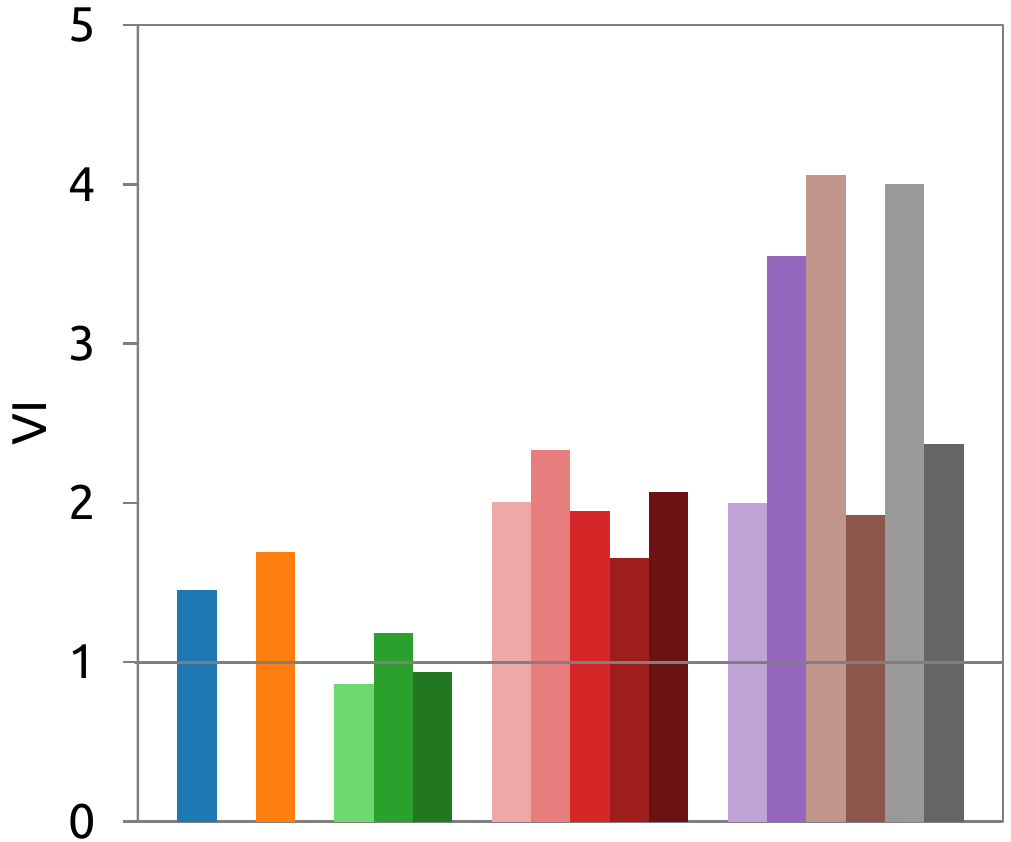}} &
			\multirow{2}{*}[33.5pt]{\raisebox{-0.5\height}{\includegraphics[width=0.115\linewidth]{figures/basic_vi_legend}}}\\
			(a) $r$$=$$1/8$ & (b) $r$$=$$2/8$ & (c) $r$$=$$3/8$ & (d) $r$$=$$4/8$ &
		\end{tabular}
	\end{center}
	\caption{Performance comparison in terms of VI for the high-frequency attack with $\epsilon=8$.}
	\label{fig:highfreq_attack_psnrratio}
\end{figure*}

\begin{figure*}[t]
	\begin{center}
		\centering
		\renewcommand{\arraystretch}{0.3}
		\renewcommand{\tabcolsep}{1.8pt}
		\footnotesize
		\begin{tabular}{ccccccccccc}
			& & \multicolumn{4}{c}{\textbf{Low-frequency attack}} & & \multicolumn{4}{c}{\textbf{High-frequency attack}} \\ \\
			& & $r$$=$$1/8$ & $r$$=$$2/8$ & $r$$=$$3/8$ & $r$$=$$4/8$ & & $r$$=$$1/8$ & $r$$=$$2/8$ & $r$$=$$3/8$ & $r$$=$$4/8$ \\ \\
			\rotatebox[origin=c]{90}{Input} & & \raisebox{-0.5\height}{\includegraphics[width=0.112\textwidth]{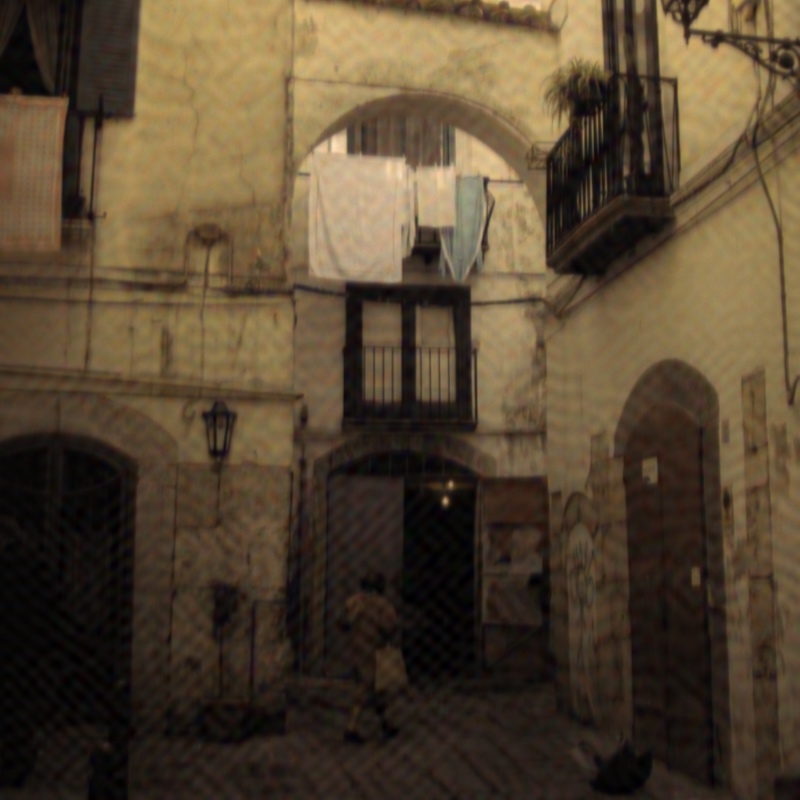}} &
			\raisebox{-0.5\height}{\includegraphics[width=0.112\textwidth]{figures/deepdeblur_3_3_ifgsm_lowfreqdct_ratio0_250_e8_attacked}} &
			\raisebox{-0.5\height}{\includegraphics[width=0.112\textwidth]{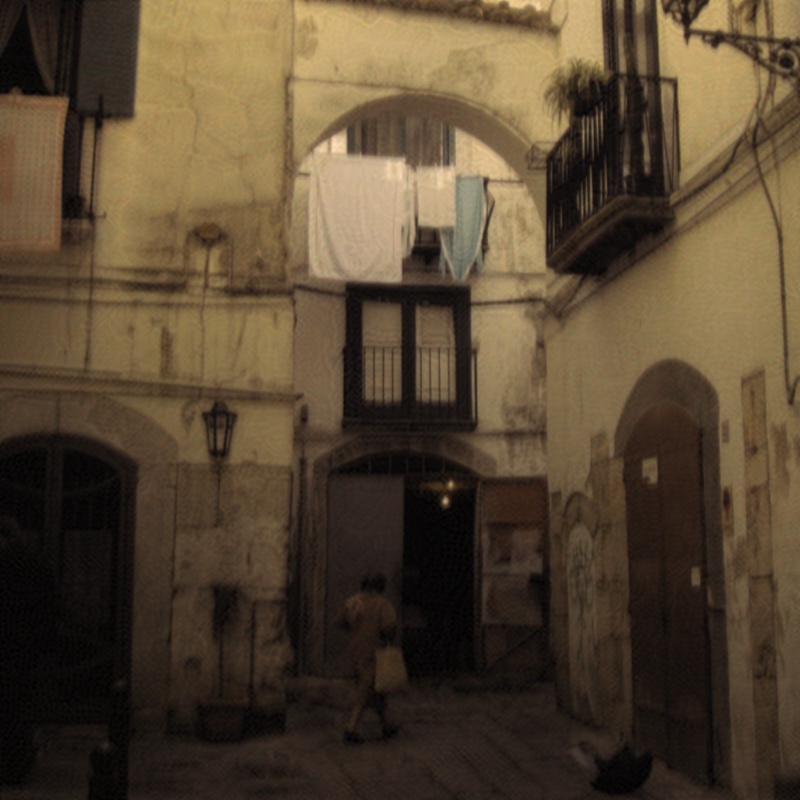}} &
			\raisebox{-0.5\height}{\includegraphics[width=0.112\textwidth]{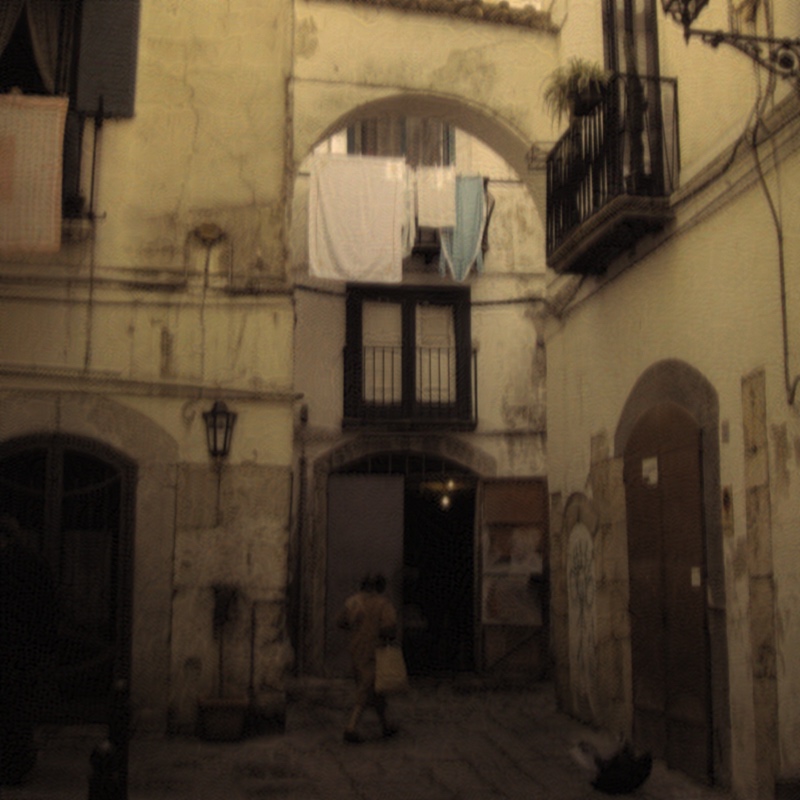}} & & \raisebox{-0.5\height}{\includegraphics[width=0.112\textwidth]{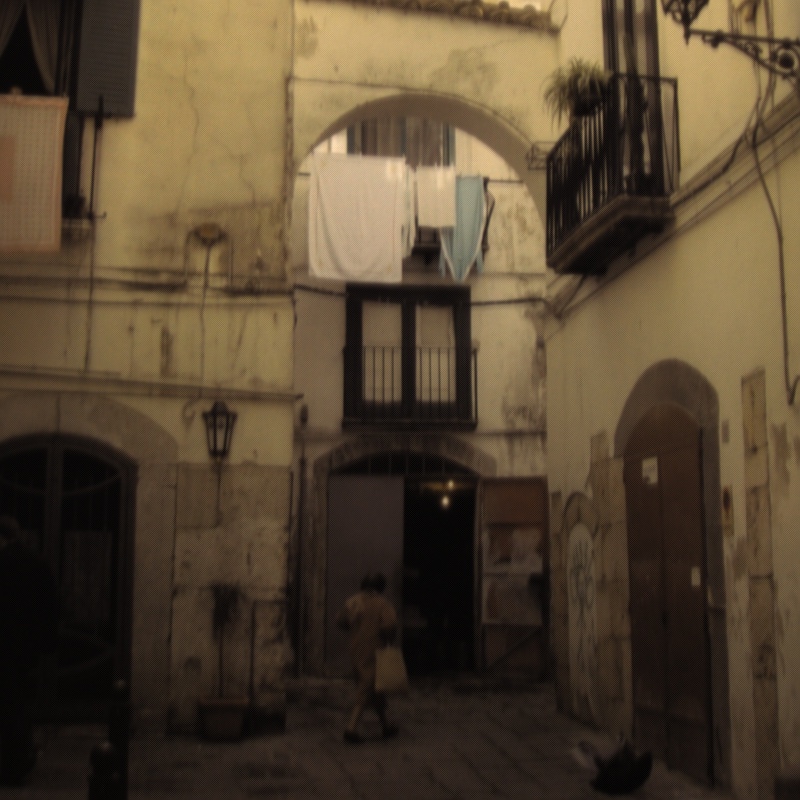}} &
			\raisebox{-0.5\height}{\includegraphics[width=0.112\textwidth]{figures/deepdeblur_3_3_ifgsm_highfreqdct_ratio0_250_e8_attacked}} &
			\raisebox{-0.5\height}{\includegraphics[width=0.112\textwidth]{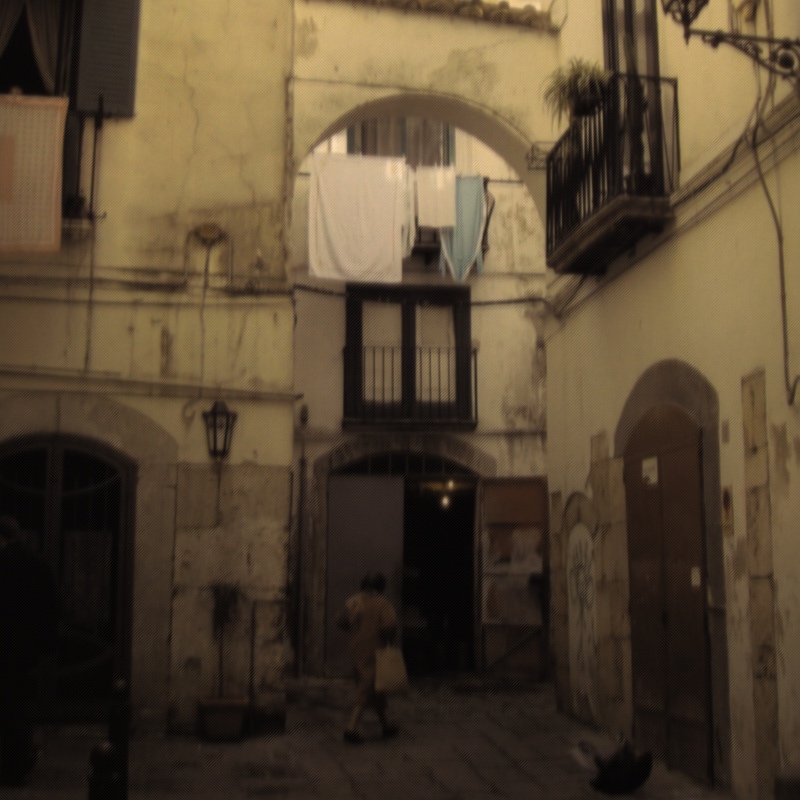}} &
			\raisebox{-0.5\height}{\includegraphics[width=0.112\textwidth]{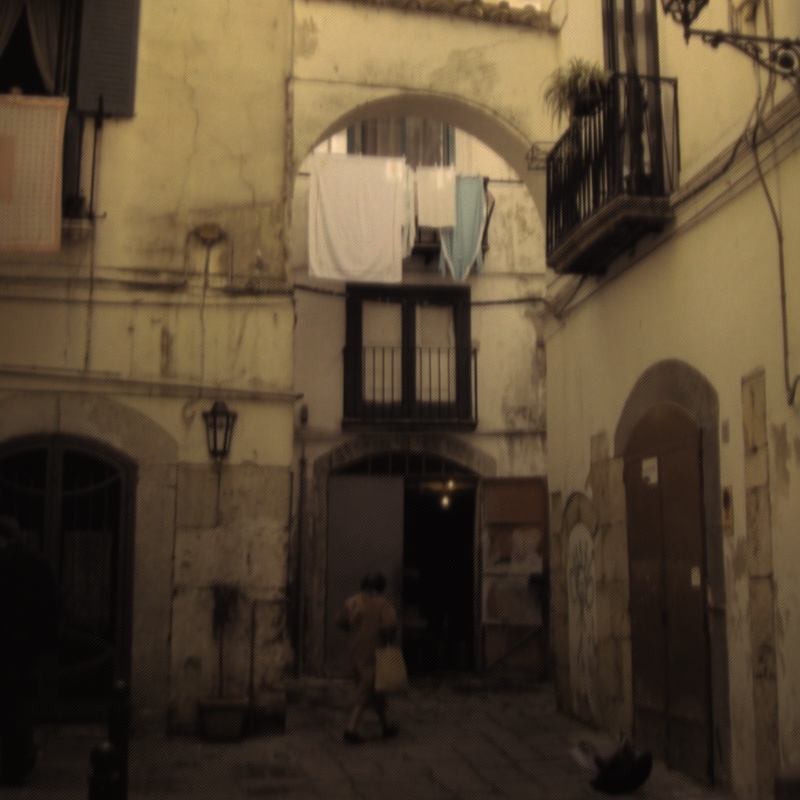}} \\ \\
			\rotatebox[origin=c]{90}{Output} & & \raisebox{-0.5\height}{\includegraphics[width=0.112\textwidth]{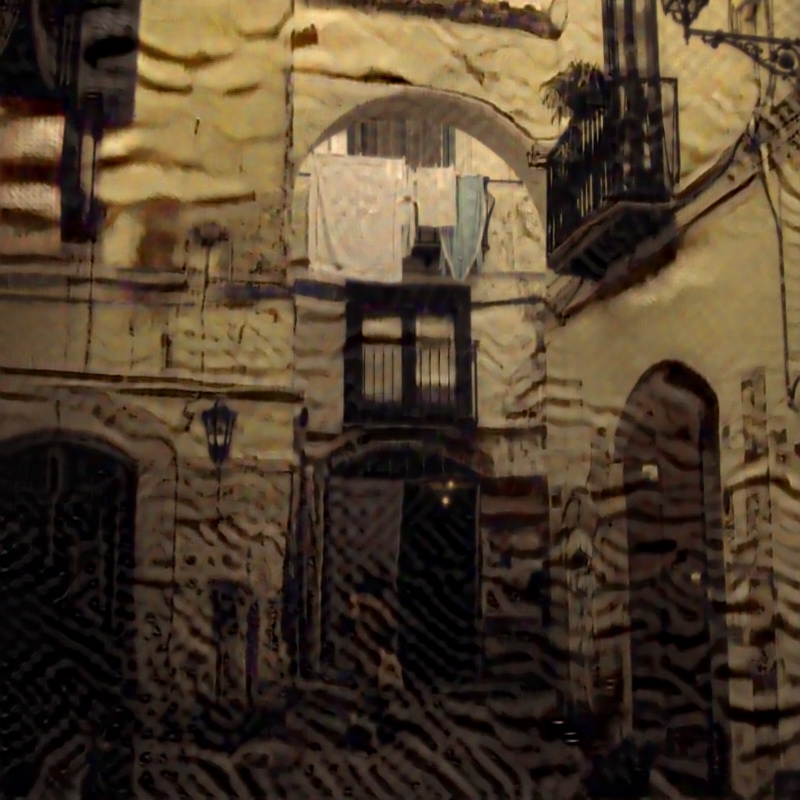}} &
			\raisebox{-0.5\height}{\includegraphics[width=0.112\textwidth]{figures/deepdeblur_3_3_ifgsm_lowfreqdct_ratio0_250_e8_output}} &
			\raisebox{-0.5\height}{\includegraphics[width=0.112\textwidth]{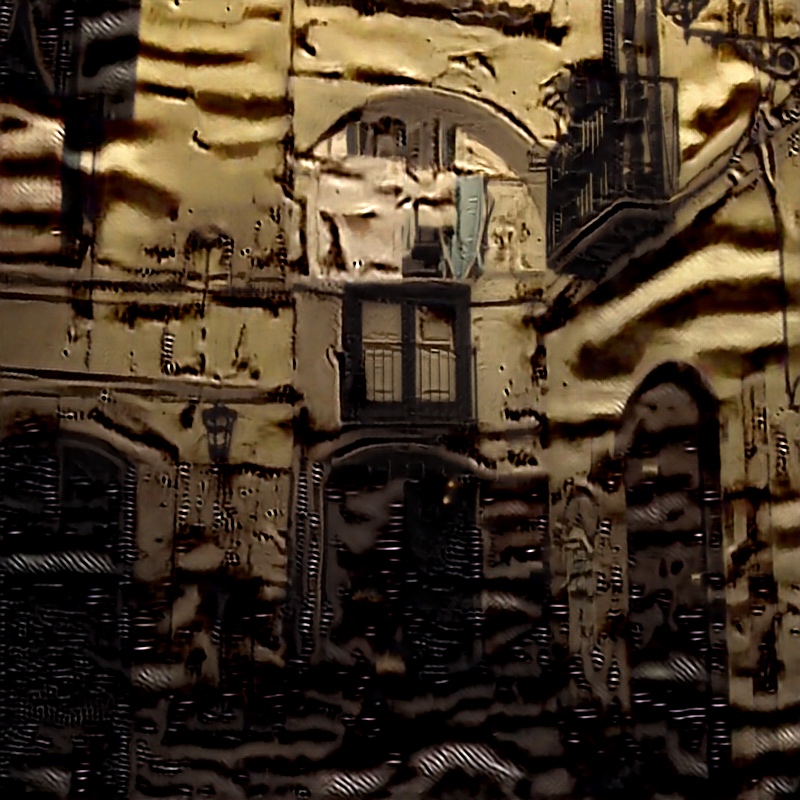}} &
			\raisebox{-0.5\height}{\includegraphics[width=0.112\textwidth]{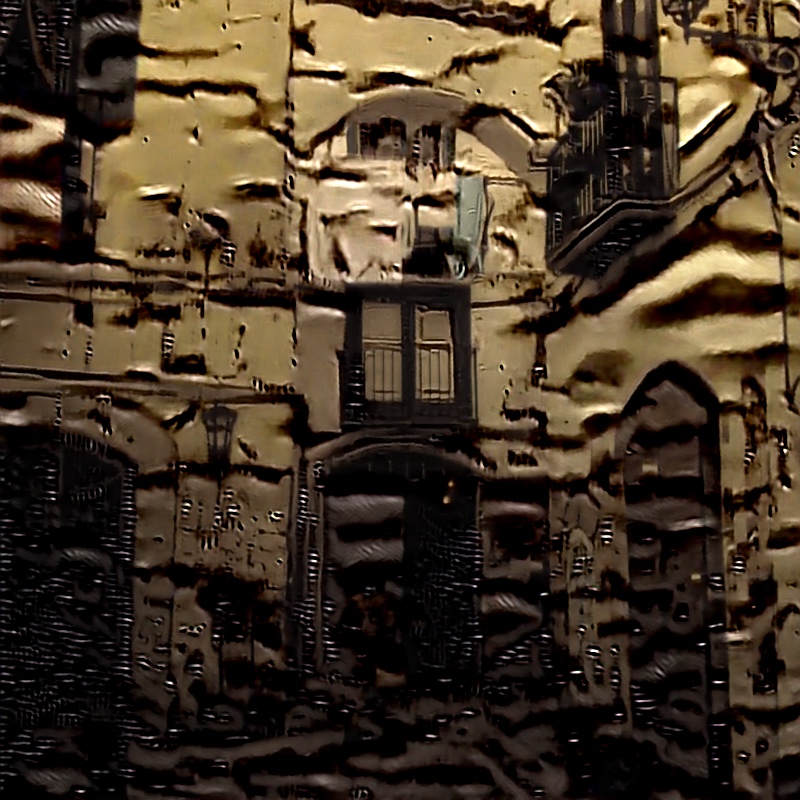}} & & \raisebox{-0.5\height}{\includegraphics[width=0.112\textwidth]{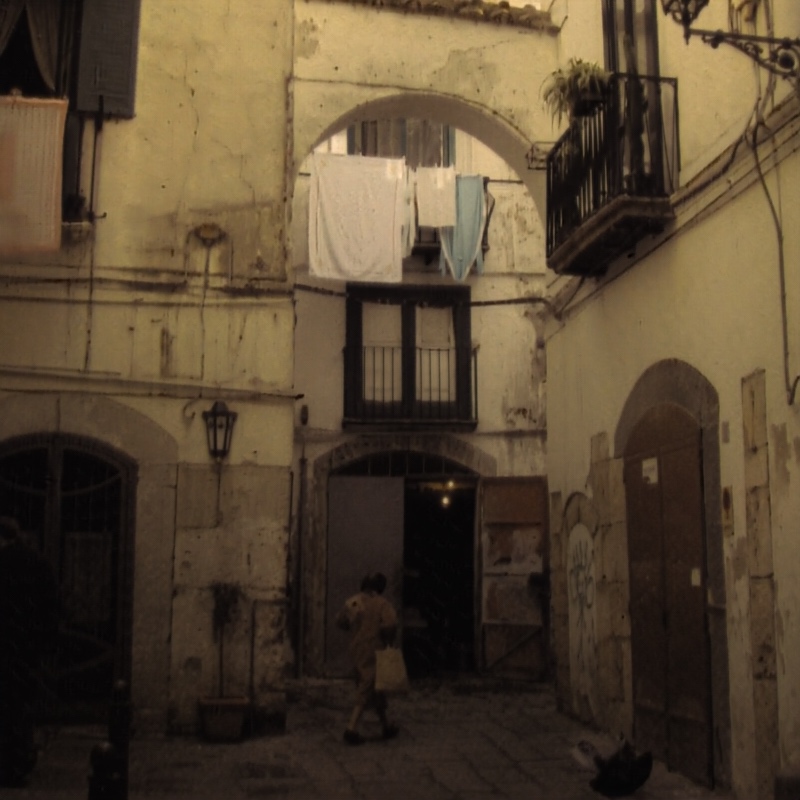}} &
			\raisebox{-0.5\height}{\includegraphics[width=0.112\textwidth]{figures/deepdeblur_3_3_ifgsm_highfreqdct_ratio0_250_e8_output}} &
			\raisebox{-0.5\height}{\includegraphics[width=0.112\textwidth]{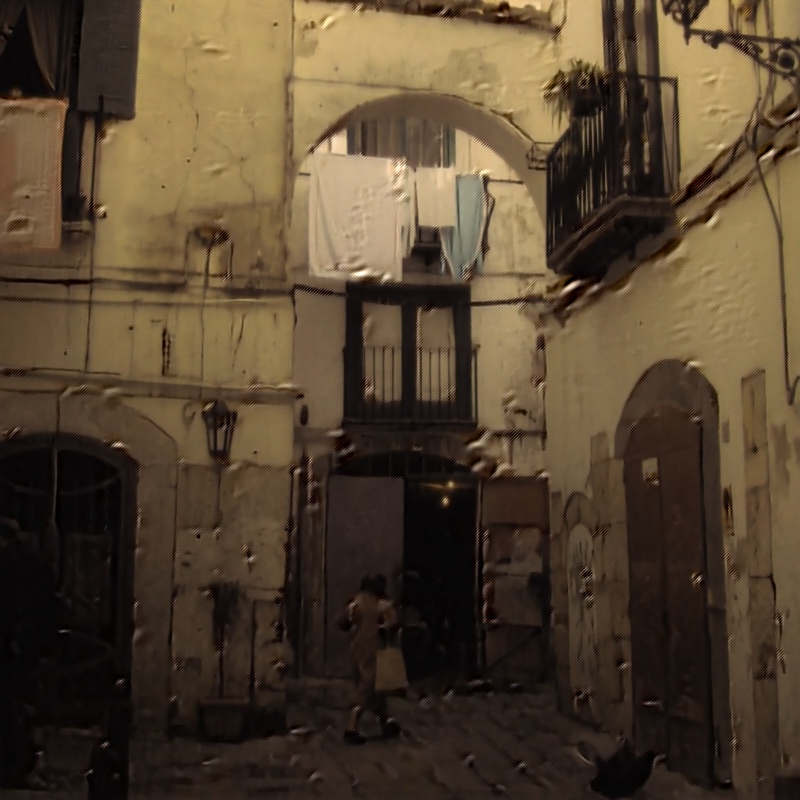}} &
			\raisebox{-0.5\height}{\includegraphics[width=0.112\textwidth]{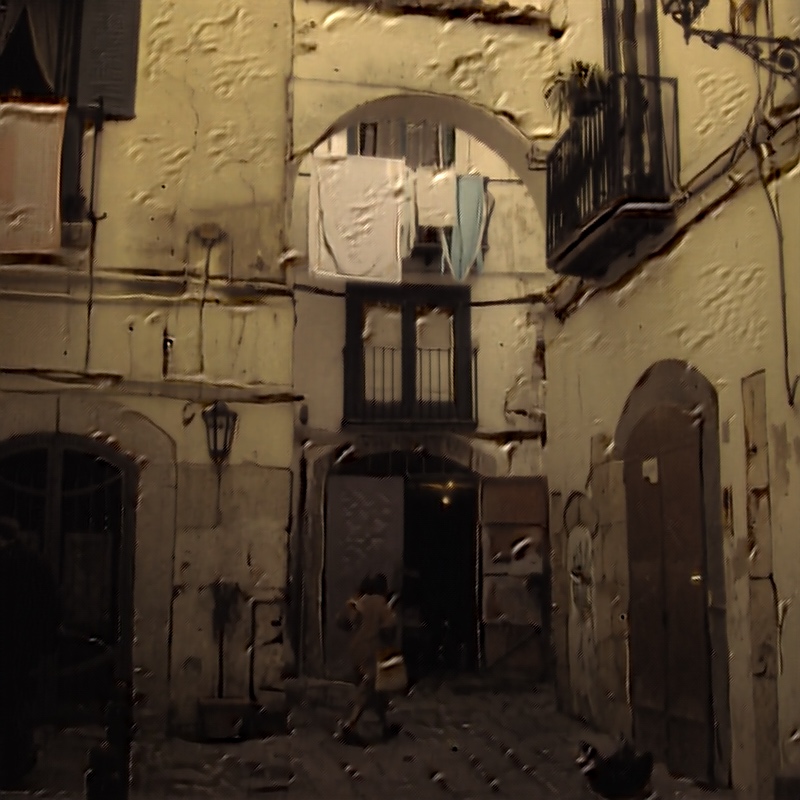}} \\ \\
			& & \multicolumn{9}{c}{\footnotesize{(a) DeepDeblur}} \\ \\
			\rotatebox[origin=c]{90}{Input} & & \raisebox{-0.5\height}{\includegraphics[width=0.112\textwidth]{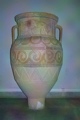}} &
			\raisebox{-0.5\height}{\includegraphics[width=0.112\textwidth]{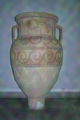}} &
			\raisebox{-0.5\height}{\includegraphics[width=0.112\textwidth]{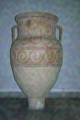}} &
			\raisebox{-0.5\height}{\includegraphics[width=0.112\textwidth]{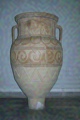}} & & \raisebox{-0.5\height}{\includegraphics[width=0.112\textwidth]{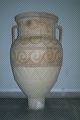}} &
			\raisebox{-0.5\height}{\includegraphics[width=0.112\textwidth]{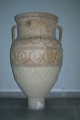}} &
			\raisebox{-0.5\height}{\includegraphics[width=0.112\textwidth]{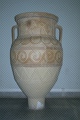}} &
			\raisebox{-0.5\height}{\includegraphics[width=0.112\textwidth]{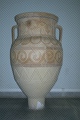}} \\ \\
			\rotatebox[origin=c]{90}{Output} & & \raisebox{-0.5\height}{\includegraphics[width=0.112\textwidth]{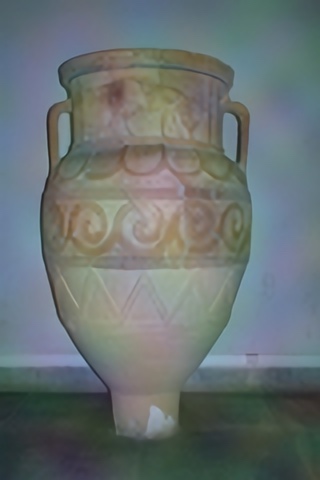}} &
			\raisebox{-0.5\height}{\includegraphics[width=0.112\textwidth]{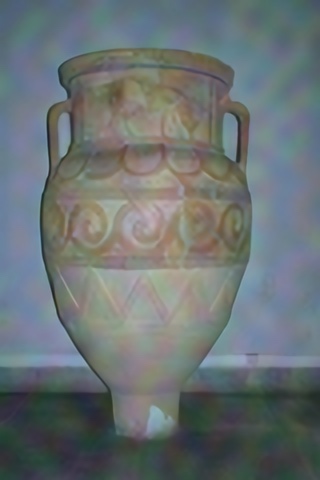}} &
			\raisebox{-0.5\height}{\includegraphics[width=0.112\textwidth]{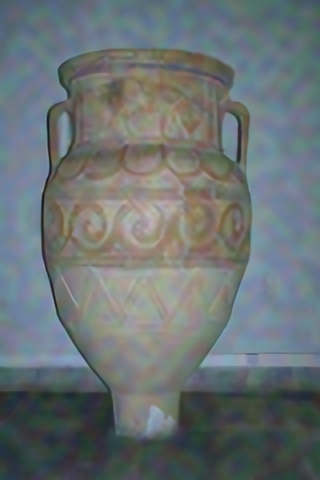}} &
			\raisebox{-0.5\height}{\includegraphics[width=0.112\textwidth]{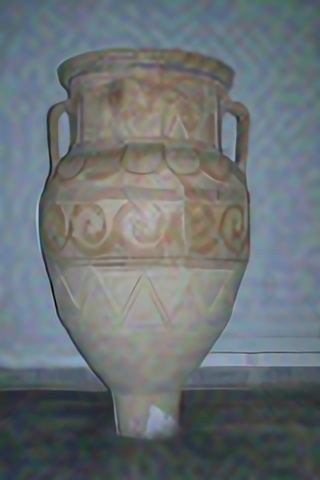}} & & \raisebox{-0.5\height}{\includegraphics[width=0.112\textwidth]{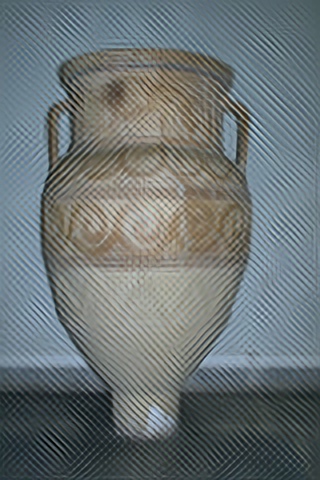}} &
			\raisebox{-0.5\height}{\includegraphics[width=0.112\textwidth]{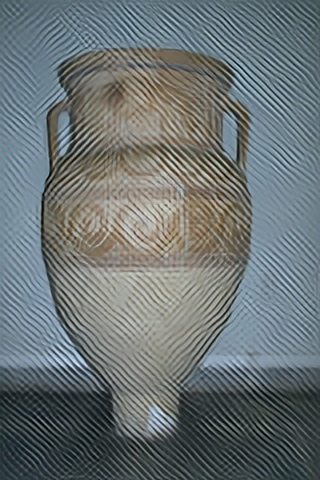}} &
			\raisebox{-0.5\height}{\includegraphics[width=0.112\textwidth]{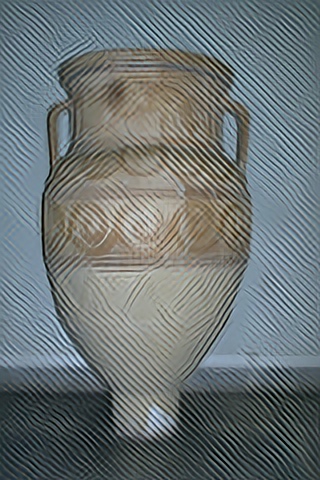}} &
			\raisebox{-0.5\height}{\includegraphics[width=0.112\textwidth]{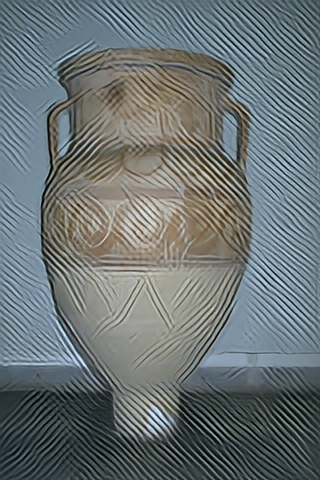}} \\ \\
			& & \multicolumn{9}{c}{\footnotesize{(b) EDSR}} \\ \\
		\end{tabular}
	\end{center}
	\caption{Images obtained from (a) DeepDeblur and (b) EDSR, where the low-frequency and high-frequency attacks are employed with $\epsilon=8$ and $r \in \{1/8, 2/8, 3/8, 4/8\}$.}
	\label{fig:frequency_attack_examples}
\end{figure*}

\clearpage

\begin{figure*}[t]
	\begin{center}
		\centering
		\renewcommand{\arraystretch}{1.0}
		\renewcommand{\tabcolsep}{0.9pt}
		\scriptsize
		\begin{tabular}{ccccccccccc}
			\multicolumn{5}{c}{\textbf{Original outputs}} & & \multicolumn{5}{c}{\textbf{I-FGSM ($\epsilon=8$)}} \\
			No defense & JPEG & Resizing & \makecell[c]{Bit\\reduction} & \makecell[c]{Self-\\ensemble} & & No defense & JPEG & Resizing & \makecell[c]{Bit\\reduction} & \makecell[c]{Self-\\ensemble} \\
			\includegraphics[width=0.095\linewidth]{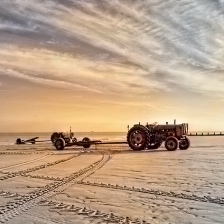} & \includegraphics[width=0.095\linewidth]{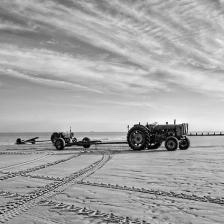} & \includegraphics[width=0.095\linewidth]{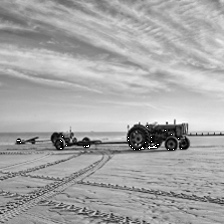} & \includegraphics[width=0.095\linewidth]{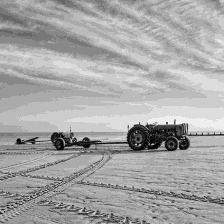} & \includegraphics[width=0.095\linewidth]{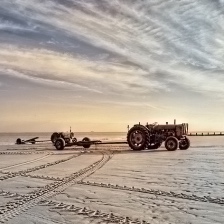} & &
			\includegraphics[width=0.095\linewidth]{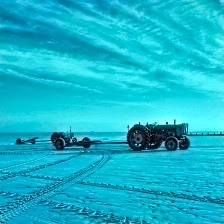} & \includegraphics[width=0.095\linewidth]{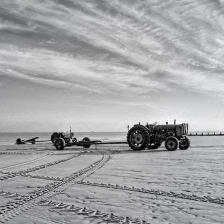} & \includegraphics[width=0.095\linewidth]{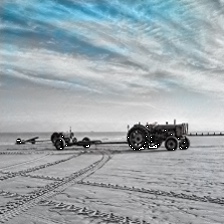} & \includegraphics[width=0.095\linewidth]{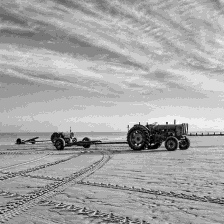} &
			\includegraphics[width=0.095\linewidth]{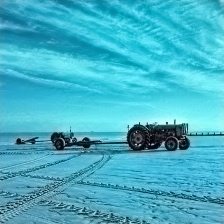}
			\\
			\multicolumn{11}{c}{\footnotesize{(a) Colorization (CIC)}} \\
			\includegraphics[width=0.095\linewidth]{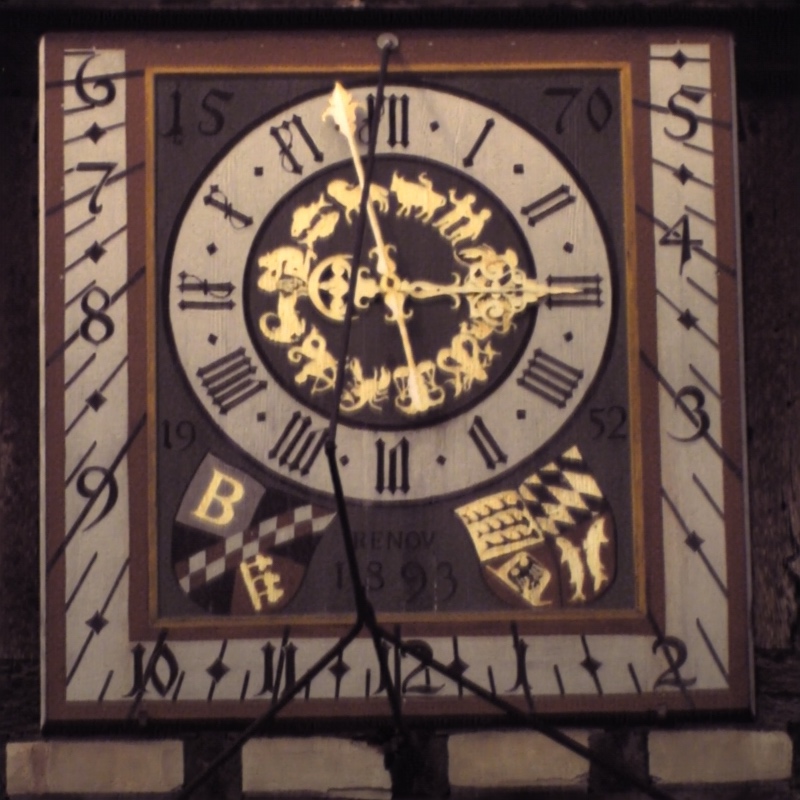} & \includegraphics[width=0.095\linewidth]{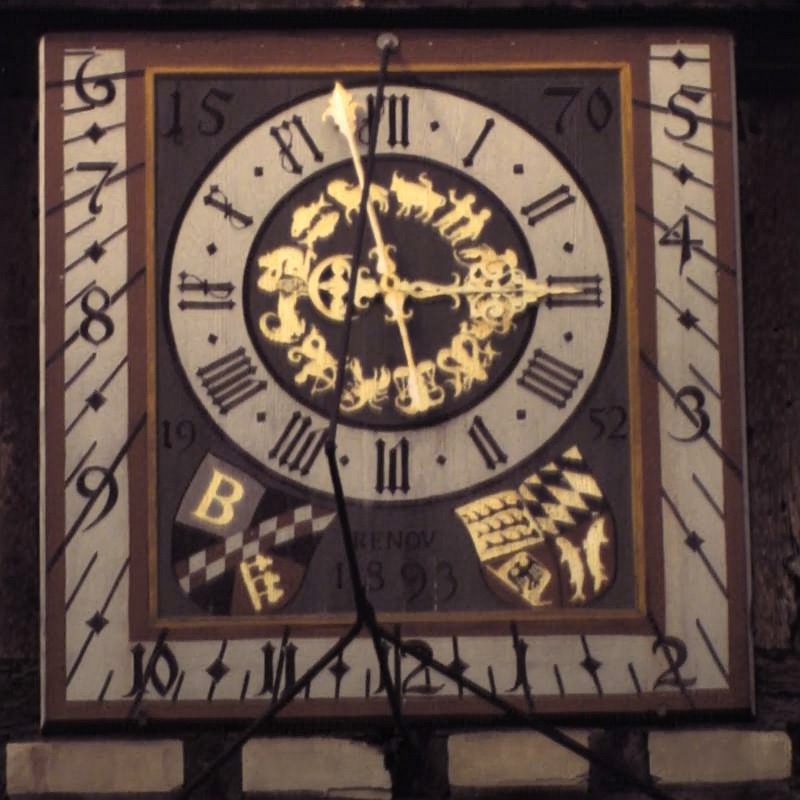} & \includegraphics[width=0.095\linewidth]{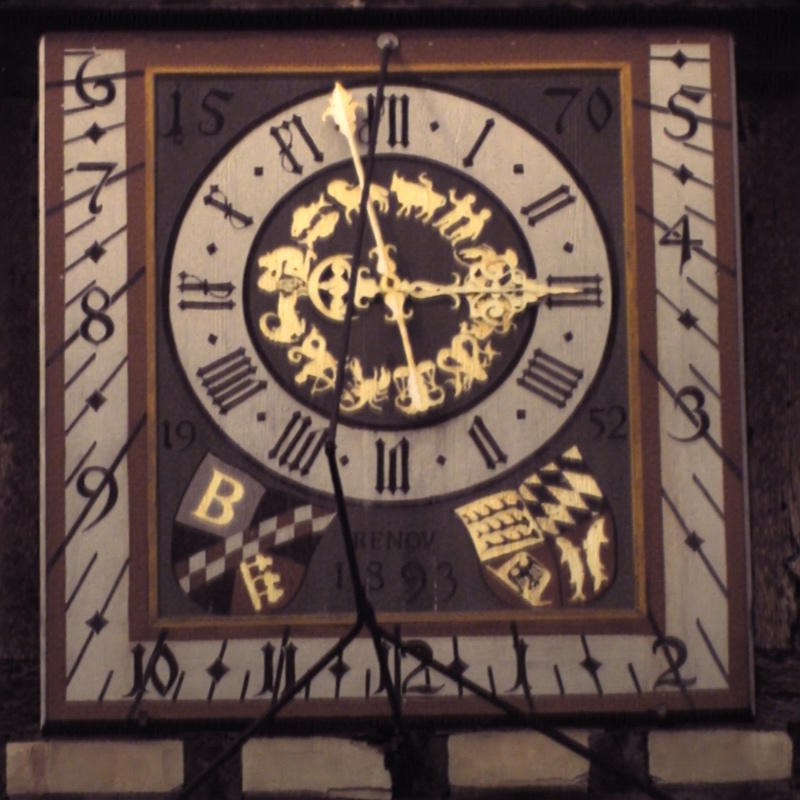} & \includegraphics[width=0.095\linewidth]{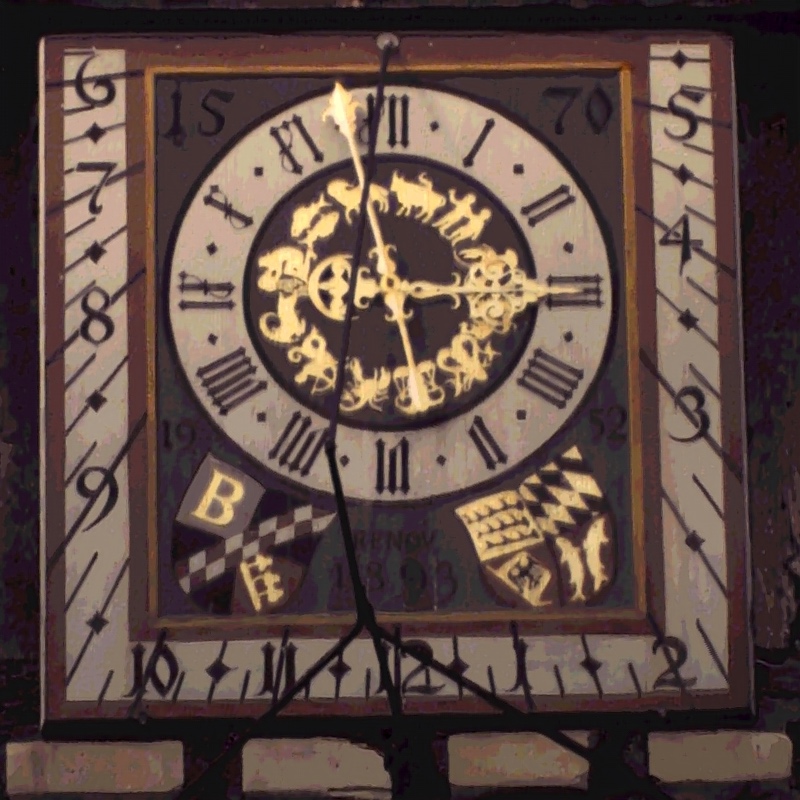} & \includegraphics[width=0.095\linewidth]{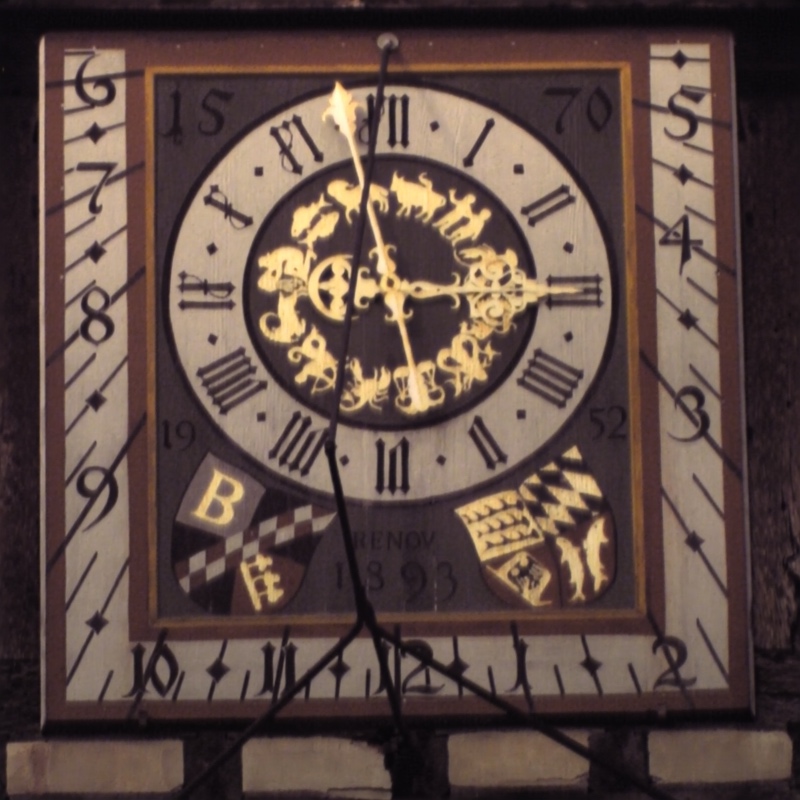} & &
			\includegraphics[width=0.095\linewidth]{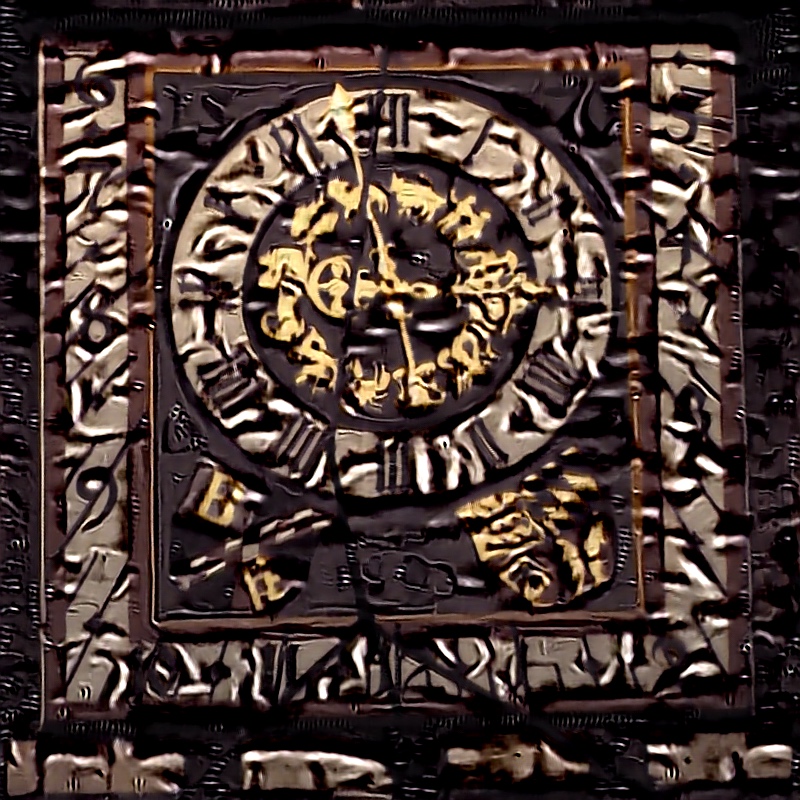} & \includegraphics[width=0.095\linewidth]{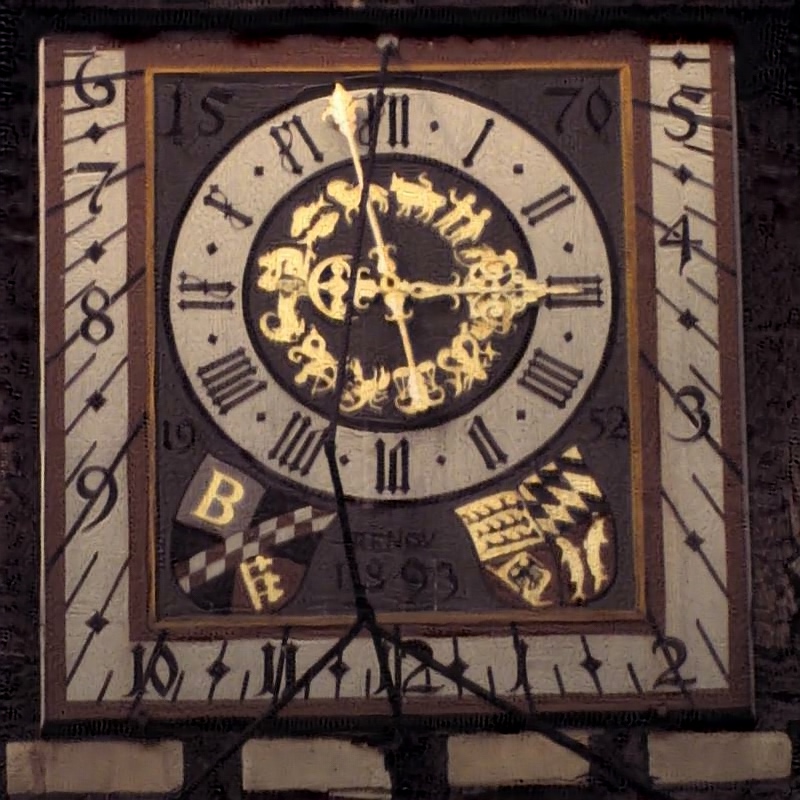} & \includegraphics[width=0.095\linewidth]{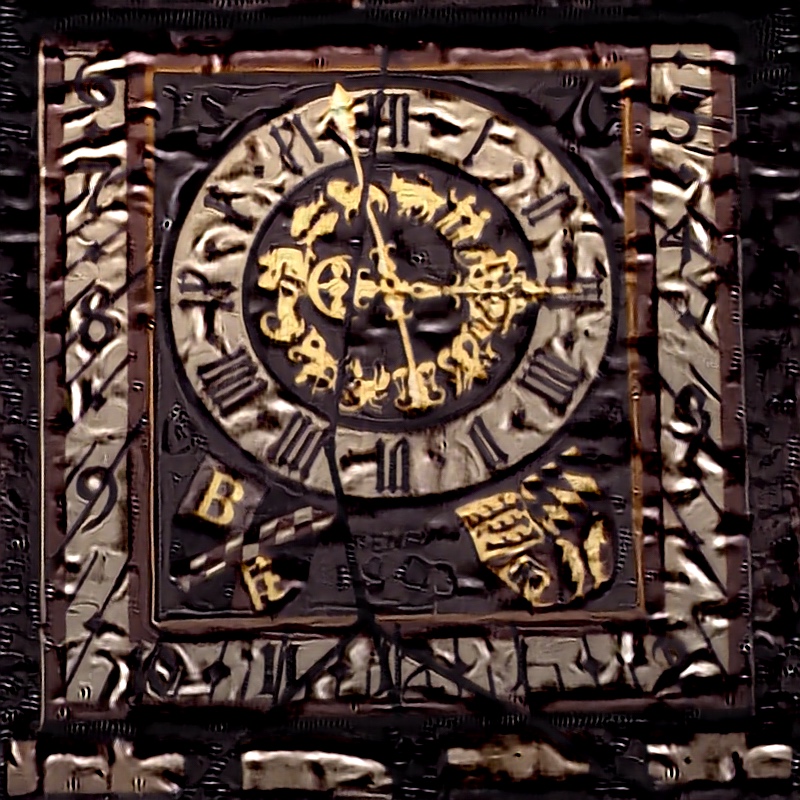} & \includegraphics[width=0.095\linewidth]{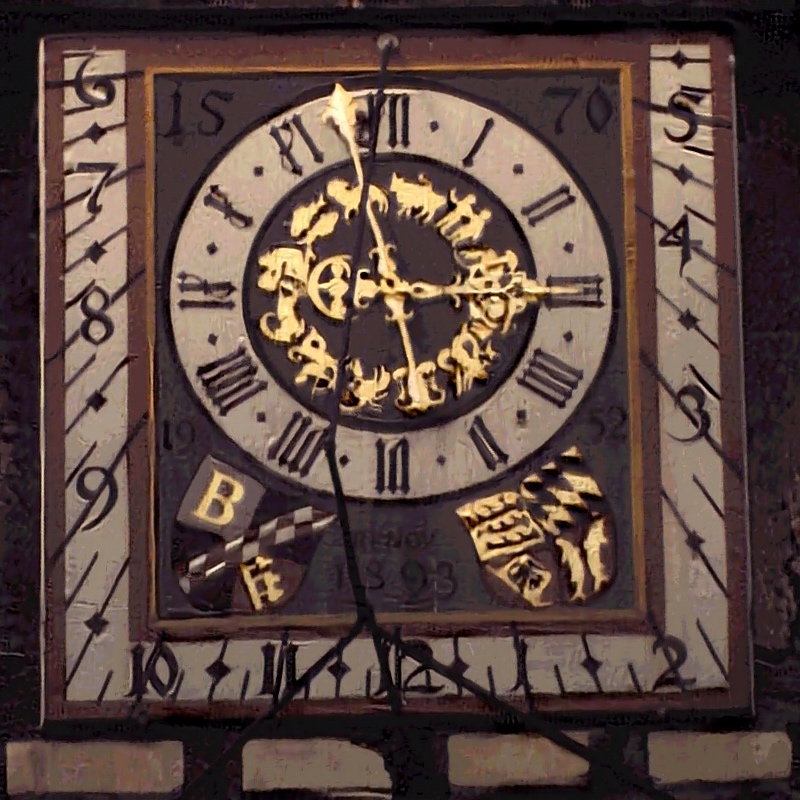} & \includegraphics[width=0.095\linewidth]{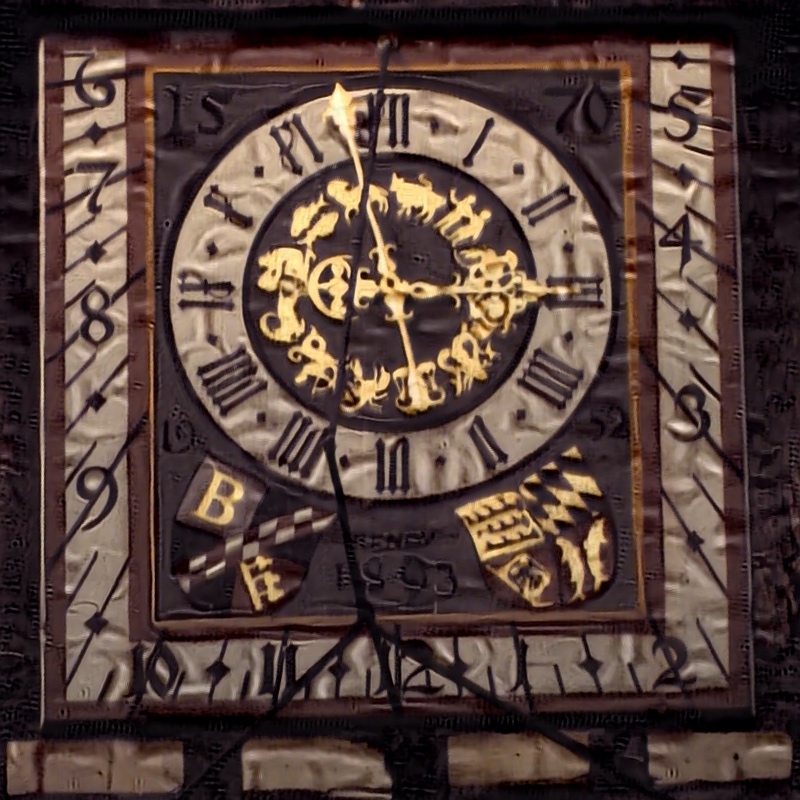}
			\\
			\multicolumn{11}{c}{\footnotesize{(b) Deblurring (DeepDeblur)}} \\
			\includegraphics[width=0.095\linewidth]{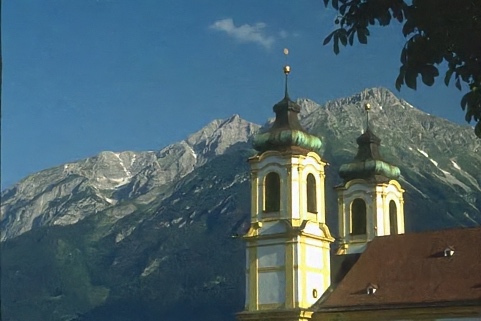} & \includegraphics[width=0.095\linewidth]{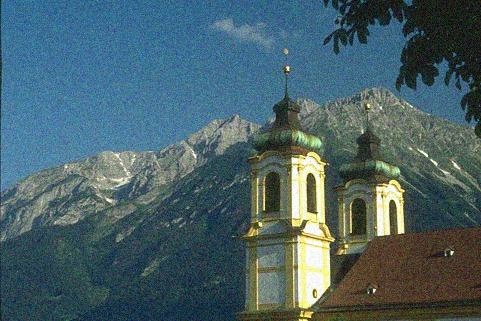} & \includegraphics[width=0.095\linewidth]{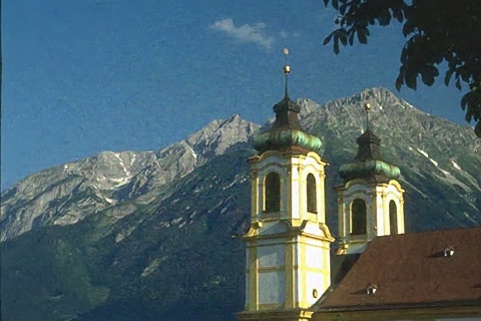} & \includegraphics[width=0.095\linewidth]{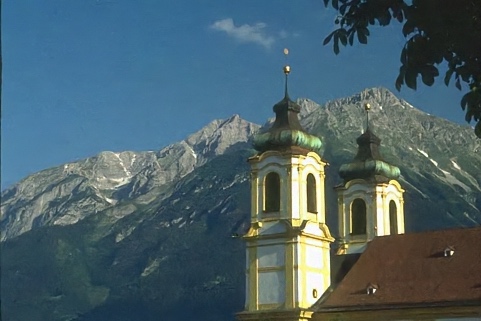} & \includegraphics[width=0.095\linewidth]{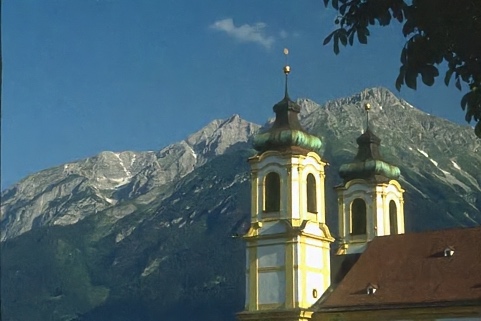} & &
			\includegraphics[width=0.095\linewidth]{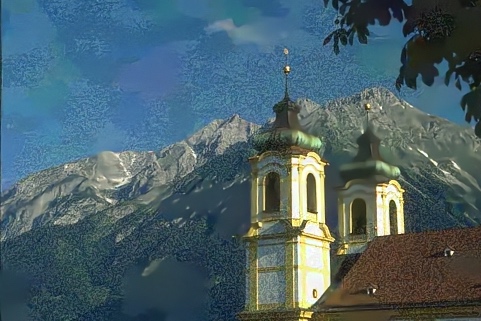} & \includegraphics[width=0.095\linewidth]{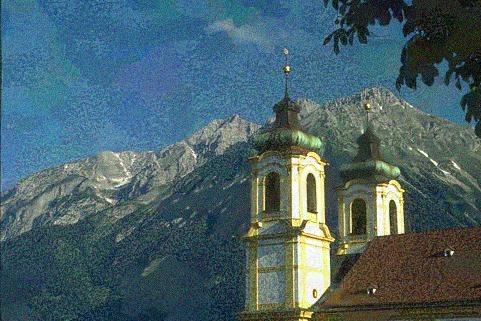} & \includegraphics[width=0.095\linewidth]{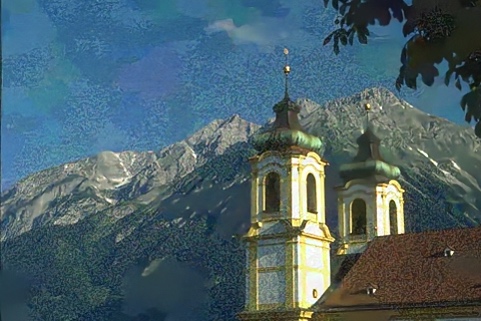} & \includegraphics[width=0.095\linewidth]{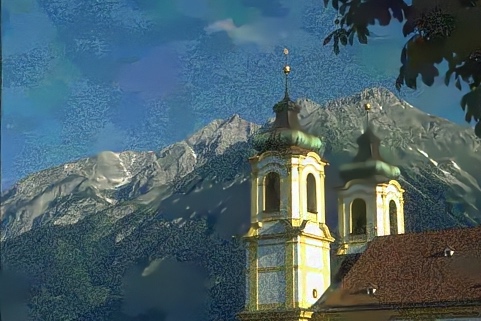} & \includegraphics[width=0.095\linewidth]{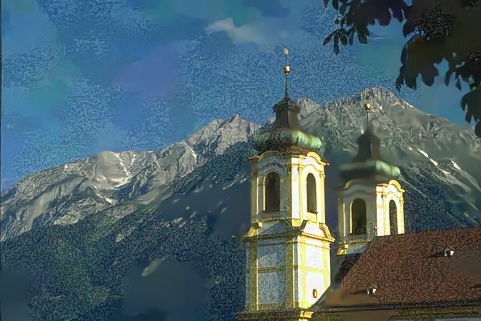}
			\\
			\multicolumn{11}{c}{\footnotesize{(c) Denoising (DnCNN (color, multiple $\sigma$))}} \\
			\includegraphics[width=0.095\linewidth]{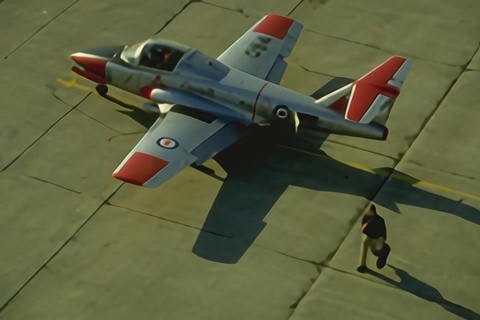} & \includegraphics[width=0.095\linewidth]{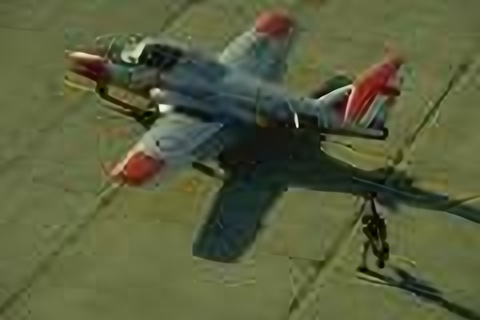} & \includegraphics[width=0.095\linewidth]{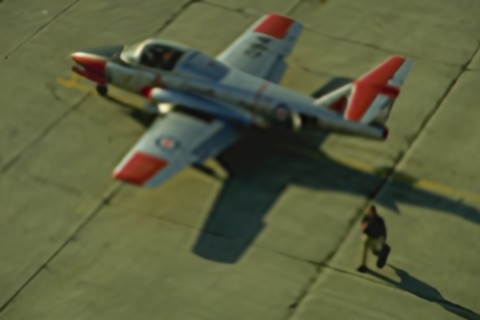} & \includegraphics[width=0.095\linewidth]{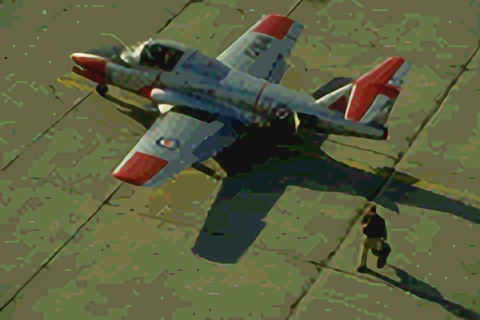} & \includegraphics[width=0.095\linewidth]{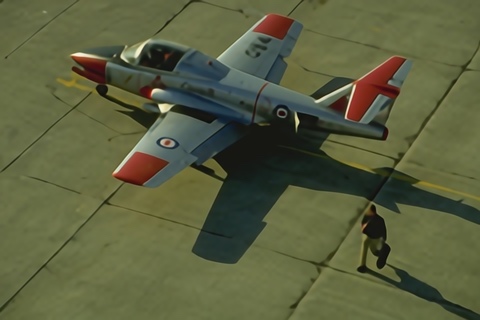} & &
			\includegraphics[width=0.095\linewidth]{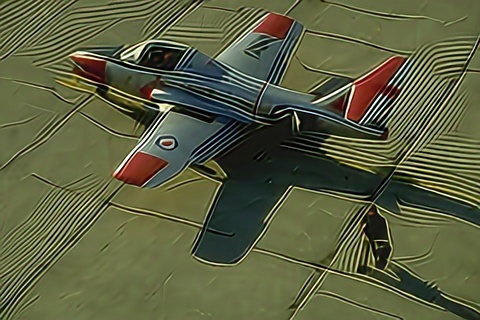} & \includegraphics[width=0.095\linewidth]{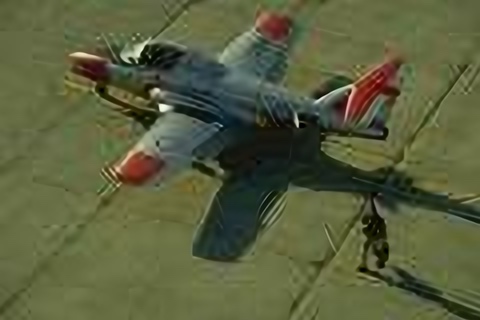} & \includegraphics[width=0.095\linewidth]{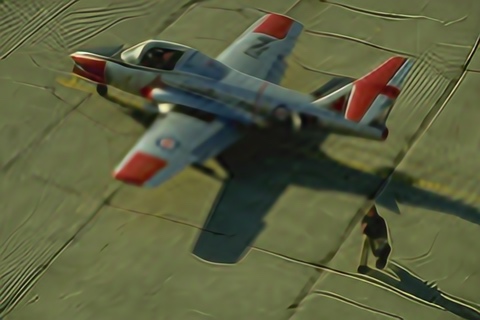} & \includegraphics[width=0.095\linewidth]{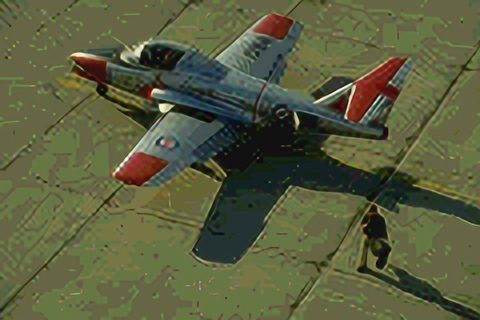} & \includegraphics[width=0.095\linewidth]{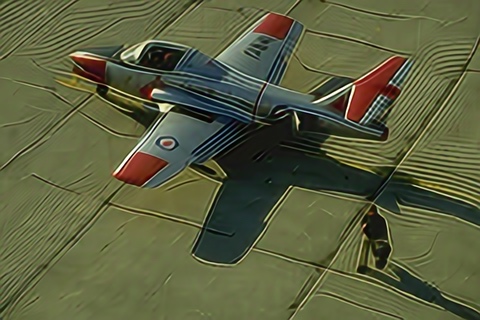}
			\\
			\multicolumn{11}{c}{\footnotesize{(d) Super-resolution (RCAN)}} \\
			\includegraphics[width=0.095\linewidth]{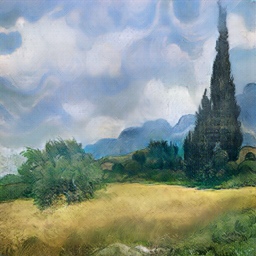} & \includegraphics[width=0.095\linewidth]{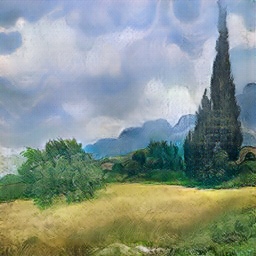} & \includegraphics[width=0.095\linewidth]{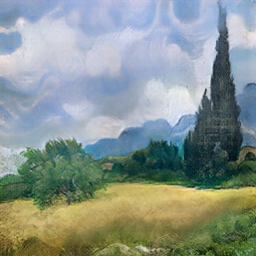} & \includegraphics[width=0.095\linewidth]{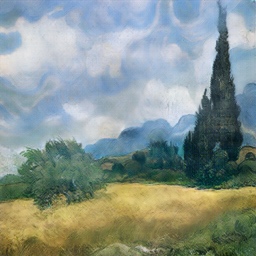} & \includegraphics[width=0.095\linewidth]{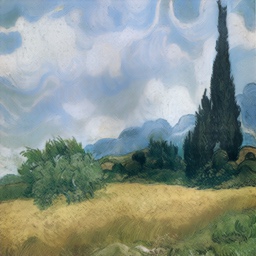} & &
			\includegraphics[width=0.095\linewidth]{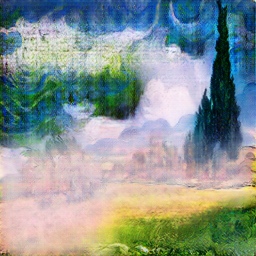} & \includegraphics[width=0.095\linewidth]{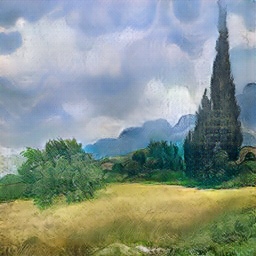} & \includegraphics[width=0.095\linewidth]{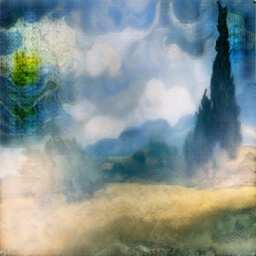} & \includegraphics[width=0.095\linewidth]{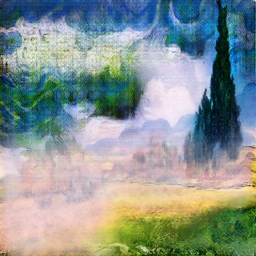} & \includegraphics[width=0.095\linewidth]{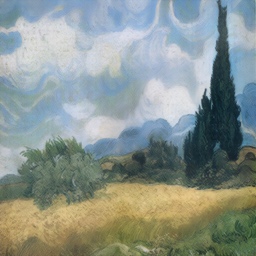}
			\\
			\multicolumn{11}{c}{\footnotesize{(e) Translation (CycleGAN, Van Gogh$\rightarrow$Photo)}}
		\end{tabular}
	\end{center}
	\caption{Example output images of the transformation-based defenses.}
	\label{fig:defense_transformation_images}
\end{figure*}

\begin{figure*}[t]
	\begin{center}
		\centering
		\renewcommand{\arraystretch}{1.0}
		\renewcommand{\tabcolsep}{1.5pt}
		\scriptsize
		\begin{tabular}{cccccccccccc}
			\multicolumn{2}{c}{\textbf{Original outputs}} & & \multicolumn{2}{c}{\textbf{I-FGSM ($\epsilon=8$)}} & & & \multicolumn{2}{c}{\textbf{Original outputs}} & & \multicolumn{2}{c}{\textbf{I-FGSM ($\epsilon=8$)}} \\
			No defense & \makecell[c]{With adversarial\\training} & & No defense & \makecell[c]{With adversarial\\training} & & & No defense & \makecell[c]{With adversarial\\training} & & No defense & \makecell[c]{With adversarial\\training} \\
			\includegraphics[width=0.115\linewidth]{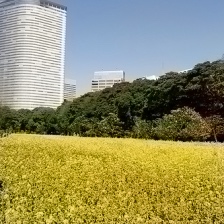} & \includegraphics[width=0.115\linewidth]{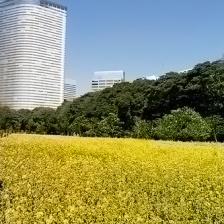} & &
			\includegraphics[width=0.115\linewidth]{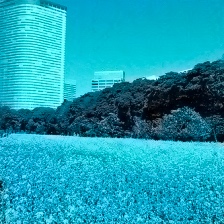} & \includegraphics[width=0.115\linewidth]{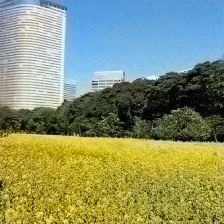} & & &
			\includegraphics[width=0.115\linewidth]{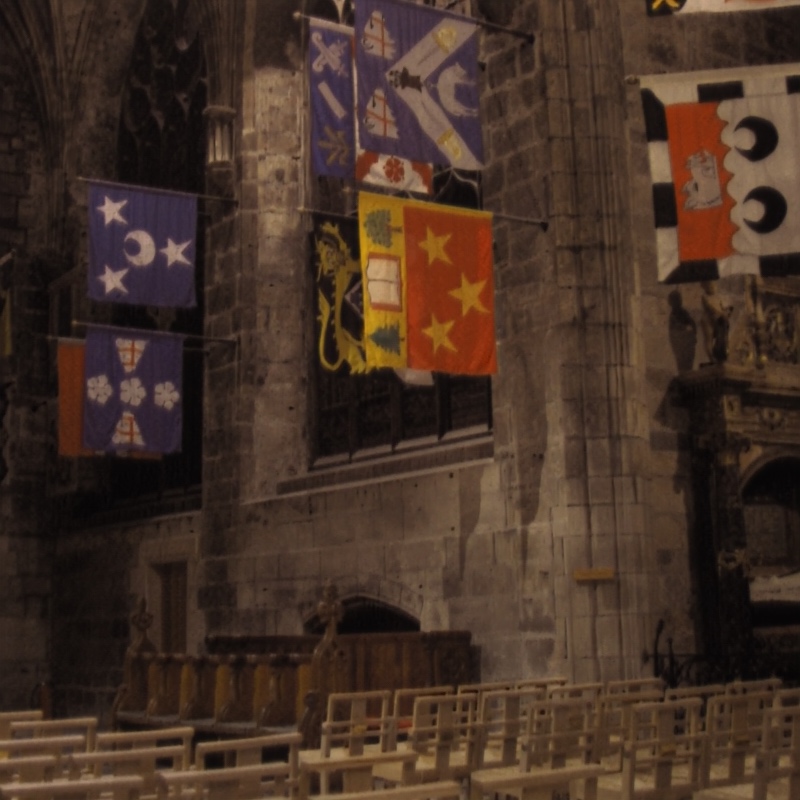} & \includegraphics[width=0.115\linewidth]{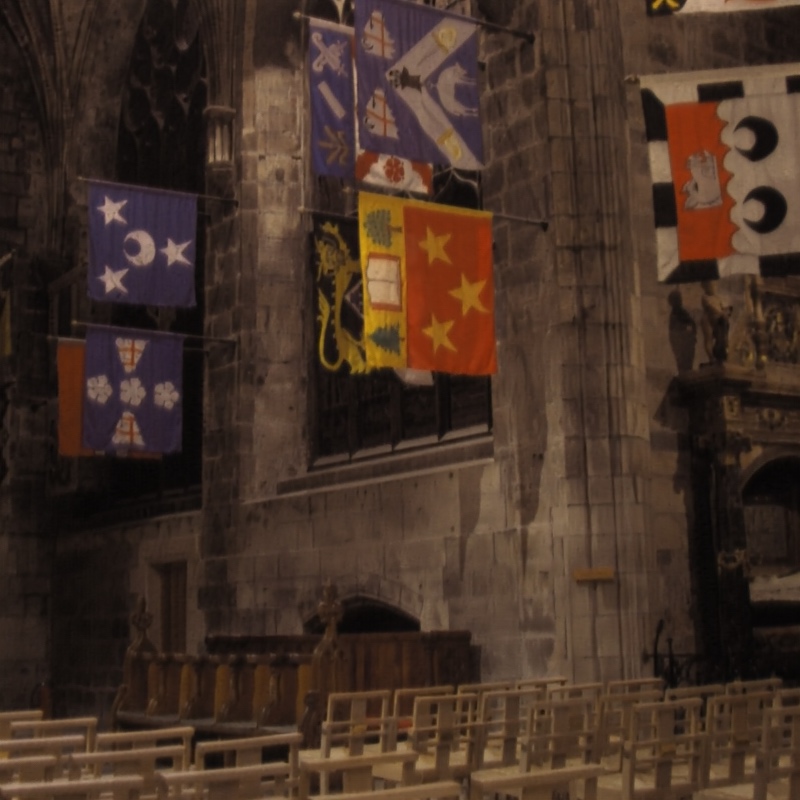} & &
			\includegraphics[width=0.115\linewidth]{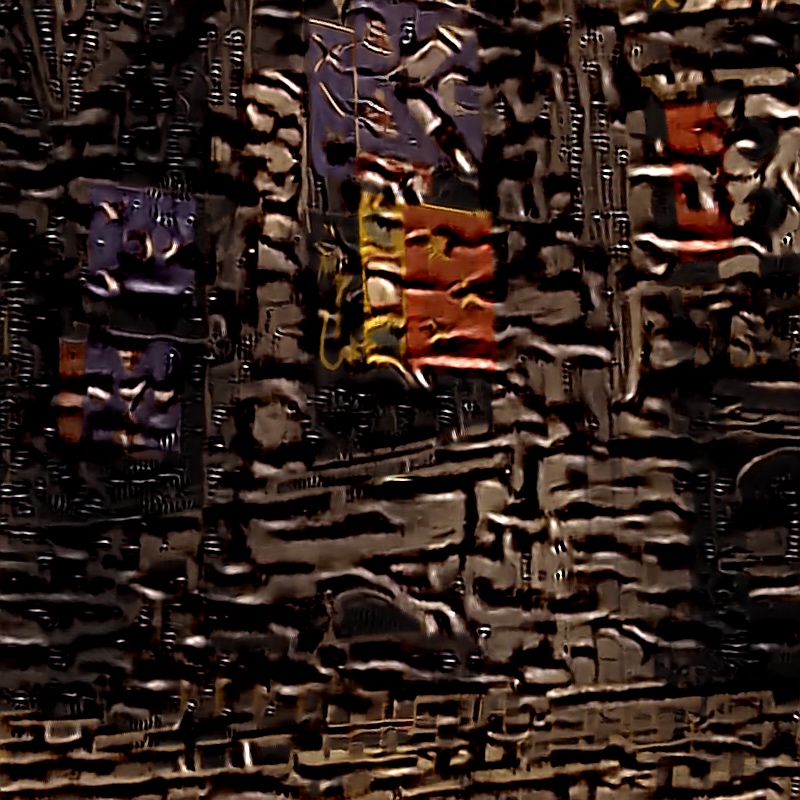} & \includegraphics[width=0.115\linewidth]{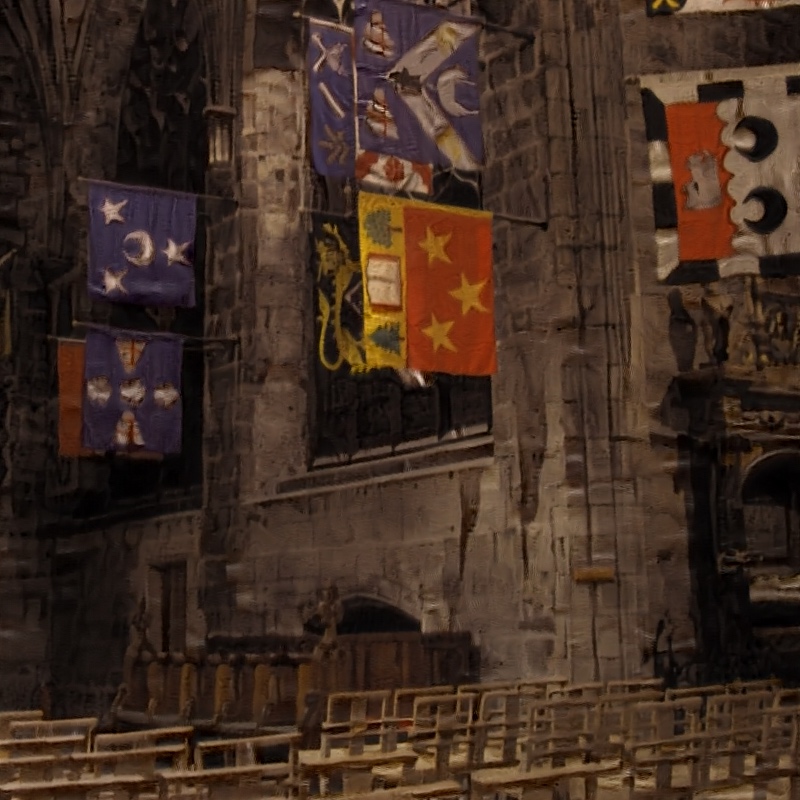} \\
			\multicolumn{5}{c}{\footnotesize{(a) Colorization (CIC)}} & & & \multicolumn{5}{c}{\footnotesize{(b) Deblurring (DeepDeblur)}} \\
			\includegraphics[width=0.115\linewidth]{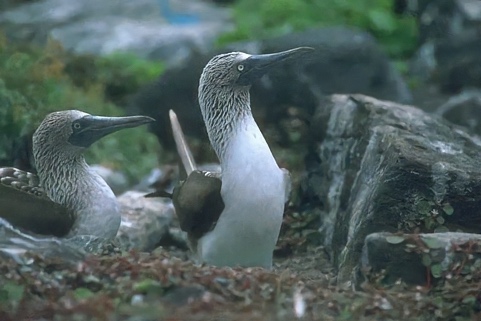} & \includegraphics[width=0.115\linewidth]{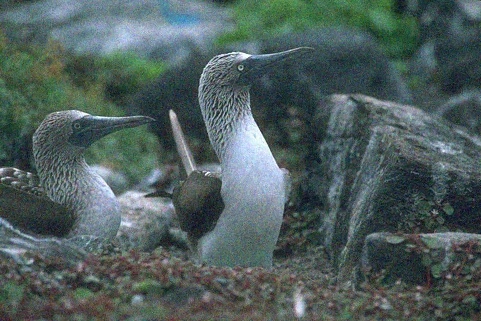} & &
			\includegraphics[width=0.115\linewidth]{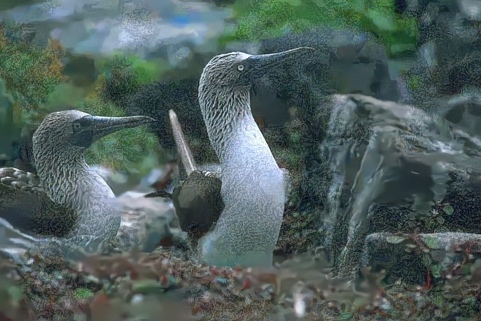} & \includegraphics[width=0.115\linewidth]{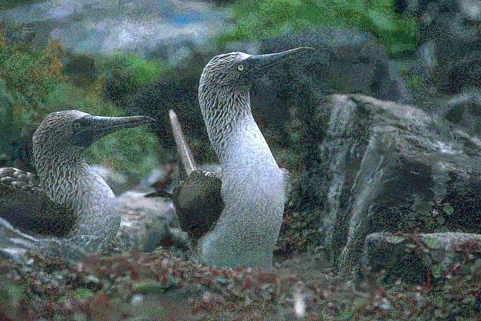} & & &
			\includegraphics[width=0.115\linewidth]{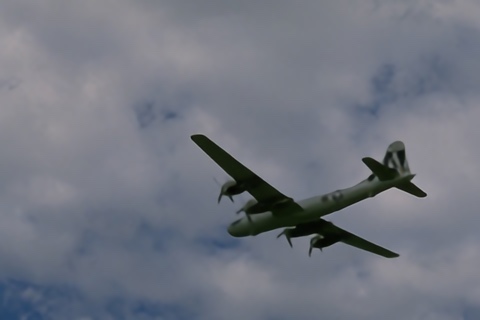} & \includegraphics[width=0.115\linewidth]{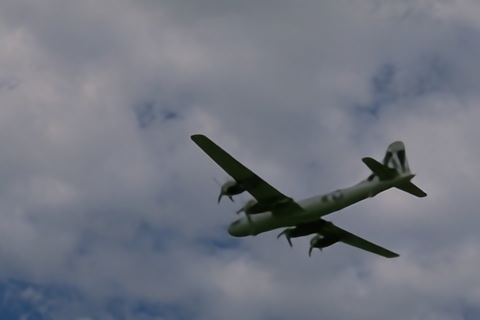} & &
			\includegraphics[width=0.115\linewidth]{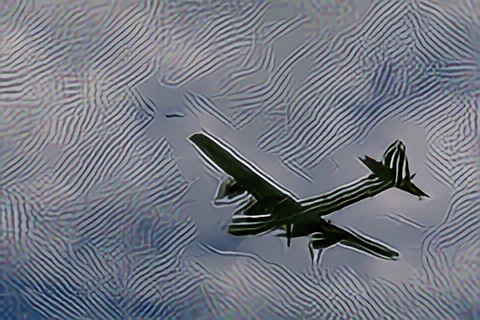} & \includegraphics[width=0.115\linewidth]{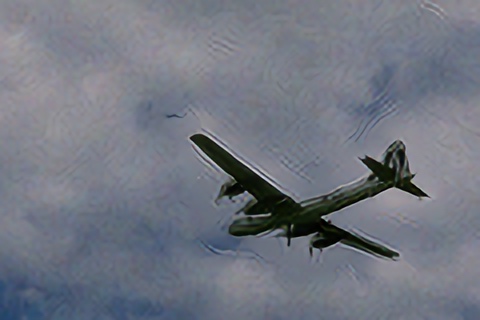} \\
			\multicolumn{5}{c}{\footnotesize{(c) Denoising (DnCNN (color, multiple $\sigma$))}} & & & \multicolumn{5}{c}{\footnotesize{(d) Super-resolution (CARN)}} \\
			\includegraphics[width=0.115\linewidth]{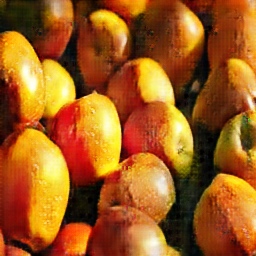} & \includegraphics[width=0.115\linewidth]{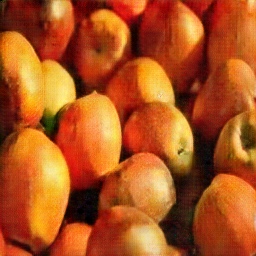} & &
			\includegraphics[width=0.115\linewidth]{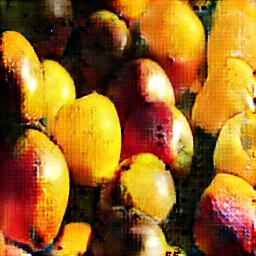} & \includegraphics[width=0.115\linewidth]{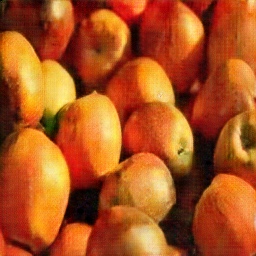} & & &
			\includegraphics[width=0.115\linewidth]{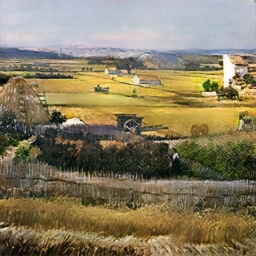} & \includegraphics[width=0.115\linewidth]{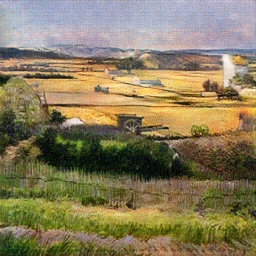} & &
			\includegraphics[width=0.115\linewidth]{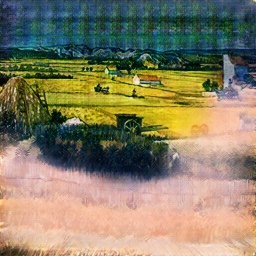} & \includegraphics[width=0.115\linewidth]{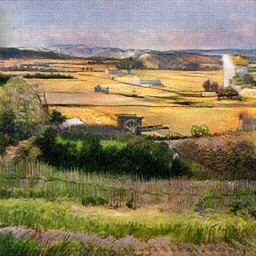} \\
			\multicolumn{5}{c}{\footnotesize{(e) Translation (CycleGAN, Apple$\rightarrow$Orange)}} & & & \multicolumn{5}{c}{\footnotesize{(f) Translation (CycleGAN, Van Gogh$\rightarrow$Photo)}} \\
		\end{tabular}
	\end{center}
	\caption{Example output images without and with the adversarial training.}
	\label{fig:defense_adversarial_training_images}
\end{figure*}

\clearpage



\bibliographystyle{IEEEtran}
\bibliography{did_arxiv}
%



\end{document}